\RequirePackage[l2tabu, orthodox]{nag} %

\documentclass[oneside]{um}  %
\usepackage{amsmath}
\usepackage{comment}

\usepackage{booktabs} 
\usepackage{enumitem}
\usepackage{algorithm}
\usepackage{algpseudocode}
\usepackage{xcolor}
\usepackage{tikz}
\usetikzlibrary{arrows,positioning,automata,calc,shapes,snakes,fit,backgrounds,svg.path}

\usepackage{array}
\usepackage{adjustbox}
\usepackage{graphicx, import}
\usepackage{pgfplots}
\usepackage{pgfplotstable}
\usepackage{placeins}

\usepackage{multirow}
\usepackage{color,soul} %
\usepackage{scalerel}
\usepackage{makecell}
\usepackage{dsfont}
\newcommand{\argmin}{\arg\!\min} %
\usepackage[utf8]{inputenc}
\inputencoding{utf8}
\definecolor{blue1}{HTML}{808080}
\usepackage{lineno}

\usepackage{amsmath}
\usepackage{tabularx}
\usepackage{xfrac}

\usepackage{multicol}

\newcommand*\circled[1]{\tikz[baseline=(char.base)]{
		\node[shape=circle,draw,inner sep=1pt] (char) {#1};}}

\newcommand{\redbullet}{{\color{UMRed}\scriptsize$\blacksquare$}}

\newcommand{\blankblock}{\tikz[baseline=(X.base)]\node[draw=black,rectangle,inner sep=0.7pt,rounded corners=0.4pt](X){\color{white}X};}

\makeatletter
\newcommand\footnoteref[1]{\protected@xdef\@thefnmark{\ref{#1}}\@footnotemark}
\makeatother

\newcommand{\equalss}{\text{=} \mkern2mu}
\newcommand{\crudespace}{ ${}$\ ${}$\ ${}$\ ${}$\ ${}$\ ${}$\ ${}$\ ${}$\ ${}$\ ${}$\ ${}$\ ${}$\ ${}$\ ${}$\ ${}$\ ${}$\ ${}$\ ${}$\ ${}$\ ${}$\ ${}$\ ${}$\ ${}$\ ${}$\ ${}$\ ${}$\ ${}$\ ${}$\ ${}$\ ${}$\ ${}$\ }
\newcommand\independent{\protect\mathpalette{\protect\independenT}{\perp}}
\def\independenT#1#2{\mathrel{\rlap{$#1#2$}\mkern3mu{#1#2}}}

\usepackage[makeroom]{cancel}
\usepackage{amssymb}  %
\usepackage{trimclip}

\usepackage{blindtext}
\usepackage{pdfpages}
\usepackage{bibentry}
\usepackage{epsdice}

\definecolor{orcidlogocol}{HTML}{A6CE39}
\tikzset{
	orcidlogo/.pic={
		\fill[orcidlogocol] svg{M256,128c0,70.7-57.3,128-128,128C57.3,256,0,198.7,0,128C0,57.3,57.3,0,128,0C198.7,0,256,57.3,256,128z};
		\fill[white] svg{M86.3,186.2H70.9V79.1h15.4v48.4V186.2z}
		svg{M108.9,79.1h41.6c39.6,0,57,28.3,57,53.6c0,27.5-21.5,53.6-56.8,53.6h-41.8V79.1z M124.3,172.4h24.5c34.9,0,42.9-26.5,42.9-39.7c0-21.5-13.7-39.7-43.7-39.7h-23.7V172.4z}
		svg{M88.7,56.8c0,5.5-4.5,10.1-10.1,10.1c-5.6,0-10.1-4.6-10.1-10.1c0-5.6,4.5-10.1,10.1-10.1C84.2,46.7,88.7,51.3,88.7,56.8z};
	}
}

\newcommand\orcidicon[1]{\href{https://orcid.org/#1}{\mbox{\scalerel*{
				\begin{tikzpicture}[yscale=-1,transform shape]
					\pic{orcidlogo};
				\end{tikzpicture}
			}{|}}}}

\usepackage{caption}
\definecolor{UMRedDark}{RGB}{80,10,20}
\newcommand{\capfnt}[1]{{\small\color{UMRed}\textbf{#1}}}
\newcommand{\acrofnt}[1]{{\color{UMRed}\textbf{#1}}}

\makeatletter
\newcommand\fs@nobottomruled{\def\@fs@cfont{\bfseries}\let\@fs@capt\floatc@ruled
	\def\@fs@pre{\hrule height.8pt depth0pt \kern2pt}%
	\def\@fs@post{}%
	\def\@fs@mid{\kern2pt\hrule\kern2pt}%
	\let\@fs@iftopcapt\iftrue}
\makeatother

\floatstyle{nobottomruled}
\restylefloat{algorithm}

\graphicspath{{./chap1/images/}{./figs}} 
\makeindex

\begin{document}	
\hypersetup{%
	colorlinks=false,
	pdfborder={0 0 0},
	pdfborderstyle={/S/U/W 0} %
}

\frontmatter
\thispagestyle{empty}
\includepdf[pages=-,pagecommand={}]{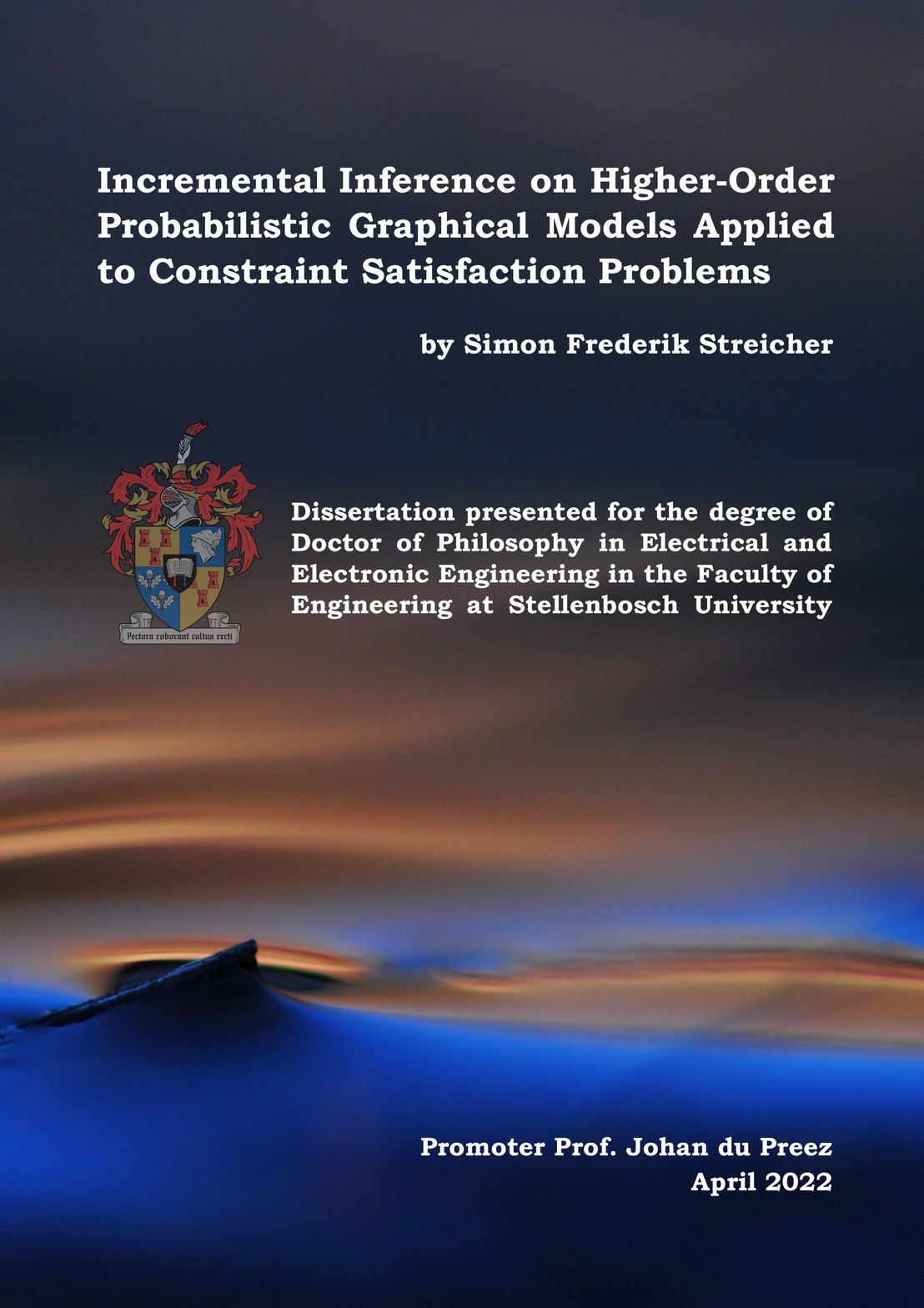}

\pagenumbering{roman}
\thispagestyle{plain}
\begin{center}
 {\Large\bfseries\sffamily\textcolor{UMRed}{Declaration}}\vspace{0.3em}
\end{center}

By submitting this dissertation electronically, I declare that the entirety of the work contained   therein is my own, original work, that I am the sole author thereof (save to the extent explicitly  otherwise stated), that reproduction and publication thereof by Stellenbosch University will not infringe any third party rights and that I have not previously in its entirety or in part submitted it for obtaining any qualification.

This dissertation includes three original papers published in peer-reviewed journals or books. %
The development and writing of the papers (published and unpublished) were the principal responsibility of myself and, for each of the cases where this is not the case, a declaration is included in the dissertation indicating the nature and extent of the contributions of co-authors.

$ $\\
\noindent { Date: April 2022}

{\color{white} \cite{streicher, landcoverdata, streicher2021strengthening, streichermasters}}

~\vfill
\begin{center}
\small
\vspace{0.3em}

Copyright \textcopyright\ \the\year\ Stellenbosch University %

All rights reserved
\end{center}
\newpage

\clearpage
\begin{abstract}
\thispagestyle{plain}
\markboth{\hspace{2.5em}Abstract}{\hspace{2.5em}Abstract}
Probabilistic graphical models (PGMs) are used extensively in the probabilistic reasoning domain. They are powerful tools for solving systems of complex relationships over a variety of probability distributions, such as
medical and fault diagnosis,
predictive modelling, 
object recognition, 
localisation and mapping,
speech recognition,
and language processing~\cite{
wemmenhove2001inference, %
rish2005distributed,     %
zhang2015ground,         %
dellaert2021factor,      %
glarner2016factor,       %
bernier2009fast,         %
taylor2021albu}.
Furthermore, constraint satisfaction problems (CSPs) can be formulated as PGMs and solved with PGM inference techniques. However, the prevalent literature on PGMs shows that suboptimal PGM structures are primarily used in practice and a suboptimal formulation for constraint satisfaction PGMs.

This dissertation aimed to improve the PGM literature through accessible algorithms and tools for improved PGM structures and inference procedures, specifically focusing on constraint satisfaction. To this end, this dissertation presents three published contributions to the current literature:
\begin{itemize}
    \item a comparative study to compare cluster graph topologies to the prevalent factor graphs~\cite{streicher},
    \item an application of cluster graphs in land cover classification in the field of cartography~\cite{landcoverdata}, and
    \item a comprehensive integration of various aspects required to formulate CSPs as PGMs and an algorithm to solve this formulation for problems too complex for traditional PGM tools~\cite{streicher2021strengthening}.
\end{itemize}

First, we present a means of formulating and solving graph colouring problems with probabilistic graphical models. In contrast to the prevailing literature that mostly uses factor graph configurations, we approach it from a cluster graph perspective, using the general-purpose cluster graph construction algorithm, LTRIP. Our experiments indicate a significant advantage for preferring cluster graphs over factor graphs, both in terms of accuracy as well as computational efficiency.

Secondly, we use these tools to solve a practical problem: land cover classification. This process is complex due to measuring errors, inefficient algorithms, and low-quality data. We proposed a PGM approach to boost geospatial classifications from different sources and consider the effects of spatial distribution and inter-class dependencies (similarly to graph colouring).
Our PGM tools were shown to be robust and were able to produce a diverse, feasible, and spatially-consistent land cover classification even in areas of incomplete and conflicting evidence. 

Lastly, in our third publication, we investigated and improved the PGM structures used for constraint satisfaction. It is known that tree-structured PGMs always result in an exact solution~\cite[p355]{koller}, but is usually impractical for interesting problems due to exponential blow-up. 
We, therefore, developed the ``purge-and-merge'' algorithm  to incrementally approximate a tree-structured PGM. 
This algorithm iteratively nudges a malleable graph structure towards a tree structure by selectively \textit{merging} factors. The merging process is designed to avoid exponential blow-up through sparse data structures from which redundancy is \textit{purged} as the algorithm progresses. This algorithm is tested on constraint satisfaction puzzles such as Sudoku, Fill-a-pix, and Kakuro and manages to outperform other PGM-based approaches reported in the literature~\cite{BaukeH, GoldbergerJ, KhanS}. Overall, the research reported in this dissertation contributed to developing a more optimised approach for higher-order probabilistic graphical models.
Further studies should concentrate on applying purge-and-merge on problems closer to probabilistic reasoning than constraint satisfaction and report its effectiveness in that domain.

\end{abstract}
\newpage
\renewcommand{\abstractname}{\Large\bfseries\sffamily\textcolor{UMRed}{Uittreksel}}
\begin{abstract}
\thispagestyle{plain}
\markboth{\hspace{2.5em}Uittreksel}{\hspace{2.5em}Uittreksel}

Grafiese waarskynlikheidsmodelle (PGM) word wyd gebruik vir komplekse waar\-skynlikheidsprobleme. Dit is kragtige gereedskap om sisteme van komplekse verhoudings oor ‘n versameling waarskynlikheidsverspreidings op te los, soos die mediese en foutdiagnoses, voorspellingsmodelle, objekherkenning, lokalisering en kartering, spraakherkenning en taalprosessering~\cite{
wemmenhove2001inference, %
rish2005distributed,     %
zhang2015ground,         %
dellaert2021factor,      %
glarner2016factor,       %
bernier2009fast,         %
taylor2021albu}.
Voorts kan beperkingvoldoeningsprobleme (CSP) as PGM’s geformuleer word en met PGM ge\-volg\-trek\-king\-teg\-nieke opgelos word. Die heersende literatuur oor PGM’s toon egter dat sub-optimale PGM-strukture hoofsaaklik in die praktyk gebruik word en ‘n sub-optimale PGM-formulering vir CSP's.

Die doel met die verhandeling is om die PGM-literatuur deur toeganklike algoritmes en gereedskap vir verbeterde PGM-strukture en gevolgtrekking-prosedures te verbeter deur op CSP toepassings te fokus. Na aanleiding hiervan voeg die verhandeling drie gepubliseerde bydraes by die huidige literatuur: 
\begin{itemize}
	\item ‘n vergelykende studie om bundelgrafieke tot die heersende faktorgrafieke te vergelyk~\cite{streicher},
	\item ‘n praktiese toepassing vir die gebruik van bundelgrafieke in “land-cover''-klassifikasie in die kartografieveld~\cite{landcoverdata}  en
	\item ‘n omvattende integrasie van verskeie aspekte om CSP’s as PGM’s te formuleer en ‘n algoritme vir die formulering van probleme te kompleks vir tradisionele PGM-gereedskap~\cite{streicher2021strengthening}.
\end{itemize}

Eerstens bied ons ‘n wyse van formulering en die oplos van grafiekkleurprobleme met PGM's. In teenstelling met die huidige literatuur wat meestal faktorgrafieke gebruik, benader ons dit van ‘n bundelgrafiek-perspektief deur die gebruik van die automatiese bundel\-grafiek\-kon\-struk\-sie-algo\-ritme, LTRIP. Ons eksperimente toon ‘n beduidende voorkeur vir bundelgrafieke teenoor faktorgrafieke, wat akkuraatheid asook berekende doeltreffendheid betref.

Tweedens gebruik ons die gereedskap om ‘n praktiese probleem op te los: ``land-cover''-klassifikasie. Die proses is kompleks weens metingsfoute, ondoeltreffende algoritmes en lae-gehalte data. Ons stel ‘n PGM-benadering voor om die geo-ruimtelike klassifikasies van verskillende bronne te versterk, asook die uitwerking van ruimtelike verspreiding en interklas-afhanklikhede (soortgelyk aan grafiek\-kleur\-probleme). Ons PGM-gereedskap is robuus en kon ‘n diverse, uitvoerbare en ruim\-telik-konsekwente ``land-cover''-klassifikasie selfs in gebiede van onvoltooide en konflikterende inligting bewys.

Ten slotte het ons in ons derde publikasie die PGM-strukture vir CSP's ondersoek en verbeter. Dit is bekend dat boomstrukture altyd tot ‘n eksakte oplossing lei~\cite[p355]{koller}, maar is weens eksponensiële uitbreiding gewoonlik onprakties vir interessante probleme. Ons het gevolglik die algoritme, purge-and-merge, ontwikkel om inkrementeel ‘n boomstruktuur na te doen.

Die algoritme hervorm ‘n bundelgrafiek stapsgewys in ‘n boomstruktuur deur faktore selektief te ``merge''. Die saamsmeltproses is ontwerp om eksponensiële uitbreiding te vermy deur van yl datastrukture gebruik te maak waarvan die waar\-skein\-lik\-heidsruimte  ge-``purge'' word namate die algoritme vorder. Die algoritme is ge\-toets op CSP-speletjies soos Sudoku, Fill-a-pix en Kakuro en oortref ander PGM-gegronde benaderings waaroor in die literatuur verslag gedoen word \cite{BaukeH, GoldbergerJ, KhanS}. In die geheel gesien, het die navorsing bygedra tot die ontwikkeling van ‘n meer geoptimaliseerde benadering vir hoër-orde PGM's. Verdere studies behoort te fokus op die toepassing van purge-and-merge op probleme nader aan waarskynlikheidsredenasie-probleme as aan CSP's en moet  sy effektiwiteit in daar\-die domein rapporteer.

\end{abstract}
\newpage

\thispagestyle{plain}
\begin{center} {\Large\bfseries\sffamily\textcolor{UMRed}{Acknowledgements}}\vspace{0.3em} \end{center}
    
\begin{itemize}
\item First and foremost, I would like to thank my promoter, Prof.\ Johan du Preez, for your help, encouragement, and most of all, your friendship. Furthermore, thanks for using your beautiful photograph as the cover page of this dissertation.
\item Secondly, I would like to thank my wife, Tina Streicher, for your support through the years of writing, especially for the last stretch. Thank you for your love and support; none of this would have been possible without your encouragement and perseverance.
\item I would like to thank Lloyd Hughes and Ekaterina Chuprikova for collaborating with me.
\item Also, thank you, Tarl Berry, Petro Wagner, Dirk Streicher, Jaco Briers, Elretha Britz, Jacques Smidt, and Francois Kamper for direct help and technical support in completing this work, Malan Kriel for patiently waiting for my full involvement in Auto Actuary, my parents for their continual encouragement, and God for overseeing it all.
\item  For funding a large portion of this research through bursaries, I would also like to thank Stone Three.
\end{itemize}

\newpage

\thispagestyle{plain}
\renewcommand{\thefootnote}{\fnsymbol{footnote}}
\begin{center}
	{\Large\bfseries\sffamily\textcolor{UMRed}{Declaration of publications}\footnote[2]{\label{footnotelable}Declaration with signatures is in possession of dissertation author and promoter.}}
\end{center}

With regard to the publications within this dissertation, the contributions of author S.\ Streicher\hspace{0.1em}\raisebox{0.27em}{\scalebox{0.9}{\orcidicon{0000-0001-9281-0411}}}\hspace{0.1em} were as follows:
\begin{description}
	\item[\hspace{1.3em}\cite{streicher}] Chapter~\ref{gc-chapter}:\ \ S. Streicher and J. du Preez, “Graph Coloring: Comparing Cluster Graphs to Factor Graphs,” in \textit{Proceedings of the ACM Multimedia 2017 Workshop on South African Academic Participation}. SAWACMMM ’17. New York, NY, USA: ACM, 2017, pp. 35–42.
	\vspace{0.2em}
	\\
	${}$\hspace{-0.5em}\redbullet\ \capfnt{90\% S.\ Streicher:} {\small
		Majority of the contributions. Paper writing and revision.
	}
	\\
	${}$\hspace{-0.5em}\redbullet\ \capfnt{10\% J.\ du Preez:} {\small
		Idea and discovery of LTRIP algorithm. Critical revision of the paper.
	}
	\item[\hspace{1.3em}\cite{landcoverdata}] Chapter~\ref{lc-chap}:\ \  L. H. Hughes, S. Streicher, E. Chuprikova, and J. du Preez. ``A Cluster Graph Approach to Land Cover Classification Boosting.'' \textit{Data}, 4(1), 2019. ISSN 2306-5729.
	\vspace{0.2em}
	\\${}$\hspace{-0.5em}\redbullet\  \capfnt{40\% L.\ H.\ Hughes:} {\small
		Idea, methodological, and experimental
		formulations. Data handling and execution
		of experiments. Paper writing and revision.
	}
	\\${}$\hspace{-0.5em}\redbullet\ \capfnt{35\% S.\ Streicher:} {\small
		The formulation, design, and execution of the PGMs used in this work, specifically Section~\ref{lc-sec:cluster-graph},  Section~\ref{lc-sec:lccpgm}, and the design of the potential functions in Section~\ref{lc-sec:priors}. Paper writing.
	}
	\\${}$\hspace{-0.5em}\redbullet\ \capfnt{20\% E.\ Chuprikova:} {\small
		Idea, data preprocessing, and evaluation of
		results in comparison to existing
		approaches. Paper writing.
	}
	\\${}$\hspace{-0.5em}\redbullet\ \capfnt{5\% J.\ du Preez:} {\small
		Critical revision of the paper.
	}
	\item[\hspace{1.3em}\cite{streicher2021strengthening}] Chapter~\ref{csp-chap}:\ \  S. Streicher and J. du Preez, “Strengthening Probabilistic Graphical Models: The Purge-and-merge Algorithm,” \textit{IEEE Access}, vol. 9, pp. 149 423–149 432, 2021.
	\vspace{0.2em}
	\\
	${}$\hspace{-0.5em}\redbullet\ \capfnt{90\% S.\ Streicher:} {\small
		Majority of the contributions. Paper writing and revision.
	}
	\\
	${}$\hspace{-0.5em}\redbullet\ \capfnt{10\% J.\ du Preez:} {\small
		Research supervision and technical expertise in guiding the scope. Cri\-tical revision of the paper.
	}
	
\end{description}

\begin{center}
	\vspace{1em}
	{\Large\bfseries\sffamily\textcolor{UMRed}{Declaration by co-authors}}{\large\textsuperscript{\ref{footnotelable}}}
\end{center}

{\small
	The undersigned hereby confirm that (1) the declaration above accurately reflects the nature and extent of the contributions of the candidate and the co-authors, (2) no other authors contributed besides those specified above, and (3) potential conflicts of interest have been revealed to all interested parties and that the necessary arrangements have been made to use the material in this dissertation.}

\renewcommand{\thefootnote}{\arabic{footnote}}

\newpage
{
    \clearpage           %
    \thispagestyle{empty}%
    \vspace*{\stretch{1}}%
    \itshape             %
    \raggedleft          %
    \vspace{-0.8em}\textcolor{UMRed}{\large{To Tina}}\ \ \  \ \ \ ${}$\\
    \vspace{-0.8em}\includegraphics[scale=0.25, bb= 0in 0in 70em -10em]{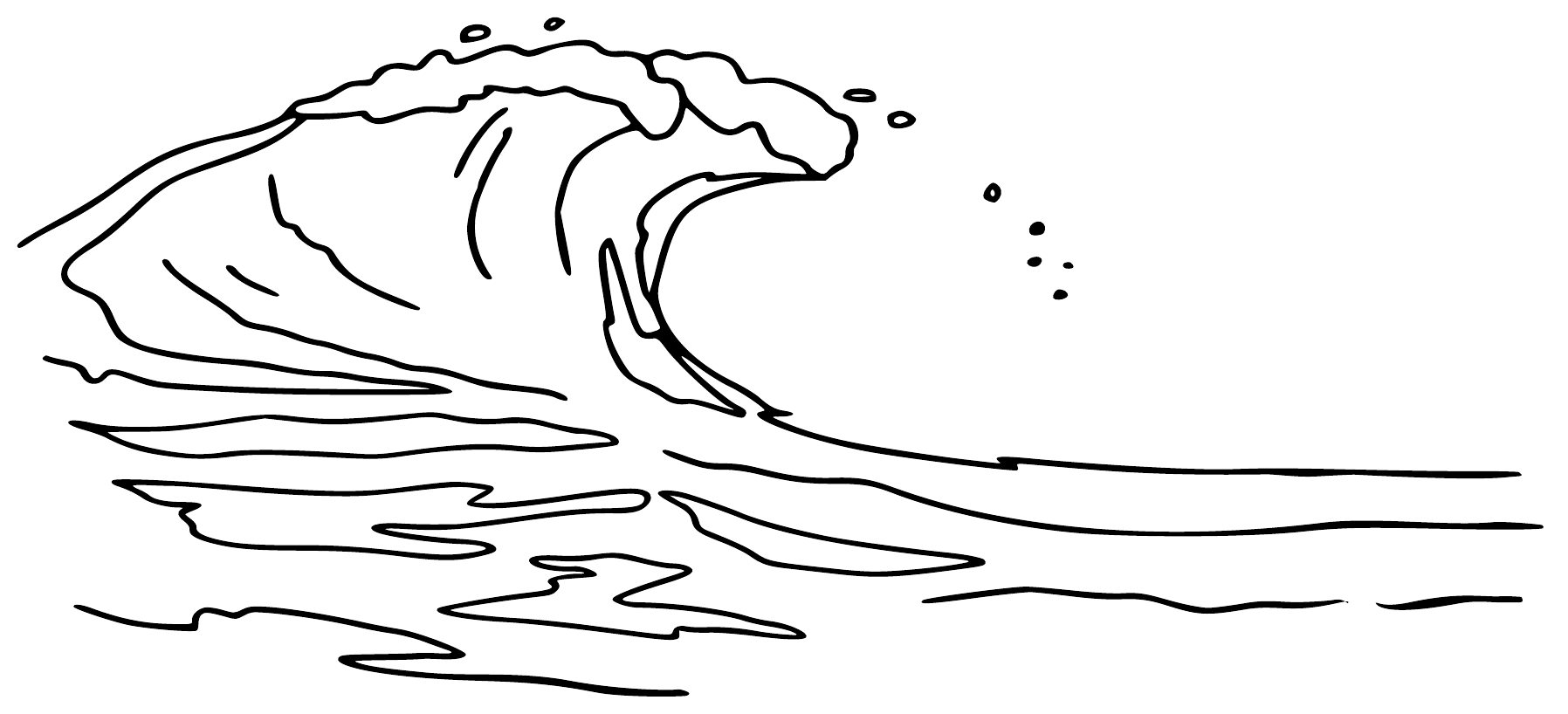} \\
     for your faith in my work when I had none left\\
    \par %
    \vspace{\stretch{3}} %
    \if@openright\cleardoublepage\else\clearpage\fi
}

\tableofcontents*
\if@openright\cleardoublepage\else\clearpage\fi

{
\chapter*{}
\begin{flushright}
	\vspace{-96pt}{\Huge\bfseries\sffamily\textcolor{UMRed}{Abbreviations}}
\end{flushright}
\vspace{12pt}
\begin{description}[itemsep=0.3pt,parsep=1pt]
	\item[\acrofnt{AC}]  arithmetic circuit 
	\item[\acrofnt{ACE}]  AC compilation and evaluation
	\item[\acrofnt{AI}]  artificial intelligence 
	\item[\acrofnt{ASTER}]  Advanced Space-borne Thermal Emission and Reflection
	\item[\acrofnt{ATKIS}]  Amtliches Topographisch-Kartographisches Informations System
	\item[\acrofnt{BP}]  belief propagation
	\item[\acrofnt{BU}]  belief update
	\item[\acrofnt{CART}]  classification and regression tree 
	\item[\acrofnt{CLC2006}]  CORINE Land Cover 2006
	\item[\acrofnt{CNF}]  conjugate normal form 
	\item[\acrofnt{CNN}]  convolutional neural network
	\item[\acrofnt{CORINE}]  Coordination of Information on the Environment
	\item[\acrofnt{CSP}]  constraint satisfaction problem
	\item[\acrofnt{DEM}]  digital elevation model 
	\item[\acrofnt{LSTM}]  long short-term memory 
	\item[\acrofnt{LTRIP}]  layered trees for the running intersection property
	\item[\acrofnt{ML}]  maximum likelihood 
	\item[\acrofnt{MODIS}]  Moderate Resolution Imaging Spectroradiometer 
	\item[\acrofnt{OLI}]  Operational Land Imager 
	\item[\acrofnt{OSM}]  OpenStreetMap 
	\item[\acrofnt{PGM}]  probabilistic graphical model
	\item[\acrofnt{RF}]  random-forest 
	\item[\acrofnt{RIP}]  running intersection property 
	\item[\acrofnt{SDD}]  sentential decision diagram
	\item[\acrofnt{SER}]  Sudoku explainer rating 
	\item[\acrofnt{SVM}]  support vector machine
	\item[\acrofnt{VGI}]  volunteered geographic information 
\end{description}
}

{
\chapter*{}
\begin{flushright}
	\vspace{-96pt}{\Huge\bfseries\sffamily\textcolor{UMRed}{Glossary}}
\end{flushright}
\vspace{12pt}
\markboth{Glossary}{Glossary}
\begin{description}[itemsep=0.25pt, parsep=1pt]
	
	\item[\acrofnt{arithmetic circuit}]
	An alternative representation of a Bayes network for rapid conditional and marginal queries -- see ACE~\cite{acewebsite} for a software implementation of such a system.
	
	\item[\acrofnt{ACE}]
	Software that compiles a Bayes network into an arithmetic circuit to answer conditional and marginal queries -- available at ACE~\cite{acewebsite}.
	
	\item[\acrofnt{Bayes network}]
	 A PGM structure representing a set of variables and their conditional dependencies via a directed acyclic graph -- see Section \ref{sec:factorising-pgms} and Figure~\ref{fig:SOL-example-Bayes-network}.
	
	\item[\acrofnt{belief propagation}]
	A message-passing algorithm for performing inference on graphical models -- well-defined on tree-structured graphs and results in exact inference. See Sections \ref{sec:chap1-intro-pgms} and \ref{sec:belief-propegation} and Algorithm~\ref{alg:sumproducttree}.
	
	\item[\acrofnt{belief update}]
	An equivalent algorithm to belief propagation that requires fewer calculations -- introduced by \citet{lauritzen1988local}.
	
	\item[\acrofnt{Calcudoku}]
	A similar puzzle to Killer Sudoku, but with the custom regions having both a number and an operator $+$, $-$, $\times$ or $\div$, such that applying a region's operator to its cells (e.g.\ $\text{\blankblock}\times\text{\blankblock}\times\text{\blankblock}$\hspace{0.09em}) must yield the number attached to the region.
	
	\item[\acrofnt{cardinality}]
	Used in this dissertation in the context of discrete random variables, where it refers to the size of the domain of the variable -- e.g.\ a random variable representing a die roll has a cardinality of 6.
	
	\item[\acrofnt{classifications boosting}]
	A method for combining multiple classifications into a stronger classification by compensating for the weaknesses of the individual classifications -- see Chapter~\ref{lc-chap}.
	
	\item[\acrofnt{cluster graph}]
	A PGM structure that allows for multivariate message passing -- see Sections~\ref{sec:chap1-intro-graphs}, \ref{gc-sec:clustergraphs}, \ref{gc-sec:ltrip}, and \ref{csp-sec:ltrip}.
	
	\item[\acrofnt{constraint satisfaction problem}]
	The problem of assigning values to variables under a given set of constraints over those variables -- see Sections~\ref{sec:chap1-intro-csp} and \ref{csp-sec-CSP}.

	\item[\acrofnt{determinism}]
	Used in this dissertation in the context of constraint satisfaction and potential tables, where it refers to zero-probability entries -- i.e.\ states within a system that are deterministically impossible to be part of a solution.
	
	\item[\acrofnt{domain}]
	Used in this dissertation in the context of discrete random variables, where it refers to the set of possible outcomes of a random variable -- e.g.\ a random variable representing a die roll can be represented by the domain $\{\text{\epsdice{1}},\text{\epsdice{2}},\text{\epsdice{3}},\text{\epsdice{4}},\text{\epsdice{5}},\text{\epsdice{6}}\}$.
	
	\item[\acrofnt{error correction code}]
	Formulation for adding redundant information to data in order to detect possible errors later on and correct those errors -- see Hamming (7,4) in Section~\ref{sec:SOL-hamming-window-example} for an example.
	
	\item[\acrofnt{exponential blow-up}]
	An informal term to indicate a non-linear, exponential growth of a solution space with regard to an increase in variables.
	
	\item[\acrofnt{factor graph}]
	A simple PGM structure that supports univariate message passing -- see Sections~\ref{sec:chap1-intro-graphs} and \ref{gc-sec:factorgraphs}.
	
	\item[\acrofnt{Fill-a-pix}]
	A paper-and-pencil adaptation of the classic computer game Minesweeper, where cells in a grid are pre-filled with digits to indicate how many neighbouring cells (including the cell with the digit) are to be painted in, in order to reveal an underlying pixel art image.
	
	\item[\acrofnt{generalised arc consistency}]
	A type of consistency within a constraint satisfaction system, if a state shared by two variables are deterministic in one factor, it should be deterministic in all other factors -- see Section~\ref{csp-sec:factorpurging} and \citet{dechter2010on}.
	
	\item[\acrofnt{graph colouring}]
	 The colouring (or labelling) of nodes in an undirected graph such that adjacent nodes do not have the same colour -- see Section~\ref{gc-sec:graph_coloring_description} and Figure~\ref{csp-fig-mapcolouring}.

	\item[\acrofnt{junction tree}]
	A PGM structure for exact inference, similar to a cluster graph but with the condition that the graph is tree-structured -- see Figure~\ref{fig:SOL-cluster-tree} for an example and Koller~\cite[p287]{koller} for variable elimination as a construction algorithm.
	
	\item[\acrofnt{Kakuro}]
	A constraint puzzle similarly shaped to a crossword, where squares are to be filled in with digits $1$ to $9$ in order to sum up to an indicated number but without repeating a digit.
	
	\item[\acrofnt{Killer Sudoku}]
	A similar puzzle to Sudoku but with additional custom regions and associated numbers, such that the sum of all cells within a region must sum to the associated number.
	
	\item[\acrofnt{Kullbach-Leibler divergence}]
	A measure of the dissimilarity of one probability distribution compared to another distribution -- see Section~\ref{sec:loopy-belief-propagation}.
	
	\item[\acrofnt{land cover classification}]
	Classifying a map according to land cover classes, such as ``forest'', ``grassland'', ``water'', ``artificial surface'', and many others -- see Section~\ref{sec:landcover-intro}.
	
	\item[\acrofnt{LTRIP}]
	An algorithm to construct a cluster graph from set of factors -- introduced in Section~\ref{gc-sec:ltrip} and summarised in Section~\ref{csp-sec:ltrip}.
	
	\item[\acrofnt{loopy graph}]
	A graph topology that allows for multiple paths between any two nodes within the graph -- i.e.\ a graph that contains cycles.
	
	\item[\acrofnt{max marginalisation}]
	Similar to marginalisation over a probability distribution but instead of summing over grouped values, the maximum value in the group is taken. This operation can replace marginalisation during message passing if only the most likely assignment over the system is required.
	
	\item[\acrofnt{maximal clique}]
	A subset of nodes from an undirected graph, where every node is adjacent to all other nodes and the subset is not extendable any further.
	
	\item[\acrofnt{maximum spanning tree}]
	Remove edges from a loopy graph with weighted edges such that the result is a tree structure with maximum possible weight -- see the Prim-Jarn\'{i}k algorithm~\cite{prim} used by LTRIP in Section~\ref{gc-sec:ltrip}.

	\item[\acrofnt{probabilistic graphical model}]
	Models that encode complex joint multivariate probability distributions using graphs -- see Section~\ref{sec:chap1-intro-pgms} and Chapter~\ref{pgm-chapter}.
	
	\item[\acrofnt{purge-and-merge}]
	An algorithm to systematically simplify inference on constraint satisfaction problems -- introduced in Section~\ref{csp-sec:purgeandmerge-main} and outlined in Algorithm~\ref{csp-alg:purgeandmerge}.
	
	\item[\acrofnt{random variable}]
	A variable with an uncertain value from a specific domain of possible values. See Section~\ref{sec:PRO-probability-distributions}.
	
	\item[\acrofnt{region graph}]
	A more general PGM structure than factor graphs and cluster graphs, see Sections~\ref{sec:chap1-intro-graphs}.
	
	\item[\acrofnt{running intersection property}]
	A necessary property for valid cluster graphs -- for any two clusters sharing a variable, there must exist a unique path of clusters and sepsets between them containing the same variable. See Sections~\ref{gc-sec:clustergraphs} and \ref{csp-sec:ltrip} and Koller~\cite[p347]{koller}.
	
	\item[\acrofnt{Shannon diversity index}]
	A metric based on Claude Shannon's formula for entropy for the variety of land uses in an area --  see \citet{duvsek2017theoretical} for a discussion on the usage and validity of this metric.
	
	\item[\acrofnt{Sudoku}]
	A constraint puzzle with a partially filled $9 \times 9$ grid, where the digits $1$ to $9$ are to be assigned to each cell such that every digit appears only once in a region, with the regions as the nine rows, nine columns, and nine non-overlapping $3 \times 3$ sub-grids.
	
	\item[\acrofnt{sum marginalisation}]
	The correct form of marginalisation over a probability distribution, where the sum is taken over grouped values.
	
	\item[\acrofnt{tree-structured graph}]
	A graph topology where there is a unique path between any two nodes -- i.e.\ there are no cycles in the graph.
	
	\item[\acrofnt{volunteered geographical information}]
	The collection, analysis, and sharing of geographic information provided by individuals.
\end{description}
}

\mainmatter 
\pagestyle{umpage}
\chapter{Introduction}

\section{Background}
\subsection{Probabilistic graphical models} \label{sec:chap1-intro-pgms}
Probabilistic graphical models (PGMs) originated with the discovery of belief propagation. \citet{pearl1986fusion} first proposed belief propagation as an exact inference procedure on trees. Then, after some investigation for its merit on loopy structures, he concluded that a loopy version of belief propagation can converge, but that ``this asymptotic equilibrium is not coherent, in the sense that it does not represent the posterior probabilities of all nodes in the network’’\cite{pearl1988probabilistic}.

Independent discoveries in belief propagation were unknowingly made with turbo codes~\cite{berrou1993near} as an iterative scheme for solving complex error codes. This came as a practical advancement in coding theory as it achieved near Shannon limit error-correcting coding and decoding but did not provide a general framework or a theoretical justification.  \citet{mceliece1998turbo} later discovered that turbo codes are simply an application of loopy belief propagation on a loopy Bayes network that captures the error coding and channel noise. 

By establishing the potential for PGMs in practical applications, a new interest in the field was developed for approximate inference via belief propagation schemes~\cite{rish1998empirical, murphy1999loopy, yedidia2000generalized, weiss2001on}. This can be attributed to both the effectiveness of PGMs as well as the expressiveness. Intricate problems with multiple dependencies can easily be formulated into graphs and solved via PGM inference. This is useful for a variety of complex scenarios, where PGMs can be used to
\begin{itemize}
	\item fill in a system's blind spot if a problem is sparsely defined,
	\item integrate an additional complexity level, such as measurement-noise parameters and other uncertainties, and
	\item solve problems with inverse relationships that are difficult to derive directly from calculus.
\end{itemize}
Thus, it is not surprising that PGMs are integral to various applications, such as 
medical and fault diagnosis,
predictive modelling, 
object recognition, 
localisation and mapping,
speech recognition,
and language processing~\cite{
wemmenhove2001inference, %
rish2005distributed, %
zhang2015ground, %
dellaert2021factor, %
glarner2016factor, %
bernier2009fast, %
taylor2021albu}.

\subsection{Graph structures}\label{sec:chap1-intro-graphs}
\begin{figure}[h!]
    \centering
    \includegraphics[width=\columnwidth]{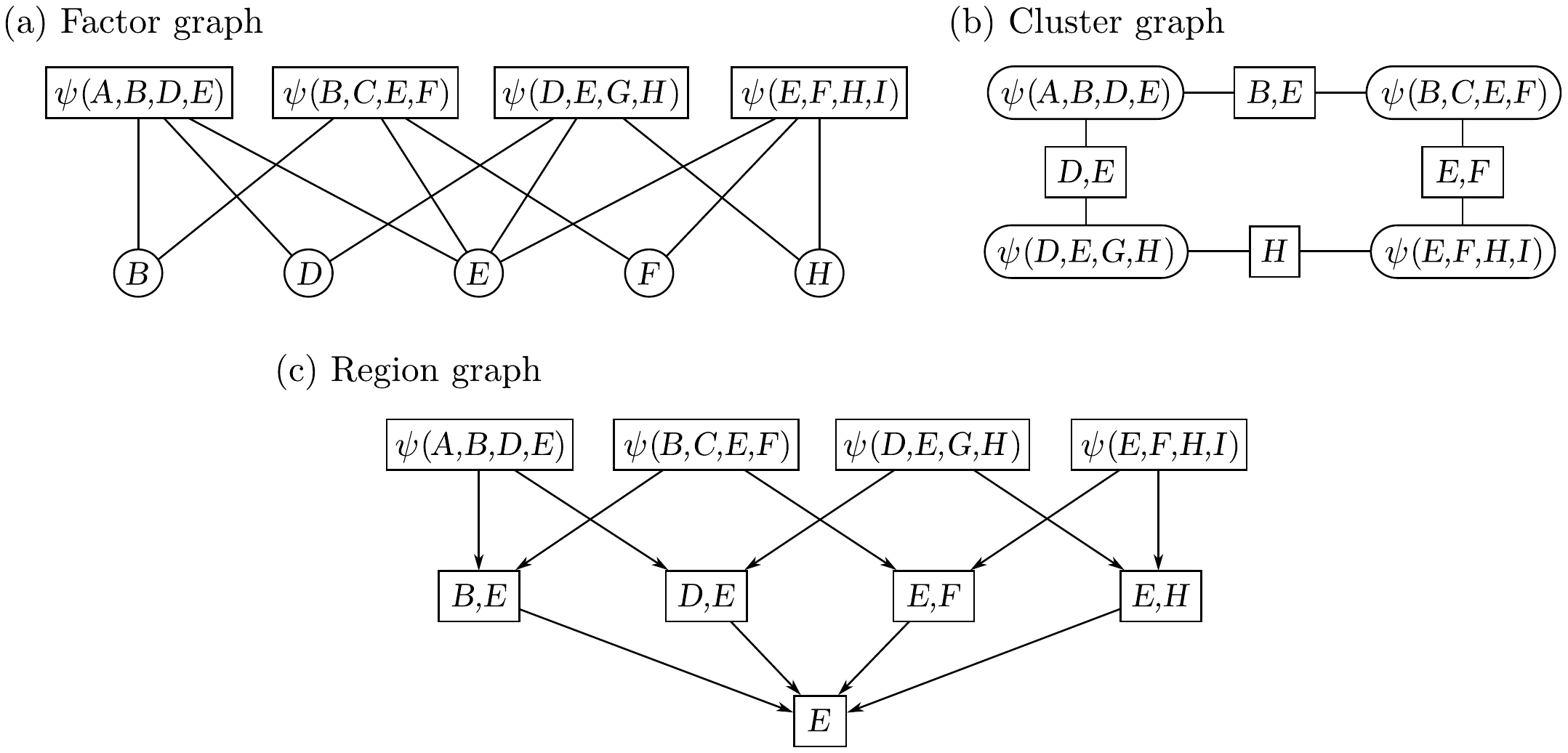}
    \caption[]{
        Example of three common types of PGM graphs configured from the same input factors $\psi(A, B, D, E)$, $\psi(B, C, E, F)$, $\psi(E, F, H, I)$, and $\psi(D, E, G, H)$. 
        The region graph in (c) can be made equivalent to the cluster graph in (b) by replacing region
        \tikz[baseline=(X.base)]\node[draw=black,rectangle,inner sep=2pt,rounded corners=0.4pt](X){$E, H$};
        with
        \tikz[baseline=(X.base)]\node[draw=black,rectangle,inner sep=2pt,rounded corners=0.4pt](X){$H$};
        and removing region
        \tikz[baseline=(X.base)]\node[draw=black,rectangle,inner sep=2pt,rounded corners=0.4pt](X){$E$};
        along with all links to it.
    }\label{fig:factor_cluster_region_graph}
\end{figure}

Many advancements have been made in the representation and formulation of probabilistic reasoning problems; this is especially true in the choice of graph structures. For example, junction trees emerged from the variable elimination algorithm~\cite{pearl1988probabilistic} and allow for belief propagation to determine exact marginal distributions. When allowing for cycles in the graph, junction trees can be extended to cluster graphs, and belief propagation can be extended to an approximate reasoning method called loopy belief propagation. Another structure is the factor graph, adapted from the physics literature~\cite{kschischang2001factor}. Loopy belief propagation on this specific structure can be translated to minimising what is referred to in statistical physics as the Bethe free energy~\cite{heskes2003stable}. \citet{yedidia2005constructing} developed this concept further and provided a generalisation of these structures and methods in the form of region graphs and generalised belief propagation (relevant to minimising what is referred to as Kikuchi free energy). To illustrate the different graph structures for further discussion, an example of each of these graph types is presented in Figure~\ref{fig:factor_cluster_region_graph}, with the order of generality as factor graph $\leq$ cluster graph $\leq$ region graph.

Graph design is one of the key elements in building an effective PGM. Factor graphs have been used in most practical applications where automatically generated graphs are required. Their deterministic structure allows for easy replication, and their relation to physics models and their simplicity allows for a more straightforward study of their behaviour. However, despite the importance of their energy-based physics relationship, they do not provide a guarantee on the accuracy obtained by loopy belief propagation~\cite[p529]{koller}. Also, due to univariate messages, pairwise correlations between variables are not as sufficiently propagated as in more general graph structures~\cite[p415]{koller}. Factor graphs are, therefore, less suited than cluster graphs and region graphs for loopy belief propagation.

A cluster graph is equivalent to a region graph of only two layers. A valid cluster graph (and factor graph) construction must adhere to the running intersection property~\cite[p347]{koller}. While factor graphs trivially achieve this property, the extra freedom afforded to cluster graphs may result in multiple valid configurations for the same input factors. For constructing a cluster graph, \citet{yedidia2005constructing} suggest fully creating all variable links between clusters and then, through some search heuristic, removing variables until the running intersection property is satisfied. Koller~\cite{koller} elaborated on their experimental results along with the findings of \citet{yedidia2005constructing} and \citet{welling2004on} and concluded that
\begin{itemize}
    \item different graphs from the same input factors can lead to wildly different solutions~\cite[p404]{koller},
    \item the choice of cluster graph is a trade-off between computational cost and accuracy~\cite[p404]{koller}, and
    \item that it is not obvious how to formulate a general-purpose cluster graph construction procedure~\cite[p429]{koller}.
\end{itemize}

Region graphs have a far greater degree of freedom in their design than cluster graphs and factor graphs. Consequently, they adhere to a generalisation of the running intersection property based on nested containment and the calibration of so-called counting numbers. A counting number is associated with each region in order to weigh the contribution of factors and variables to the system. \citet{yedidia2005constructing} provide insight into efficient graph structures along with some methods to construct them. However, they clarify that ``the problem of generating region graphs with highly accurate marginals is still an open research problem''\cite{yedidia2005constructing}. \citet{welling2004on} proposed the ``regional pursuit'' algorithm as a sequential approach to designing region graphs from three operations they call ``split’’, ``merge’’,  and ``death’’. 

It is unclear to what extent region graphs are superior to cluster graphs since an efficient algorithm for constructing cluster graphs is still mostly unexplored. Cluster graphs do have some advantages over deeply nested region graphs because of their structural simplicity. For example, they can accommodate the simplified formulation of belief propagation as defined on junction trees. Furthermore, a cluster graph will have fewer or equal regions than a region graph constructed from the same factors. This will result in fewer marginals to compute. Cluster graph construction already falls within a large space of optimisation. Therefore, it might be useful to explore cluster graph construction thoroughly before relying on region graph inference.

\subsection{Constraint satisfaction}\label{sec:chap1-intro-csp}
A constraint satisfaction problem (CSP) involves a set of variables, a domain over each variable, and a set of constraints over these variables. A CSP solver is tasked with finding a solution (i.e.\ an assignment) over these variables. \citet{dechter2010on} investigated the relationship between PGMs and constraint satisfaction. They proved that loopy belief propagation on cluster graphs with determinism in the network (i.e.\ zero-potentials) is reduced to an algorithm for generalised arc consistency. Any determinism propagated through the system cannot be reverted, and all determinism propagates within a finite number of iterations. This provides at least one formal justification for using belief propagation on loopy structures. Their investigation also shows that PGMs can be ``flattened'', such that sparse structures represent the factors in the system, thereby hiding zero-potentials and allowing determinism to reduce the potential space. Furthermore, replacing the marginalisation operations in belief propagation with max-marginalisation results in the same zero propagation but with less overall computation.

Although the constraint satisfaction properties of PGMs have been studied, the PGM literature does not provide a general-purpose CSP solver. One particular system for testing CSPs is Sudoku puzzles. 

\citet{MoonT} investigated belief propagation on Sudoku puzzles using a factor graph to represent the constraint network. The system could, reportedly, not solve all Sudoku puzzles it encountered. 

\citet{GoldbergerJ} further explored the difference between belief propagation using max-marginalisation or sum-marginalisation. They report that a single round of belief propagation does not reliably solve Sudoku puzzles. For failed attempts, sum-marginalisation offers approximate marginals over the Sudoku factors that can be used to find the maximum likelihood.
In contrast, the beliefs obtained through max-mar\-gin\-al\-isation will have an equal weighting for all non-zero potentials. 

\citet{KhanS} combine belief propagation with Sinkhorn balancing. As with \citet{GoldbergerJ}, they aimed for a maximum likelihood estimation and not an exact solver. They reported an improvement but did not reliably solve Sudokus of moderate difficulty.  

The PGM-based constraint satisfaction approaches listed here are all limited in one way or another. They are either inefficient in purging all the redundant potentials from the system or rely on heuristics to pick a maximum likelihood solution.

Other advances have been made to solve PGM-based constraint networks with tools from other domains. These approaches would typically convert a PGM into another domain and then solve it with domain-specific tools. Some examples include converting PGMs to Boolean satisfiability problems (such as conjugate normal form~\cite{cnfbool}), sentential decision diagrams~\cite{sddchoi}, and arithmetic circuits~\cite{acewebsite}. For example, the arithmetic-circuit compilation and evaluation (ACE) software~\cite{acewebsite} can compile a factor graph (or a Bayes network) into an arithmetic circuit, which can be queried to infer a solution. These approaches are helpful, but they rely on tools outside the scope of traditional PGM structures and belief propagation algorithms.

\section{Objectives}
This dissertation aims to explore some of the unresolved themes in the established PGM literature, specifically with regard to constraint satisfaction. We aim to develop inference techniques to express and solve constraint satisfaction problems more reliably. Our exact objectives are listed as
\begin{itemize}
    \item  exploring efficient cluster graph construction and comparing cluster graphs to the prevalent factor graph structure,
    \item investigating constraint satisfaction using PGMs and developing methods to solve high-order CSPs,
    \item applying our findings to a practical problem to verify our approach and the PGM approach in general.
\end{itemize}

\section{Contributions}
This dissertation contributed to the following publications
\begin{description}
    \item[\hspace{1.3em}\cite{streicher}] S. Streicher and J. du Preez, “Graph Coloring: Comparing Cluster Graphs to Factor Graphs,” in \textit{Proceedings of the ACM Multimedia 2017 Workshop on South African Academic Participation}. SAWACMMM ’17. New York, NY, USA: ACM, 2017, pp. 35–42,
    \item[\hspace{1.3em}\cite{landcoverdata}] L. H. Hughes, S. Streicher, E. Chuprikova, and J. du Preez. ``A Cluster Graph Approach to Land Cover Classification Boosting.'' \textit{Data}, 4(1), 2019. ISSN 2306-5729, and
    \item[\hspace{1.3em}\cite{streicher2021strengthening}]  S. Streicher and J. du Preez, “Strengthening Probabilistic Graphical Models: The Purge-and-merge Algorithm,” \textit{IEEE Access}, vol. 9, pp. 149 423–149 432, 2021.
\end{description}

\noindent The direct contributions from author S.\ Streicher are
\begin{itemize}
    \item a comparative study between cluster graphs and factor graphs, in which cluster graphs show great promise in comparison to factor graphs in~\cite{streicher},
    \item comprehensive integration of various aspects required to formulate graph colouring problems into PGMs in~\cite{streicher}, further expanded to general constraint satisfaction in~\cite{streicher2021strengthening},
    \item a practical application of these principles in a land cover classification problem in the field of cartography in~\cite{landcoverdata},
    \item the purge-and-merge algorithm, which makes it possible to simplify inference on complex CSPs systematically, developed as a combination of constraint satisfaction, LTRIP, belief propagation, and factor multiplication in~\cite{streicher2021strengthening}, and
    \item an illustration of these tools in solving some very challenging puzzles, namely Sudoku, Fill-a-pix, Kakuro, and Calcudoku puzzles in~\cite{streicher2021strengthening}.
\end{itemize}

Furthermore, a significant contribution published via this work is the general-purpose cluster graph construction algorithm, LTRIP in \cite{streicher}; however, S.\ Streicher cannot be credited for discovering the basic algorithm. Nevertheless, the author can be credited for abstracting the ``Connection-Weights'' subroutine to allow for alternative minimisation functions.

\chapter{Probabilistic graphical model basics} \label{pgm-chapter}

\section{Introduction}
For the sake of completeness, this chapter describes basic underlying PGM techniques. These techniques will serve as building blocks for the later chapters of this dissertation, which covers incremental inference on higher-order PGMs. We, therefore, approach this chapter as a collection of relevant theory necessary to replicate the PGMs used in this work. First, we investigate the exponential blow-up found in the complexity of multivariate probabilities. Secondly, we introduce the concept of probability distributions and show how conditional independencies can be used to factorise large probabilistic spaces. Thirdly, we provide a computational model for representing factors and for computing factor operations. Finally, we introduce a technique for formulating a problem probabilistically using a cluster graph and solving the problem using belief propagation. Our example on the Hamming (7,4) code indicates that the system is an effective probabilistic inference tool that yields practical results.

\section{Why consider probabilistic graphical models}
Graphical models emerged from the successes of the 1960s for the use of graph inference in Kalman filtering~\cite{huang1994automatic}, error correction codes~\cite{mceliece1998turbo}, and hidden Markov models~\cite{ronen1995parameter}. 
Graphical models are a resourceful combination of graph theory and probabilistic inference techniques; they provide powerful tools for optimisation and probabilistic computation. They are used to encode dependencies among interacting variables to model a system’s underlying probabilistic structure.

At the heart of PGMs lies the problem of exponential blow-up, where a problem's slight growth can quickly lead to a computationally intractable large space. This can occur when finding a joint probability over even a modest number of variables.

\begin{figure}[h]
    \centering
    \def\svgscale{0.9}
    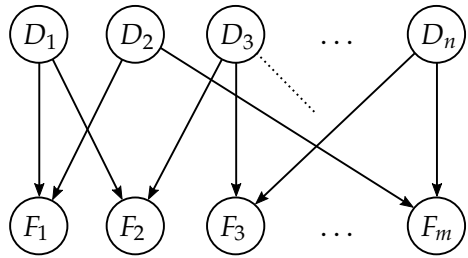
    \caption{Bayes network showing example dependencies between diseases  $D_j$ and findings $F_i$ in a medical network. Edges are only present where there is a direct link between a disease and a finding.}
    \label{fig:medicalnet}
\end{figure}

A good example of this phenomenon is described in \citet{medical-network} and in~\citet[p773]{theodoridis2015machine}. A system for disease hypothesis inference on patients is explored by using a PGM for encoding diseases and symptoms. The model operates on
\begin{itemize}
    \item $n$ disease hypotheses $D_1, \ldots , D_n$,
    \item $m$ symptom findings $F_1, \ldots , F_m$,
    \item a disease being either present or absent in a patient, and 
    \item each finding being either observed or unobserved in a patient -- an approximation for a symptom being present or not.
\end{itemize}
To understand the connections between these variables, see Figure~\ref{fig:medicalnet}, where the diseases and findings are represented as a Bayes network and an edge between disease $D_i$ and finding $F_j$ represents a corresponding link in the disease profile database. A simplification arising from this formulation is that the disease hypotheses are independent, and symptoms are conditionally independent given a disease hypothesis.

The goal of this model is to predict the presence of a number of diseases, given the presence of a set of findings. This can be achieved by calculating the joint probability distribution over all the variables $P(F_1, \ldots, F_m, D_1, \ldots, D_n)$ and then, for each patient, condition on the patient's findings to obtain the disease profile of that patient. The joint distribution can be expressed by (established later in Section~\ref{sec:factorising-pgms})
\begin{equation}
    P(F_1, \ldots, F_m, D_1, \ldots, D_n) = \prod_{i=1}^{n} P(F_i | \mathbf{D}_i) \prod_{j=1}^{m} P({D}_j), \nonumber
\end{equation}
where $\mathbf{D}_i$ is the set of all associated diseases related to finding ${F}_i$. We like to point out here that the resulting joint distribution has a space of size $2^{n+m}$ and can easily grow to a computationally intractable scale.

When focusing on a single disease hypothesis, however, some reduction in the complexity can be obtained. We can then rephrase the problem as follows, given a set of findings $\mathbf{F}'=\mathbf{f}'$, what is the probability of a specific disease $D_j=d_j$? By defining $\mathbf{D}'$ as the set of all diseases related to $\mathbf{F}'$, we can use Equations~\ref{eq:list-of-prob-rules} (established later in Section~\ref{sec:factorising-pgms}) to formulate the problem as
\begin{align}
    P(D_j | \mathbf{F}') 
    &= \frac{P(\mathbf{F}'| D_j)P(D_j)
    }{P(\mathbf{F}')} \nonumber
    \\
    & = \frac{\sum_{D=d \forall D \in \mathbf{D}' \setminus \{D_j\}} \prod_{F_i \in \mathbf{F}'} P(F_i| \mathbf{D}_i) \prod_{D_k \in \mathbf{D}' } P(D_k)}{\sum_{D = d \forall D \in \mathbf{D}'} \prod_{F_i \in \mathbf{F}'} P(F_i| \mathbf{D}_i) \prod_{D_k \in \mathbf{D}' } P(D_k)}. \nonumber
\end{align}
If no additional factorisation is applied, the summation in the denominator will involve $2^{|\mathbf{D}'|}$ terms. This calculation is still computationally intractable for even a modest set of diseases $|\mathbf{D}'|=500$.  It is, therefore, clear that additional tools are required to reason about these problems computationally. The following sections will introduce efficient PGM structures and show how problems like these can be formulated and solved effectively using PGMs.

\section{Probability distributions} \label{sec:PRO-probability-distributions}
A random variable is a variable with an uncertain value. This value can be from a continuous domain (such as measuring the temperature in $^{\circ}$C) or a discrete domain (such as rolling a die to determine a one-in-six outcome). In this work, we are primarily interested in discrete random variables and will mainly express probabilities in terms of a discrete probability mass function. 
The distribution over $X$, i.e.\ $P(X)$, is represented by a mass function $p_{X}(X\equalss x)$, which we will mainly write in shorthand as either $p(X=x)$, or $p(x)$, whenever the meaning is clear from the surrounding context. Furthermore, we will liberally switch between expressing ideas in terms of probability distributions or mass functions.

A probability distribution represents the uncertainty inherent in one or more random variables, $P(X_1$, $\ldots$, $X_n)$, and can be expressed by a joint mass function $p(x_1, \ldots , x_n)$. The mass function is a mapping between the states over a group of variables and the probability of each state. 

We can classify different types of distributions according to their context. The following structures are most important to our work: joint distributions, marginal distributions, conditional distributions, and potential functions~\cite{streichermasters}.

\begin{itemize}
    \item
	A joint distribution describes the combined probability of two or more random variables. For example the jointly distributed variables $X_1, X_2, \ldots, X_n$ have a joint distribution $P(X_1, X_2, \ldots, X_n)$ with mass function $p(x_1,   \ldots, x_n)$.
    
    \item
    A marginal distribution is a distribution over a subset of the variables of a joint distribution, calculated by summing over unwanted variables, e.g. the marginal probability distribution $P(X_3, \ldots, X_n)$ with regard to $P(X_1, X_2, X_3, \ldots, X_n)$ is calculated as
    \begin{equation} 
        P(X_3, \ldots, X_n) =  \sum_{X_1 =x_1, X_2 = x_2}   P(X_1, X_2, X_3, \ldots, X_n). \nonumber
    \end{equation}
    
    \item
    A conditional distribution is a distribution with some of the variables observed (i.e.\ the values of those variables are certain), and the other random variables remain unobserved. The conditional distribution 
    \begin{equation}
        P(X_3, \ldots , X_n | X_1\equalss x_1, X_2\equalss x_2) \nonumber
    \end{equation}
    is equivalent to setting $P(X_1, \ldots, X_n)|_{X_1\equalss x_1 \text{ and }  X_2\equalss x_2}$ and normalising the result to have the sum of all its values equal to 1.

    \item
    A potential function is a function similar in structure to a probability mass function, often used as the intermediate result in a chain of computations. A potential function $\phi(X_1\equalss x_1, \ldots, X_n \equalss x_n)$, shorthand $\phi(x_1, \ldots, x_n)$, is a function over random variables $X_1$, $\ldots$, $X_n$ with all its values greater than zero, i.e.\ $\phi(x_1, \ldots, x_n) > 0$. Therefore, a potential function $\phi(x_1, \ldots, x_n)$ is a more general case of a mass function $p(x_1, \ldots, x_n)$ without requiring that the sum of its values equals 1.
\end{itemize}

\section{Probability theory}\label{sec:factorising-pgms}
We will start this discussion by listing the common theorems applicable to PGM literature (using $\independent$  as the symbol for independence):
\begin{align}
    &\text{\redbullet\ \ Product rule:} & &\mkern -17mu 
    P(X_1, X_2) = P(X_1|X_2)P(X_2) \nonumber \\
    &\text{\redbullet\ \ Chain rule:} & &\mkern -17mu 
    P(X_1,\ldots,X_n) = \prod_{i=1}^{n} P(X_i|X_{i+1}\equalss x_{i+1},\ldots,X_n\equalss x_n) \nonumber \\
    &\text{\redbullet\ \ Bayes rule:} & &\mkern -17mu 
    P(X_1| X_2)  = \frac{P(X_2|X_1) P(X_1)}{P(X_2)} \nonumber \\
    &\text{\redbullet\ \ Independence:} & &\mkern -17mu 
    P(X_1, X_2) = P(X_1)P(X_2) \text{ iff } X_1 \independent  X_2 \nonumber \\
    &\text{\redbullet\ \ Conditional independence:} & &\mkern -17mu 
    P(X_1,X_2|X_3) = P(X_1|X_3)P(X_2|X_3) \text{ iff } X_1 \independent X_2 | X_3 \nonumber \\
    &\text{\redbullet\ \ Marginalisation:} & &\mkern -17mu 
    \sum_{X_2=x_2} P(X_1,X_2) = P(X_1) \label{eq:list-of-prob-rules}
\end{align}

By establishing these equations, we can now easily discuss the basic principles on which PGMs are built. We will show how some distributions can be factorised into a lower-dimensional representation and how some computational sequences can result in fewer calculations than others.

Given a joint distribution with $n$ random variables $P(X_1, \ldots, X_n)$, we can use the chain rule from Equations~\ref{eq:list-of-prob-rules} and factorise the distribution as
\begin{equation} \label{eq:prob-factorisation}
    P(X_1, \ldots, X_n) = P(X_1|X_{2},\ldots,X_n )P(X_{2}|X_{3},\ldots,X_n)\ldots P(X_n).
\end{equation}

Now consider two extreme cases about the underlying dependencies between the random variables: scenario 1, where no independencies exist between the variables involved and scenario 2, where all the variables are mutually independent -- such as a series of coin flips.

For scenario 1, to marginalise with respect to one variable $X_1$ we sum over the other variables $X_2 \ldots X_n$ as
\begin{equation} \label{eq:extreme1}
    p(x_1) = \sum_{x_2} \sum_{x_3} \ldots \sum_{x_n} p(x_1 , x_2, x_3 \ldots , x_n).
\end{equation}  
This is the general case where the structure cannot be exploited to reduce the computational cost. The number of summations is $\mathcal{O}(\text{card}(X_2)\cdot\text{card}(X_3)\ldots\text{card}(X_n))$, or in a simplified case $\mathcal{O}(k^{n-1})$, when the cardinality of all random variables is equal to $k$. Such a system can easily lead to an intractable computation, for even a modest variable count of $n=100$ and cardinality of $k=2$.

For scenario 2, we can employ the product rule to write the joint probability as
\begin{equation} \label{eq:extreme2}
    P(X_1 , X_2 , \ldots, X_n) = P(X_1)P(X_2)\ldots P(X_n),
\end{equation}
and the marginalisation over $X_1$ as
\begin{equation}
    p(x_1) = P(x_1) \sum_{x_2} p(x_2) \sum_{x_{3}} p(x_3 ) \ldots \sum_{x_n} P(x_n).
\end{equation}
This reduces the complexity from $\mathcal{O}(k^{n-1})$ to $\mathcal{O}(k (n-1))$. For $n=101$ and $k=2$, such as in the case of $101$ independent coin flips, the number of summations is reduced from $2^{100}$ down to $100$. In reality, though, no summation is necessary for this example since all the $\sum_{x_i}P(x_i)$ terms are equal to $1$.

\subsection*{Marginal posteriors}

\begin{figure}[h!]%
    \centering
    
    \begin{tikzpicture}[
        thick,scale=0.84, every node/.style={transform shape},auto
        ]
        \node (x1) [draw, circle] {$X_1$};
        \node (x7) [draw, circle, right =3.3cm of x1] {$X_7$};
        \node (x4) [draw, circle, above =0.8cm of x7] {$X_4$};
        \node (xx) [] at ($(x1)!0.5!(x7)$) {};
        \node (x10) [draw, circle, below =0.47cm of xx] at ($(x1)!0.5!(x7)$) {\hspace{-0.11em}$X_{10}$\hspace{-0.11em}};
        \node (x2) [draw, circle, above =0.8cm of x10] {$X_2$};
        \node (x3) [draw, circle, above =0.8cm of x2] {$X_3$};
        \node (x5) [draw, circle, above =0.8cm of x4] {$X_5$};
        \node (x8) [draw, circle, above =.8cm of x3] {$X_8$};
        \node (x9) [draw, circle, above =.8cm of x8] {$X_9$};
        \node (x6) [draw, circle, above =.8cm of x5] {$X_6$};
        \draw[->] (x2) to (x10);
        \draw[->] (x1) to (x10);
        \draw[->] (x2) to (x1);
        \draw[->] (x3) to (x1);
        \draw[->] (x2) to (x7);
        \draw[->] (x3) to (x7);
        \draw[->] (x4) to (x7);
        \draw[->] (x5) to (x4);
        \draw[->] (x6) to (x5);
        \draw[->] (x8) to (x3);
        \draw[->] (x9) to (x8);
    \end{tikzpicture}
    
    \caption{Example Bayes network depicting dependencies between random variables $X_1, \ldots, X_{10}$. A factorisation of this graph can be found in Equation~\ref{eq:clustergraph-example-factorisation} in the main text.}
    \label{fig:SOL-example-Bayes-network}
\end{figure}
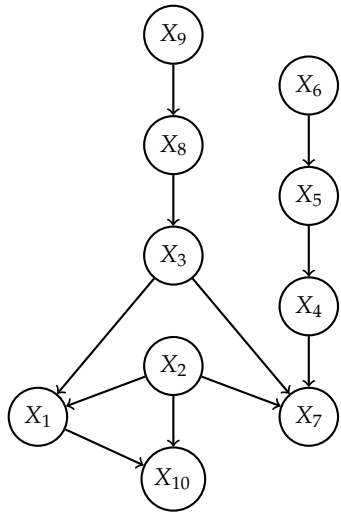

Since the two scenarios given above are extreme cases, most of the problems practically encountered will fall somewhere between these two extremes. For instance, given a distribution with the dependencies captured by the Bayes network in Figure~\ref{fig:SOL-example-Bayes-network}, we will show how the marginal distributions can be calculated more directly than first calculating the joint distribution and then the marginals from that result. 

In a Bayes network, the edges are drawn from a factor’s right-hand side conditional variables to the left-hand side variables. For example, the network {\footnotesize \circled{$Z$}$\leftarrow$\circled{$Y$}$\leftarrow$\circled{$X$}} would represent the conditional distribution relationships $P(Y|X)$ and $P(Z|Y)$ and the full factorisation $P(X,Y,Z) = P(Z|Y)P(Y|X)P(X)$.

The following is, therefore, a valid factorisation of the Bayes network in Figure~\ref{fig:SOL-example-Bayes-network}
\begin{align}
    P(X_1, \ldots, X_{10}) =\, & \nonumber \\[-3pt]
    &\mspace{-80mu} 
        P(X_1|X_2, X_3) \cdot 
        P(X_{10}|X_2, X_1) \cdot 
        P(X_7| X_2, X_3, X_4) \cdot 
        P(X_2) \cdot 
        \nonumber \\[-3pt]
    &\mspace{-80mu}
        P(X_4| X_5) \cdot
        P(X_5|X_6) \cdot 
        P(X_6) \cdot 
        P(X_3 | X_8) \cdot 
        P(X_8| X_9) \cdot 
        P(X_9). \label{eq:clustergraph-example-factorisation}
\end{align}

A calculation of the marginals $P(X_1, X_2, X_3)$ can be made more efficient than simply calculating the product of all the factors and then applying the marginalisation. 
Instead, the system's independencies can be exploited to allow for a piecewise marginalisation and multiplication sequence, such as
\begin{align}
    P(X_1, X_2, X_{3}) =\, & \nonumber \\[-3pt]
    &\mspace{-100mu} 
    P(X_1|X_2, X_3) 
    \sum_{x_{10}}
    P(X_{10}|X_2, X_1)
    \sum_{x_4, x_7}
    P(X_7| X_2, X_3, X_4)  
    P(X_2)  
    \nonumber \\[-3pt]
    &\mspace{-100mu}
    \sum_{x_5} P(X_4| X_5) 
    \sum_{x_6} P(X_5|X_6) P(X_6)
    \sum_{x_8} P(X_3 | X_8)
    \sum_{x_9} P(X_8| X_9)
    P(X_9). \label{eq:marginalising-clustergraph-example}
\end{align}

Furthermore, the extraction of not only $P(X_1,X_2,X_3)$ but of all the marginal distribution can be achieved by using a single calculation flow. For this, we require an algorithm such as variable elimination or belief propagation on a junction tree~\cite{cozman2000generalizing}. A junction tree is a graph representation that acts as a map for the order of a marginalisation sequence. A junction tree avoids cycles by clustering distributions together that cannot be marginalised independently. For ill-structured problems, this can lead back to the original issue and result in a single joint distribution without an optimised marginalisation sequence.

In Section~\ref{sec:belief-propegation}, we provide a formulation for belief propagation on tree-structured graphs. Section~\ref{sec:loopy-belief-propagation} introduces a heuristic for performing belief propagation on some ill-structured problems by allowing the PGM graph structure to contain cycles.

\section{Factors}
A factor $\psi(X_1, \ldots, X_n)$ over random variables $X_1, \ldots, X_n$ describes the knowledge we have of those variables within a system -- usually captured by a mass or potential function. In our work, the term factor is generally used in the context of factorising a system over groups of random variables. A mass function or potential function is generally used to represent (and work with) the underlying information. Therefore, in this work, the term factor is more linked to context than having a strictly distinct definition from the potential function it represents.

To use factors in a computational model, we need to represent their underlying data in terms of a data structure that can be stored and manipulated programmatically. For example, since a potential function is a mapping between states and potentials, this mapping can be stored in any data structure that can capture this relationship, such as an $n$-dimensional array~\cite{harris2020array}, a NamedArray~\cite{NamedArrays}, or a map or dictionary type~\cite{cpp,van1995python}.

\begin{figure}[H]
    \centering
    \def\svgscale{0.9}
    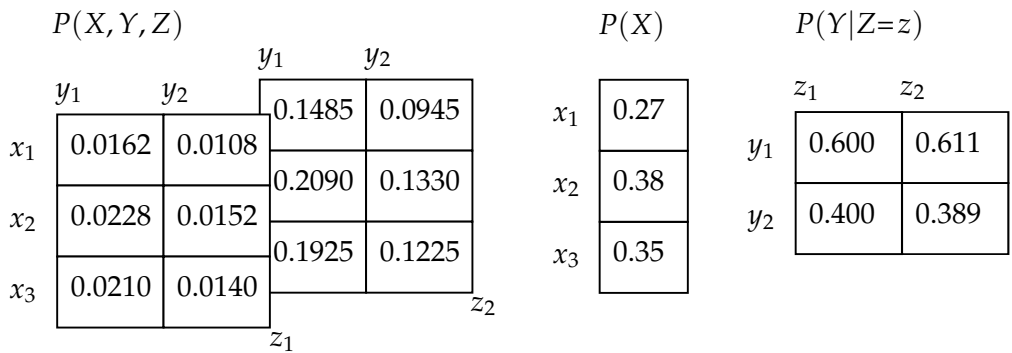
    \vspace{-0.2em}
    \caption{Example joint distribution $P({X}, {Y}, {Z})$, univariate distribution $P({X})$, and conditional distribution $P(Y|Z)$. They are illustrated as 3D, 1D, and 2D tables respectively to highlight their dimensionality. Adapted from~\cite{streichermasters}.}\label{fig:PROB-tables-intro}
\end{figure}

\begin{figure}[H]
    \centering
    \vspace{-0.5em}
    \def\svgscale{0.9}
    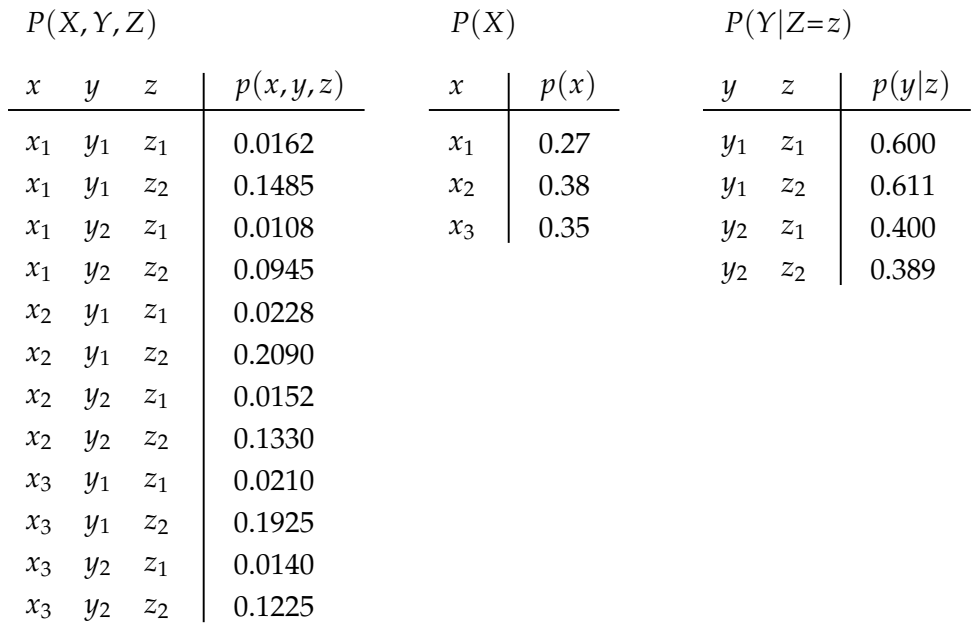
    \caption[]{Example joint distribution $P({X}, {Y}, {Z})$, univariate distribution $P({X})$, and conditional distribution $P({Y|Z})$, illustrated as tables.
    \vspace{-1em}}\label{fig:PROB-tables-intro-sparse}
\end{figure}

The choice of data structure can significantly affect the implementations of various factor operations, and great care needs to be taken to allow for applying labels to dimensions, re-ordering variables, aligning shared-scope between factors, and applying correct broadcasting rules. Figures~\ref{fig:PROB-tables-intro} and~\ref{fig:PROB-tables-intro-sparse} illustrate the same probability distributions but with two different data layouts.


Note that some intricacies are involved in working with calculations on potential and mass functions, such as redundant dimensions. For example, if the  distribution $P(X|Y)$ is represented as a table, it will require a two-dimensional structure. However, if $X \independent Y$, then $P(X|Y) = P(X)$, which can be represented as a one-dimensional structure. Intricacies like these can be dealt with on a case-by-case basis or be programmatically baked into the data structures or factor operations.

\section{Factor operations} \label{chap:basic-concepts}
When encountering an equation such as  $p(x_1, x_2, x_3) = \frac{1}{Z} \phi_1(x_1, x_2) \phi_2(x_2) \phi_3(x_1, x_3)$, it might not yet be clear how to perform the underlying mathematical operations. This section acts as an implementation guide for different factor operations used in the rest of our work. These operations are discussed by Koller~\cite{koller} as factor multiplication, division, marginalisation, reduction, damping, and normalisation~[\citenum{koller}; Defs.\ 4.2, 10.7, 13.12, and 4.5; Eq.\ 11.14; and Ch.\ 4].

Using the factorisation $P({X},{Y},{Z}) = P({X})  P({Y}|{Z})  P({Z})$ (with subsequent independencies ${X} \independent {Y}, {Z}$) from Figure~\ref{fig:PROB-tables-intro} as a baseline, we will show examples for each factor operation mentioned above.

\subsection{Multiplication} 
\begin{figure}[H]%
        \centering
        \vspace{-0.17em}
        \def\svgscale{0.9}
        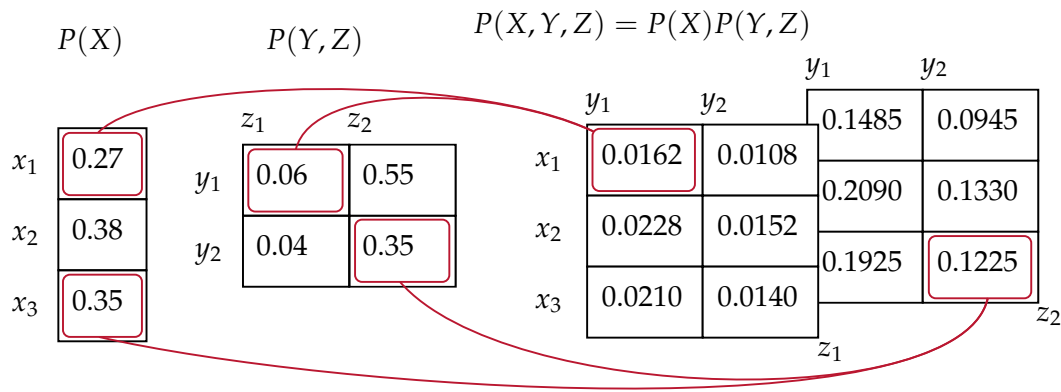
    \caption{The product of two distributions. Note that the result has a structure of three dimensions, a total state space of 12 potentials, and $X \independent {Y, Z}$. Adapted from~\cite{streichermasters}.\vspace{-1em}}
    \label{fig:PROB-tables-multiply1}
\end{figure}

\begin{figure}[H]%
        \centering
        \def\svgscale{0.9}
        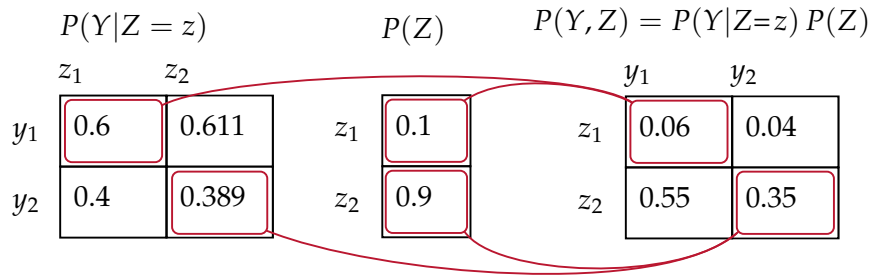
    \caption[]{The product of two distributions with common scope $Z$. Note that the conditional distribution is conditioned on all possible outcomes of $Z$. Adapted from~\cite{streichermasters}.}
    \label{fig:PROB-tables-multiply2}
\end{figure}

Our first example is the product $P({X},{Y},{Z}) = P({X})  P({Y}, {Z})$, and our second example is the product $P(Y,Z)=P(Y|Z)P(Z)$. The procedure is illustrated in  Figures~\ref{fig:PROB-tables-multiply1} and \ref{fig:PROB-tables-multiply2} as the product of each combination of state over the intersecting variables of the two distributions.

\subsection{Division}
\begin{figure}[H]%
        \centering
        \def\svgscale{0.9}
        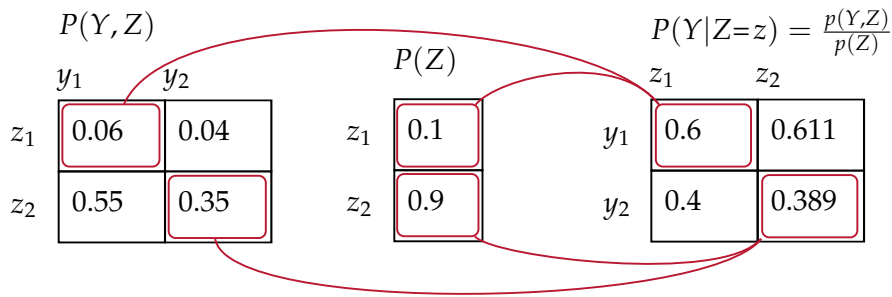
    \caption[]{The division of two distributions. This is the inverse of the multiplication in Figure~\ref{fig:PROB-tables-multiply2}. Adapted from~\cite{streichermasters}.}
    \label{fig:PROB-tables-divide}
\end{figure}

This procedure is the inverse of multiplication. It is, therefore, required that the variables in the denominator are a subset of the variables in the numerator. The calculation 
$P(Y|Z) = \frac{P({Y},{Z})}{P({Z})}$ is displayed in Figure~\ref{fig:PROB-tables-divide}. It is accomplished by dividing matching variables in a component-wise fashion. Note that $\frac{0}{0}$ is defined as $0$ in this context.

\subsection{Marginalisation} 
\begin{figure}[H]%
        \centering
        \def\svgscale{0.9}
        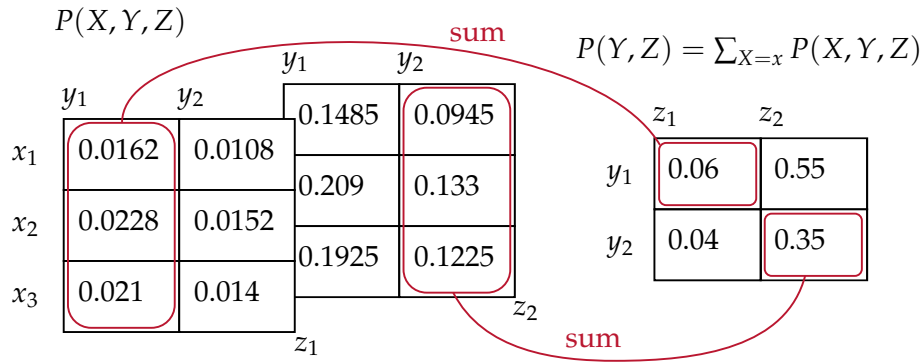
    \caption[]{The marginalisation of a distribution. The variable $X$ is removed from the table by summing over its domain. Adapted from~\cite{streichermasters}.}
    \label{fig:PROB-tables-marginalise}
\end{figure}

A marginal distribution is accomplished by summing over the joint distribution states not represented by the marginal. 
For example, the marginal distribution $P(Y,Z)$, calculated from the distribution $P({X}, {Y}, {Z})$, would require summing over $X$ as $P({Y},{Z}) = \sum_{{X=x}}{P({X}, {Y}, {Z})}$. See Figure~\ref{fig:PROB-tables-marginalise}.

\subsection{Conditioning (reduction)}
\begin{figure}[H]%
        \centering
        \def\svgscale{0.9}
        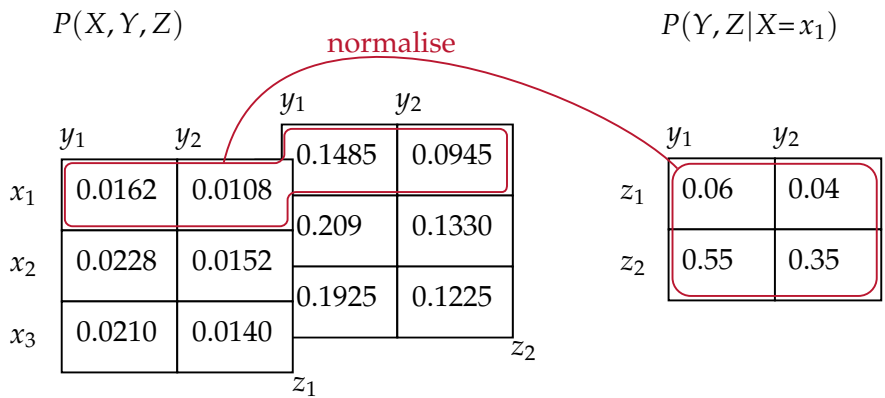
    \caption[]{The reduction of a distribution, by observing $X\equalss x_1$. Adapted from~\cite{streichermasters}.}
    \label{fig:PROB-tables-condition-value}
\end{figure}

When a random variable is reduced to a specific value, the resulting table is sliced (and normalised) according to that outcome. The distribution is thereby reduced to the dimensions of the variables with uncertain values. For example, see the process for extracting the distribution $P(X,Y |X\equalss x_1)$ in Figure~\ref{fig:PROB-tables-condition-value}.

\begin{figure}[H]%
        \centering
        \def\svgscale{0.9}
        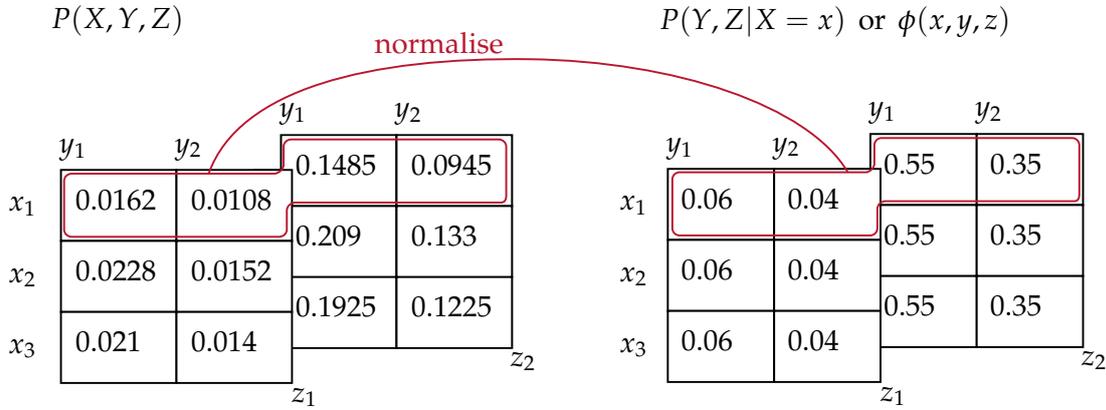
    \caption[]{A conditional table capturing the random distributions for all conditioned states $X\equalss x_1$, $X\equalss x_2$, and $X\equalss x_3$ as a single potential function. Adapted from~\cite{streichermasters}.}
    \label{fig:PROB-tables-condition}
\end{figure}

Furthermore, we can generalise the conditioning to any value, $X\equalss x$, such that all possible conditionals over $X$ are captured by a single table. This can be represented by the potential function $\phi(x, y, z)=P(Y\equalss y,Z\equalss z|X\equalss x)$ as illustrated in Figure~\ref{fig:PROB-tables-condition}.

\subsection{Damping (element-wise averaging)}
Damping is not a fundamental factor operation but a specific biasing technique used during non-exact message passing as described in Section~\ref{sec:loopy-belief-propagation}. However, for the sake of compiling these operations in one place, we present it here.

\begin{figure}[h]
    \centering
    \def\svgscale{0.9}
    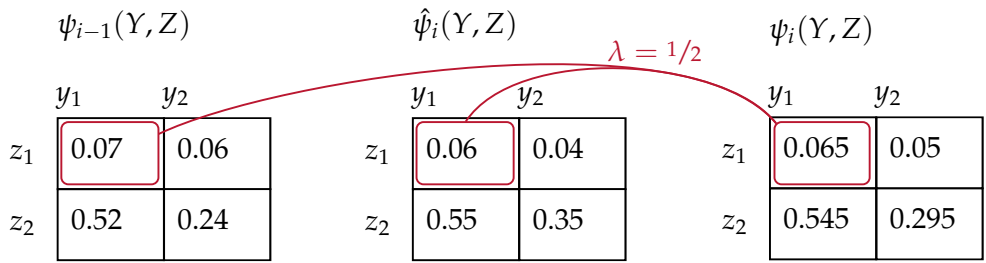
    \caption[]{Damping as weighted element-wise averaging between the values of two factors, used in the context of message updates. Note that the initial update factor is indicated as $\hat{\psi}_{i}$, and the resulting dampened factor is indicated as ${\psi}_i$.}
    \label{fig:factor-damping}
\end{figure}

Given an optimisation problem with factor $\psi_{i-1}(y,z)$ as an intermediate result at step $i{-}1$ and $\psi_{i}(y,z)$ as an update at step $i$, it is common in some systems for such updates to overshoot. As a result, the system can be steered away from a good optimisation. To combat this, we can dampen the update by replacing it with an interpolation between itself and the previous result at step $i{-}1$. This can be achieved with a component-wise weighted average $\lambda{\psi}_{i}(y,z) + (1-\lambda)\psi_{i-1}(y,z)$, where $0 \leq \lambda \leq 1$. This process is illustrated in Figure~\ref{fig:factor-damping}. Note that for damping to be correctly weighted, the potentials need to be normalised first.

\section{Probabilistic formulation} \label{sec:SOL-hamming-window-example}
We explain probabilistic formulation with the help of an example: the Hamming (7,4) code~\cite{streichermasters}. %
This idea is to encode a 4-bit message into a 7-bit sequence by introducing three extra parity bits. The goal is to include enough redundancy in the sequence to allow for message validation and error correction. Therefore, if the 7-bit sequence Hamming encoded sequence gets corrupted, the original 4-bit message can be recovered within a level of confidence.

The four message bits are $b_1, b_2, b_3, b_4$, the three parity bits are $b_5, b_6, b_7$, and their mathematical relationships are
\begin{align}
    b_5 &= b_1\oplus b_2\oplus b_3, \nonumber\\
    b_6 &= b_2 \oplus b_3 \oplus b_4, \nonumber\\
    b_7 &= b_1 \oplus b_3 \oplus b_4, \label{eq:SOL-b1plusb2plusb3}
\end{align}
with $\oplus$ as the XOR operator.

\begin{figure}[h!]%
    \centering
    
    \begin{tikzpicture}[
        thick,scale=0.84, every node/.style={transform shape},
        auto
        ]
        
        \node (r1) [draw, circle, double]                         {$R_1$};
        \node (r4) [draw, circle, double, right =5cm of r1]       {$R_4$};
        \node (r2) [draw, circle, double] at ($(r1)!0.333!(r4)$)  {$R_2$};
        \node (r3) [draw, circle, double] at ($(r1)!0.666!(r4)$)  {$R_3$};
        
        \node (b1) [draw, circle, below =0.8cm of r1]     {$B_1$};
        \node (b4) [draw, circle, below =0.8cm of r4]     {$B_4$};
        \node (b2) [draw, circle] at ($(b1)!0.333!(b4)$)  {$B_2$};
        \node (b3) [draw, circle] at ($(b1)!0.666!(b4)$)  {$B_3$};
        
        \node (b5) [draw, circle, below =2cm of b1]       {$B_5$};
        \node (b7) [draw, circle, below =2cm of b4]       {$B_7$};
        \node (b6) [draw, circle] at ($(b5)!0.500!(b7)$)  {$B_6$};

        \node (r5) [draw, circle, double, below =.8cm of b5]      {$R_5$};
        \node (r7) [draw, circle, double, below =.8cm of b7]       {$R_7$};
        \node (r6) [draw, circle, double] at ($(r5)!0.500!(r7)$)  {$R_6$};
        \draw[->] (b1) to (b5); 
        \draw[->] (b2) to (b5); 
        \draw[->] (b3) to (b5); 
        \draw[->] (b2) to (b6); 
        \draw[->] (b3) to (b6); 
        \draw[->] (b4) to (b6); 
        \draw[->] (b1) to (b7); 
        \draw[->] (b3) to (b7); 
        \draw[->] (b4) to (b7); 
        
        \draw[->] (b1) to (r1); 
        \draw[->] (b2) to (r2); 
        \draw[->] (b3) to (r3); 
        \draw[->] (b4) to (r4); 
        \draw[->] (b5) to (r5); 
        \draw[->] (b6) to (r6); 
        \draw[->] (b7) to (r7); 
    \end{tikzpicture}
    \caption[]{A Bayes network for the Hamming (7,4) probabilistic formulation, with message bits $B_1, \ldots, B_4$, parity bits $B_5, B_6, B_7$, and received bits $R_1, \ldots, R_7$. The double-lined circles indicate observed variables, e.g.\ $R_1 \equalss r_1, \ldots R_7 \equalss r_7$.}
    \label{fig:SOL-hamming-bayes}
\end{figure}
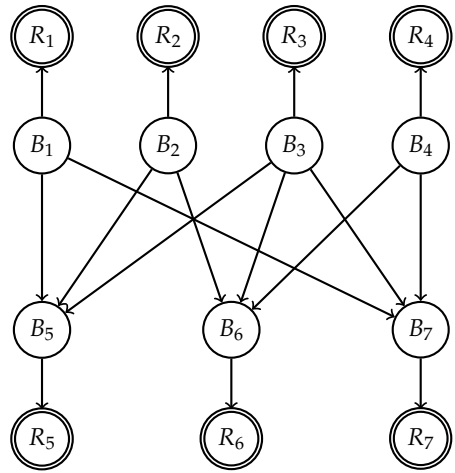

Given a message $b_1,  \ldots, b_7 $ is transmitted and received as $r_1,\ldots,r_7$, the receiver can extract the underlying transmitted sequence with a confidence related to the level of discrepancy within Equations~\ref{eq:SOL-b1plusb2plusb3}. We can represent this problem probabilistically by assigning the random variables $B_1, \ldots, B_7$ and $R_1, \ldots, R_7$ to the underlying bits of the system. Then, we can draw the relationships between the variables as a Bayes network, see Figure~\ref{fig:SOL-hamming-bayes}, and factorise the joint distribution using the chain rule:
\begin{align}
    P(B_1, \ldots, B_7, R_1, \ldots, R_7) =\, \nonumber \\[-3pt]
    &\mspace{-108mu} P(R_1|B_1)\cdot 
    P(R_2|B_2)\cdot 
    P(R_3|B_3)\cdot  
    P(R_4|B_4)\cdot 
    P(R_5|B_5)\cdot  
    P(R_6|B_6)\cdot  
    P(R_7|B_7)\cdot \nonumber \\[-3pt]
    &\mspace{-108mu} P(B_5| B_1, B_2, B_3)\cdot 
    P(B_6| B_2, B_3, B_4)\cdot 
    P(B_7| B_1, B_3, B_4)\cdot \nonumber \\[-3pt]
    &\mspace{-108mu} P(B_1)\cdot 
    P(B_2)\cdot 
    P(B_3)\cdot 
    P(B_4). \label{eq:SOL-chainrule}
\end{align}

To formulate this as a reasoning problem: we would like to estimate the values of the message bits $B_1,\ldots,B_7$, given the observations $r_1,\ldots,r_7$, i.e.\ determining the values $b_1,\ldots,b_7$ that maximises the distribution
\begin{equation}
    P(B_1,\ldots,B_7 | R_1\equalss r_1,\ldots,R_7\equalss r_7).
\end{equation}
This can naïvely be accomplished by multiplying all the factors together, applying the reductions $R_i\equalss r_i$, and marginalising to the univariate distributions $P(B_i)$. Although this is viable for a problem with a small scope (such as this), it can easily become computationally intractable for problems with even a modest number of random variables. For instance, a $9\times9$ Sudoku grid has only $81$ variables, each with a domain of $9$ possibilities, resulting in a joint distribution of $9^{81}$ potentials.

\subsection*{Factorisation}\label{sec:intro-cluster-graph}
To formulate the problem into a PGM structure, we first need to find a suitable factorisation for the problem. Ideally, a system is simply factorised by applying the chain rule and using the resulting distributions as PGM factors. However, in practice, most factorisations are derived by using the available inputs, data, and logic. Thus, the problem is usually then worked backwards to a suitable factorisation.

For the Hamming (7,4) example, our choice of factors is the three mathematical relationships in Equation~\ref{eq:SOL-b1plusb2plusb3} and seven correlations between receive and send bits $P(R_i | B_i)$:
\begin{gather}
    \psi_1(B_5, B_1, B_2, B_3),\ \psi_2(B_6, B_2, B_3, B_4),\ \psi_3(B_7, B_1, B_3, B_4), \nonumber \\
    \psi_4(R_1, B_1),\ \psi_5(R_2, B_2), \ldots,\ \psi_{10}(R_7, B_7).    \label{eq:SOL-list-of-factors}
\end{gather}
Note that the univariate factors $P(B_1),\ldots ,P(B_4)$ are omitted since they carry no prior information in this case.

These factors can now be constructed into a suitable graph to assist with belief propagation. In Chapter~\ref{gc-chapter}, we investigate two graph configurations and find cluster graphs ideal for PGM tasks. For further discussion, see Section~\ref{gc-sec:topology}, where the benefits of cluster graphs are compared to factor graphs, and the graph construction algorithm LTRIP is provided (Section~\ref{gc-sec:ltrip}). As a result, a cluster graph is configured by using the factors as nodes and linking them up with sepsets, a connecting set of variables to facilitate message passing between nodes -- see Figure~\ref{fig:SOL-clustergraph}.

\begin{figure}[h!]
    \centering
    \begin{tikzpicture}[
        thick,scale=0.85, every node/.style={transform shape},
        auto
        ]
        
        \node (c2)    [draw, rounded rectangle]                           {$B_6, B_2, B_3, B_4$};
        \node (c3)    [draw, rounded rectangle, right=2.5cm    of c2  ]     {$B_7, B_1, B_3, B_4$};
        \node (mid23) [] at ($(c2)!0.500!(c3)$)                           {};
        \node (c1)    [draw, rounded rectangle, below=0.866*2.5cm    of mid23]    {$B_5, B_1, B_2, B_3$};
        \node (r6)    [draw, rounded rectangle] at ($(c2)+(150:3)$)       {$R_6, B_6$};
        \node (r4)    [draw, rounded rectangle] at ($(c3)+( 30:3)$)       {$R_4, B_4$};
        \node (r7)    [draw, rounded rectangle] at ($(c3)+(-30:3)$)       {$R_7, B_7$};

        \node (r1)    [draw, rounded rectangle] at ($(c1)+(-150:3)$)      {$R_1, B_1$};
        \node (r2)    [draw, rounded rectangle] at ($(c1)+(-110:2.5)$)    {$R_2, B_2$};
        \node (r3)    [draw, rounded rectangle] at ($(c1)+(-70 :2.5)$)    {$R_3, B_3$};
        \node (r5)    [draw, rounded rectangle] at ($(c1)+(-30 :3)$)      {$R_5, B_5$};
        
        \node (s13) [draw] at ($(c1)!0.500!(r3)$) {$B_3$};
        \node (s26) [draw] at ($(c2)!0.550!(r6)$) {$B_6$};
        \node (s37) [draw] at ($(c3)!0.560!(r7)$) {$B_7$};      
        \node (s34) [draw] at ($(c3)!0.560!(r4)$) {$B_4$};
        \node (s11) [draw] at ($(c1)!0.550!(r1)$) {$B_1$};      
        \node (s12) [draw] at ($(c1)!0.500!(r2)$) {$B_2$};      
        \node (s15) [draw] at ($(c1)!0.550!(r5)$) {$B_5$}; 
        \node (sc12)[draw] at ($(c1)!0.500!(c2)$) {$B_2, B_3$};
        \node (sc23)[draw] at ($(c2)!0.500!(c3)$) {$B_3, B_4$}; 
        \node (sc31)[draw] at ($(c3)!0.500!(c1)$) {$B_1$};      
        
        \draw[-] (c1) -- (s13) ; \draw[-] (r3) --  (s13) ;
        \draw[-] (c2) -- (s26) ; \draw[-] (r6) --  (s26) ;
        \draw[-] (c3) -- (s37) ; \draw[-] (r7) --  (s37) ;
        \draw[-] (c3) -- (s34) ; \draw[-] (r4) --  (s34) ;
        \draw[-] (c1) -- (s11) ; \draw[-] (r1) --  (s11) ;
        \draw[-] (c1) -- (s12) ; \draw[-] (r2) --  (s12) ;
        \draw[-] (c1) -- (s15) ; \draw[-] (r5) --  (s15) ;
        
        \draw[-] (c1) -- (sc12);  \draw[-] (c2)  -- (sc12);
        \draw[-] (c2) -- (sc23);  \draw[-] (c3)  -- (sc23);
        \draw[-] (c3) -- (sc31);  \draw[-] (c1)  -- (sc31);
        
    \end{tikzpicture}
    
    \caption[]{A cluster graph formulation for the recovery of a Hamming (7,4)  encoded message. This graph is produced from the factors in Equation~\ref{eq:SOL-list-of-factors} using the LTRIP algorithm from Section~\ref{gc-sec:topology}. Note the sepset
    \tikz[baseline=(X.base)]\node[draw=black,rectangle,inner sep=2pt,rounded corners=0.4pt](X){$B_1$};
    between
    \tikz[baseline=(X.base)]\node[draw=black,rounded rectangle,inner sep=2pt](X){$B_5, B_1, B_2, B_3$};
    and
    \tikz[baseline=(X.base)]\node[draw=black,rounded rectangle,inner sep=2pt](X){$B_7, B_1, B_3, B_4$};
    is not a full intersection since LTRIP enforces the running intersection property~\cite[p347]{koller}.
    }\label{fig:SOL-clustergraph}
\end{figure}

\section{Belief propagation}\label{sec:belief-propegation}
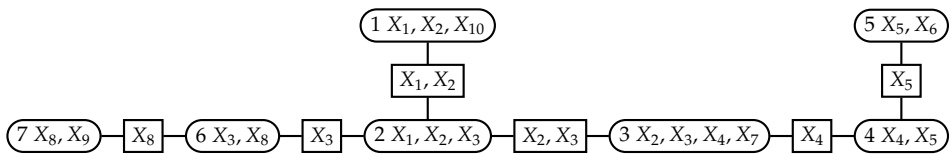
\begin{figure}[H]%
    \centering
    \begin{tikzpicture}[
        thick,scale=0.72, every node/.style={transform shape},
        auto
        ]
        
        \node (c1)    [draw, rounded rectangle                        ]  {$1\ X_1, X_2, X_{10}$};
        \node (c1c2)  [draw                   , below=0.4cm    of c1  ]  {$X_1, X_2$};
        \node (c2)    [draw, rounded rectangle, below=0.4cm    of c1c2]  {$2\ X_1, X_2, X_3$};
        \node (c2c3)  [draw                   , right=0.4cm    of c2  ]  {$X_2, X_3$};
        \node (c3)    [draw, rounded rectangle, right=0.4cm    of c2c3]  {$3\ X_2, X_3, X_4, X_7$};
        \node (c3c4)  [draw                   , right=0.4cm    of c3  ]  {$X_4$};
        \node (c4)    [draw, rounded rectangle, right=0.4cm    of c3c4]  {$4\ X_4, X_5$};
        \node (c4c5)  [draw                   , above=0.4cm    of c4  ]  {$X_5$};
        \node (c5)    [draw, rounded rectangle, above=0.4cm    of c4c5]  {$5\ X_5, X_6$};               
        \node (c2c6)  [draw                   , left =0.4cm    of c2  ]  {$X_3$};
        \node (c6)    [draw, rounded rectangle, left =0.4cm    of c2c6]  {$6\ X_3, X_8$};               
        \node (c6c7)  [draw                   , left =0.4cm    of c6  ]  {$X_8$};
        \node (c7)    [draw, rounded rectangle, left =0.4cm    of c6c7]  {$7\ X_8, X_9$};
        
        \draw[-] (c1) -- (c1c2) ; \draw[-] (c2) --  (c1c2) ;
        \draw[-] (c2) -- (c2c3) ; \draw[-] (c3) --  (c2c3) ;
        \draw[-] (c3) -- (c3c4) ; \draw[-] (c4) --  (c3c4) ;
        \draw[-] (c4) -- (c4c5) ; \draw[-] (c5) --  (c4c5) ;
        \draw[-] (c2) -- (c2c6) ; \draw[-] (c6) --  (c2c6) ;
        \draw[-] (c6) -- (c6c7) ; \draw[-] (c7) --  (c6c7) ;
    \end{tikzpicture}
	\vspace{0.4em}
    \caption[]{A tree-structured cluster graph %
    produced from the 
    factorisation of the Bayes network in Figure~\ref{fig:SOL-example-Bayes-network}. See the main text for an example of applying belief propagation to this structure.
    }
    \label{fig:SOL-cluster-tree}
\end{figure}

\citet{pearl1988probabilistic} first discovered belief propagation as an algorithm for exact inference on tree-structured networks. We will, therefore, first demonstrate how to formulate this scheme on a tree structure and then return to the Hamming (7,4) example to show how to extend this scheme into an approximate reasoning algorithm for loopy structured PGMs. For now, we reuse the factorisation in Equation~\ref{eq:clustergraph-example-factorisation} to produce the structure in Figure~\ref{fig:SOL-cluster-tree} in order to assist our example.

Belief propagation is built on two concepts: $\beta_i(\mathbf{C}_i)$ is the cluster belief for cluster $i$, and $\delta_{i \rightarrow j}(\mathbf{S}_{i,j})$ is the sepset belief for the sepset located between clusters $i$ and $j$. We often denote these simply as $\beta_i$ and $\delta_{i \rightarrow j}$ when it is clear from the context. $\beta_i$ is designed to be the posterior distribution of factor $\psi_i$ and $\delta_{i \rightarrow j}$ is designed to propagate information from cluster $i$ to cluster $j$.

Performing belief propagation on a tree structure is similar to calculating the marginal distribution of each factor using an efficient factorisation and marginalisation sequence. It is essentially a generalisation of the example in Equation~\ref{eq:marginalising-clustergraph-example}. The following two equations formulate these beliefs:
\begin{align}
    \beta_i &= \psi_i  \prod_{k \in \text{Adj}(i)} \delta_{k \rightarrow i}, \text{\ \ and } \label{eq:calc-beliefs}\\
    \delta_{i \rightarrow j} &= \sum_{\mathbf{C}_i \setminus \mathbf{S}_{i,j}}  \psi_i   \prod_{k \in (\text{Adj}(i) \setminus \{j\})} \delta_{k \rightarrow i }, 
\end{align}
with $k \in \text{Adj}(i)$ depicting neighbouring indices, and $\psi_i$ the factor associated with $\mathbf{C}_i$, i.e.\ the prior.
For example, this procedure is designed to steer a cluster such as ${\{X_1, X_2, X_3\}}$, with $\psi(X_1, X_2, X_3) = P(X_1 | X_2, X_3)$, to the posterior belief $\beta(X_1, X_2, X_3) = P(X_1, X_2, X_3)$.

The message order for belief propagation on a tree structure is captured by Algorithm~\ref{alg:sumproducttree} (sum-product message passing~\cite[p348]{koller}). For example, applying this algorithm to the graph structure in Figure~\ref{fig:SOL-cluster-tree} results in the following message passing order~\cite{streichermasters}:
\begin{enumerate}
    \item we first approach end node (5),
    
    $\delta_{5 \rightarrow 4} = \sum_{X_6}\, \psi_5$
    
    $\delta_{4 \rightarrow 3} = \sum_{X_5}\, \psi_4   \,\delta_{5 \rightarrow 4}$
    
    $\delta_{3 \rightarrow 2} = \sum_{X_4, X_7}\, \psi_3   \,\delta_{4 \rightarrow 3}$

    \item then proceed to end node (7),
    
    $\delta_{7 \rightarrow 6} = \sum_{X_9}\, \psi_7$
    
    $\delta_{6 \rightarrow 2} = \sum_{X_8}\, \psi_6   \,\delta_{7 \rightarrow 6}$
    
    \item  then to end node (1),
    
    $\delta_{1 \rightarrow 2} = \sum_{X_{10}}\, \psi_1$
    
    \item and finally, the order is reversed to propagate consensus back across the nodes,
    
    $\delta_{2\rightarrow 1} = \sum_{X_3}\,     \psi_2   \,\delta_{3 \rightarrow 2}    \,\delta_{6 \rightarrow 2}$
    
    $\delta_{2\rightarrow 6} = \sum_{X_1, X_2}\,\psi_2   \,\delta_{1 \rightarrow 2}    \,\delta_{3 \rightarrow 2}$
    
    $\delta_{6\rightarrow 7} = \sum_{X_3}\,     \psi_6   \,\delta_{2 \rightarrow 6} $
    
    $\delta_{2\rightarrow 3} = \sum_{X_4, X_7}\,\psi_2   \,\delta_{1 \rightarrow 2}    \,\delta_{6 \rightarrow 2}$
    
    $\delta_{3\rightarrow 4} = \sum_{X_5}\,     \psi_3   \,\delta_{2 \rightarrow 3} $
    
    $\delta_{4\rightarrow 5} = \sum_{X_6}\,     \psi_4   \,\delta_{3 \rightarrow 4} $.
\end{enumerate}

\noindent We can now calculate the cluster beliefs by applying Equation~\ref{eq:calc-beliefs}. For example the marginal distribution $P(X_1, X_2, X_3) = \beta_2(\mathbf{C}_2)$ is calculated as
\begin{equation}
    \beta_2 = \psi_2 \,  \delta_{1 \rightarrow 2}\,   \delta_{6 \rightarrow 2}\,  \delta_{3 \rightarrow 2}. \nonumber
\end{equation}
\\\vspace{-1.7em}

\begin{algorithm}[h!]
    \caption{\ Belief propagation. Adapted from~\cite{streichermasters}.}
    \label{alg:sumproducttree}
    \vspace{0.5em}
    \capfnt{Input:}\ \, Cluster tree $\mathcal{T}$ (with clusters $\mathbf{C}_i$, edges $\mathcal{E}$, and sepsets $\mathcal{S}$).
    \vspace{0.5em}    
    \begin{algorithmic}[1]
        
        \State {\color{blue1}// Calculate sepset beliefs}
        \While{any $\mathbf{C}_i$ is ready to pass a message to $\mathbf{C}_j$} \label{alg:sumproducttree-transmit-ready}
        \State{$ \delta_{i \rightarrow j}$ := $\sum_{\mathbf{C}_i \setminus \mathbf{S}_{i,j}}  \psi_i   \prod_{k \in (\text{Adj}(i) \setminus \{j\})} \delta_{k \rightarrow i }$}
        \EndWhile  
        
        \State {\color{blue1}// Calculate cluster beliefs}
        \For{each cluster $\mathbf{C}_i$}
        \State{$\beta_i$ := $\psi_i   \prod_{k \in \text{Adj}(i)} \delta_{k \rightarrow i}$}
        \EndFor
    \end{algorithmic}
    \begin{description}
        \item[Line \ref{alg:sumproducttree-transmit-ready}] Cluster $\mathbf{C}_i$ is ready to pass a message to $\mathbf{C}_j$ when $\mathbf{C}_i$ has received all neighbouring messages except for $\delta_{j \rightarrow i}$.
    \end{description}
\end{algorithm}

There are slight variants on message passing and belief propagation. One such variation is called belief update message passing, also known as the Lauritzen-Spiegelhalter algorithm~\cite{lauritzen1988local}. It is  mathematically equivalent to belief propagation~\cite[p368]{koller} but is preferred in many cases, as it requires fewer calculations than belief propagation.

\section{Loopy belief propagation}\label{sec:loopy-belief-propagation}
If a cluster graph contains loops, a cyclic dependency between messages will result in a deadlock. However, we can heuristically solve this by passing messages iteratively until the beliefs converge. Note that it is proven that such a heuristic will not always converge and that the posterior beliefs can deviate significantly from the true marginals of the system~\cite[p407]{koller}. However, these schemes typically return practical results if implemented effectively~\cite[p407]{koller}.

Loopy belief propagation introduces a new problem in the form of deriving a suitable message scheduling scheme. This is an active research problem~\cite{verburgh2020parallelising} with many different implementations and trade-offs. However, a robust message passing scheme is essential to improving convergence and accuracy~\cite[p408]{koller}.

Our focus is on a message passing scheme that prioritises the parts of the system that underwent significant updates. Our implementation is as follows: after a message $\delta_{i\rightarrow j}$ is propagated, all messages $\delta_{j\rightarrow k} \forall \text{Adj}(j) \setminus \{i\}$ are added to a priority queue using an error metric as weights. This allows us to prioritise the parts of the system that are furthest from converging. The adjustments necessary for belief propagation to accompany loopy graphs are shown in Algorithm~\ref{alg:sumproduct} as loopy belief propagation~\cite[p397]{koller}.

As an error metric, we use the Kullback-Leibler divergence as an approximation of the error between consecutive messages. However, note that the Kullback-Leibler divergence $D_{KL}(P||Q)$ measures the loss of information when a distribution $P$ is approximated by a distribution $Q$, given as
\begin{equation}
    D_{KL}(P||Q) = \sum_i P(i) \, \ln\frac{P(i)}{Q(i)} .
\end{equation}
It is a nonsymmetric function, i.e.\ $D_{KL}(P||Q) \ne  D_{KL}(Q||P)$, and is, therefore, not a true distance or error metric. It does, however, report information about the similarity between the two distributions and is, in practice, still exceptionally efficient for these kinds of applications~\cite{ihler2005loopy}.

Since the algorithm is defined on potential functions and not on probability distributions directly, it is important to note the relationship between 
them, that is,
\begin{equation}
    p(x_1, \ldots , x_n) = \frac{1}{Z} \phi(x_1, \ldots, x_n),
\end{equation}
where $Z$ is the normalisation constant $\sum_{i}\phi(i)$. Numerical stability is known to be a problem in long-running loopy belief propagation cycles~\cite{mooij2012sufficient}. It is, therefore, advisable to keep messages and beliefs normalised. \newpage

\begin{algorithm}[H]
    \caption{\ Loopy belief propagation. Adapted from~\cite{streichermasters}.}
    \label{alg:sumproduct}
    \vspace{0.5em}
    \capfnt{Input:}\ \, Cluster graph $\mathcal{T}$ (with clusters $\mathbf{C}_i$, edges $\mathcal{E}$, and sepsets $\mathcal{S}$).
    \vspace{0.5em}
    \begin{algorithmic}[1]
        
        \State{$q := [\ ]$}
        \For{each edge $(i, j)$ in $\mathcal{E}$} \label{alg:sumproduct-forloopone}
        \State {\color{blue1}// Initialise queue with miniscule priorities and a bias towards leaf nodes}
        \State{priority $:= 10^{-10}(|\text{Adj}(i)|+|\text{Adj}(j)|)^{-1}$}
        \State{$q.$push$($priority $\rightarrow (i,j))$}
        \State{$q.$push$($priority $\rightarrow (j,i))$}
        \State {\color{blue1}// Initialise cluster messages as vacuous}
        \State{$\delta_{i\rightarrow j} := \mathbf{1}$}
        \State{$\delta_{j\rightarrow i} := \mathbf{1}$}
        \EndFor  
        \State {\color{blue1}//  Message passing}
        \While{$q$ is not empty}\label{alg:sumproduct-graphcalibrated}
        \State $(i,j) := q.$pop\_highest$()$
        \State $\delta_{\text{prev}} := \delta_{i\rightarrow j}$
        \State{$ \delta_{i \rightarrow j}:= \text{norm}\mkern-3mu\left(\sum_{\mathbf{C}_i \setminus \mathbf{S}_{i,j}}  \psi_i   \prod_{k \in (\text{Adj}(i) - \{j\})} \delta_{k \rightarrow i }\, \right)$} \ \  {\color{blue1}// Update message \label{alg:sumproduct-update}}
        \State {\color{blue1}//  Propagate priorities}
        \State priority $:= \text{error}(\delta_{\text{prev}},  \delta_{i \rightarrow j})$ \label{eq:error-metric}
        \For{$k \in \text{Adj}(j) - \{i\}$ }
        \If{ $(j, k) \in  q $}
        \State remove $(j, k)$ from $q$
        \EndIf
        \If{$\text{priority} > $ chosen threshold} \label{alg:sumproduct-appendcondition}
        \State $q$.push$($priority$ \rightarrow (j, k))$
        \EndIf
        \EndFor
        \EndWhile
        \State {\color{blue1}//  Calculate posterior beliefs}
        \For{each cluster $\mathbf{C}_i$} \label{alg:sumproduct-beliefs}
        \State{$\beta_i := \psi_i   \prod_{k \in \text{Adj}(i)} \delta_{k \rightarrow i}$}
        \EndFor
        
    \end{algorithmic}
    \begin{description}\setlength{\itemsep}{1pt}\setlength{\parsep}{0pt}
        \item[Line 8--9] An alternative initialisation is $\delta_{i\rightarrow j} := \text{norm}\mkern-3mu\left(\sum_{\mathbf{C}_i \setminus \mathbf{S}_{i,j}} \psi_i \right)$.
        \item[Line \ref{alg:sumproduct-update}] Message damping can be applied by adding the line:\\
        ${}$\hspace{70pt}$\delta_{i \rightarrow j}:= (1-\lambda)(\delta_{\text{prev}}) + (\lambda)(\delta_{i \rightarrow j})$,\\
        where $\lambda$ is the damping factor.
        \item[Line~\ref{alg:sumproduct-appendcondition}] Local convergence is reached when all message updates are below a chosen threshold.
    \end{description}
    
\end{algorithm}

\section{Example}
\subsubsection*{Hamming (7,4) model}

We now have the tools to solve the Hamming (7,4) code as a PGM. We initialise the cluster graph factors of Equation~\ref{eq:SOL-list-of-factors} using the parity bit logic of Equation~\ref{eq:SOL-b1plusb2plusb3}. Factor $\psi_1(B_5, B_1, B_2, B_3)$ represents the conditional distribution $P(B_5| B_1, B_2, B_3)$ via
\begin{equation}
    \psi_1(B_5, B_1, B_2, B_3) = \left\{
    \begin{array}{ll}
        1, &\text{where }\,  b_5 = b_1  \oplus b_2  \oplus b_3 \\
        0, &\text{where }\,  b_5 \neq b_1  \oplus b_2  \oplus b_3
    \end{array} 
    \right. . \nonumber
\end{equation}
We similarly define the factors $\psi_2(B_6, B_2, B_3, B_4)$ and $\psi_3(B_7, B_1, B_3, B_4)$.

The factors $\psi_4(R_1, B_1), \ldots, \psi_{10}(R_7, B_7)$ are represented by $P(R_1| B_1), \ldots, P(R_7| B_7)$. If we assume a $10\%$ chance for each bit to be erroneously flipped, we can set a probability of $90\%$ for a transmitted bit being equal to a received bit and $10\%$ otherwise:
\begin{equation}
    \psi_4(R_1, B_1) = \left\{
    \begin{array}{ll}
        0.9, &\text{where }\,  b_1 =  r_1 \\
        0.1, &\text{where }\,  b_1 \neq r_1
    \end{array} 
    \right. .\nonumber
\end{equation}
We assign the potential functions for $\psi_5, \ldots, \psi_{10}$ similarly.

 Now that we have factor potentials, we can solve a Hamming (7,4) message by applying observations and running loopy belief propagation.

\subsubsection*{Example}
Given that a message ``1010'' is to be sent over a noisy channel using a Hamming (7,4) code, the message is first encoded as ``1010010'' and then transmitted. After transmission, the message is recorded as ``1110010'', with an unknown bit-flip at $b_2$. We are now going to explore the use of belief propagation for detecting and recovering bit-flips.

First, we integrate the observations
\begin{equation}
    R_1\equalss 1,\ R_2\equalss 1,\ R_3\equalss 1,\ R_4\equalss 0,\, R_5\equalss 0,\ R_6\equalss 1,\ R_7\equalss 0 \nonumber
\end{equation}
by replacing the factors $\psi_{4}(R1, B1),\ldots,\psi_{10}(R7,B7)$ with reduced versions of themselves,  $\psi\prime_4(B_1),\ldots,\psi\prime_{10}(B_7)$, with their potentials as
\begin{align}
    &\psi\prime_4(1)=0.9, & &
    \psi\prime_5(1)=0.9, & &
    \psi\prime_6(1)=0.9, & &
    \psi\prime_7(1)=0.1, \nonumber \\ 
    &\psi\prime_8(1)=0.1, & &
    \psi\prime_9(1)=0.9, & &
    \psi\prime_{10}(1)=0.1. & & & \nonumber 
\end{align}

\begin{figure}[h] 
    \centering
    \def\svgscale{0.9}
    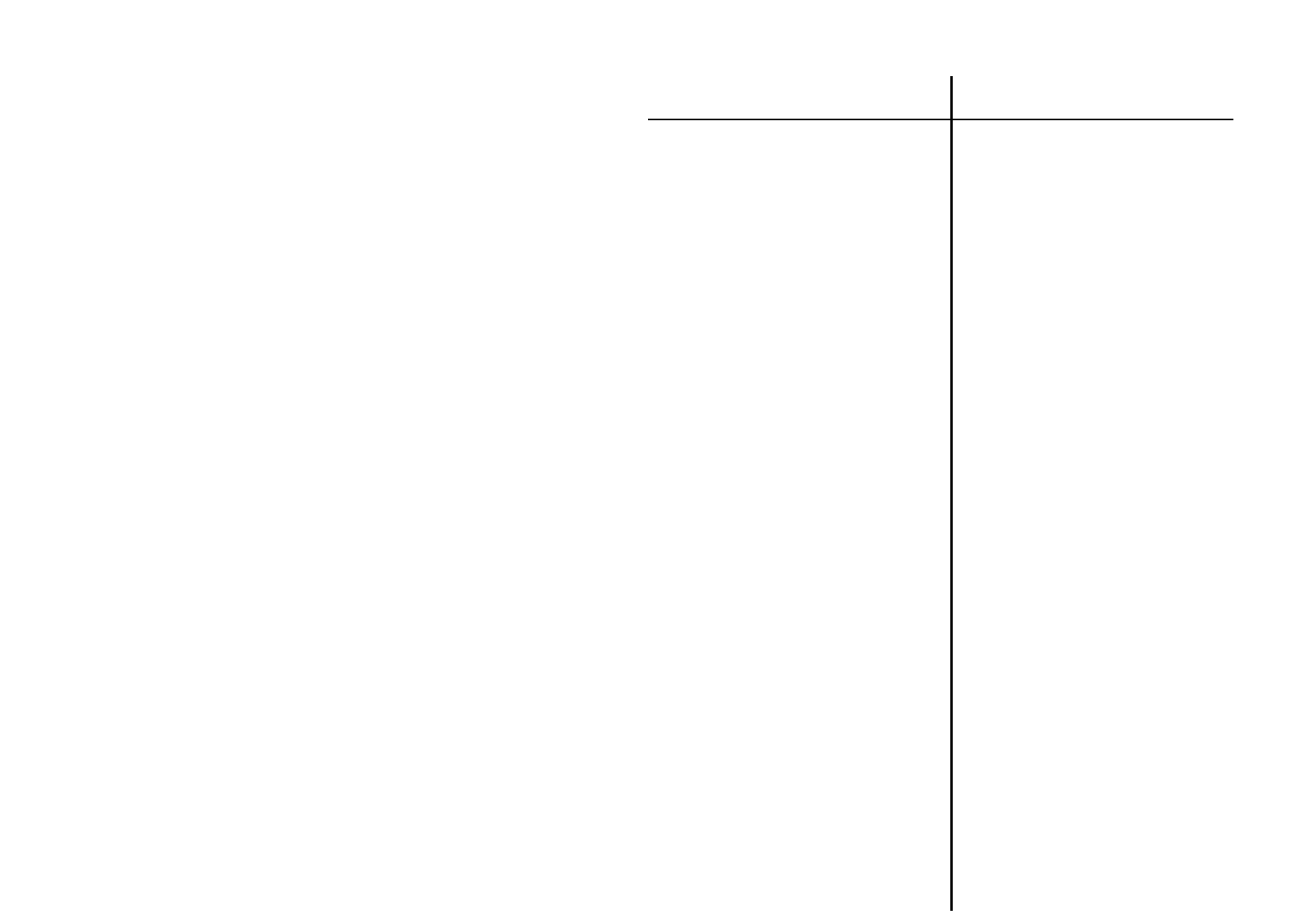
    \caption{Probability distributions obtained from performing belief propagation on an erroneous Hamming (7,4) message of ``1110010``. Loopy belief propagation managed to identify and correct a bit-flip in $b_2$ with a $~98\%$ certainty, recovering the correct message of ``1010010``.}
    \label{fig:beliefs-results}
\end{figure}

We can then run loopy belief propagation (Algorithm~\ref{alg:sumproduct}) on the cluster graph of Figure~\ref{fig:SOL-clustergraph} and obtain posterior beliefs for each factor. By querying the posterior distributions, the system can extract the original message as ``1010010'' with very high confidence. To illustrate, posterior distributions from a resulting run of loopy belief propagation~\cite{streichermasters} are shown in Figure~\ref{fig:beliefs-results}.

\section{Conclusion}
The objective of this chapter was to present a basis for formulating and solving probabilistic reasoning problems using PGMs. We supplied all the theory and tools necessary to formulate, build, and solve the Hamming (7,4) code as a graphical model. The goal was to provide basic underlying PGM techniques as building blocks for the later chapters.

\chapter{Graph colouring: comparing cluster graphs to factor graphs}\label{gc-chapter}

\section*{Preface}
\renewcommand{\thefootnote}{\fnsymbol{footnote}}

This chapter presents \textit{Graph Colouring: Comparing Cluster Graphs to Factor Graphs}~\cite{streicher}\footnote[2]{Sections of this work have been published in:
    \\
    S. Streicher and J. du Preez, “Graph Coloring: Comparing Cluster Graphs to Factor Graphs,”
    in \textit{Proceedings of the ACM Multimedia 2017 Workshop on South African Academic Participation}. SAWACMMM ’17. New York, NY, USA: ACM, 2017, pp. 35–42.
}, first presented at the ACM 25th International Conference on Multimedia (ACM MM 2017) at the Computer History Museum in Silicon Valley.

This publication is the first instalment of three to serve as the framework for our research on incremental inference on higher-order PGMs, applied to constraint satisfaction.
The main contributions under author S.\ Streicher are
\begin{itemize}
    \item a comparative study between cluster graphs and factor graphs, in which cluster graphs show great promise in comparison to factor graphs, and
    \item a comprehensive integration of various aspects required to formulate graph colouring problems into PGMs (further expanded to general constraint satisfaction in Chapter~\ref{csp-chap}).
\end{itemize}

Furthermore, a significant contribution is the general-purpose cluster graph construction algorithm, LTRIP. Credit for the discovery of the algorithm goes to author J.\ du Preez, and credit for formulating the algorithm and abstracting the ``Connection-Weights'' subroutine goes to S.\ Streicher.

By establishing the groundwork for graph colouring and cluster graph formulation, this chapter forms the basis for our larger project aimed to develop efficient PGM solutions to constraint satisfaction.
Chapter~\ref{lc-chap} contributes to a cartography classification problem by using the region-based formulation established in Section~\ref{gc-sec:fourcolorproblem} and applying the PGM-based inference found in Section~\ref{gc-sec:messagepassing}. 
Chapter~\ref{csp-chap} introduces a higher-order PGM technique called purge-and-merge that combines LTRIP with factor multiplication to solve problems too difficult for other belief-propagation approaches, overcoming the limitations presented in Section~\ref{gc-sec:experiments}.

\renewcommand{\thefootnote}{\arabic{footnote}}
\section*{Abstract}
We present a formulation for solving graph colouring 
problems with probabilistic graphical models. In contrast to
the prevailing literature that uses factor graphs for this
purpose, we instead approach it from a cluster graph
perspective. Noting the lack of algorithms to
automatically construct valid cluster graphs, we provide such an algorithm (termed LTRIP).  Our experiments indicate a
significant advantage for preferring cluster graphs over
factor graphs, both in terms of accuracy as well as
computational efficiency.

\section{Introduction}

Due to their learning, inference, and pattern-recognition abilities,
machine learning techniques such as neural networks, probabilistic
graphical models (PGMs), and other inference-based algorithms have
become quite popular in artificial intelligence research. PGMs can
easily express and solve intricate problems with many dependencies,
making it a good match for problems such as graph colouring. The PGM
process is similar to aspects of human reasoning, such as the process
of expressing a problem by using logic and observation and applying
inference to find a reasonable conclusion. With PGMs, it is often
possible to express and solve a problem from easily formulated
relationships and observations, without the need to derive complex
inverse relationships. This can be an aid to problems with many
inter-dependencies that cannot be separated into independent parts to
be approached individually and sequentially.

Although the \emph{cluster} graph topology is well established in the
PGM literature~\cite{koller, dechter2010on}, the overwhelmingly dominant
topology encountered in literature is the \emph{factor} graph. We
speculate that this is at least partially due to the absence of algorithms
to \emph{automatically} construct valid cluster graphs, whereas factor
graphs are trivial to construct. To address this, we detail a general-purpose construction algorithm termed LTRIP (layered trees for the running
intersection property). We have been covertly experimenting with this
algorithm for a number of years~\cite{myprasapaper, charldev, streichermasters, daniekphd}.

The graph colouring problem originated from the literal
colouring of planar maps. It started with the four-colour map theorem,
first noted by Francis Guthrie in 1852. He conjectured that four
colours are sufficient to colour neighbouring counties differently for
any planar map. It was ultimately proven by Kenneth Appel and Wolfgang
Haken in 1976 and is notable for being the first major mathematical
theorem with a computer-assisted proof. In general, the graph colouring
problem deals with the labelling of nodes in an undirected graph such
that adjacent nodes do not have the same label. The problem is core to
a number of real-world applications, such as scheduling timetables for
university subjects or sporting events, assigning taxis to customers,
and assigning computer programming variables to computer
registers~\cite{lewis2015guide,timetables,briggs1992register}. As
graphical models became popular, message passing provided an exciting new
approach to solving graph colouring and (the closely related)
constraint satisfaction
problems~\cite{MoonT,kroc2009counting}. For constraint
satisfaction, the survey propagation message passing technique seems to
be particularly
effective~\cite{braunstein2005survey,maneva2004survey,kroc2012sprevisited,knuth2011art4b}.
These techniques are primarily based on the factor graph PGM topology.

The work reported here forms part of a larger project aimed at
developing an efficient alternative for the above message passing
solutions to graph colouring. Cluster graphs and their efficient
configuration are important in that work -- hence our interest in those
aspects here. Although we also provide basic formulations for
modelling graph colouring problems with PGMs, this is not the primary
focus of this chapter but instead only serves as a vehicle for
comparing topologies.

The rest of this chapter is structured as follows. Section~\ref{gc-sec:graph_coloring}
shows how the constraints of a graph colouring problem can be
represented as ``factors''. Furthermore, it is shown how these factors
are linked up into graph structures on which inference can be
applied. Section~\ref{gc-sec:topology} discusses the factor graph and
cluster graph topologies and algorithms for automatically
configuring them. The former is trivial; for the latter, we provide the
LTRIP algorithm in Section~\ref{gc-sec:ltrip}. Section~\ref{gc-sec:sudoku}
then integrates these ideas by expressing the well-known Sudoku puzzle
(an instance of a graph colouring problem) as a PGM. The experiments in
Section~\ref{gc-sec:experiments} show that the cluster graph approach is simultaneously faster and more
accurate, especially for complex
cases. The last two sections consider possible future exploration
and final conclusions.

\section{Graph colouring with PGMs} \label{gc-sec:graph_coloring}

This section provides a brief overview of graph colouring and PGMs,
along with techniques for formulating a graph colouring problem as
a PGM. We also explore the four-colour map theorem and illustrate how to solve these and similar problems through an example.

\subsection{A general description of graph colouring problems} \label{gc-sec:graph_coloring_description}
Graph colouring problems are NP-complete -- easily defined and verified
but can be difficult to invert and solve. The problem is of
significant importance as it is used in a variety of combinatorial and
scheduling problems.

The general graph colouring problem deals with attaching labels (or
``colours'') to nodes in an undirected graph, such that (a) no two
nodes connected by an edge may have the same label, and (b) the number
of different labels that may be used is minimised. Our focus is mainly
on the actual labelling of a graph.

\begin{figure}[h]
  \centering
  \includegraphics[width=0.85\columnwidth]{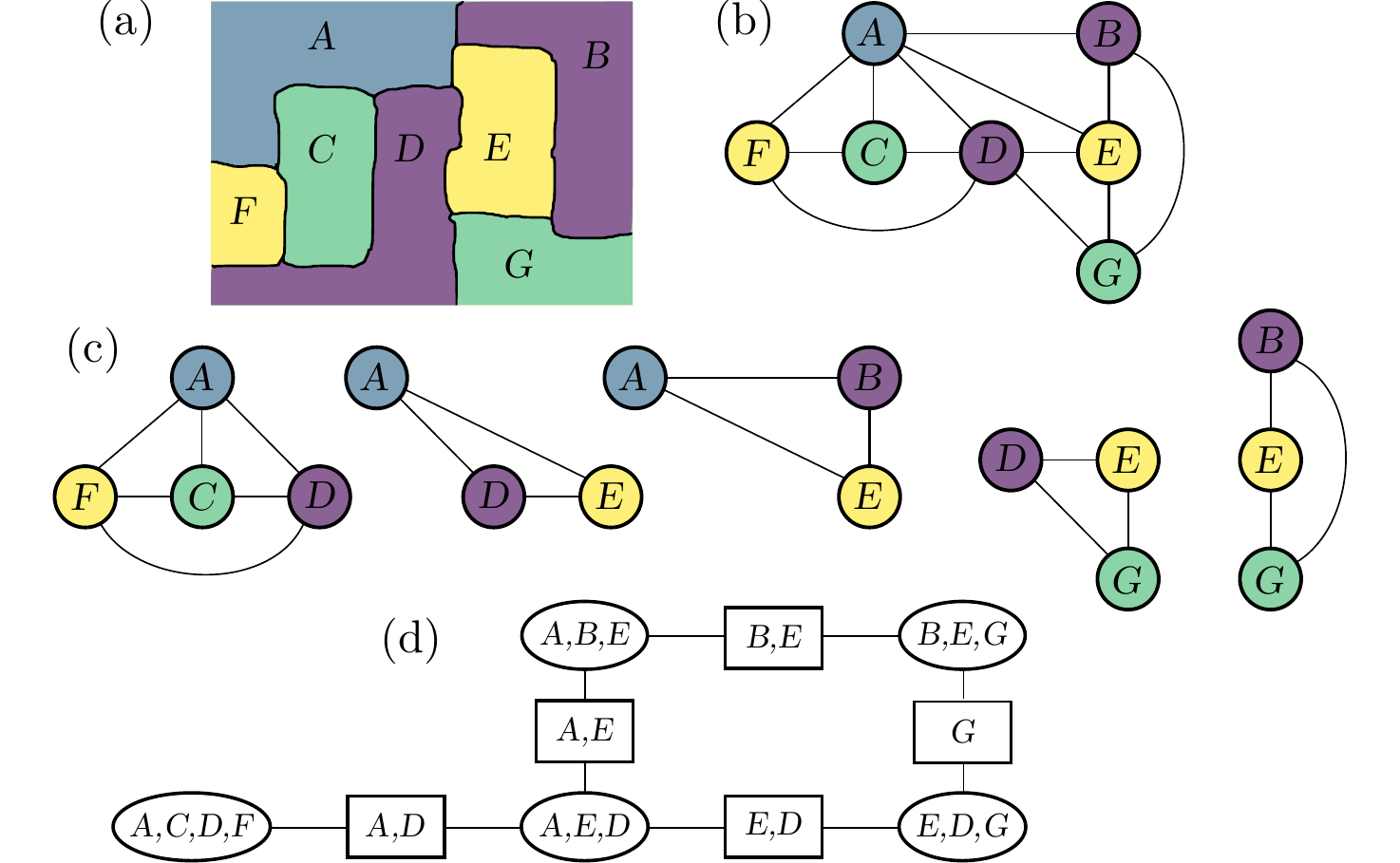}
  \caption{ (a) A four-colour graph problem containing regions $A$ to
    $G$, with (b) its graph colouring representation, (c) the maximal
    cliques within the graph, and (d) a cluster graph configuration
    for this problem. The ellipses represent the clusters and the
    boxes the sepsets -- see the main text for more detail.
  }\label{gc-fig:fourcolormap}
\end{figure}

A practical example of such a graph colouring is the classical four-colour map problem that gave birth to the whole field: a cartographer
is to colour the regions of a planar map such that no two adjacent
regions have the same colour. To present this problem as graph
colouring, an undirected graph is constructed by representing each
region in the map as a node and each boundary between two regions as
an edge connecting those two corresponding nodes. Once the problem is
represented in this form, a solution can be approached by any typical
graph colouring algorithm.  An example of this parametrisation can be
seen in Figure~\ref{gc-fig:fourcolormap}~(a) and (b); we refer to (c) and
(d) later on.

\subsection{PGMs to represent graph colouring problems} \label{gc-sec:gcwithpgms}
PGMs are used as a tool to reason about large-scale probabilistic
systems in a computationally feasible manner. They are known for their
powerful inference over problems with many interdependencies. It is
often useful for problems that are difficult to approach
algorithmically, with graph colouring being a specific example.

In essence, a PGM is a compact representation of a probabilistic space
as the product of smaller, conditionally independent distributions
called factors.  Each factor defines a probabilistic relationship over
the variables within its associated cluster -- a cluster being a set
of random variables. For discrete variables, this results in a discrete
probability table over all possible outcomes of these
variables. Instead of explicitly calculating the product of these factors
(which typically is not computationally feasible), a PGM connects them
into an appropriate graph structure. We apply inference by passing
messages over the links in this structure until
convergence is reached (see Section~\ref{sec:loopy-belief-propagation}). In combination with the initial factor
distributions, these converged messages can then be used to obtain the
(approximate) posterior marginal distributions over subsets of
variables.

To factorise a graph colouring problem, we first need to parametrise
the problem probabilistically. This is achieved by allowing each node
in the graph to be represented by a discrete random variable $X_i$,
that can take on a number of states. For graph colouring, these states
are the available labels for the node, e.g.\ four
colours in the case of the four-colour map problem.

Now that we have the random variables of our system and the domains of these variables, we need to capture the relationship between them to represent it as factors in our PGM.
For graph colouring, no two adjacent nodes may have the same colour;
therefore, their associated random variables may not have the same
state. One representation of this system would then be to capture this
relationship using factors with a scope of two variables, each taken
as an adjacent pair of nodes from the colouring graph. Although this fully represents the solution space, there is still a trade-off
between accuracy and cluster size (cardinality)~\cite{mateescu2010join}. Fortunately, some configurations allow for larger clusters.

A clique is defined as a set of nodes that are all adjacent to each
other within the graph, and a maximal clique is defined as a clique that is not a subset of any other clique. In order to maximise factor scope, we prefer to define our factors directly on the maximal
cliques of the graph. (We use the terms clique and cluster more or
less interchangeably.)  We can then set the discrete probability tables of
these factors to only allow states where all the variables are
assigned different labels. In the next section, we show an example of
this.

After finalising the factors, we can complete the PGM by linking these
factors in a graph structure. There are several valid structure
variants to choose from -- in this work, we specifically focus on
factor graph and cluster graph structures. In the resulting graph
structure, linked factors exchange information with each other about
\emph{some}, and not necessarily all, of the random variables they
have in common. These variables are known as the separation set, or
``sepset'' for short, on the particular link of the graph. Whichever
graph structure we choose must satisfy the so-called running
intersection property (RIP)~\cite[p347]{koller}. This property
stipulates that for all variables in the system, any occurrence of a
particular variable in two distinct clusters should have a unique
(i.e.\ precisely one) path linking them up via a sequence of sepsets that
all contain that particular variable. Several examples of this are
evident in Figure~\ref{gc-fig:fourcolormap}~(d). In particular, note the
absence of the $E$ variable on the sepset between the $\{B,E,G\}$ and
$\{E,D,G\}$ clusters. If $E$ were to be included, there would have been two
distinct sepset paths containing $E$ between those two
clusters. This would be invalid, broadly, because it causes a type of
positive feedback loop.

After establishing the factors and linking them in a graph
structure, we can apply inference by using one of several belief
propagation algorithms available (see Section~\ref{sec:loopy-belief-propagation}).

\subsection{Example: The four-colour map problem}\label{gc-sec:fourcolorproblem}
We illustrate the above by means of the four-colour map problem. The
example in Figure~\ref{gc-fig:fourcolormap} can be expressed by the seven
random variables $A$ to $G$, grouped into five maximal cliques as
shown. There will be no clique with more than four variables
(otherwise, four-colours would not be sufficient, resulting in a
counter-example to the theorem). These maximal cliques are represented
as factors with uniform distributions over their valid
(i.e.\ non-conflicting) colourings. We do so by assigning either a
possibility or an impossibility to each joint state over the factor's
variables. More specifically, we represent the potential function $\phi(A,C,D,F)$ with a discrete table, 
assigning a ``1'' for outcomes where all variables have differing colours
and a ``0'' for cases with duplicate colours.

For example, the factor belief $\phi(A,C,D,F)$ for the
puzzle in Figure~\ref{gc-fig:fourcolormap} is shown in
Table~\ref{gc-tab:probtable}.  These factors are connected into a graph
structure -- such as the cluster graph in
Figure~\ref{gc-fig:fourcolormap}~(d). We can use belief propagation
algorithms on this graph to find posterior beliefs.
\begin{table}[h!]
  \centering
  \begin{tabular}{r c c c c|l l}
    Random variables $\rightarrow$ & $A$ & $C$ & $D$ & $F$ &  &\\
    \hline
    State $\rightarrow$ & 1 & 2 & 3 & 4 & 1 &  $\leftarrow$ $\phi(A, C, D, F)$\\
    & 1 & 2 & 4 & 3 & 1 &  \\
    & 1 & 3 & 2 & 4 & 1 &  \\
    & 1 & 3 & 4 & 2 & 1 &  \\
    &   &   & $\vdots$  &   &   & \\
    & 4 & 3 & 2 & 1 & 1  & \\\cline{2-6}
    & \multicolumn{4}{c|}{elsewhere} & 0 &
  \end{tabular}
  \caption{A discrete table representing $\phi(A,C,D,F)$, where all possible combinations of outcomes for $\{A, C, D, F\}$ are captured.}
  \label{gc-tab:probtable}
\end{table}

We successfully tested this concept on various planar maps of size
$100$ up to $8000$ regions. These were created by generating
superpixels on arbitrary images using the SLIC algorithm~\cite{slic} to serve as the
initially uncoloured regions.

PGMs configured to utilise only binary probabilities always preserve all possible solutions~\cite{dechter2010on}. The underlying reason is that a state considered as possible within a particular factor will always be retained unless a message from a neighbouring factor flags it as impossible. In that case, it is quite correct that it should be removed from the spectrum of possibilities. \citet{dechter2010on} proved this by showing that applying belief propagation on this type of configuration results in a constraint satisfaction algorithm for generalised arc consistency.

\begin{figure}[h]
  \begin{center}
    \includegraphics[width=0.85\columnwidth]{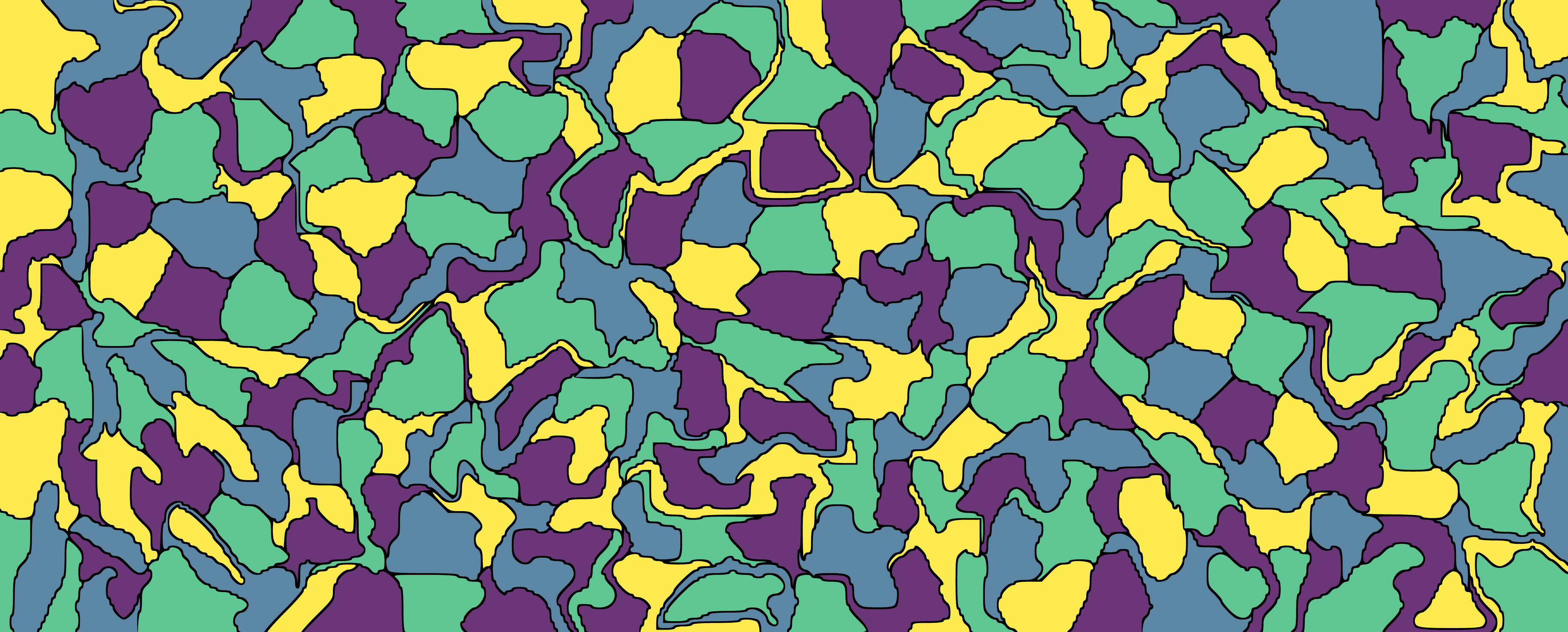}
  \end{center}
  \caption{ A generated planar map with its colouring results from a PGM labelling the 250 regions into one out of four colours.  }\label{gc-fig:fourcolor}
\end{figure}

However, in this four-colour map example, the space of solutions can, in
principle, be prohibitively large. We force our PGM to instead find a
particular unique solution, by firstly fixing the colours in the
largest clique and, secondly, by very slightly biasing the other factor
probabilities towards initial colour preferences. This makes it
possible to pick a particular unique colouring as the most likely
solution. An example of a graph of 250 regions can be seen in
Figure~\ref{gc-fig:fourcolor}.

\section{Factor vs cluster graph topologies} \label{gc-sec:topology}
The graph structure of a PGM can make a big difference in the speed
and accuracy of inference convergence. That said, factor graphs are
the predominant structure in literature -- surprisingly so, since we
found them to be inferior to a properly structured cluster
graph. Cluster graphs allow for passing multivariate messages between
factors, thereby maintaining some of the inter-variable correlations
already known to the factor. This is in contrast to factor graphs,
where information is only passed through univariate messages, thereby
implicitly destroying such correlations.

A search on scholar.google.com (conducted on June 28, 2017) for
articles relating to the use of factor graphs versus cluster graphs in
PGMs returned the following counts:
\begin{itemize}[leftmargin=1.4em]
\item 5590 results for \textit{probabilistic graphical models "factor graph"},
\item 661 results for \textit{probabilistic graphical models "cluster graph"}, and
\item 49 results for \textit{probabilistic graphical models "factor graph" "cluster graph"}.
\end{itemize}
Among the latter 49 publications (excluding four items authored at our
university), no cluster graph constructions are found other than for
Bethe / factor graphs, junction trees, and the clustering of Bayes
networks. We speculate that this relative scarcity of cluster graphs
points to the absence of an automatic and generic procedure for
constructing good RIP satisfying cluster graphs.

\subsection{Factor graphs}\label{gc-sec:factorgraphs}
A factor graph, built from clusters $\mathbf{C}_i$, can be expressed
in cluster graph notation as a Bethe graph $\mathcal{F}$. For each
available random variable $X_j$, $\mathcal{F}$ contains an additional
cluster $\mathbf{C}_j=\{X_j\}$. Their associated factors are all
uniform (or vacuous) distributions and, therefore, do not alter the
original products' distributions.  Each cluster containing $X_j$ is
linked to this vacuous cluster $\mathbf{C}_j$. This places $\mathbf{C}_j$ at the hub of
a star-like topology, with all the various $X_j$ subsets radiating
outwards from it. Due to this star-like topology, the RIP requirement is
trivially satisfied.

\begin{figure}[h]
  \centering
  \vspace{0.3em}
  \includegraphics[width=0.85\columnwidth]{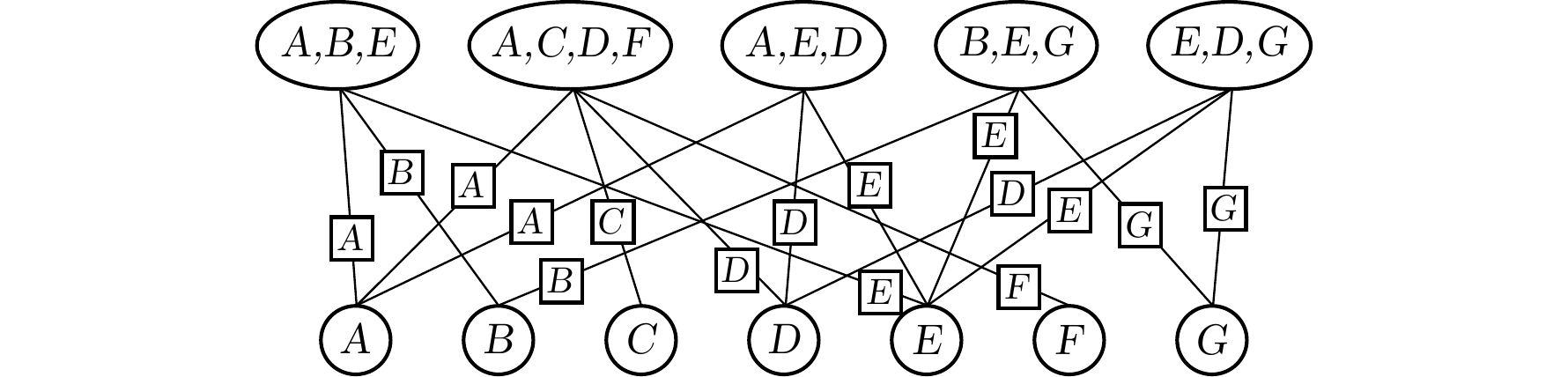}
  \caption{ The Bethe factor graph topology applicable to
    Figure~\ref{gc-fig:fourcolormap}. Note the univariate sepsets are arranged
    in a star-like topology.  }\label{gc-fig:bethegraph}
\end{figure}

The setup of a factor graph from this definition is straightforward,
the structure is deterministic, and the placements of sepsets are well
defined. Figure~\ref{gc-fig:bethegraph} provides the factor graph for the
factors shown in Figure~\ref{gc-fig:fourcolormap}. %

\subsection{Cluster graphs}\label{gc-sec:clustergraphs}
A cluster graph $\mathcal{T}$, built from clusters $\mathbf{C}_i$, is
a non-unique undirected graph, where
\begin{enumerate}%
\item no cluster is a subset of another cluster, $\mathbf{C}_i
  \nsubseteq \mathbf{C}_j$ for all $i \neq j$,
\item the clusters are used as the nodes,
\item the nodes are connected by non-empty sepsets $\mathbf{S}_{i,j}
  \subseteq \mathbf{C}_i \cap \mathbf{C}_j$, and
\item the sepsets satisfy the running intersection property.
\end{enumerate}
Point (1) is not strictly necessary (see the factor graph
structure) but provides convenient computational savings. Moreover, it can
always be realised by simply assimilating non-obliging clusters into a
superset cluster via factor multiplication. Refer to
Figure~\ref{gc-fig:fourcolormap}~(d) for an example of a typical cluster
graph.

Although Koller~\cite[p404]{koller} provides extensive theory on cluster graphs, no general solution for its construction
is provided. Indeed, they state that ``the choice
of cluster graph is generally far from obvious, and it can make a
significant difference to the [belief propagation] algorithm.''
Furthermore, the need for such a construction algorithm is made clear
from their experimental evidence, which indicates that faster
convergence and an increase in accuracy can be obtained from better
graph structuring. Therefore, since cluster graph theory is well
established, an efficient and uncomplicated cluster graph construction
algorithm will be useful. We provide the LTRIP algorithm for this
purpose.

\subsection{Cluster graph construction via LTRIP}\label{gc-sec:ltrip}

The LTRIP algorithm is designed to satisfy the running intersection
property for a cluster graph $\mathcal{T}$ by layering the
interconnections for each random variable separately into a tree
structure and then superimposing these layers to create the combined
sepsets. More precisely, for each random variable ${X}_i$
available in $\mathcal{T}$, all the clusters containing ${X}_i$
are interconnected into a tree structure -- this is then the layer
for ${X}_i$. After finalising all these layers, the sepset
between cluster nodes $\mathbf{C}_i$ and $\mathbf{C}_j$ in $\mathcal{T}$ is the union of all the individual variable connections over all these layers.

\begin{figure}[h]
    \centering
    \includegraphics[width=0.95\columnwidth]{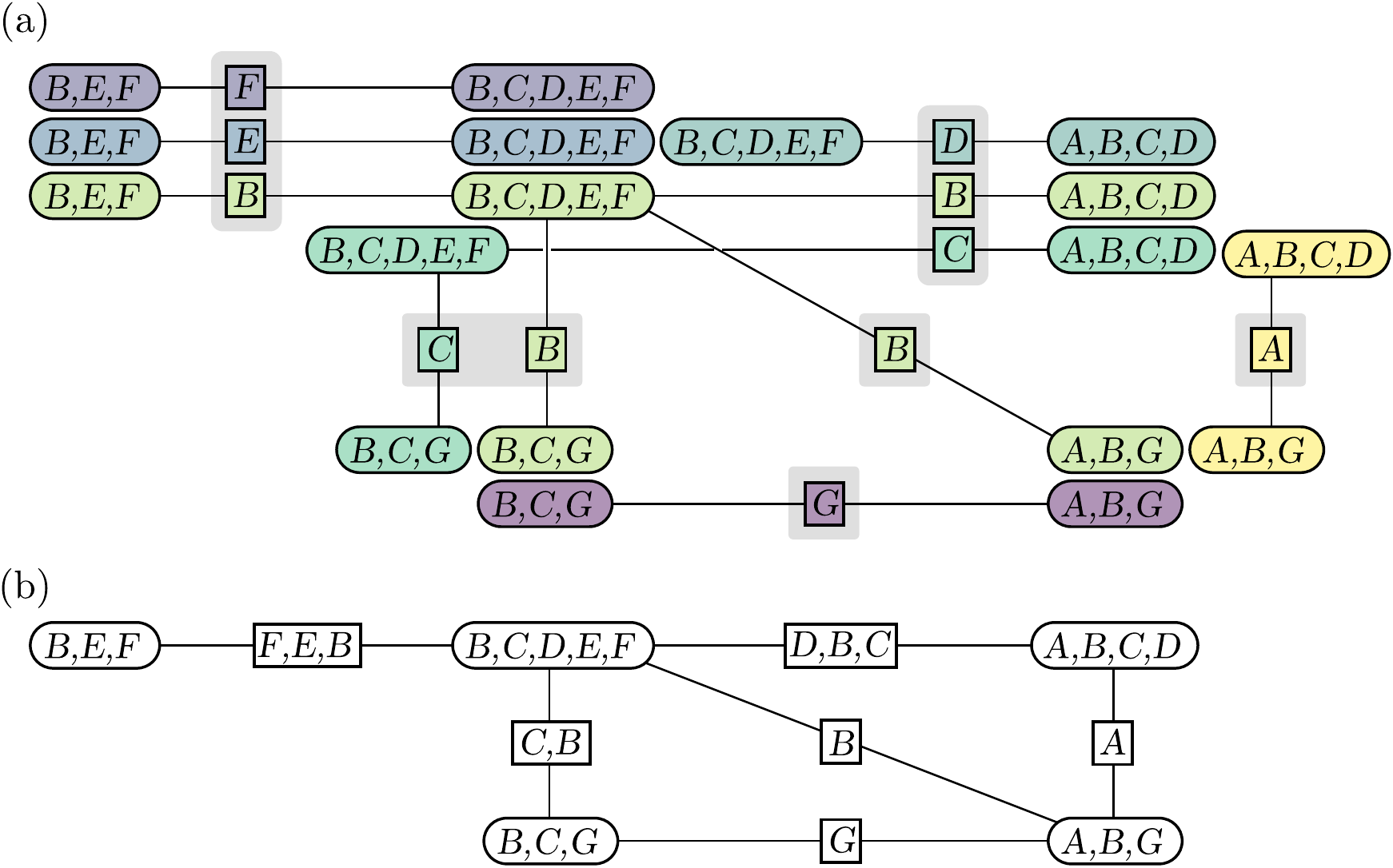}
    \caption{The intermediate tree structures resulting from the LTRIP procedure in Algorithm~\ref{gc-alg:ltrip}. The five clusters to be linked are $\{B,C,D,E,F\}$, $\{A,B,C,D\}$, $\{B,E,F\}$, $\{B,C,G\}$, and $\{A,B,G\}$. (a) The LTRIP procedure yielded seven intermediate tree structures, each a contribution from one of the seven variables $A$, $B$, $C$, $D$, $E$, $F$, and $G$. (b)~The resulting cluster graph is simply a superposition of these tree structures.\vspace{-1em}}\label{gc-fig:ltrip-trees}
\end{figure}

\begin{figure}[h]
    \centering
    \includegraphics[width=0.85\columnwidth]{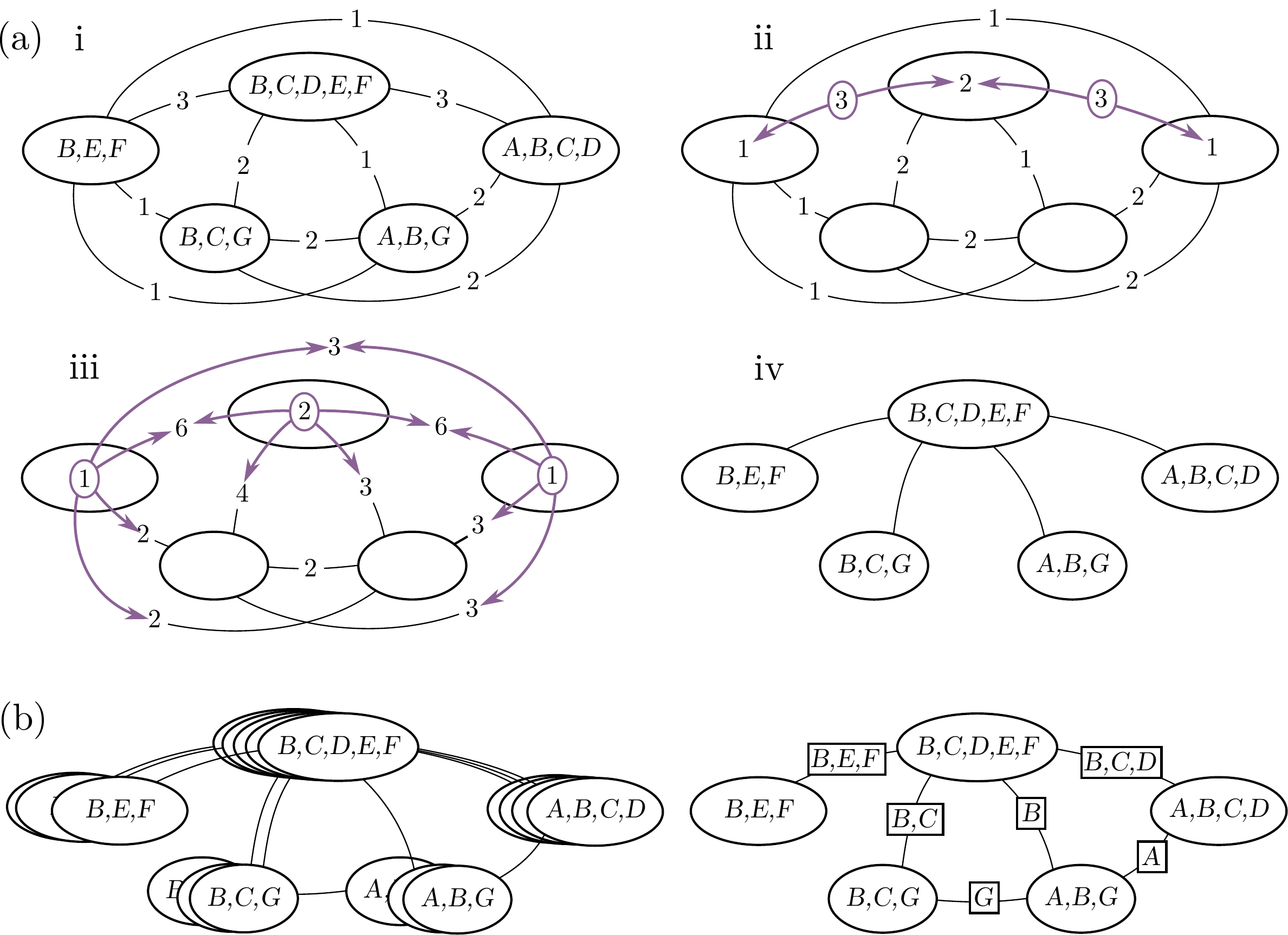}
    \caption[]{Illustration of constructing a cluster graph via the LTRIP
        procedure. The five clusters to be linked are $\{B,C,D,E,F\}$,
        $\{A,B,C,D\}$, $\{B,E,F\}$, $\{B,C,G\}$ and $\{A,B,G\}$.
        \vspace{0.25em}\\
		\capfnt{Figure \ref{gc-fig:ltrip} (a):}
        The procedure for joining up all clusters containing
        variable $B$ into a tree. In sub-step (i), we set the initial
        connection weights as the number of variables shared by each
        cluster pair. In sub-step (ii), we identify the current maximal
        connection weight to be $m = 3$. In sub-step (iii), we note for
        each cluster how many of its links have maximal weight $m$.  This
        number is added to all its connection weights. This emphasises
        clusters that are strongly connected to others. In sub-step (iv),
        we use these connection weights to form a maximal spanning tree,
        connecting all occurrences of variable $B$.
        \vspace{0.25em}\\
        \capfnt{Figure \ref{gc-fig:ltrip} (b):}
        Similarly constructed connection trees for all other variables are
        superimposed to yield the final cluster graph and sepsets.}\label{gc-fig:ltrip}
\end{figure}

While this procedure guarantees to satisfy the RIP requirement, there
is still considerable freedom in exactly how the tree structure on
each separate layer is connected. We were guided by the
assumption that it is beneficial to prefer linking clusters with a
high degree of mutual information. We, therefore, chose to create trees
that maximise the size of the sepsets between clusters. The full
algorithm is detailed in Algorithm~\ref{gc-alg:ltrip}. In addition, an example for constructing all the intermediate trees can be found in Figure~\ref{gc-fig:ltrip-trees}, and an
illustration of constructing a tree for a single variable in Figure~\ref{gc-fig:ltrip}. Furthermore, a reiteration and summary of 
the algorithm can be found in Section~\ref{csp-sec:ltrip}.

\def\NoNumber#1{{\def\alglinenumber##1{}\State #1}\addtocounter{ALG@line}{-1}}
\begin{algorithm}[h!]
    \caption{\ LTRIP}
    \label{gc-alg:ltrip}
    \vspace{0.5em}
    \capfnt{Input:}\ \, Set of clusters $\mathcal{V} = \{\mathbf{C}_1, \ldots, \mathbf{C}_N \}$, with subsets already assimilated into their su-
    \\${}$\hspace{2em} persets; i.e.\ $\mathbf{C}_i \nsubseteq \mathbf{C}_j \, \forall \, i \neq j$.
    \vspace{0.5em}
    \begin{algorithmic}[1]
        
        \State{\color{blue1}// {Empty set of sepsets}}
        \State{$\mathcal{S} := \{\}$ %
        }
        \For{each random variable $X$ found within $\mathcal{V}$}
        \State{\color{blue1}// {This inner loop procedure is illustrated in Figure~\ref{gc-fig:ltrip} (a)}}
        \State $\mathcal{V}_{X}$ := set of clusters in $\mathcal{V}$ containing ${X}$
        \State{$\mathcal{W}_{X} := \text{Connection-Weights}(\mathcal{V}_X)$}
        \State{\color{blue1}// {Add $X$ to the appropriate sepsets}}
        \State $\mathcal{P}_{X}$ := max spanning tree over $\mathcal{V}_{X}$ using weights $\mathcal{W}_{X}$ \label{alg:ltrip-spanningtree}
        \For{each edge $(i,j)$ in $\mathcal{P}_{X}$}
        \If{sepset $\mathbf{S}_{i,j}$ already exists in $\mathcal{S}$}
        \State{$\mathbf{S}_{i,j}\mkern1mu$.\,insert$(X)$}
        \Else
        \State{$\mathcal{S}\mkern1mu$.\,insert$(\mkern2mu \mathbf{S}_{i,j} \mkern-3.5mu = \mkern-3.5mu \{ {X} \} \mkern2mu )$}
        \EndIf
        \EndFor
        \EndFor\vspace{-0.1em}
        \State{$\mathcal{T}:=$ cluster graph of $\mathcal{V}$ connected with sepsets $\mathcal{S}$}\vspace{-0.1em}
        \NoNumber{} %
        \Function{}{}\hspace{-0.1em} Connection-Weights\hspace{0.1em}($\mathcal{V}_X$)
        \State $\mathcal{W}_{X}$ := $\{w_{i,j} = \left|\mathbf{C}_i \cap \mathbf{C}_j \right|$ for $ \mathbf{C}_i, \mathbf{C}_j \in \mathcal{} \mathcal{V}_X, \ i \neq j \}$
        \State{\color{blue1}// {Emphasise nodes strongly connected to multiple %
                nodes}}
        \State $m := \text{max}(\mathcal{W}_{X})$
        \For{$i$}
        \State{\color{blue1}// {Number of maximal edges on this node}}
        \State{$t_i$ := %
            number of adjacent nodes $j$ for which $w_{i,j} = m$
        }
        \State{\color{blue1}// {Add to each edge touching this node}}
        \For{j}
        \State{$w_{i,j}$ += $t_i$ }
        \EndFor
        \EndFor
        \State \Return $\mathcal{W}_{X}$
        \EndFunction
    \end{algorithmic}
    \begin{description} \setlength{\itemsep}{1pt} \setlength{\parsep}{0pt}
    \item[Line \ref{alg:ltrip-spanningtree}] For constructing a maximum spanning tree, see the Prim-Jarn\'{i}k algorithm~\cite{prim}.
    \end{description}
\end{algorithm}


Note that other (unexplored) alternatives are possible for the ``Connection-Weights'' function in the algorithm. In particular, it would be interesting to evaluate
information-theoretic considerations as criteria.

\section{Modelling Sudoku via PGMs} \label{gc-sec:sudoku}
\begin{figure}[h]
  \centering    
  \includegraphics[width=0.85\columnwidth]{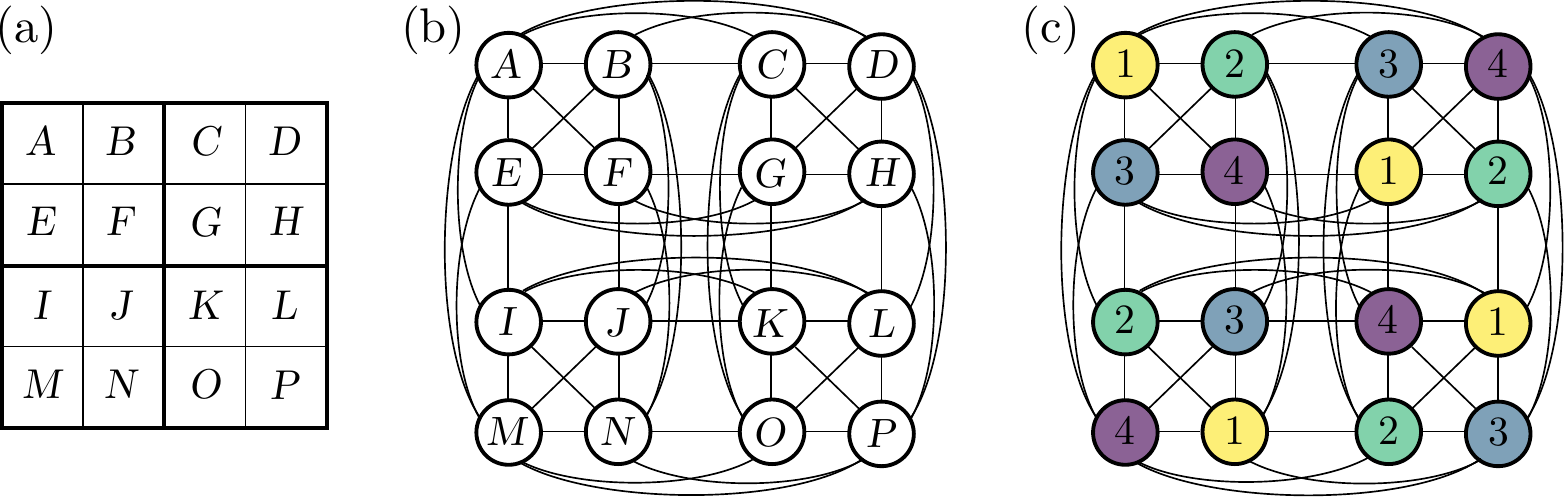}
  \caption{ (a) An example of a $4\times 4$ scaled Sudoku grid, with
    (b) its colouring graph, and (c) a non-unique colouring solution.
  }\label{gc-sudoku}
\end{figure}

The Sudoku puzzle is a well-known example of a graph colouring
problem. A player is required to label a $9 \times 9$ grid using the
integers ``1'' to ``9'', such that 27 selected regions have no
repeated entries. These regions are the nine rows, nine columns, and
nine non-overlapping $3 \times 3$ sub-grids of the puzzle. Each label
is to appear exactly once in each region.  Multiple solutions are possible if a Sudoku puzzle is under-constrained, i.e.\ too few of the values are known beforehand. A well-defined puzzle should have
only one unique solution.  We illustrate these constraints with a
scaled-down $4\times4$ Sudoku (with $2\times 2$ non-overlapping
sub-grids) in Figure~\ref{gc-sudoku}~(a).

We use the Sudoku puzzle as a proxy for testing graph colouring via
PGMs since this is a well-known puzzle with many freely available
examples. However, it should be kept in mind that solving Sudoku
puzzles per se is \emph{not} a primary objective of this chapter, and in Chapter~\ref{csp-chap}, we explore constraint satisfaction solvers in more detail. We now show how to construct a PGM for a Sudoku puzzle by following the same approach described for the four-colour map problem.

\subsection{Probabilistic representation}

For the graph colouring and probabilistic representation of the Sudoku
puzzle, each grid entry is taken as a node, and all nodes that are
prohibited from sharing the same label are connected with edges, as seen
in Figure~\ref{gc-sudoku}~(b). It is apparent from the graph that each of
the Sudoku's ``no-repeat regions'' is also a maximal clique within
the colouring graph.

The probabilistic representation for the scaled-down $4\times 4$
Sudoku is, therefore, $16$ random variables $A$ to $P$, each
representing a cell within the puzzle. The factors of the system are
set up according to the $12$ cliques present in the colouring graph.
Three examples of these factors, a row constraint, a column constraint, 
and a sub-grid constraint, are respectively $\{A, B, C, D\}$, $\{A, E,
I, M\}$, and $\{A, B, E, F\}$. The entries for the discrete table of
$\{A, B, C, D\}$ are precisely the same as those of
Table~\ref{gc-tab:probtable}. The proper $9{\times}9$-sized Sudoku puzzle used in our experiments is set up in exactly the same manner as the scaled-down version but using $27$ cliques, each of size nine.

Also note, in the case of Sudoku puzzles, some variables are already assigned a value. To integrate this into the system, we formally ``observe'' those variables. There are
various ways to deal with this, one of which is to purge all the
discrete distribution states not in agreement with the
observations. Following this, the variables can be purged from all
factor scopes altogether.

\subsection{Graph structure for the PGM}

\begin{figure}[h]
  \centering
  \includegraphics[width=0.85\columnwidth]{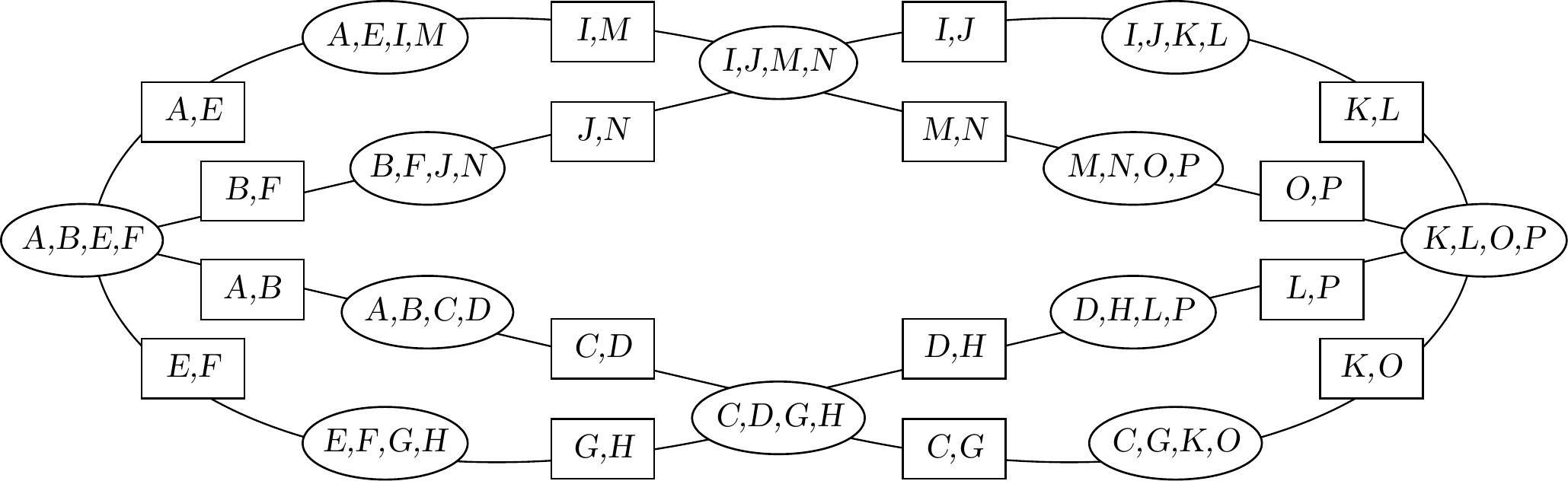}
  \caption{A cluster graph construction for the $4\times 4$ Sudoku clusters.
  }\label{gc-fig:sudoku_construction}
\end{figure}

We have shown how to parametrise the Sudoku puzzle as a colouring graph and how to parametrise this graph probabilistically. This allows capturing the relationships between the
variables of the system via discrete probability distributions. The
next step is to link the factors into a graph structure. We outlined
factor graph construction in Section~\ref{gc-sec:factorgraphs} and cluster graph construction via LTRIP in Section~\ref{gc-sec:ltrip}. Finally, we apply these two
construction methods directly to the Sudoku clusters, thereby creating
structures such as the cluster graph of
Figure~\ref{gc-fig:sudoku_construction}.

\subsection{Message passing approach}\label{gc-sec:messagepassing}

For a full discussion on belief propagation, see Sections~\ref{sec:belief-propegation} and~\ref{sec:loopy-belief-propagation}. The specific design choices for our PGM implementation are as follows:
\begin{itemize}
\item For the inference procedure, we used the belief \emph{update} procedure, also known as the Lauritzen-Spiegelhalter
algorithm~\cite{lauritzen1988local}.
\item The convergence of the system and the message passing schedule
  are determined according to Kullback-Leibler divergence between the
  newest and immediately preceding sepset beliefs.
\item Max-normalisation and max-marginalisation are used in order to
  find the maximum posterior solution over the system.
\item All  discrete distributions support sparse representations in order to make efficient use of memory and processing resources.
\end{itemize}

\section{Experimental investigation}\label{gc-sec:experiments}

As stated earlier, factor graphs are the dominant PGM graph structure
encountered in the literature. However, this seems like a compromise since
cluster graphs have traits that should enable superior performance. This section investigates the efficiency of cluster graphs compared
to factor graphs by using Sudoku puzzles as test cases.

\subsection{Databases used}

For our experiments, we constructed test examples from two sources,
(a) 50 $9 \times 9 $ Sudoku puzzles ranging in difficulty taken from
Project Euler~\cite{hughes2012problem}, and (b) the ``95 hardest
Sudokus sorted by rating'' taken from Sterten~\cite{sterten}. All
these Sudoku problems are well-defined (with unique solutions), and their
solutions are available for verification.

\subsection{Purpose of experiment}

The goal of our experiments is to investigate both the accuracy and efficiency of cluster graphs compared to factor graphs. We hypothesise that properly connected cluster graphs, as
constructed with the LTRIP algorithm, will perform better during
loopy belief propagation than a factor graph constructed with the same
factors.

Mateescu~\cite{mateescu2010join} shows that inference behaviour
differs with factor complexities: a graph with large clusters is likely
to be computationally more demanding than a graph with smaller clusters
(when properly constructed from the same problem). However, the posterior
distribution is likely to be more precise. We, therefore, want to also
test the performance of cluster graphs compared to factor graphs over
a range of cluster sizes.

\subsection{Design and configuration of the experiment}

Our approach is to set up Sudoku tests with both factor graphs and
cluster graphs using the same initial clusters. With regard to setting
up the PGMs, we follow the construction methodology outlined in
Section~\ref{gc-sec:sudoku}.

In order to generate graphs with smaller cluster sizes, we strike a
balance between clusters of size two, using every adjacent pair of
nodes within the colouring graph as described in
Section~\ref{gc-sec:gcwithpgms} and using the maximal
cliques within the graph, also described in that section. We do so by
generating $M$-sized clusters from an $N$-sized clique (where $M
\leq N$). We split the cliques by sampling all $M$-combination of
variables from the $N$ variable clique and keeping only a subset of
the samples, such that every pair of adjacent nodes from the clique is
represented at least once within one of the samples.

For experiments using the Project Euler database, we construct Sudoku
PGMs with cluster sizes of three, five, seven, and nine variables in
this manner. This results in graphs of $486$, $189$, $108$, and $27$
clusters, respectively. We compare the run-time efficiency and solution
accuracy for both factor and cluster graphs constructed from
the same set of clusters.

On the much harder Sterten database, PGMs based on clusters with less than nine variables were wildly inaccurate. We, therefore, limit those experiments to only clusters with nine variables.

\subsection{Results and interpretation}\label{gc-sec:resultsandinterpretation}

\begin{figure}[h]
	\centering
	\includegraphics[width=\textwidth]{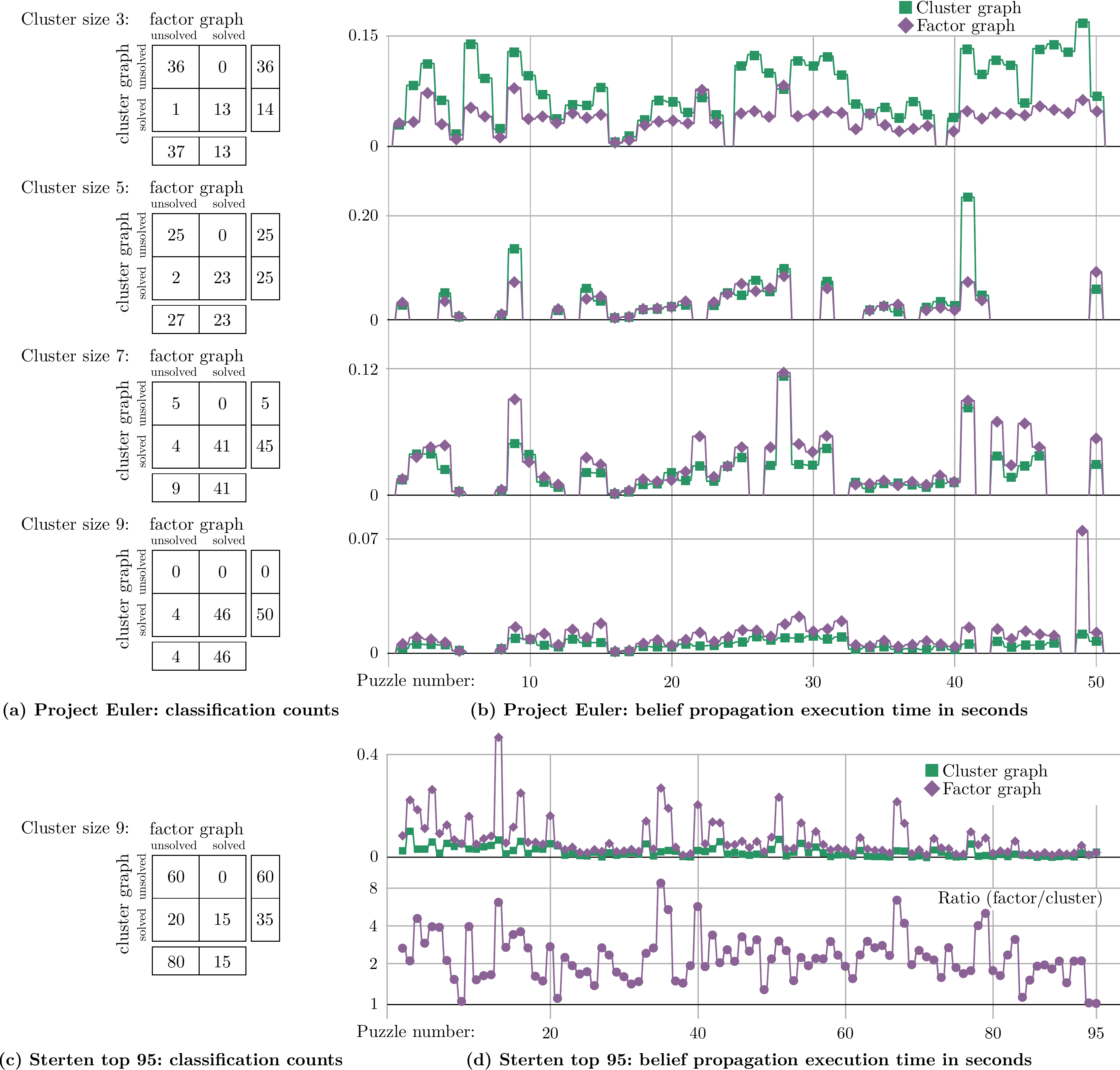}
	\caption{
The results of our test cases.  Note that whenever a cluster graph failed to obtain a valid solution, the corresponding factor graph also failed.  Also, note that (b) includes only points where the factor graph and cluster graph posteriors are equivalent, whereas (d) includes all results (otherwise the plot would mostly be empty).
	}\label{gc-fig:results}
\end{figure}

The results are reported in Figure~\ref{gc-fig:results}.
Cluster graphs showed superior accuracy for all the available test
cases.  Note that from our results, whenever a cluster
graph failed to obtain a valid solution, the corresponding factor graph
also failed. However, it happened regularly that a cluster graph
succeeded where a factor graph failed, especially in more complex configurations.

In the case of small clusters, factor graphs are faster
than cluster graphs. This is unexpected since cluster graphs built from small clusters are getting closer to factor graphs in terms of sepset sizes. We expected the execution speed to also get
closer to each other in this case.

As the cluster sizes increase (especially when the problem domain becomes more difficult), the cluster graphs clearly outperform the factor graphs in terms of execution speed. Two explanations come
to mind. Firstly, with the larger sepset sizes, the cluster graph
needs to marginalise out fewer random variables when passing messages
over that sepset. Since marginalisation is one of the expensive
components in message passing, this should result in computational
savings. Secondly, the larger sepset sizes allow factors to pass
richer information to their neighbours. This speeds up the convergence
rate, once again resulting in computational savings.

\section{Future work}

The LTRIP algorithm is shown to produce well-constructed graphs. However,
the criteria for building the maximal spanning trees in each layer can
probably benefit from further refinement. In particular. we suspect that
taking the mutual information between factors into account might prove useful.

Our graph colouring parametrisation managed to solve certain Sudoku
puzzles successfully, as well as assigning colours to the four-colour
map problem. This is a good starting point for developing more advanced
techniques for solving graph colouring problems.

In this chapter, we evaluated our cluster graph approach on a limited
set of problems. We hope that the LTRIP algorithm will enhance
the popularity of these problems as well as other related problems.
This should provide evaluations from a richer set of conditions,
contributing to a better understanding of the merits of this
approach.

\vspace{1.5em} %
\section{Conclusion}

The objective of this study was to illustrate how graph colouring
problems can be formulated with PGMs, to provide a means for
constructing proper cluster graphs, and to compare the performance of these graphs to the ones prevalent in the current literature.

The main contribution of this chapter is certainly LTRIP, our proposed cluster
graph constructing algorithm. %
The cluster graphs produced by LTRIP show great promise compared
to the standard factor graph approach, as demonstrated by our experimental results.

\chapter{A cluster graph approach to land cover classification boosting}\label{lc-chap}

\definecolor{Pantone283}{RGB}{152,198,234}
\definecolor{Pantone542}{RGB}{100,160,200}
\definecolor{IvoryTUM}{RGB}{218,215,203}
\definecolor{OrangeTUM}{RGB}{227,114,34}
\definecolor{GreenTUM}{RGB}{162,173,0}

\newcolumntype{x}[1]{>{\centering\arraybackslash\hspace{0pt}}p{#1}}
\newcommand\Tstrut{\rule{0pt}{2.6ex}}         %
\newcommand\Bstrut{\rule[-0.9ex]{0pt}{0pt}}   %
\newcommand{\correction}[1]{#1}

\section*{Preface}
\renewcommand{\thefootnote}{\fnsymbol{footnote}}

In the previous chapter, \textit{Graph Colouring: Comparing Cluster Graphs to Factor Graphs}, we illustrated how to formulate graph colouring problems as PGMs. We also presented this formulation via examples such as Sudoku puzzles and the four-colour map problem (where a planar map's neighbouring regions are coloured differently using only four colours). Furthermore, we provide a general-purpose cluster graph construction algorithm named LTRIP. Our experimental results show that the cluster graphs produced by LTRIP have superior characteristics compared to factor graphs, both in terms of speed and accuracy. The publication presented here subsequently illustrates how these contributions are used to solve a region-based classification problem in cartography, similar to the four-colour map problem.

In this chapter, we present \textit{A Cluster Graph Approach to Land Cover Classification Boosting}~\cite{landcoverdata}\footnote[2]{Sections of this work have been published in:
\\
L. H. Hughes, S. Streicher, E. Chuprikova, and J. du Preez. ``A Cluster Graph Approach to Land Cover Classification Boosting.'' \textit{Data}, 4(1), 2019. ISSN 2306-5729.
}, a collaborative effort between the Electronic Engineering department of Stellenbosch University and the Department of Civil, Geo and Environmental Engineering of the Technical University of Munich. Author S.\ Streicher’s contribution to this publication is the formulation, design, and execution of the PGMs used in this work; specifically (but not exclusively) Section~\ref{lc-sec:cluster-graph}, Section~\ref{lc-sec:lccpgm}, and the design of the potential functions in Section~\ref{lc-sec:priors}.

This publication sets out to solve a practical problem in cartography and classification, where multiple (and contradictory) classifications for pieces of land are combined and boosted to form a unified, more accurate classification.

The main contribution of this publication is in the form of a concrete application of the tools developed in Chapters~\ref{pgm-chapter} and~\ref{gc-chapter}; notably, approaching a region-based problem statement as a probabilistic reasoning problem and configuring the problem into a cluster graph using LTRIP. This approach produced a feasible, diverse, and spatially-consistent boosted land cover classification. In addition, a validation study reported an overall accuracy improvement compared to a reference dataset.

In conclusion, this paper serves as a practical demonstration of the power of cluster graphs. It establishes a substantial application of LTRIP for use in fields outside of Computer Science and Engineering. A solution to the land cover classification problem is made highly accessible through the use of LTRIP and the general PGM approach described in previous chapters.

\renewcommand{\thefootnote}{\arabic{footnote}}
\section*{Abstract}
Land cover classification is complex due to algorithmic errors, the spatio-temporal heterogeneity of Earth observation data, variation in data availability, and reference data quality.  This article proposes a probabilistic graphical model approach in the form of a cluster graph, to boost geospatial classifications and produce a more accurate and robust classification and uncertainty product. Cluster graphs can be used for reasoning about geospatial data such as land cover classifications by considering the effects of spatial distribution and inter-class dependencies in a computationally efficient manner. To assess the capabilities of our proposed cluster graph boosting approach, we apply it to the field of land cover classification. We make use of existing land cover products, GlobeLand30 and CORINE Land Cover, along with data from volunteered geographic information (VGI), namely OpenStreetMap (OSM), to generate a boosted land cover classification and the respective uncertainty estimates. Our approach combines qualitative and quantitative components by applying our probabilistic graphical model to data inputs as well as subjective expert judgments. Evaluating our approach on a test region in Garmisch-Partenkirchen, Germany, our approach boosted the overall land cover classification accuracy by 1.4\% compared to an independent reference land cover dataset. Our approach was shown to be robust and was able to produce a diverse, feasible, and spatially-consistent land cover classification in areas of incomplete and conflicting evidence. On an independent validation scene, we demonstrated that our cluster graph boosting approach was generalisable even when initialised with poor prior assumptions.

\section{Introduction}\label{sec:landcover-intro}

The rapid development of remote sensing techniques and data processing algorithms has led to the availability of large amounts of Earth observation data at regular intervals. However, the usefulness of this data is limited to a small community of experts. Therefore, practical data analysis methods and applications based on remote sensing data have become an active area of research, specifically thematic mapping and land cover classification~\cite{arino2012glcm, ChenPOK2015, Giri2013}. Land cover has been recognised as a key variable in environmental studies for deforestation, climate assessment, food and water security, and urban growth \cite{Ban2015, Chen2015}. 

Due to its high relevance, the demand for accurate and timely production of land cover classification has grown rapidly as governments push towards large-scale, and frequent monitoring of agricultural and urban environments \cite{martino2008new}. While a magnitude of approaches exists for creating land cover classifications, the quality, resolution, and reclassification periods for these vary drastically \cite{Foody2002, gislason2006random, pal2005support, Russwurm2018}. Additionally, many of these approaches make use of supervised learning where accurately labelled training data is required. This data is often manually annotated and thus suffers from human biases and varying accuracy. Furthermore, as the quality and quantity of this data directly affect the accuracy of the final classifier, the usability of these approaches is often limited to regions that depict similar features to the training dataset. 

Overall, land cover classification is an inherently nuanced problem and has a large scope for error. These errors can largely be attributed to automated classification algorithm errors, temporal and spatial heterogeneity of Earth observation data, variation in the availability and quality of reference data, and the need for human intervention in labelling. Therefore, trade-offs are required to produce \textit{good enough} land cover classifications. Thus, it becomes clear that no single land cover classification product can be produced to cover all use cases with a sufficient spatial resolution, coverage extent, accuracy, and granularity. Due to these reasons, there has been a growing interest in fusing different land cover classifications, including crowd-sourcing information, into a single more complete data product~\cite{Fritz2009, PerezHoyos2012, Gengler2016, Chen2017}. These methods are based on various data boosting approaches that play off each dataset's strengths to gain a more complete classification and minimise errors. While these approaches have seen various successes, spatially-weighted land cover prediction and the assessment of its related uncertainties have received limited attention. 

In this study, we address these challenges with four objectives: 
\begin{enumerate}
	\item to explore the potential of a \correction{probabilistic graphical model approach, using cluster graphs, towards producing a more accurate} land cover classification,
	\item to exploit the potential of expert knowledge for probabilistic reasoning in land cover classification,
	\item to perform uncertainty analysis on the outcome using the Shannon diversity index as a measure of uncertainty, and 
	\item to potentially contribute to OpenStreetMap data with missing land cover information.
\end{enumerate}

To this end, we propose an efficient approach using cluster graphs for boosting spatial heterogeneous data. We then investigate the effect of priors (in the form of expert knowledge) on the accuracy of the proposed inference process. Finally, we analyse the accuracy and uncertainty of the boosted classification. We demonstrate the applicability of our method, using three major datasets, namely GlobeLand30~\cite{Brovelli2015}, Coordination of Information on the Environment (CORINE)~\cite{aune2010corine}, and volunteered geographic information (VGI) based on OpenStreetMap (OSM)~\cite{haklay2008openstreetmap}.

The study proceeds as follows: First, in Section~\ref{lc-sec:related}, we review the relevant literature on land cover boosting algorithms and accuracy assessment methods. Section~\ref{lc-sec:method} describes our proposed approach and architectural design. In Section~\ref{lc-sec:results}, we report on results of the cluster graph approach for land cover boosting applied to data of Garmisch-Partenkirchen (Bavaria, Germany). Finally, we conclude with an overview of our findings.

\section{Related work}\label{lc-sec:related}

Over the years, many approaches have been proposed for land cover classification, of which the majority of these rely on satellite imagery and remote sensing techniques. Many of these approaches are based on simple linear methods and use hand-crafted features and simple classification schemes to create land cover maps. The most common of which is the maximum likelihood (ML) classifier. This approach has maintained steady popularity due to its availability and easy to use nature. However, as \citet{pal2003assessment} discuss, this approach does not provide as high-quality results as decision tree approaches, such as \citet{gislason2006random}.

\citet{gislason2006random} proposed using a random-forest (RF) classifier in the form of a classification and regression tree (CART) to extract land cover classifications from multi-source remote sensing data. While CART remains a commonly applied method for land cover classification (due to its low computational complexity and high interpretability), the produced classification tree does not generalise to vastly different environments. Based on this shortfall of RF approaches, \citet{pal2005support} proposed using support vector machines (SVMs) for remote sensing data classification. While they found that SVMs provided significantly better results, they also noted the sensitivity of the approach to datasets, parameter selection and class separability. These findings were reiterated in various studies into existing linear land cover classification approaches \cite{shao2012comparison,huang2002assessment}.

More recently, there has been a departure from linear approaches and into the domain of non-linear classifiers, with a specific focus on deep learning techniques. Deep learning has gained considerable popularity due to the success of convolutional networks in image classification tasks. Most notably, the success of AlexNet in achieving state-of-the-art performance on the ImageNet classification task \cite{krizhevsky2012imagenet}. Based on the success of deep learning in conventional computer vision applications, remote sensing practitioners have turned their focus to exploiting these advancements for land cover classification \cite{marmanis2016deep,kussul2017deep,castelluccio2015land}. Notably, \citet{castelluccio2015land} proposed using convolutional neural networks (CNNs) for land cover classification. Following this CNN approach, \citet{marmanis2016deep} used existing CNNs, which were pretrained using ImageNet data, and applied these networks to remote sensing classification tasks. Following a different approach, \citet{Russwurm2018} proposed using temporal data and a long short-term memory (LSTM) model for learning land cover classifications from underlying phenological features.

While existing techniques have all shown success in producing land cover classifications, they still suffer from a wide array of drawbacks that limit their overall usability and usefulness in large-scale and generalised land cover mapping applications \cite{Foody2002}. For this reason, several studies have adopted data boosting approaches that combine the relative strengths of various classifiers to produce an improved land cover classification.

\citet{Chen2017} suggested an approach to create a high-quality land cover classification using data fusion based on data from the Landsat-8 Operational Land Imager (OLI), Moderate Resolution Imaging Spectroradiometer (MODIS), China Environment 1A series (HJ-1A), and Advanced Space-borne Thermal Emission and Reflection (ASTER) digital elevation model (DEM). While this approach yields more accurate land cover classification results, it relies on the fusion of raw data and, thus, does not exploit well known and trusted land cover classification schemes or existing archives of land cover data.

\citet{PEREZHOYOS201272} approached this problem using existing land cover datasets, such as CLC2006, CLC2000, MODIS, and GlobeCover, to create a synergistic land cover map of Europe. The thematic overlap was performed based on knowledge of the data quality and the formation of a common set of classes. Affinity scores between various classes in different datasets were then calculated, and a hybrid land cover classification was produced.

Following on from this approach, \citet{tahaclassifier} proposed the use of a classifier ensemble to improve the accuracy of land cover classification approaches. RapidEye remote sensing imagery was classified using numerous well-known land cover classification approaches, such as SVM, CART, and \correction{artificial neural networks}. The resulting classification maps were then combined using a bagging and boosting approach to generate a single boosted classification map.

The approaches towards boosting taken by \citet{PEREZHOYOS201272} and \citet{tahaclassifier} resulted in boosted land cover classifications, showing a small overall accuracy improvement but consistent class-wise classification accuracy improvement over the initial classifications. However, neither approach inherently exploits expert knowledge or VGI data to support the boosting process. These two sources can provide critical prior information to resolve ambiguities and sensitivities when combining existing land cover classification datasets. Furthermore, the analysis of the uncertainties present in the final land cover classifications has largely been overlooked, despite its importance in understanding and utilising land cover classifications at a large scale.

Various measures have been proposed to describe data uncertainty quantitatively. These include scalar values like probability, error percentage, distance (e.g. from ground truth), standard deviation \cite{Griethe2006}, and the Shannon diversity index, a quantitative estimator of complexity~\cite{OIK14714}. These measures have been used within the framework of various probabilistic approaches, such as Bayes networks, belief functions, interval sets, and fuzzy set theory \cite{Correa2009}. However, few of these approaches have been applied to land cover classification boosting and, thus, this field remains predominantly unexplored. For this reason, we take inspiration from literature based within the realm of probabilistic graphical models (namely cluster graphs), graph colouring, and image processing. Our proposed approach follows the same general assumptions as problems where neighbouring relationships play a large role, such as the four-colour map problem (see Section~\ref{gc-sec:fourcolorproblem}), image segmentation \cite{imageseg2010}, and image de-noising \cite{Bishop2006,pgmsapp2017}.

\section{Land cover classification boosting with PGMs}\label{lc-sec:method}

This section introduces the concept of a cluster graph approach and describes how to formulate our land cover classification boosting problem as a graphical model. Furthermore, we describe how to apply inference on this model to perform classification boosting for our land cover problem.
 
\subsection{Cluster graphs}\label{lc-sec:cluster-graph}
Probabilistic graphical models (PGMs) are a resourceful combination of graph theory and probabilistic inference techniques. A cluster graph is a type of PGM known for performing inference over problem spaces with many inter-dependencies and complex relationships. It can, therefore, be used to approach problems that are otherwise too difficult to define and solve algorithmically. In a general sense, cluster graphs can be seen as a method whereby a large system is broken down and clustered into smaller sections, such that these sections can be connected in a graph structure. The graph structure allows these smaller systems to communicate about their combined outcome and, thus, perform inference. Therefore, cluster graphs can be described as a tool to reason about large-scale probabilistic systems in a computationally efficient manner.

Cluster graphs provide a compact representation of a probabilistic space as the product of conditionally independent distributions referred to as factors. Each factor defines a probabilistic relationship over its associated cluster of variables. For discrete classification problems, such as our land cover classification problem, these factors will have potential functions related to prior beliefs or variable dependencies. In practice, these factors are built-up from any available knowledge and assumptions (i.e. educated guesses, or expert knowledge) about the variables and the relationships within the model.

As a means to inference, explicitly calculating the product of these factors is useful but typically not computationally feasible. A cluster graph rather connects factors into a graph structure with the factors as nodes and connections holding a set of variables, called a sepset. Information may be passed between factors through connecting sepsets in one of the many PGM inference techniques. Typically the factors of a cluster graph are initialised using the prior beliefs of the system. These beliefs are then updated by passing information about neighbouring sepset variables and factors until all the beliefs reach convergence.
This produces approximations of the marginal distributions of the system and, therefore, a solution to the problem at hand. 

A detailed discussion of cluster graph topologies is found in Section~\ref{gc-sec:topology}. The details of message passing in graphical models and determining convergence are discussed in Sections~\ref{sec:belief-propegation} and~\ref{sec:loopy-belief-propagation}.

\subsection{Proposed approach}
\label{lc-sec:lccpgm}
Given the situation where multiple, independent, non-agreeing classifications exist, we can combine these classifications to obtain a new, more accurate classification in an approach referred to as classification boosting. 

Many approaches of classification boosting exist and have varying degrees of success dependant on the problem. The most notable are naïve boosting, where the mode of the different classifications is selected as the output class, and ensemble methods, where an optimal combination of existing classifiers is learned to produce a new classification \cite{Ghimire2012}. To this end, we propose the use of cluster graphs to solve the classification boosting problem.

Cluster graphs have numerous benefits over existing approaches since they make defining variables and relationships easy and exhibit powerful inference abilities. Furthermore, unlike the naïve and ensemble methods of classification boosting, cluster graphs can extrapolate expert knowledge to reclassify regions that were probabilistically unlikely in the original classification. Therefore, it can be argued that cluster graphs are positioned somewhere between learning a new classifier and pure classification boosting. 

To model the land cover classification problem, we made the following assumptions about the nature of the land cover classification data we are using.
\begin{itemize}
\item The classifications are noisy observations that correlate to the true underlying class.
\item The underlying land cover map can be sufficiently divided into small squares, similar to pixels, with the observations sub-sampled to correspond to these locations.
\item Adapting Tobler's first law of geography~\cite{Tobler1970}, locations in close proximity have a higher likelihood to be of the same class than locations \correction{further away.}
\end{itemize}

\begin{figure}[h!]
  \centering
     \includegraphics[scale=1.27]{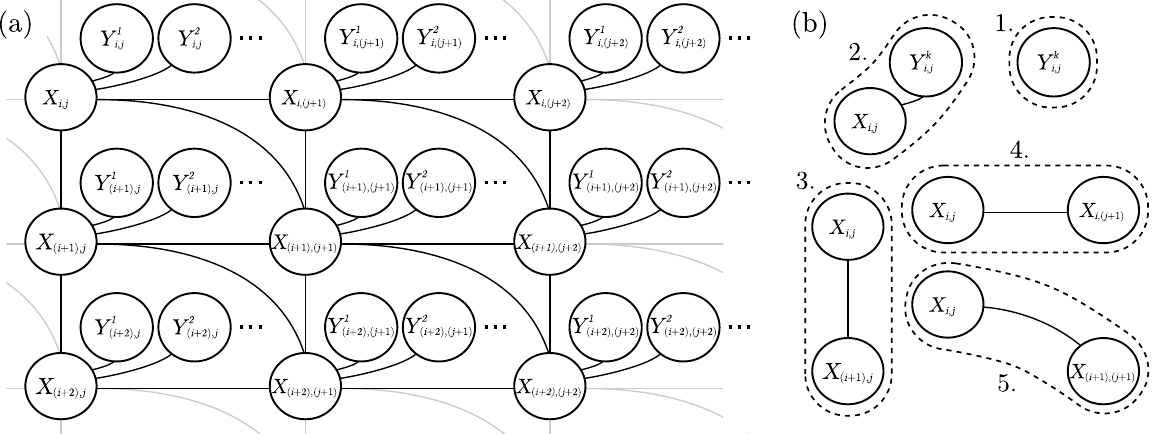}
  \caption{Graph (a) represents our relationship model for the land cover problem with nodes $X_{i,j}$ as the underlying classes and nodes $Y^k_{i,j}$ as observations taken from various classification approaches. The groupings in (b) represent our choice of factors as described in Table~\ref{lc-tab.factors}.}\label{lc-fig:undir_graph}
\end{figure}

We simplified these assumptions into the following relationship model. First, we split our underlying map into $N\times M$ pixels, and assign a random variable $X_{i,j}$ to each pixel to represent the underlying class. Then, for data taken from $K$ different approaches (in our case, $K$ different land cover classification maps), we assign the variable $Y^k_{i,j}$ as an observation correlating to $X_{i,j}$ by sampling from the classification $k$ at pixel $(i,j)$. We also assign a relationship between each $Y^k_{i,j}$ and its associated state $X_{i,j}$. Finally, we assign a relationship between all neighbouring pixels of $X_{i,j}$ to enforce Tobler's first law of geography. These variables and relationships are illustrated in Figure~\ref{lc-fig:undir_graph}(a).

For a PGM formulation of this model, we choose the factors in a manner that would capture the relationships of the model and can easily be initialised from the land cover data. \correction{This choice is outlined in Figure~\ref{lc-fig:undir_graph}(b) and Table~\ref{lc-tab.factors}, along with a description of the purpose of each factor in our model.}
\begin{table}[h!]
    \centering
    \begin{tabular}{p{0.02\linewidth} x{0.2\linewidth} x{0.2\linewidth} p{0.47\linewidth}}
    ~& Num. of factors & Factor variables & Purpose \\  \toprule
    1.& $N M K$   &   $\{Y^k_{i,j}\}$               & Capture the different classifications for each variable \\
    2.& $N M K$   &   $\{Y^k_{i,j}, X_{i,j}\}$      & Relationship between observations and underlying classes\\
    3.& $(N-1) M$       &   $\{X_{i,j}, X_{(i+1),j}\}$    & Relationship between southward neighbours \\
    4.& $N (M-1)$       &   $\{X_{i,j}, X_{i,(j+1)}\}$    & Relationship between eastward neighbours \\
    5.& $(N-1) (M-1)$   &   $\{X_{i,j}, X_{(i+1),(j+1)}\}$  & Relationship between south-eastward neighbours \\
    \end{tabular}
    \\ 
    \caption{Factor setup for the land cover PGM}
    \label{lc-tab.factors}
\end{table}

Our factors are then initialised according to the land cover classifications and assumptions in the form of expert knowledge. The idea is simple: we assign a potential function to each factor according to an underlying relationship. More specifically, we set the \correction{observation variables} 
$Y^k_{i,j}$
according to the land cover classification $k$: for hard classifications, we have 
$P(Y^k_{i,j} \hspace{-0.2em}=\hspace{-0.2em} \correction{\text{classification}^k_{i,j}})\equalss 1$ 
and 
$P(Y^k_{i,j}\hspace{-0.2em}\neq\hspace{-0.2em} \correction{\text{classification}^k_{i,j}})\equalss 0$,
and for soft classifications, we make use of joint probability tables initialised by expert knowledge about the likelihood of class co-occurrence in the defined region, and the class-specific classification quality, see Table~\ref{lc-tab.joint}. Furthermore, the variables in the relational factors are defined as more likely to have the same outcome, such as Table~\ref{lc-tab.factors} factors 2 to 5.

\begin{table}[h!]
\centering
\begin{tabular}{l|ccc}
         & \multicolumn{1}{l}{$X_{i,j} =a$} & \multicolumn{1}{l}{$X_{i,j}=b$} & \multicolumn{1}{l}{$X_{i,j}=c$} \\ \hline
$X_{i,(j+1)} = a$ & $p(a,a)$                     & $p(a,b)$                     & $p(a,c)$                     \\ 
$X_{i,(j+1)} = b$ & $p(b,a)$                     & $p(b,b)$                     & $p(b,c)$                     \\ 
$X_{i,(j+1)} = c$ & $p(c,a)$                     & $p(c,b)$                     & $p(c,c)$                     \\ 
\end{tabular}
\caption{\correction{Prior probabilities table of $P(X_{i,j}, X_{i,(j+1)})$ for three classes $a$, $b$, and $c$, capturing expert knowledge. This is easily expandable to $N$ classes, and the same table is usually applied to define relationship priors between eastward, southward and south-eastward neighbours}}
\label{lc-tab.joint}
\end{table}

Using our defined and initialised factors, we (a) construct a cluster graph, (b) obtain a posterior distribution over this graph using PGM inference techniques and (c) extract the random variables $X_{i,j}$ from the posterior as the most likely land cover classifications. Thus completing the boosting process by creating the boosted \correction{land} cover classification using the $X_{i,j}$ variables.

A summarised overview of the construction, inference, and settings used in this chapter are presented below.

\begin{enumerate}
    \item Configuring a cluster graph is not a trivial task and requires some heuristics. For the construction of our cluster graphs, we use the LTRIP procedure described in Section~\ref{gc-sec:ltrip} and further summarised in Section~\ref{csp-sec:ltrip}.
    \item We used belief update, also known as the Lauritzen-Spiegelhalter algorithm \cite{lauritzen1988local}, to perform inference over the graph.
    \item The convergence of the system, as well as the scheduling of messages, is determined according to the Kullbach-Leibler divergence between the newest and immediately preceding sepset beliefs.
    \item The distribution over a variable $X_{i,j}$ is found by locating a factor containing the variable and marginalising to that variable, i.e.\ $P(X_{i,j}) = \sum_{Y_{i,j}}P(X_{i,j}, Y_{i,j})$.
\end{enumerate}

\begin{figure}[ht!]
  \centering
     \includegraphics[scale=1]{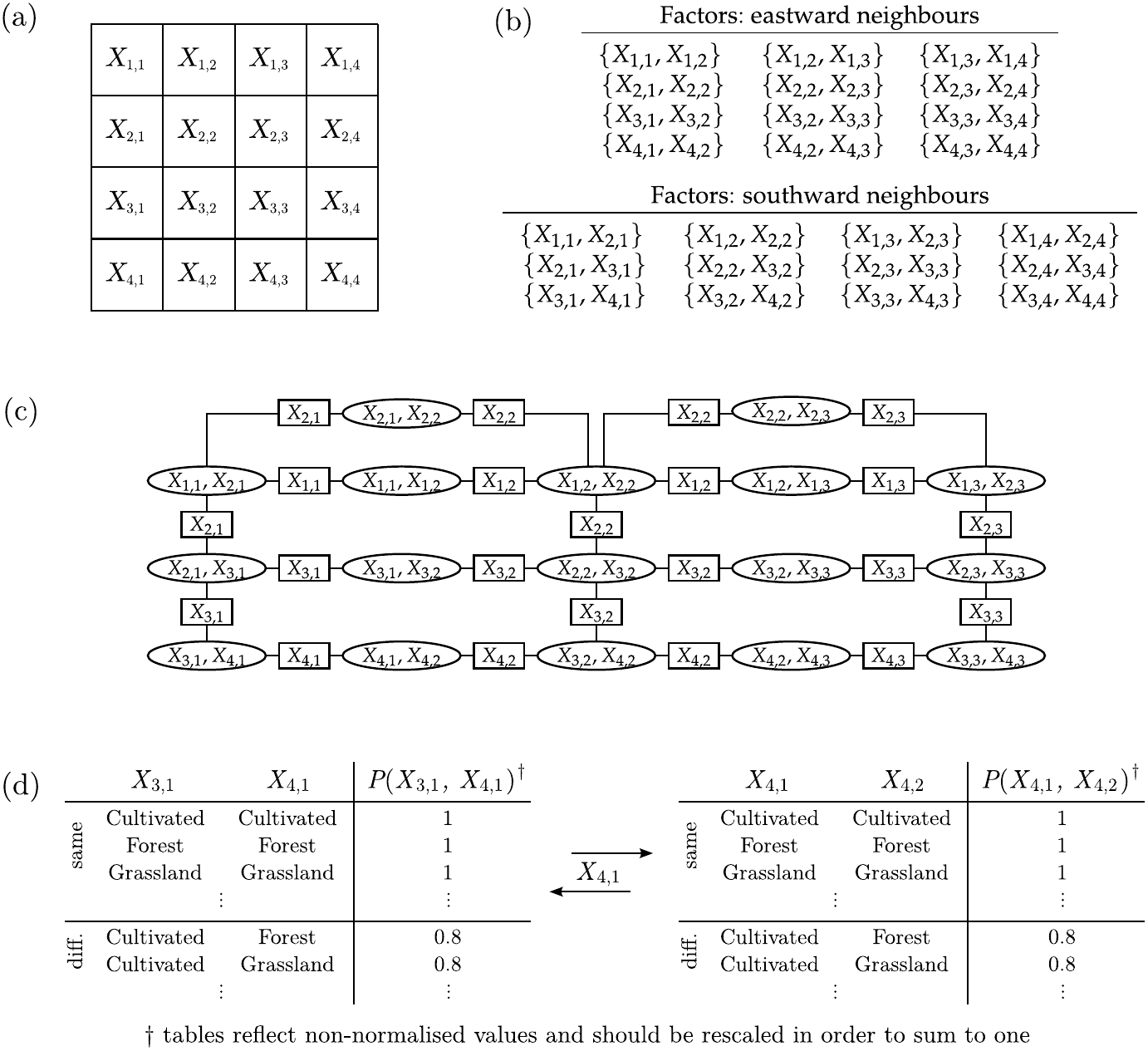}
  \caption{A simplified example of expressing our land cover classification problem as a cluster graph. In (a) we show a $4\times4$ location grid, in (b) we introduce factors describing neighbouring relationships to enforce Tobler's first law constraints, in (c) we show a cluster graph construction from these factors using LTRIP, and in (d) we highlight the use of discrete tables in message passing and the common variables through which information is passed. \correction{The joint probabilities defined here are merely for illustrative purposes, and should rather be specified by expert knowledge.}}\label{lc-fig:grid_to_graph}
\end{figure}

To better understand the proposed cluster graph architecture and inference process, see the reduced example presented in Figure~\ref{lc-fig:grid_to_graph}. For illustrative purposes we only consider the spatial factors (lines 3-5 of Table~\ref{lc-tab.factors}), reduce the location grid to $4\times 4$, and use a constant probability to define all inter-class relationships. In practice, the prior probabilities are to be specified by expert knowledge, and all factors need to be included  to boost the original land-cover datasets.

\begin{figure}[ht!]
  \centering
     \includegraphics[width=0.6\textwidth]{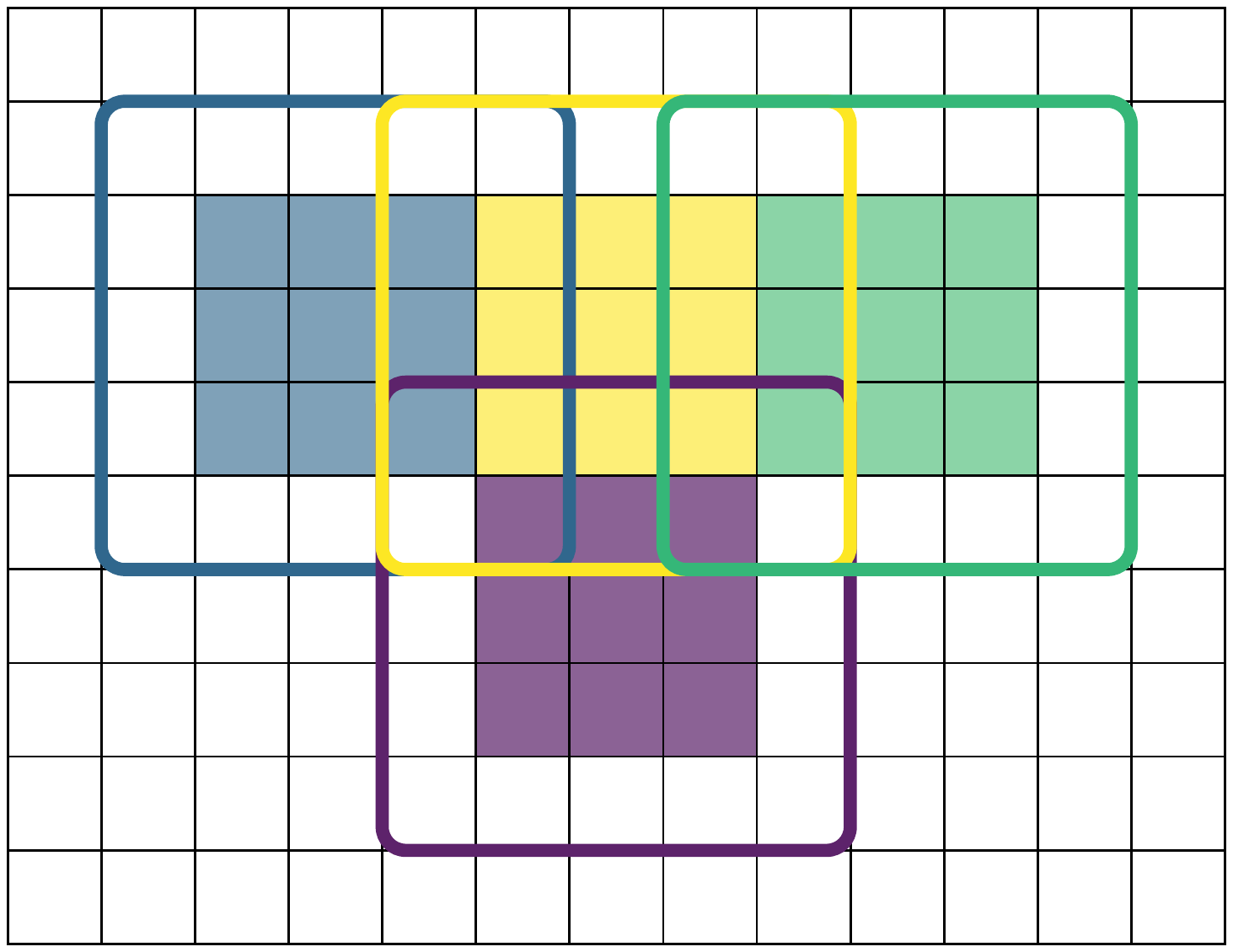}
  \caption{A simplified illustration of how a large region can be sub-divided and processed in a parallel manner while reducing the chance of artefacts along stitching boundaries.}
  \label{lc-fig:Stitching}
\end{figure}

As a final note, since the number of factors grows by order $NM$, it is useful to split the problem into smaller sections that can be processed in parallel. We found it safe to assume that regions sufficiently far apart have near-zero influence on each other. Thus, we segmented the region into non-overlapping sub-regions with overlapping boundaries to enforce smoothness along the edges. We then ran the cluster graph process on each of these sub-regions in parallel. Finally, we stitched the posteriors together along the subregion boundaries while discarding the overlapping regions (which may contain conflicting results). For a more intuitive understanding of this process, please refer to Figure~\ref{lc-fig:Stitching}.

\subsection{Datasets}
\label{lc-sec:dataset}
We selected two commonly used land cover datasets to assess our proposed approach, namely GlobeLand30~\cite{Brovelli2015} and CORINE Land Cover (CLC2006) for Germany~\cite{aune2010corine}. Additionally, we use land cover data derived from volunteered geographic information (VGI), specifically OpenStreetMap (OSM)~\cite{haklay2008openstreetmap}. The test region of Garmisch-Parten\-kir\-chen, Germany (101~224~km$^2$) was selected due to the availability of high-quality datasets and sufficient diversity in the distribution of land classes. The land cover classifications are temporally independent of one another, each derived from imagery captured over different periods of time. Since large temporal change events can lead to inconsistencies, this can be a problem for boosting applications. However, the assumption was made that little temporal change has occurred over the dataset acquisition period due to the specific nature of the test location. Thus, despite the temporal data heterogeneity, we can still evaluate our cluster graph approach for boosting land cover classifications.

\begin{figure}[h!]
\centering
\subbottom[Cultivated lands -- lands used for agricultural purposes, gardens, dry and irrigated farmlands\vspace{-0.3em}]{
\resizebox*{4cm}{!}{\includegraphics{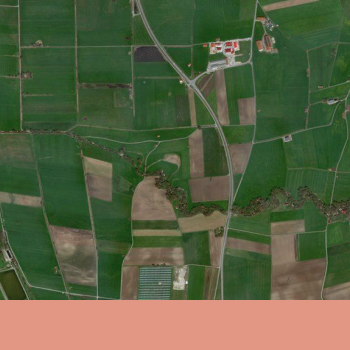}}} \hspace{0.1em}
\subbottom[Forest -- lands covered with woods by more than 30\%. \crudespace \crudespace  ]{\label{lc-fig:LCexampleAtkis}
\resizebox*{4cm}{!}{\includegraphics{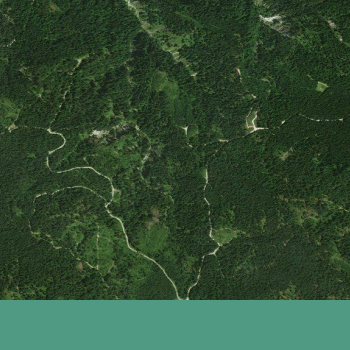}}} \hspace{0.1em}
\subbottom[Grassland -- lands covered with natural grass by more than 10\% \crudespace \crudespace ]{\label{lc-fig:LCexamplePMG}
\resizebox*{4cm}{!}{\includegraphics{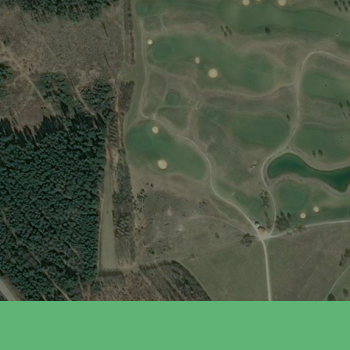}}}
\subbottom[Shrubland -- lands covered with shrubs by more than 30\%]{    
\resizebox*{4cm}{!}{\includegraphics{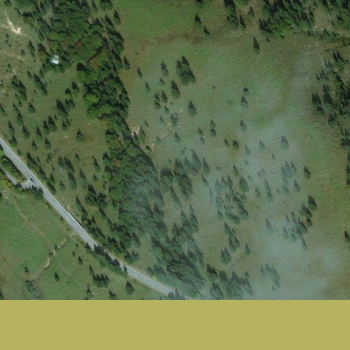}}} \hspace{0.1em}
\subbottom[Wetland \crudespace \crudespace \crudespace ]{
\resizebox*{4cm}{!}{\includegraphics{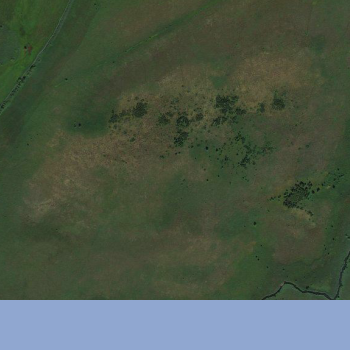}}} \hspace{0.1em}
\subbottom[Water -- rivers, lakes, natural and fish reservoir \crudespace]{\label{lc-fig:LCstudy2OSM}
\resizebox*{4cm}{!}{\includegraphics{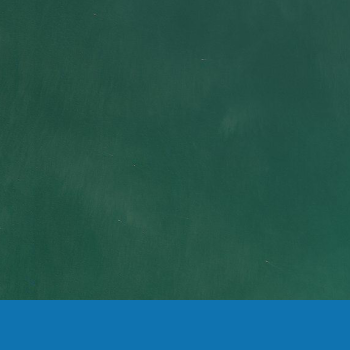}}}
\subbottom[Artificial surface -- lands modified by human activities]{\label{lc-fig:LCstudy2OSM}
\resizebox*{4cm}{!}{\includegraphics{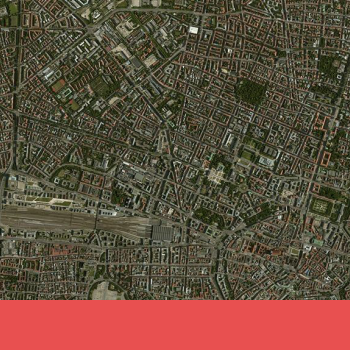}}} \hspace{0.1em}
\subbottom[Bareland -- lands with less than 10\% vegetation]{\label{lc-fig:LCstudy2OSM}
\resizebox*{4cm}{!}{\includegraphics{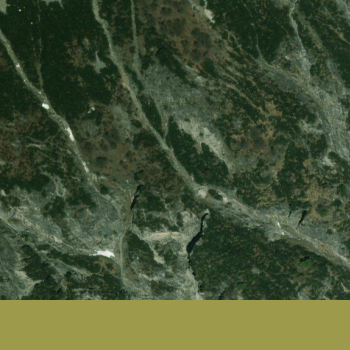}}} \hspace{0.1em}
\subbottom[Permanent snow and ice \crudespace\vspace{0.27em}]{\label{lc-fig:LCstudy2OSM}
\resizebox*{4cm}{!}{\includegraphics{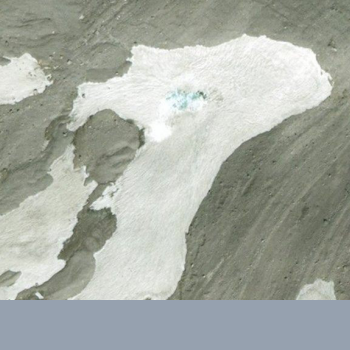}}}
\caption{Classes available for our chosen study area.} 
\label{lc-fig:LCexampleCLasses}
\end{figure}

GlobeLand30 is a global-scale land cover product of 30m resolution for two baseline years (2000 and 2010). For this study, we make use of the 2010 version of the dataset. This dataset comprises 10 major land cover classes: cultivated areas, forests, grassland, shrubland, wetland, water bodies, tundra, artificial surfaces, bare land, and permanent snow and ice. However, only nine of these classes are present in our study region, as depicted in Figure~\ref{lc-fig:LCexampleCLasses}, with examples of the visual appearance of each class.

Raw remote sensing imagery was selected to coincide with the local vegetation growth season to reduce the effects of cloud cover on the creation of the GlobeLand30 dataset \cite{ChenPOK2015}. Thus, the land cover classification was created based on a mosaic of suitable images with minimal cloud occlusions. According to the data provider, the land cover classifications of our study area were generated from a mosaic of images acquired on 31st August 2009. Previous studies indicate that the overall classification accuracy of GlobeLand30 can range from 46\%~\cite{Sun2015} and up to 80\%~\cite{Brovelli2015, ChenPOK2015, JokarArsanjani2016}. Therefore, the true accuracy of the dataset is heterogeneous.

Land cover mapping that focuses on EU countries is available through the Coordination of Information on the Environment (CORINE) program. The CORINE Land Cover 2006 dataset (CLC2006) for Germany follows common European-wide CORINE nomenclature that consists of 44 classes, where 37 classes are relevant to Germany, and 29 classes are relevant to the study area. We scaled down the class complexity to provide a consistent classification for all the data sources. This was achieved by reassigning the 44 CORINE land cover classes to the 10 classes corresponding to the GlobeLand30 classification. The details of the reclassification processes can be seen in Table~\ref{lc-tab:corine}.

\begin{table}[h!]
\centering
\caption{Reclassification of CLC2006 land cover classes relevant for the study area based on the GlobeLand30 scheme.}
\label{lc-tab:corine}
\begin{tabular}{l|cc}
{Land cover classes} & \multicolumn{1}{l}{CLC2006,
Pixel values} & \multicolumn{1}{l}{GlobeLand30,
Pixel values}  \\ \hline
Cultivated & 32 - 41                    & 10                                      \\ 
Forest & 42 - 45                     & 20                                         \\ 
Grassland & 46 - 47                   & 30  \\      
Shrubland & 48 – 49                   & 40  \\    
Wetland & 55 - 59                  & 50  \\   
Water bodies & 60 - 64                   & 60  \\ 
Tundra & -                   & 70  \\ 
Artificial surfaces & 21 – 31                   & 80  \\ 
Bareland & 50 – 53                   & 90  \\ 
Permanent ice and snow & 54                   & 100  \\ 
\end{tabular}
\end{table}

\begin{figure}[h!]
    \centering
    \includegraphics[width=0.8\textwidth]{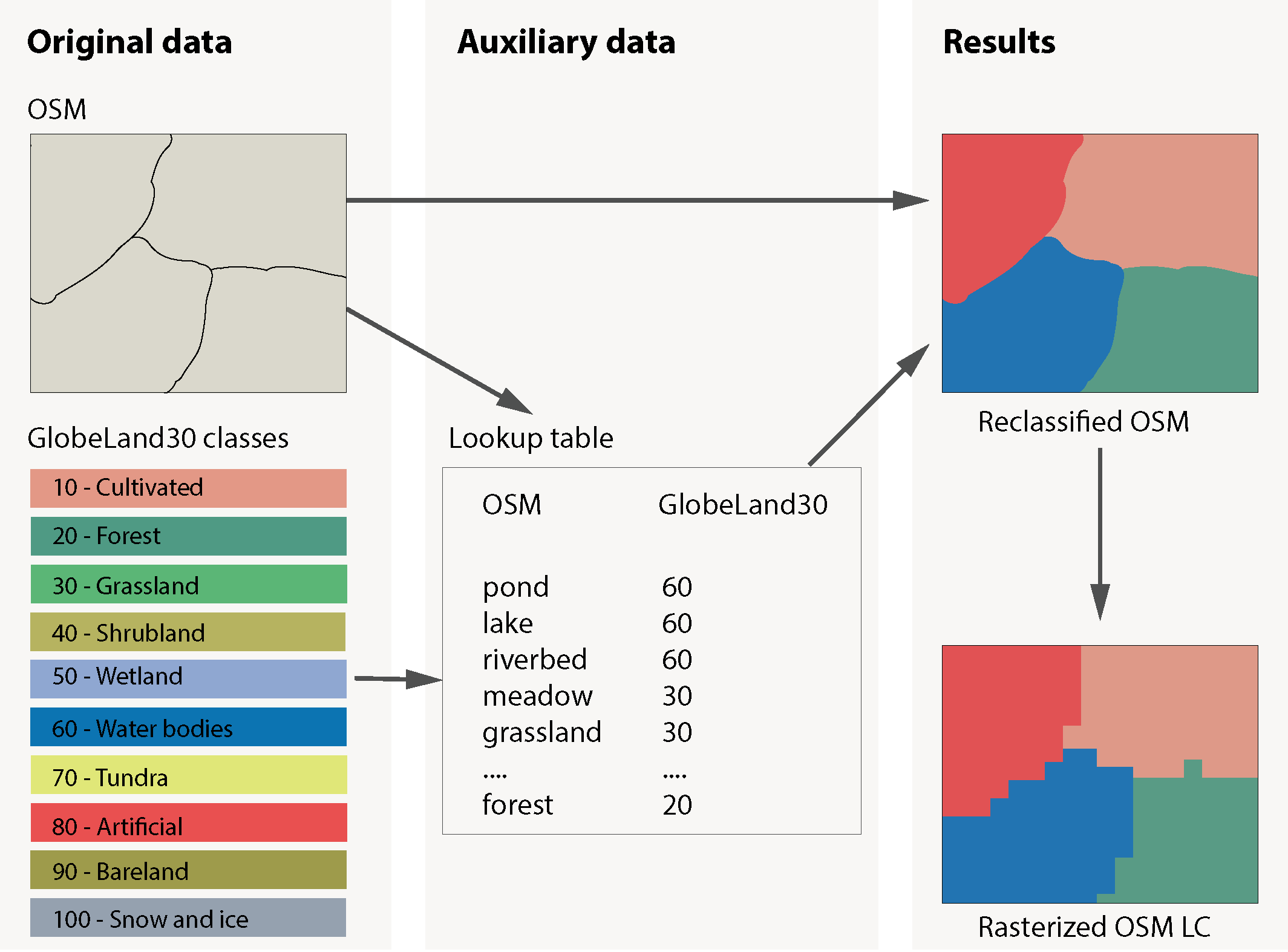}
    \caption{An overview of pre-processing steps for converting original OSM data into land cover. Source: \cite{Chuprikova}.}
    \label{lc-fig:osmWorkflow}
\end{figure}

Along with GlobeLand30 and CORINE (CLC2006), data from OSM plays an important role in this study as an auxiliary source. OSM is one of the most widespread and well-recognised VGI projects. Although the OSM data is not specifically tailored to the needs of land cover mapping and the OSM data and user community is very diverse, the data has valuable input for the land cover classification. \correction{In our research, we implemented a method suggested by \citet{Chuprikova} for deriving a land cover map from the OSM database, as shown in Figure~\ref{lc-fig:osmWorkflow}. To preserve the entire content of the database, we use a complete XML-encoded extract of the OSM database, representing our study area, instead of pre-processed Shapefiles  distributed by OSM data providers. The data pre-processing has been done in a combination of automatic and manual OSM tag annotations to the GlobeLand30 classification scheme.} For the derivation of the land cover map, a subset of the OSM tags, namely “amenity”, “building”, “historic”, “land use”, “leisure”, “natural”, “shop”, “tourism”, and “waterway” are considered. This mapping is only conducted for polygon features since point and line features do not provide immediate information about the coverage of an area. We define a mapping from the OSM attributes to the classes used in the GlobeLand30 classification scheme.

The data is then rasterised and resampled using a nearest neighbour approach to generate a land cover classification with the same resolution and spatial alignment as the GlobeLand30 dataset. The workflow for converting the OSM data into a land cover raster product is depicted in Figure~\ref{lc-fig:osmWorkflow}.

To accurately access our classification boosting approach, a ground truth dataset is required. Although it is intractable to obtain a truly accurate land cover classification, it is still practical to use a \textit{good enough} reference dataset. For our purpose, we chose a well maintained and frequently updated land cover classification dataset, the German national Amtliches Topographisch-Kartographisches Informations System (ATKIS). This dataset represents a Digital Landscape Model of scale 1:10,000 and 1:25,000 (Basis-DLM), and was provided by an official national cartographic authority (Bundesamt f{\"u}r Kartographie und Geod{\"a}sie). Our selection was further motivated by the reported use of this dataset as a  reference map by numerous other authors \cite{JokarArsanjani2016, Fan2013}.

\subsubsection*{Dataset preprocessing}
Pre-processing and harmonisation of the classes among these three heterogeneous data\-sets were performed  to simplify the construction of our cluster graphs. However, it should be noted that this step is not required and merely reduces the complexity in defining the inter-class relationships between the various datasets (by ensuring a common set of class labels exist for all the datasets).

The details of our pre-processing steps are outlined below:
\begin{enumerate}
\item All datasets are cropped to the municipal boundary of Garmisch-Partenkirchen.
\item The classes of OSM and CLC2006 are normalised to match the 10 classes specified by GlobeLand30.
\item The datasets were then rasterised with 30m pixel resolution and aligned to GlobeLand30 using the nearest neighbour resampling method as found in standard GIS software. (This ensured that each pixel was covering the same area of land. For OSM data, the process is more involved and is described in Figure~\ref{lc-fig:osmWorkflow}.)
\item Individual rasters were then sub-divided into sub-regions as per the explanation in Section~\ref{lc-sec:lccpgm}.
\end{enumerate}

\subsection{Definition of priors and parameters}
\label{lc-sec:priors}
In addition to the cluster graph implementation and dataset, our approach requires inter-class relationships, a per map confidence factor, and the parameters of our sub-regions and boundary sizes.

The inter-class relationships are defined based on expert knowledge and fundamental laws of geography. Firstly, classes that are likely to occur next to each other are assigned to a high probability. In contrast, classes unlikely to neighbour each other -- based on region, geography, and expert assumptions -- are assigned a low value. Lastly, the self occurrence probability of each class $p(n, n)$ is assigned the highest value to add dependence on Tobler's first law of geography. Due to the nature of inference on cluster graphs via a form of consensus, the prior beliefs of the system do not need to be exact probabilities but rather need to reflect the relative relationships between various classes. For instance, if $p(a, b) = p(a, c)$ and $p(a, b) >> p(a, d)$, it reflects that it is equally likely that class $b$ or $c$ could neighbour class $a$ and that it is significantly less likely that class $d$ would neighbour class $a$. The full potential function depicting the relationships used in our experiments is shown in Table~\ref{lc-tab:fulljoint}.

\begin{table}[ht!]
\centering
\caption{A potential function in the form of a discrete table for the inter-class relationships as defined by expert knowledge.}
\label{lc-tab:fulljoint}
\resizebox{\textwidth}{!}{%
\begin{tabular}{l|ccccccccc}
 & \multicolumn{1}{l}{cultivated} & \multicolumn{1}{l}{forest} & \multicolumn{1}{l}{grassland} & \multicolumn{1}{l}{shrubland} & \multicolumn{1}{l}{wetland} & \multicolumn{1}{l}{water} & \multicolumn{1}{l}{artificial} & \multicolumn{1}{l}{bareland} & \multicolumn{1}{l}{snow} \\ \hline
cultivated & 1 & 0.15 & 0.4 & 0.05 & 0.2 & 0.05 & 0.4 & 0.05 & 0.05 \\ %
forest & 0.15 & 1 & 0.05 & 0.4 & 0.2 & 0.05 & 0.15 & 0.05 & 0.05 \\ %
grassland & 0.4 & 0.05 & 1 & 0.2 & 0.15 & 0.15 & 0.05 & 0.05 & 0.05 \\ %
shrubland & 0.05 & 0.4 & 0.2 & 1 & 0.15 & 0.05 & 0.05 & 0.15 & 0.05 \\ %
wetland & 0.2 & 0.2 & 0.15 & 0.15 & 1 & 0.4 & 0.05 & 0.05 & 0.05 \\ %
water & 0.05 & 0.05 & 0.15 & 0.05 & 0.4 & 1 & 0.05 & 0.2 & 0.05 \\ %
artificial & 0.4 & 0.15 & 0.05 & 0.05 & 0.05 & 0.05 & 1 & 0.2 & 0.05 \\ %
bareland & 0.05 & 0.05 & 0.05 & 0.15 & 0.05 & 0.2 & 0.2 & 1 & 0.65 \\ %
snow & 0.05 & 0.05 & 0.05 & 0.05 & 0.05 & 0.05 & 0.05 & 0.65 & 1 \\ %
\end{tabular}%
}
\end{table}

In addition to introducing expert knowledge into the inference process through the inter-class table, we also include further expert knowledge in the form of a classification map confidence factor. This factor can weigh the confidence in each of the input datasets as a whole or in a class-wise manner. Also, the weighting factor can either be set by expert opinion or through more complex statistics.

To assess the effects of including VGI in the form of OSM data, we performed multiple experiments where the confidence in the OSM data was adjusted according to expert opinion. The confidence factors for the CLC2006 and GlobeLand30 datasets were kept constant to only assess the effect of VGI data which is often an incomplete and noisy source of land cover information.

The selected confidence factors are described in Table~\ref{lc-tab:lcconf}, and each of our experimental setups is detailed below:
\begin{enumerate}
\item \textbf{Scenario 1}: All land cover maps were assumed to be of equal quality, i.e.\\$P(Y^k_{i,j} \equalss \text{Classification}) = 1$.
\item \textbf{Scenario 2}: OSM data was assumed to be less accurate  overall, i.e.\\$P(Y^\text{OSM}_{i,j} \equalss \text{Classification}) = 0.7$.
\item \textbf{Scenario 3}: OSM data was excluded completely from the boosting process.
\item \textbf{Scenario 4}: OSM data is assumed to be less accurate overall, except for grassland. The classes are weighted as follows: overall OSM weighting 0.75, cultivated 0.7, wetland 0.6 and grassland 1.0.
\end{enumerate}

\begin{table}[ht!]
\centering
\caption{Land cover map confidence scores for adding prior information about dataset confidence, this data is captured by the observation factors $(Y_{i,j}^k, X_{i,j})$ in Table~\ref{lc-tab.factors}. Four scenarios were evaluated to determine the effects of various expert assumptions. *Scenario 4: The OSM layer confidence was not uniformly weighted per class, with cultivated=0.7, grassland=1, wetland=0.6 and remaining classes=0.75 }
\label{lc-tab:lcconf}
\begin{tabular}{l|cccc}
 & \multicolumn{1}{l}{Scenario 1} & \multicolumn{1}{l}{Scenario 2} & \multicolumn{1}{l}{Scenario 3} & \multicolumn{1}{l}{Scenario 4} \\ \hline
GlobeLand30 & 1 & 1 & 1 & 1 \\ %
CLC2006 & 1 & 1 & 1 & 1 \\ %
OSM & 1 & 0.7 & \texttimes & 0.75* \\ %
\end{tabular}%
\end{table}

Lastly, the sub-region and boundary sizes were defined such that each sub-region was square with a side length of $35px$ (1050m) and a boundary of $6px$ (180m). It was found that a total boundary width (left + right, or top + bottom), which is between 25\% and 50\% of the sub-region width or length, is appropriately large for the edge factors to have a negligible influence. In our case, this boundary was defined as $\frac{12px}{35px}\approx35\%$.

\section{Results}
\label{lc-sec:results}
\begin{figure}[h!]
  \centering
     \includegraphics[width=0.8\textwidth]{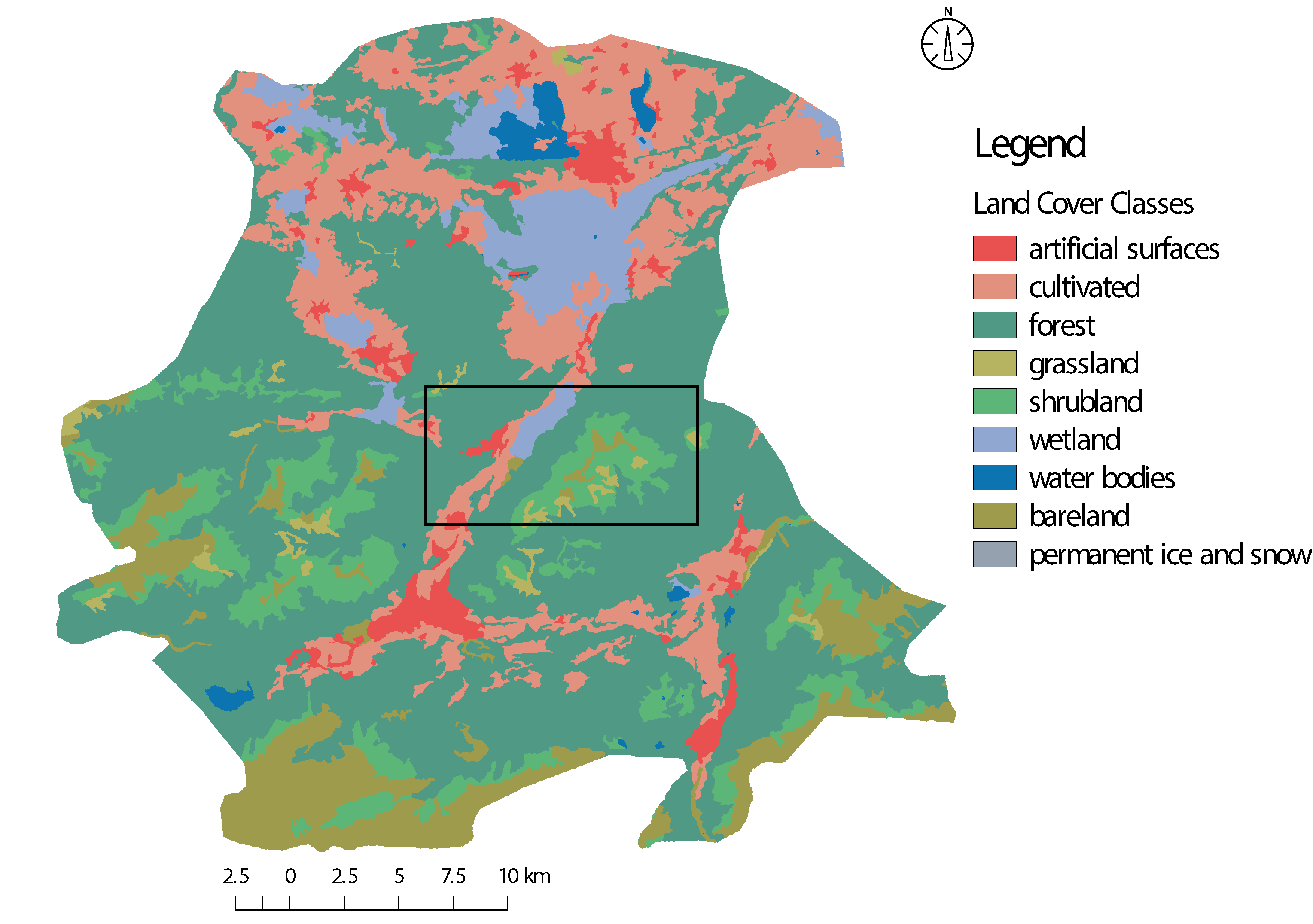}
  \caption{Overview map of the study area located in Garmisch-Partenkirchen, Germany. The land cover classifications include nine land cover classes based on the classification scheme adopted from GlobeLand30 (the tundra class is not present within the study area). The subset shows a selected area which is referred to during our more detailed discussions (i.e.\ Figure~\ref{lc-fig:LCmaps}).}
  \label{lc-fig:overviewMap}
\end{figure}

In this section, we report on the results obtained using our cluster graph land cover boosting approach. We assess the accuracy of the land cover maps produced and evaluate the uncertainty (in the form of the Shannon diversity index) extracted during the classification boosting process. To perform our assessments, we use the dataset defined in Section~\ref{lc-sec:dataset} over the study area of Garmisch-Partenkirchen, Germany (see Figure~\ref{lc-fig:overviewMap}). To asses thematic classification accuracy, we adopted the method of error matrix evaluation and provided accuracy metrics such as the overall accuracy, Kappa, and the class-wise balanced accuracy. 

The overall accuracy represents the proportion of correctly classified pixels to the reference map. The Kappa coefficient $\kappa \in [-1, 1]$ indicates how well the classification performed compared to randomly assigned values. In other words, the Kappa values represent an agreement between two classifications, where $\kappa < 0$ show no agreement, $0 \leq \kappa \leq 0.4$ represent a small degree of agreement, $0.4 < \kappa \leq 0.6$ represents a moderate agreement, $0.6  < \kappa \leq 0.8$ indicate significant agreement, and $0.8  < \kappa \leq  1$ show strong agreement. \correction{Furthermore, we use the class-wise balanced accuracy~\cite{mosley2013a} to evaluate the classification on a class-wise basis. The class-wise balanced accuracy represents correctly classified proportions for each class, which is essential as the classes are imbalanced in their distribution within the scene. This measure is favoured over the traditional consumer and producer accuracy as we are making a comparison to a reference dataset rather than to true ground-truth data. The class-wise balanced accuracy, therefore, represents both the consumer and producer accuracy as a single accuracy measure weighted according to the class distribution in the reference dataset.}

To quantitatively assess the accuracy of our boosted land cover maps, we compare our boosted land cover classifications against a reference land cover mapping. As a reference, we chose to use a national dataset called Amtliches Topographisch-Kartographisches Informations System (ATKIS), which represents a Digital Landscape Model with a scale of 1:250,000 (Basis-DLM) provided by an official cartographic authority (Bundesamt f{\"u}r Kartographie und Geod{\"a}sie). The remainder of the classification and uncertainty results are analysed qualitatively concerning the individual land cover classifications used as input to the factors of our proposed cluster graph formulation.

\subsection*{Effect of prior information}
We tested four different scenarios to assess the effects of expert knowledge and the inclusion of volunteered geographic information (VGI). In all the scenarios, the initial inter-class priors were assigned according to Table~\ref{lc-tab:fulljoint}. However, the overall confidence factor for each dataset was altered on a global and class level (as described in Table~\ref{lc-tab:lcconf} and Section~\ref{lc-sec:priors}).

\begin{figure}[h!]
\centering
\begin{tikzpicture}[scale=0.7]
  \centering
  \begin{axis}[
        ybar, axis on top,
        height=10cm, width=20cm,
        bar width=0.4cm,
        ymajorgrids, tick align=inside,
        major grid style={draw=white},
        enlarge y limits={value=.1,upper},
        ymin=0, ymax=100,
        axis x line*=bottom,
        axis y line*=right,
        y axis line style={opacity=0},
        tickwidth=0pt,
        enlarge x limits=true,
        legend style={
            at={(0.5,-0.2)},
            anchor=north,
            legend columns=-1,
            /tikz/every even column/.append style={column sep=0.5cm}
        },
        ylabel={Class-wise balanced accuracy},
        symbolic x coords={
           Cultivated, Forest, Grassland, Shrubland,
           Wetland,Water,
           Artificial,
          Bareland},
       xtick=data,
       nodes near coords={
        \pgfmathprintnumber[precision=0]{\pgfplotspointmeta}
       }
    ]
    \addplot [draw=none, fill=IvoryTUM] coordinates {
      (Cultivated, 89.45)
      (Forest, 85.18)
      (Grassland, 56.411)
      (Shrubland, 67.076)
      (Wetland, 89.933)
      (Water, 85.28)
      (Artificial, 83.272)
      (Bareland, 80.657)};
   \addplot [draw=none,fill=Pantone542] coordinates {
      (Cultivated, 90.59)
      (Forest, 86.01)
      (Grassland, 56.0776)
      (Shrubland, 68.179)
      (Wetland, 90.146)
      (Water, 86.675)
      (Artificial, 84.774)
      (Bareland, 82)};
   \addplot [draw=none, fill=GreenTUM] coordinates {
      (Cultivated, 90.09)
      (Forest, 85.51)
      (Grassland, 54.9555)
      (Shrubland, 66.273)
      (Wetland, 90.189)
      (Water, 83.351)
      (Artificial, 82.428)
      (Bareland, 81.939) };
   \addplot [draw=none, fill=OrangeTUM] coordinates {
      (Cultivated, 90.6)
      (Forest, 86.02)
      (Grassland, 56.2365)
      (Shrubland, 70.233)
      (Wetland, 90.114)
      (Water, 86.563)
      (Artificial, 84.313)
      (Bareland, 81.905)};
    \legend{Scenario 1,Scenario 2, Scenario 3, Scenario 4}
  \end{axis}
  \end{tikzpicture}
    \caption{The class-wise balanced accuracy for each of our defined scenarios when compared to the ATKIS dataset.}
    \label{lc-fig:compare}
\end{figure}
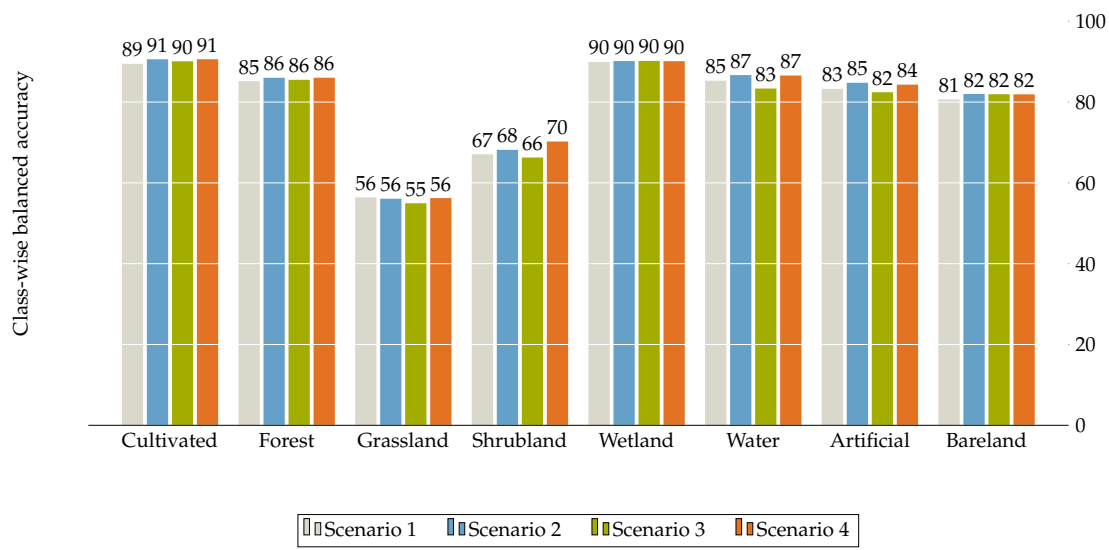

Using these expert beliefs, we performed inference over the test region. Furthermore, we compared the final boosted classification to the ATKIS dataset to determine the effects of various sets of expert knowledge and VGI on the final boosted classification. The results of these comparisons can be seen in Figure~\ref{lc-fig:compare} with a more detailed analysis of overall accuracy and Kappa in Table~\ref{lc-tab.accuracy}.

\begin{table}[h!]
    \centering
    \begin{tabular}{p{0.15\linewidth}p{0.11\linewidth}p{0.11\linewidth}p{0.11\linewidth}p{0.11\linewidth}}
        
        Estimate &  Scenario 1 & Scenario 2 & Scenario 3 & Scenario 4\\  \toprule
        Accuracy &  0.7741 & 0.7798 & 0.7648 & 0.7828 \\
        Kappa  &    0.6581 & 0.6704 & 0.6506 & 0.6748\\
        
    \end{tabular}
    \\ 
    \caption{Overall accuracy and Kappa results of various test scenarios with respect to ATKIS reference dataset}
    \label{lc-tab.accuracy}
\end{table}

\subsection*{Qualitative assessment of scenarios}
\begin{figure}[h!]
\centering
\subbottom[GlobeLand30 2010]{
\resizebox*{5cm}{!}{\includegraphics{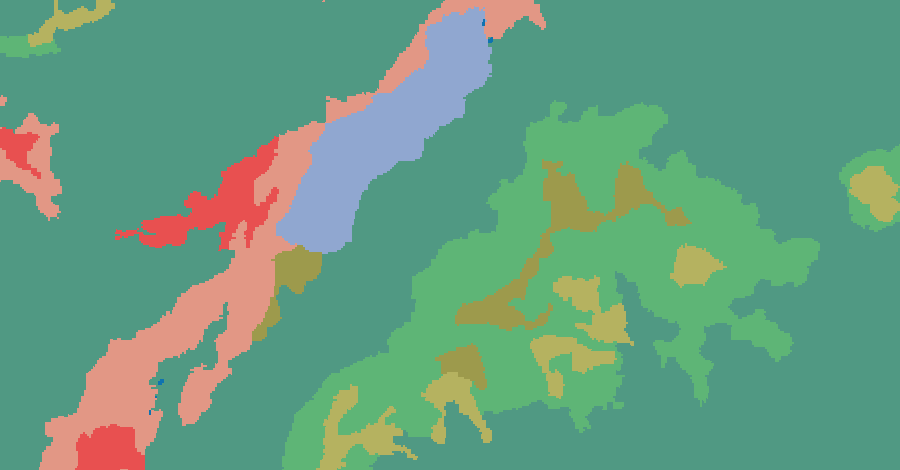}}\label{lc-fig:GLC30_fig6}}\hspace{1pt}
\subbottom[CLC2006]{
\resizebox*{5cm}{!}{\includegraphics{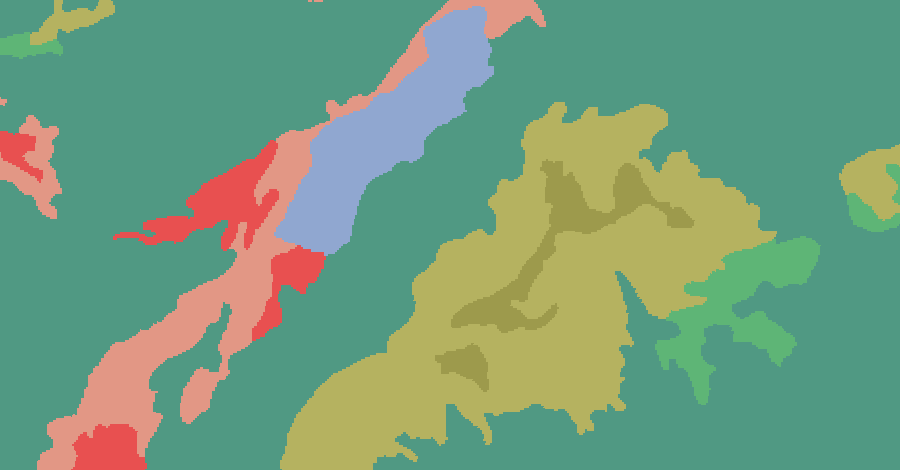}}\label{lc-fig:CORINE_fig6}}\hspace{1pt}
\subbottom[Land cover derived from OSM]{
\resizebox*{5cm}{!}{\includegraphics{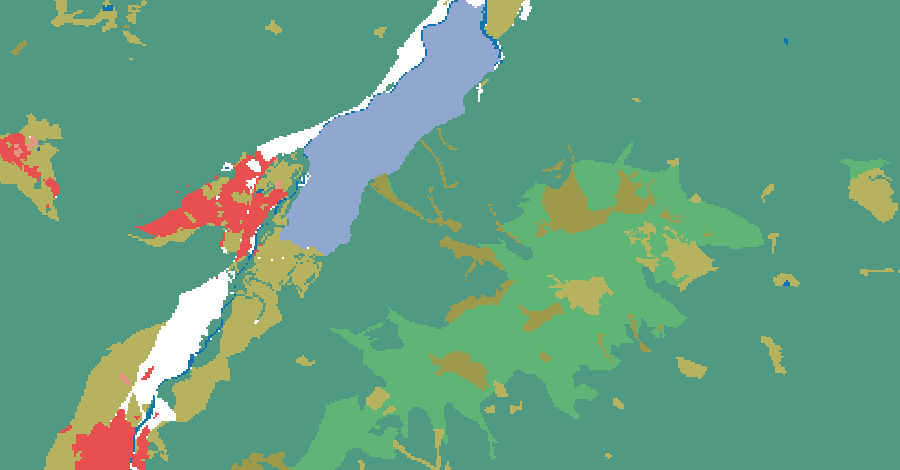}}\label{lc-fig:OSM_fig6}}\hspace{1pt}
\subbottom[Scenario 1 -- no weighting factor]{
\resizebox*{5cm}{!}{\includegraphics{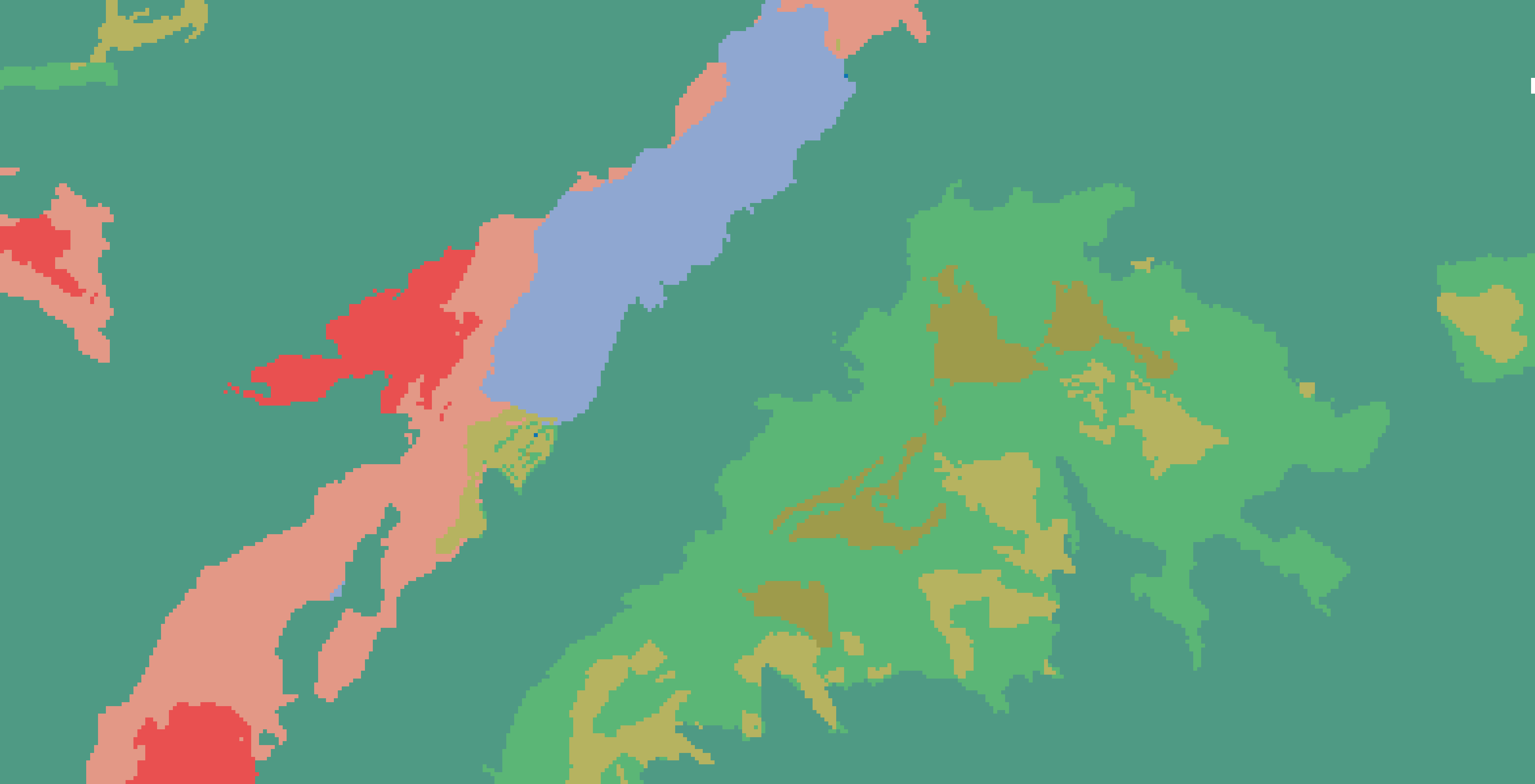}}\label{lc-fig:scenario1}}\hspace{1pt}
\subbottom[Scenario 2 -- OSM weighting factor of 0.7]{\label{lc-fig:scenario2}
\resizebox*{5cm}{!}{\includegraphics{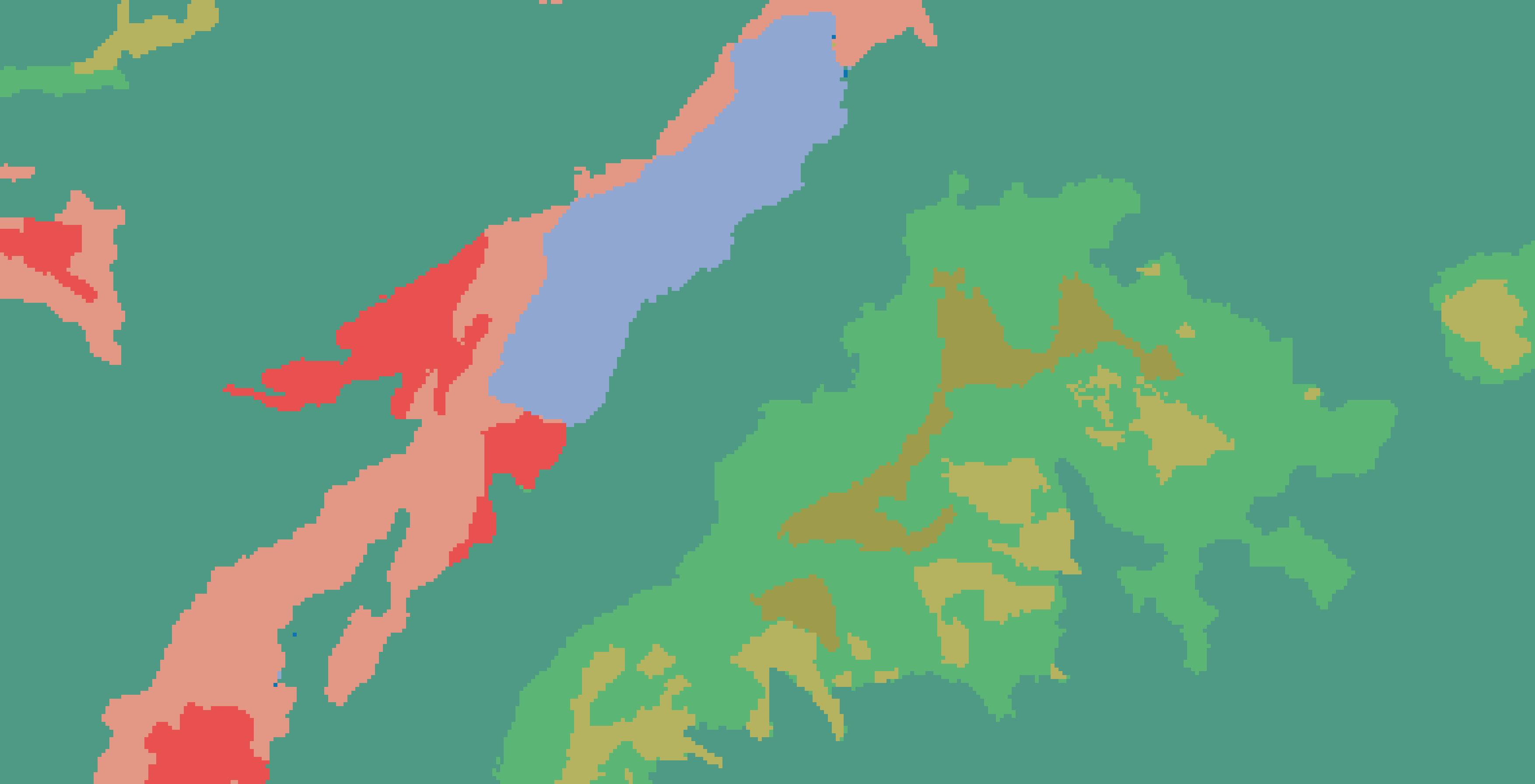}}}\hspace{1pt}
\subbottom[Scenario 3 -- without OSM \crudespace \crudespace]{\label{lc-fig:scenario3} 
\resizebox*{5cm}{!}{\includegraphics{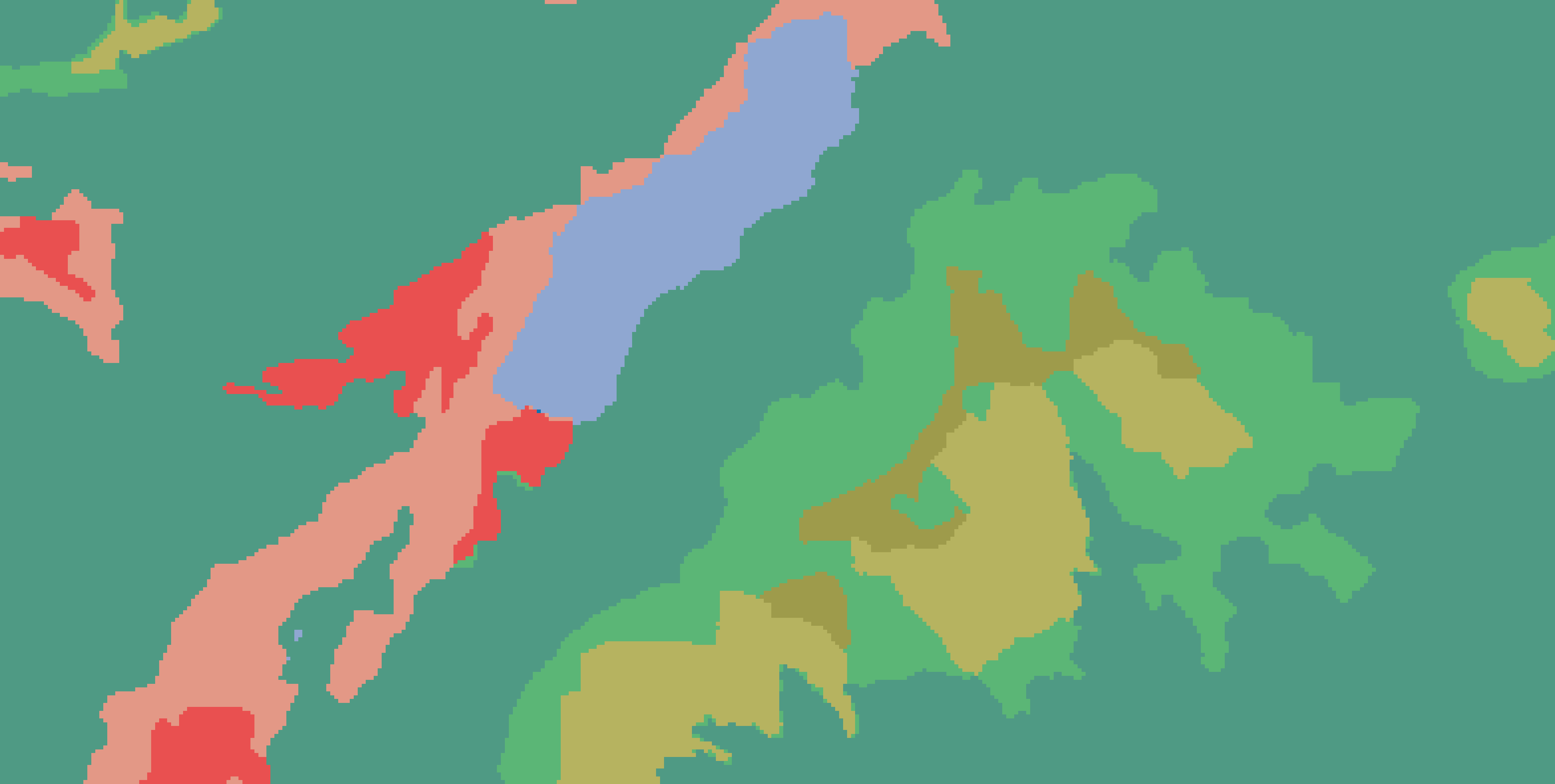}}}\hspace{1pt}
\subbottom[Scenario 4 -- weighting factor for all OSM classes is 0.75, other classifications adopt weighting factors as follows: cultivated 0.7, grassland 1.0, wetland 0.6]{\label{lc-fig:scenario4}
\resizebox*{5cm}{!}{\includegraphics{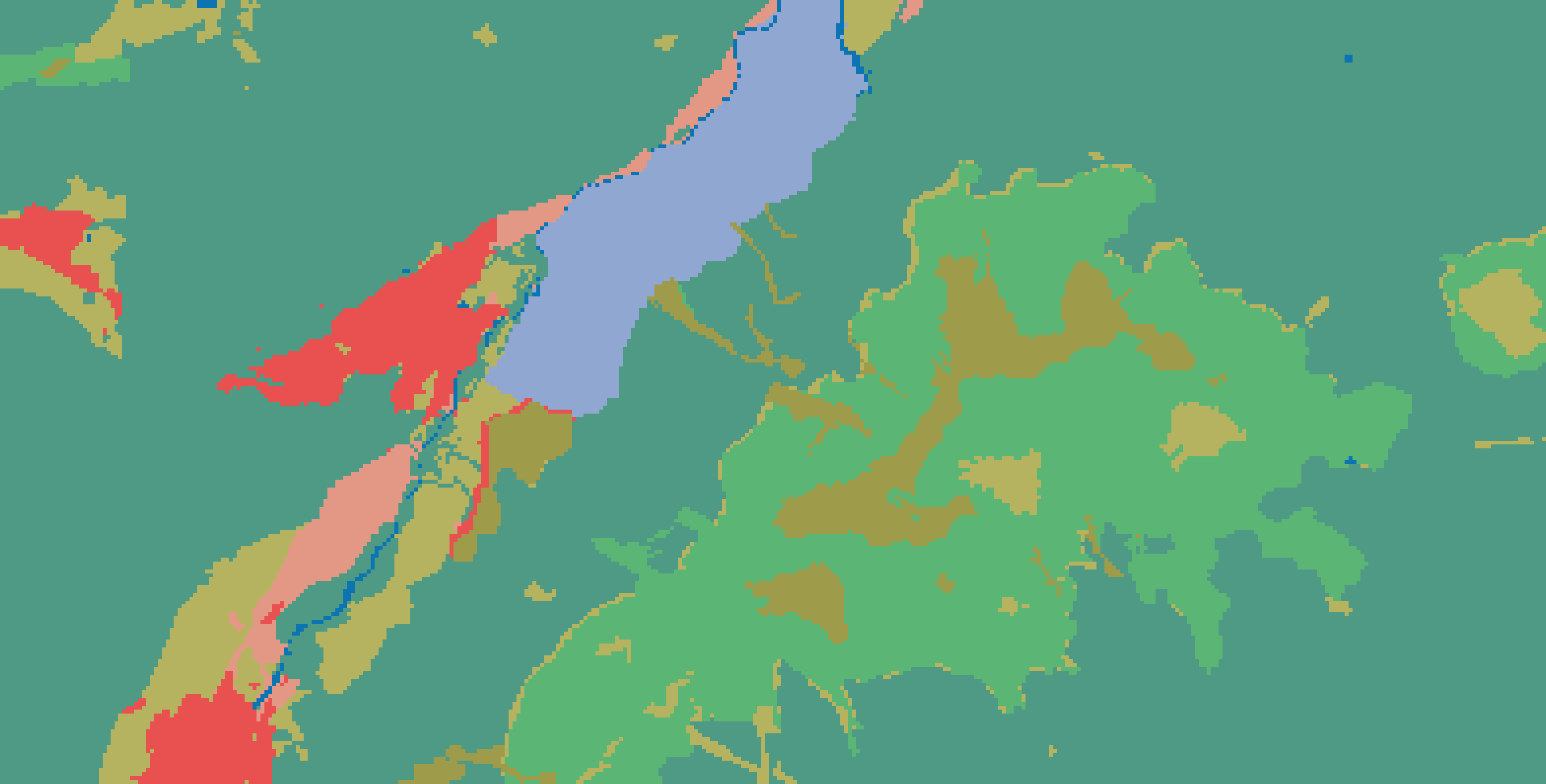}}}\hspace{1pt}
\subbottom[ATKIS \crudespace \crudespace \crudespace \crudespace \crudespace \crudespace \crudespace \crudespace \crudespace ]{\label{lc-fig:ATKIS_fig6}
\resizebox*{4.88cm}{!}{\includegraphics{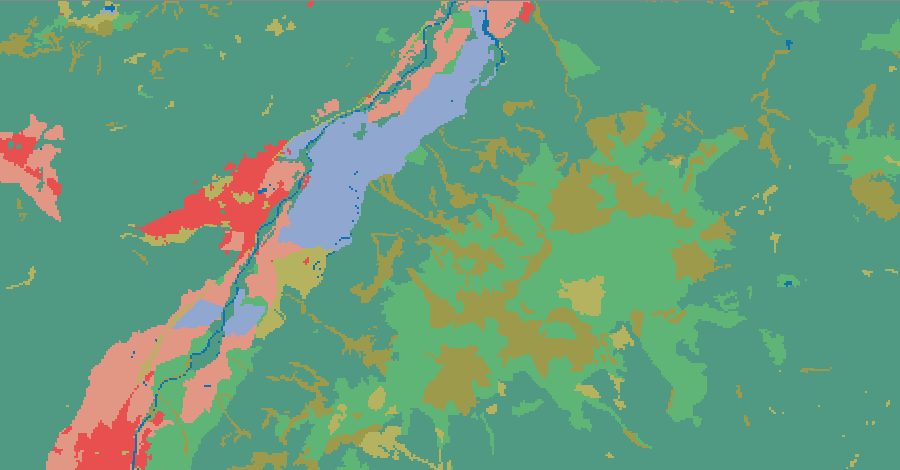}}}\hspace{1pt}
\subbottom{\label{lc-fig:ATKIS_fig6Legend}
\resizebox*{4.88cm}{!}{\includegraphics{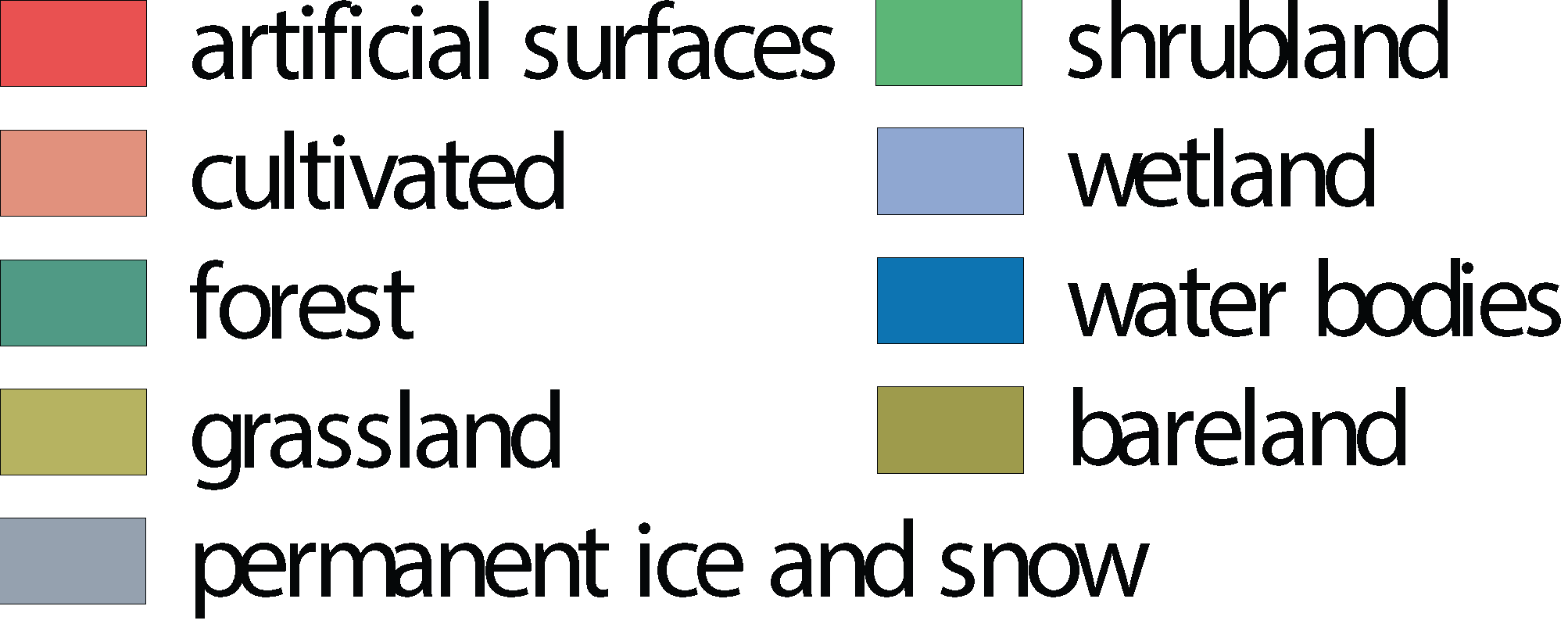}}}\hspace{1pt}
\caption{Overview of the land cover classification outputs based on the different scenarios. Scenario 1 -- no weighting factor has been applied. Scenario 2 -- OSM data is weighted by a factor of 0.7. Scenario 3 -- no OSM data is included. Scenario 4 -- the general weighting factor for OSM classes is 0.75, except for the following classes: cultivated 0.7, grassland 1.0, wetland 0.6.} 
\label{lc-fig:LCmaps}
\end{figure}

We further investigate the various scenarios in a qualitative manner by evaluating the boosted land cover maps. Figure~\ref{lc-fig:LCmaps} shows a subsection of our study area for each of our four test scenarios, as well as for the input and reference data. Using this approach, we can gain a visual understanding and intuition for the performance of our approach.

\subsection*{Comparison to individual datasets}
From Figure~\ref{lc-fig:compare} and Table~\ref{lc-tab.accuracy}, it is clear that Scenario 4, with class-wise weighting, appears to provide the best results. For this reason, we will use Scenario 4 as the output from our approach to compare to the performance of the original land cover classifications of CLC2006, OSM, and GlobeLand30.

Comparing the classifications to the ATKIS dataset, we obtain class-wise accuracy as depicted in Figure~\ref{lc-fig:compareAll} with an overall accuracy for the various land cover classifications as described in Table~\ref{lc-tab.accuracyAll}.

\begin{figure}[h!]
	\centering
	\begin{tikzpicture}[scale=0.7]
		\centering
		\begin{axis}[
			ybar, axis on top,
			height=10cm, width=20cm,
			bar width=0.4cm,
			ymajorgrids, tick align=inside,
			major grid style={draw=white},
			enlarge y limits={value=.1,upper},
			ymin=0, ymax=100,
			axis x line*=bottom,
			axis y line*=right,
			y axis line style={opacity=0},
			tickwidth=0pt,
			enlarge x limits=true,
			legend style={
				at={(0.5,-0.2)},
				anchor=north,
				legend columns=-1,
				/tikz/every even column/.append style={column sep=0.5cm}
			},
			ylabel={Class-wise balanced accuracy},
			symbolic x coords={
				Cultivated, Forest, Grassland, Shrubland,
				Wetland,Water,
				Artificial,
				Bareland},
			xtick=data,
			nodes near coords={
				\pgfmathprintnumber[precision=0]{\pgfplotspointmeta}
			}
			]
			\addplot [draw=none, fill=IvoryTUM] coordinates {
				(Cultivated, 50.5246)
				(Forest, 82.01)
				(Grassland, 66.8785)
				(Shrubland, 60.246)
				(Wetland, 86.904)
				(Water, 93.252)
				(Artificial, 91.261)
				(Bareland, 61.09)};
			\addplot [draw=none,fill=Pantone542] coordinates {
				(Cultivated, 90.35)
				(Forest, 85.79)
				(Grassland, 55.2222)
				(Shrubland, 53.6037)
				(Wetland, 89.541)
				(Water, 84.999)
				(Artificial, 82.741)
				(Bareland, 82.155)};
			\addplot [draw=none, fill=GreenTUM] coordinates {
				(Cultivated, 89.6)
				(Forest, 84.87)
				(Grassland, 54.8387)
				(Shrubland, 70.419)
				(Wetland, 90.019)
				(Water, 85.257)
				(Artificial, 82.73)
				(Bareland, 81.629)};
			\addplot [draw=none, fill=OrangeTUM] coordinates {
				(Cultivated, 90.6)
				(Forest, 86.02)
				(Grassland, 56.2365)
				(Shrubland, 70.233)
				(Wetland, 90.114)
				(Water, 86.563)
				(Artificial, 84.313)
				(Bareland, 81.905)};
			\legend{OSM, CLC2006, GlobeLand30, Proposed (Scenario 4)}
		\end{axis}
	\end{tikzpicture}
	\caption{The balanced, class-wise accuracy of our approach and the original land cover maps when compared to the ATKIS reference dataset.}
	\label{lc-fig:compareAll}
\end{figure}

\begin{table}[h!]
	\centering
	\begin{tabular}{p{0.15\linewidth}p{0.11\linewidth}p{0.11\linewidth}p{0.17\linewidth}p{0.13\linewidth}p{0.11\linewidth}}
		Estimate &  CLC2006 & OSM & GlobeLand30 & Scenario 4 \\  \toprule
		Accuracy &  0.7452 & 0.7405 & 0.7697 & 0.7828 \\
		Kappa  &    0.6228 & 0.5013 & 0.6564 & 0.6748\\
	\end{tabular}
	\caption{Accuracy assessment and comparison of the original datasets to the proposed approach.}
	\label{lc-tab.accuracyAll}
\end{table}

As \correction{mentioned}, the overall accuracy represents the proportion of correctly classified pixels to the reference map. For this reason, the measure of accuracy is relative rather than absolute, as it depends on the quality of the reference data \cite{Veregin1999}. In our case, the ATKIS dataset provides a larger variety of classes and more detailed coverage than the datasets we are boosting. Thus, the overall accuracy is relative to the level of detail on the reference map. Nevertheless, the relative nature of the accuracy is an important factor to keep in mind when assessing the results with respect to existing work, which might employ another reference map or accuracy measure.

\begin{figure}[h!]
\centering
\subbottom[Bing Map Aerial]{
\resizebox*{4cm}{!}{\includegraphics{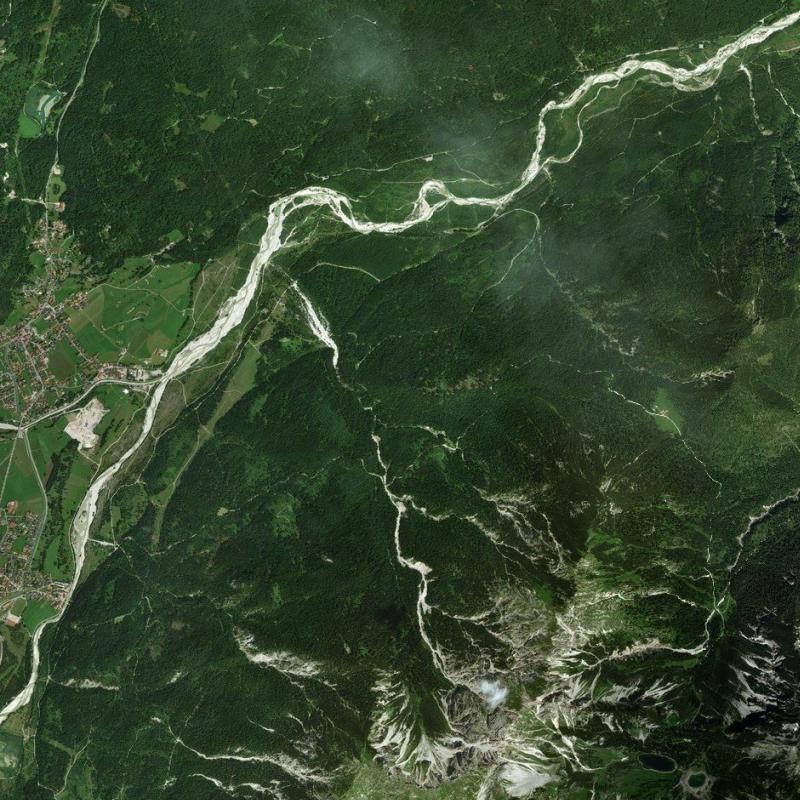}}}\hspace{1pt}
\subbottom[ATKIS]{\label{lc-fig:LCexampleAtkis}
\resizebox*{4cm}{!}{\includegraphics{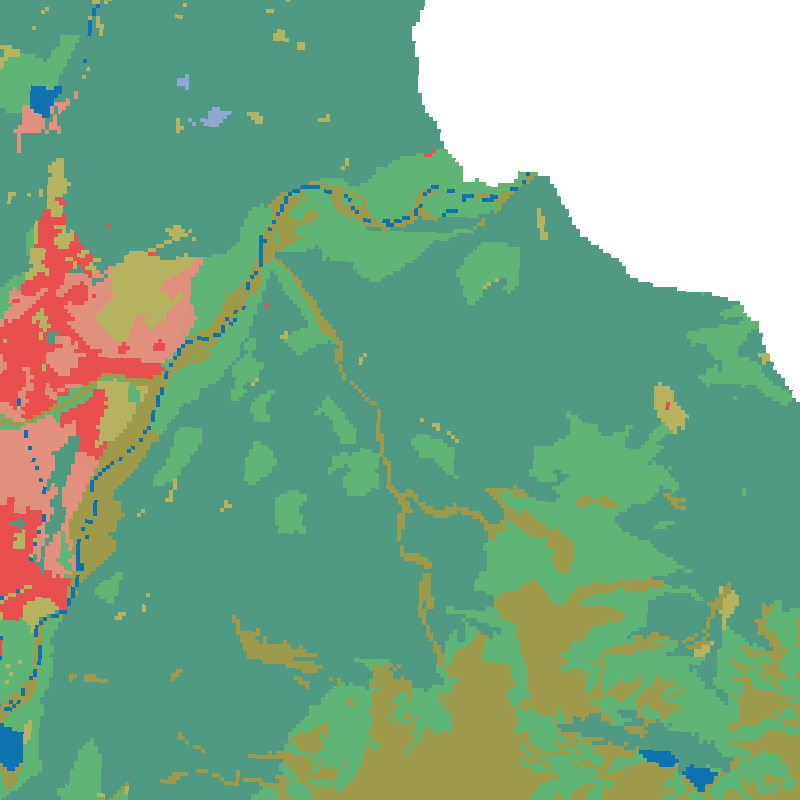}}}\hspace{1pt}
\subbottom[Scenario 4]{\label{lc-fig:LCexamplePMG}
\resizebox*{4cm}{!}{\includegraphics{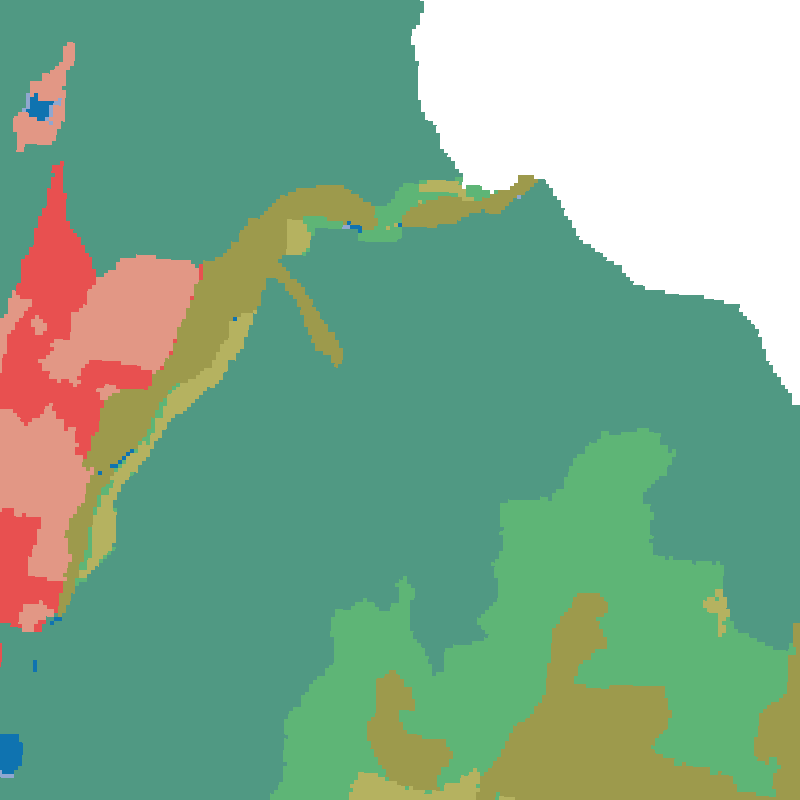}}}\hspace{1pt}
\subbottom[GlobeLand30]{    
\resizebox*{4cm}{!}{\includegraphics{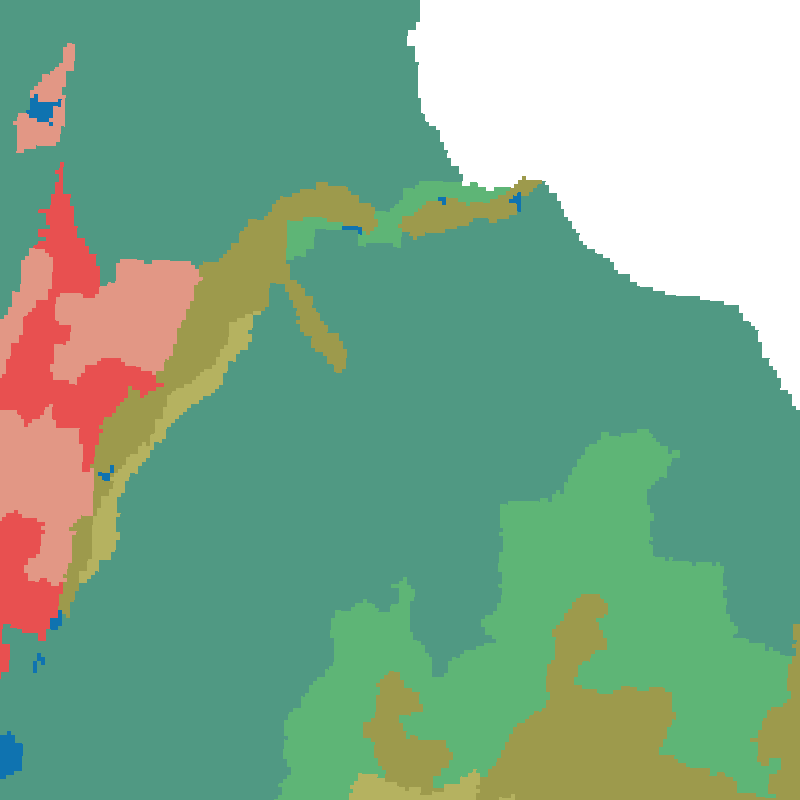}}}\hspace{1pt}
\subbottom[CLC2006]{
\resizebox*{4cm}{!}{\includegraphics{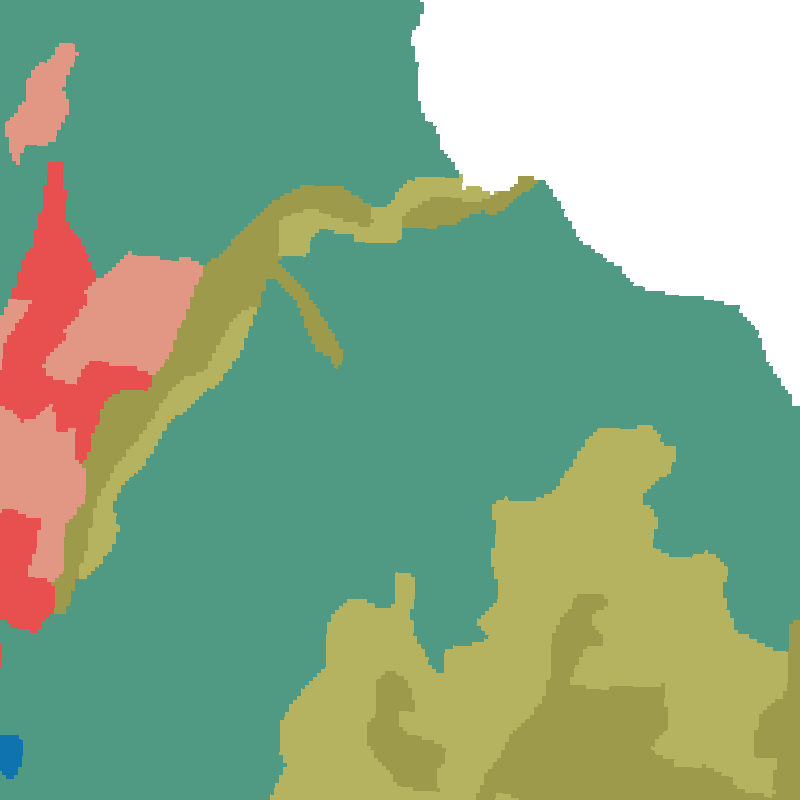}}}\hspace{1pt}
\subbottom[Land cover from OSM]{\label{lc-fig:LCexampleOSM}
\resizebox*{4cm}{!}{\includegraphics{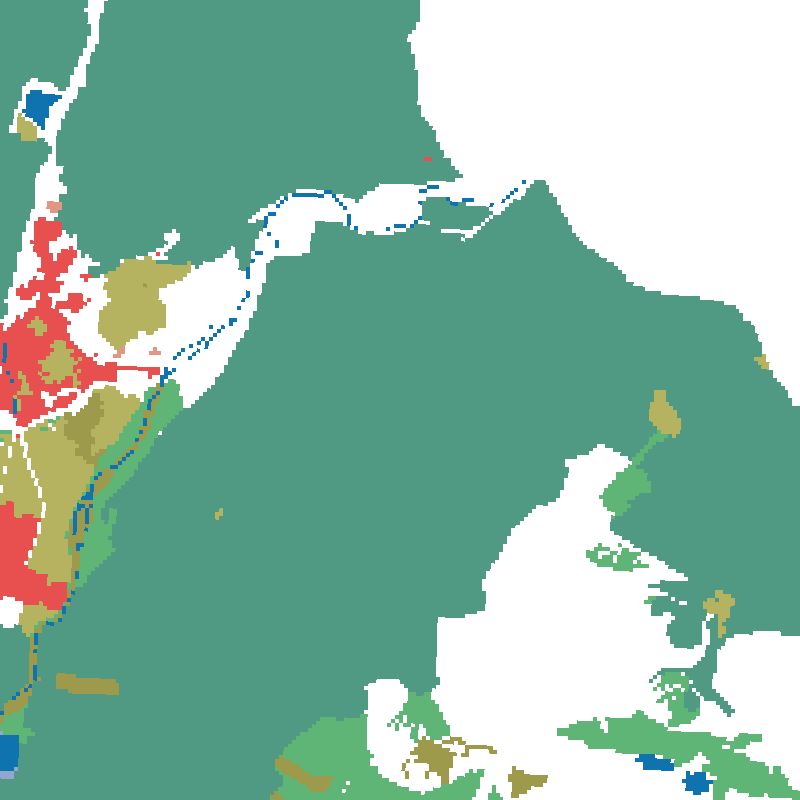}}}\hspace{1pt}
\subbottom{\resizebox*{4.88cm}{!}{\includegraphics{chap4-figs/figs/scenarioLegend2.png}}}\hspace{1pt}
\caption{Comparison of the different land cover datasets over a selected region in our test scene. Due to inherent ambiguity in definition of classes such as shrubland and grassland, some areas with conflicting patterns are observed. Furthermore, the largely incomplete OSM data in the region can also be observed as the white (no data) pixels.} 
\label{lc-fig:LCexample}
\end{figure}

Additionally, we compare the prior land cover classifications and our best boosted classification to an aerial image and the ATKIS reference data in Figure~\ref{lc-fig:LCexample}. From this figure, the heterogeneity of the datasets and the incomplete nature of the OSM dataset is clearly visible.

\subsection*{Classification uncertainty}
Due to the Bayesian framework in which cluster graphs are rooted, we obtain a measure of the probability for the likelihood of each class being present in each pixel of the boosted land cover map. Based on these probabilities, we can extract an uncertainty metric, Shannon diversity index, for each pixel in our boosted land cover classification map.

\begin{figure}[h!]
    \begin{center}
        \includegraphics[width=0.9\linewidth]{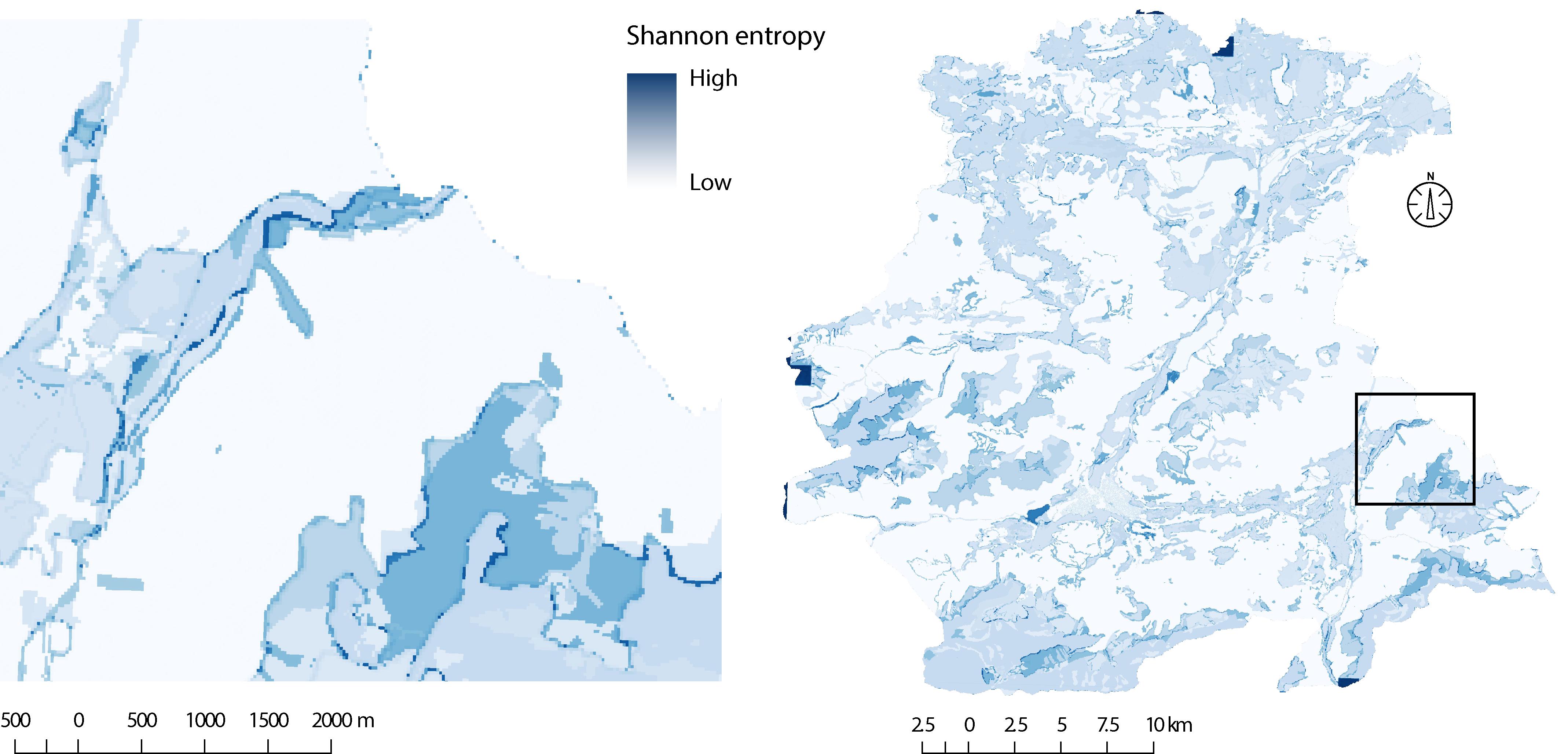}
    \end{center}
    \caption{Land cover Shannon diversity index map (Scenario 4) depicting the uncertainty in the land cover classification. Low values represent patterns with the highest degree of thematic uncertainty when the land cover class was assigned.}\label{lc-fig:LCconfidence}
\end{figure}

The uncertainty map for our approach (Scenario 4) and a zoomed-in section corresponding to the boosted region in Figure~\ref{lc-fig:LCexample} are depicted in Figure~\ref{lc-fig:LCconfidence}.

\subsection*{Validation study}
\begin{figure}[h!]
  \centering
     \includegraphics[width=0.8\textwidth]{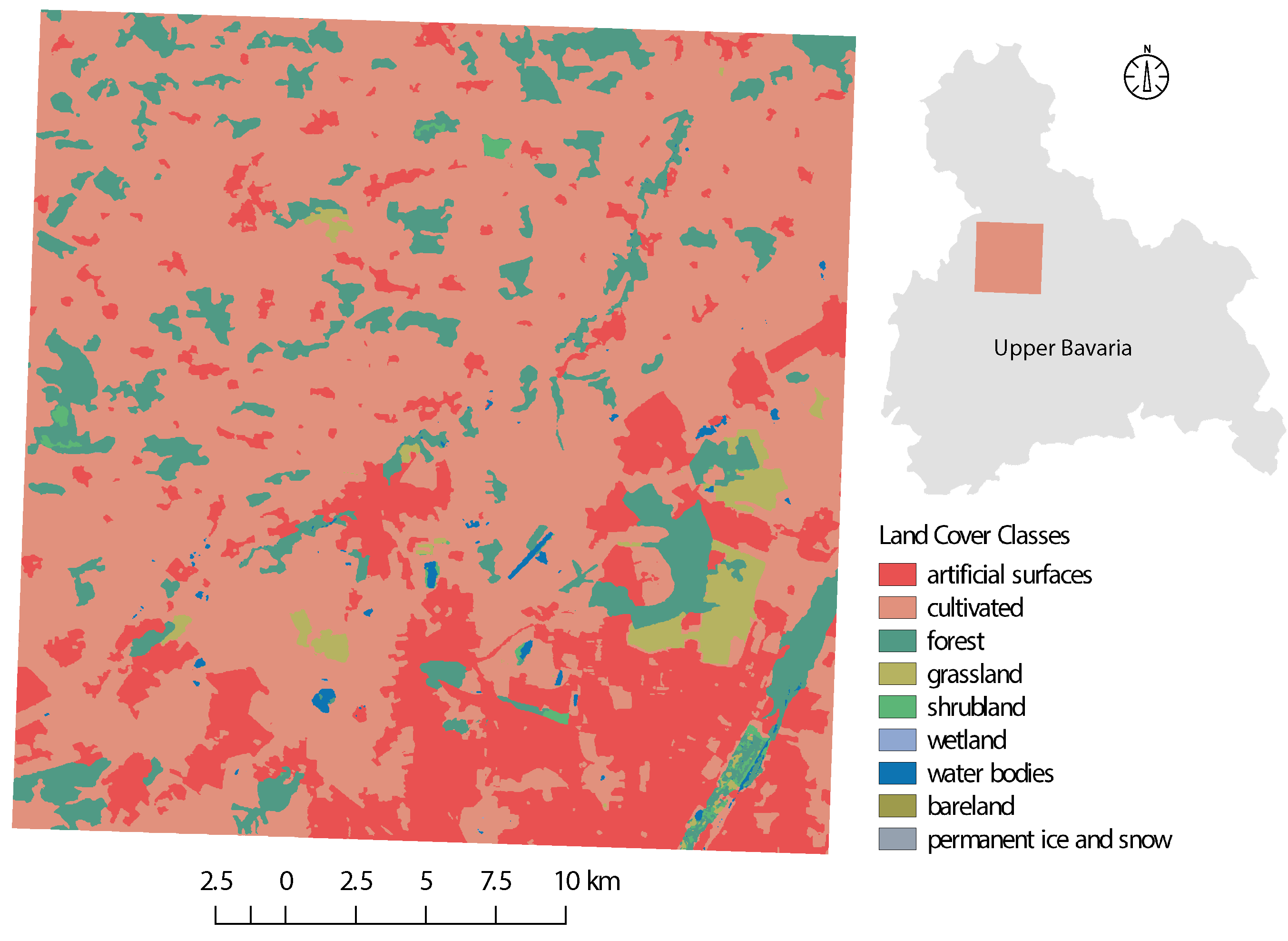}
  \caption{Overview map of the validation study area located in Upper Bavaria, Germany. The land cover classifications include six land cover classes based on the classification scheme adopted from GlobeLand30.}
  \label{lc-fig:validationMap}
\end{figure}

\correction{Unlike machine learning methods, Bayesian methods do not typically require an independent validation study as there is no differentiation between training and testing phases.} However, we performed such a study to assess our general approach and choice of factors and priors. This study was conducted using the proposed cluster graph approach to an auxiliary study area of 84~850~km$^2$ within Germany (see Figure \ref{lc-fig:validationMap}). To evaluate the generalisation of our approach with respect to the setting of priors and confidence factors, we kept these values as they were defined for the original test region (Scenario 4).

\begin{figure}[h!]
\centering
\subbottom[Bing Map Aerial]{
\resizebox*{4cm}{!}{\includegraphics{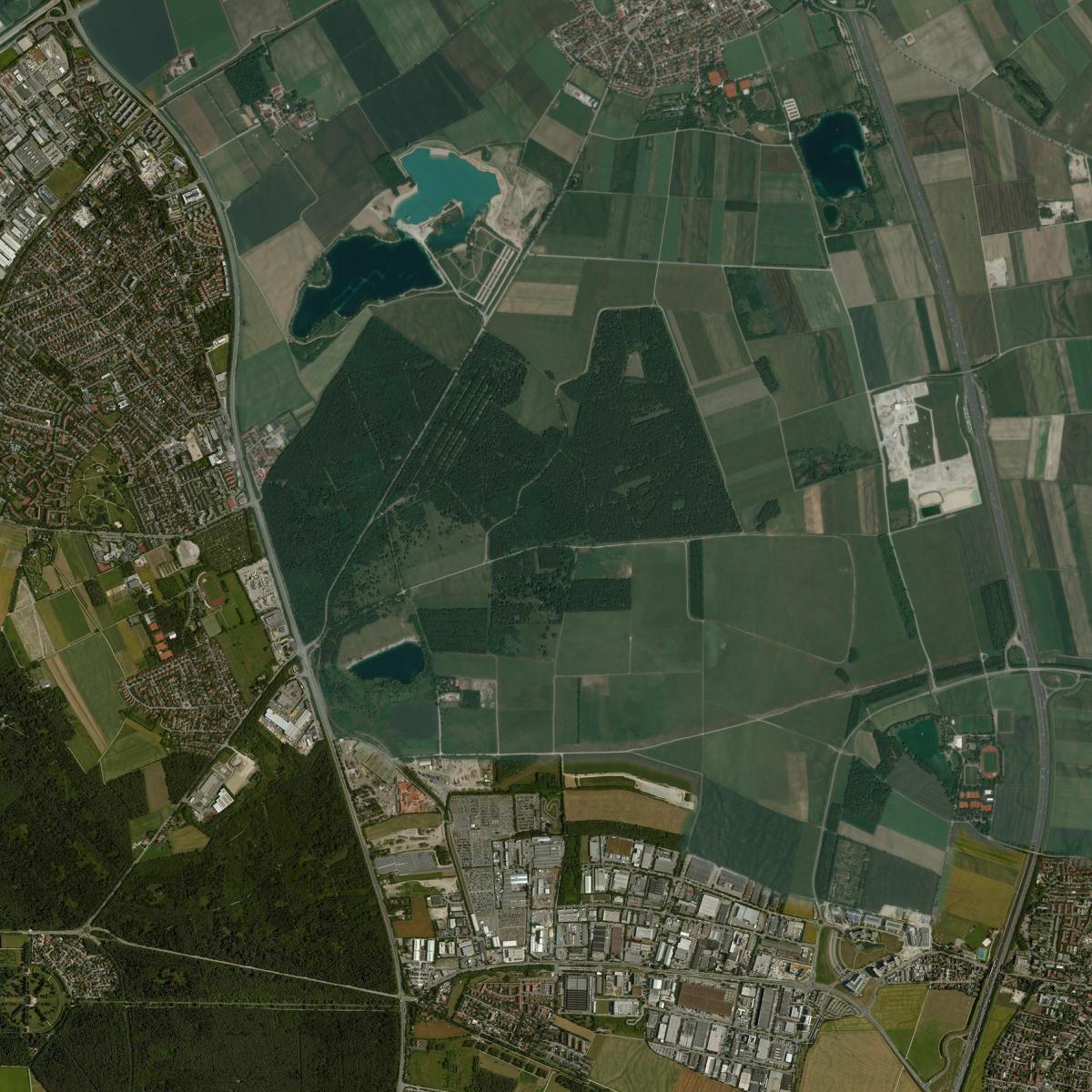}}}\hspace{1pt}
\subbottom[ATKIS]{\label{lc-fig:LCexampleAtkis}
\resizebox*{4cm}{!}{\includegraphics{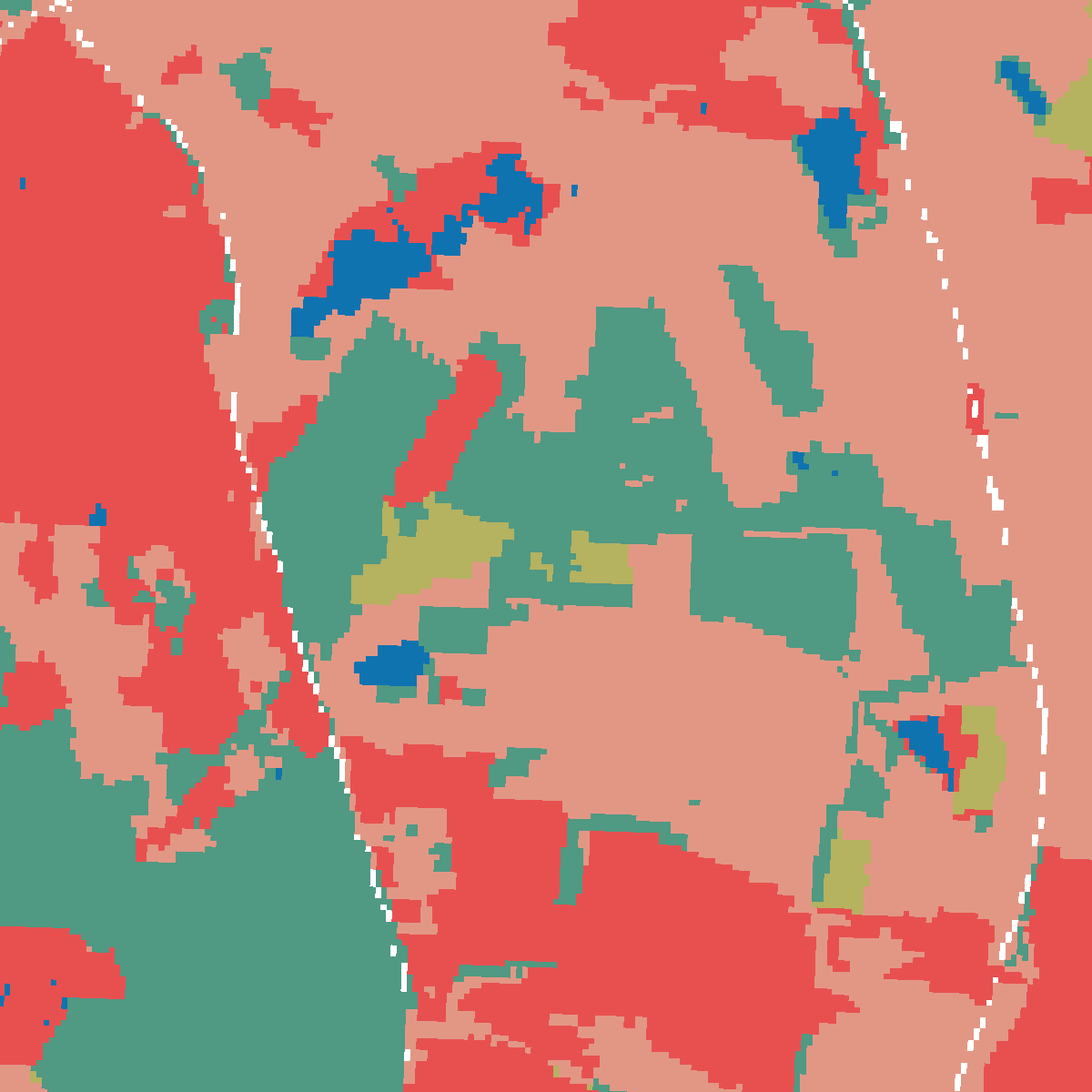}}}\hspace{1pt}
\subbottom[Our Approach]{\label{lc-fig:LCexamplePMG}
\resizebox*{4cm}{!}{\includegraphics{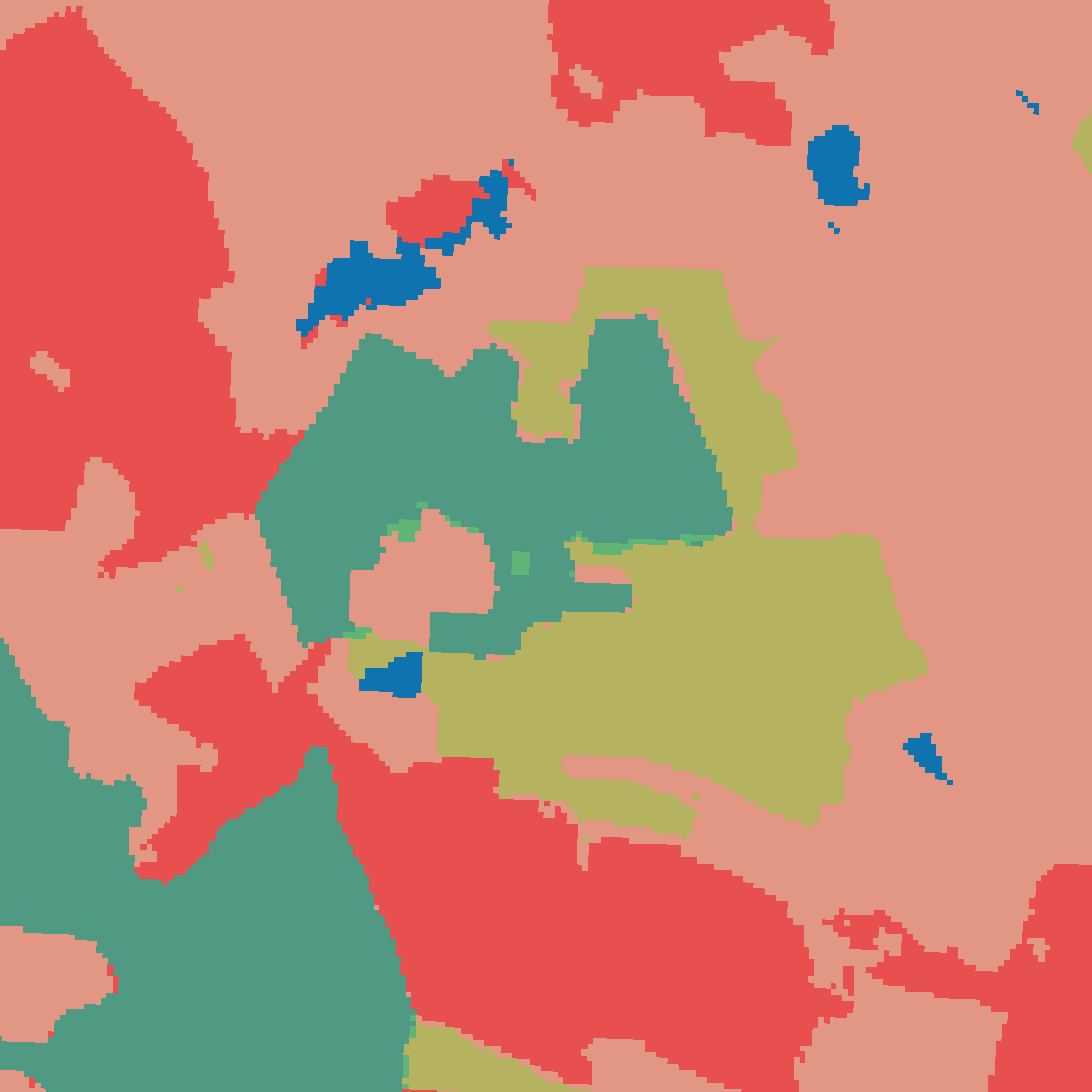}}}\hspace{1pt}
\subbottom[GlobeLand30]{    
\resizebox*{4cm}{!}{\includegraphics{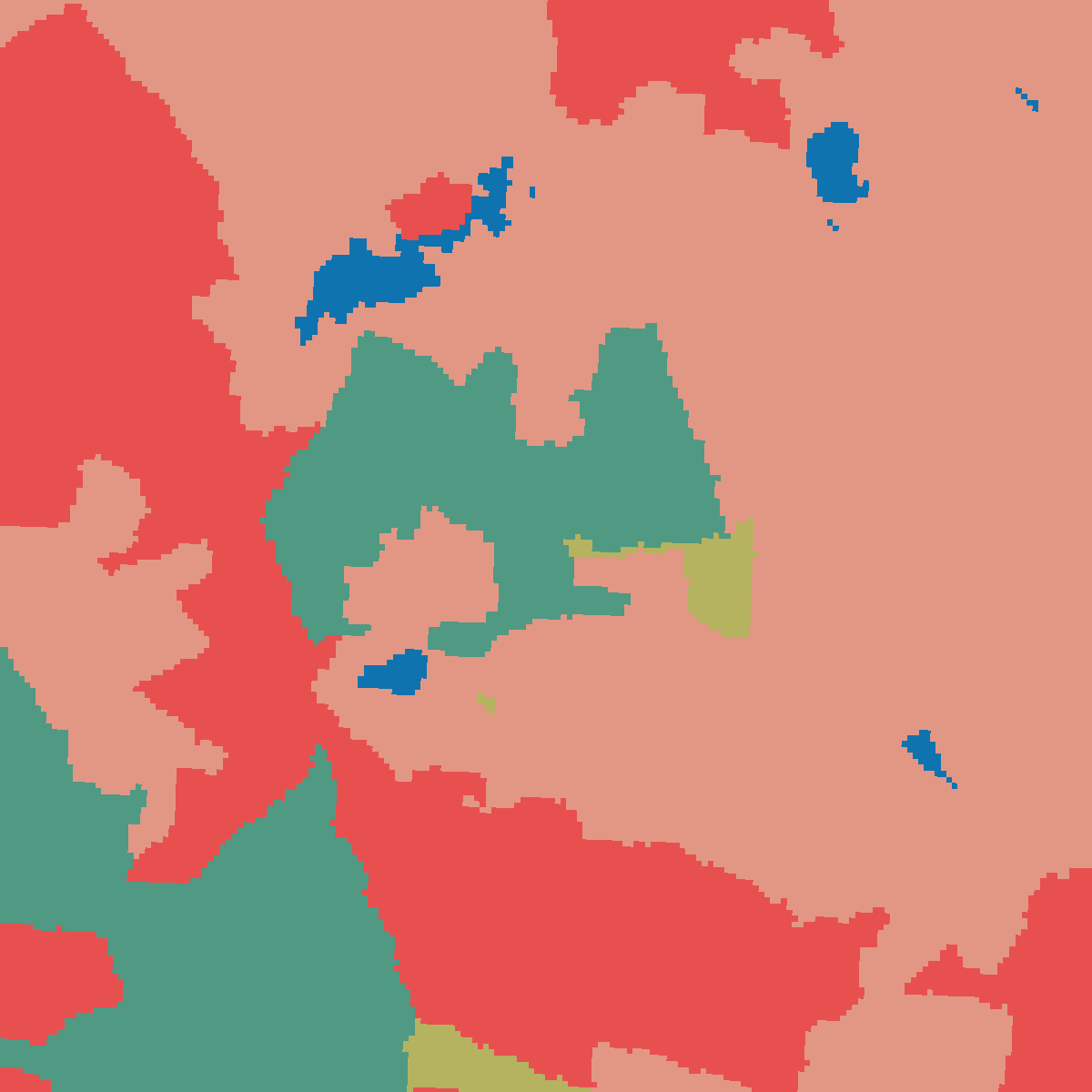}}}\hspace{1pt}
\subbottom[CLC2006]{
\resizebox*{4cm}{!}{\includegraphics{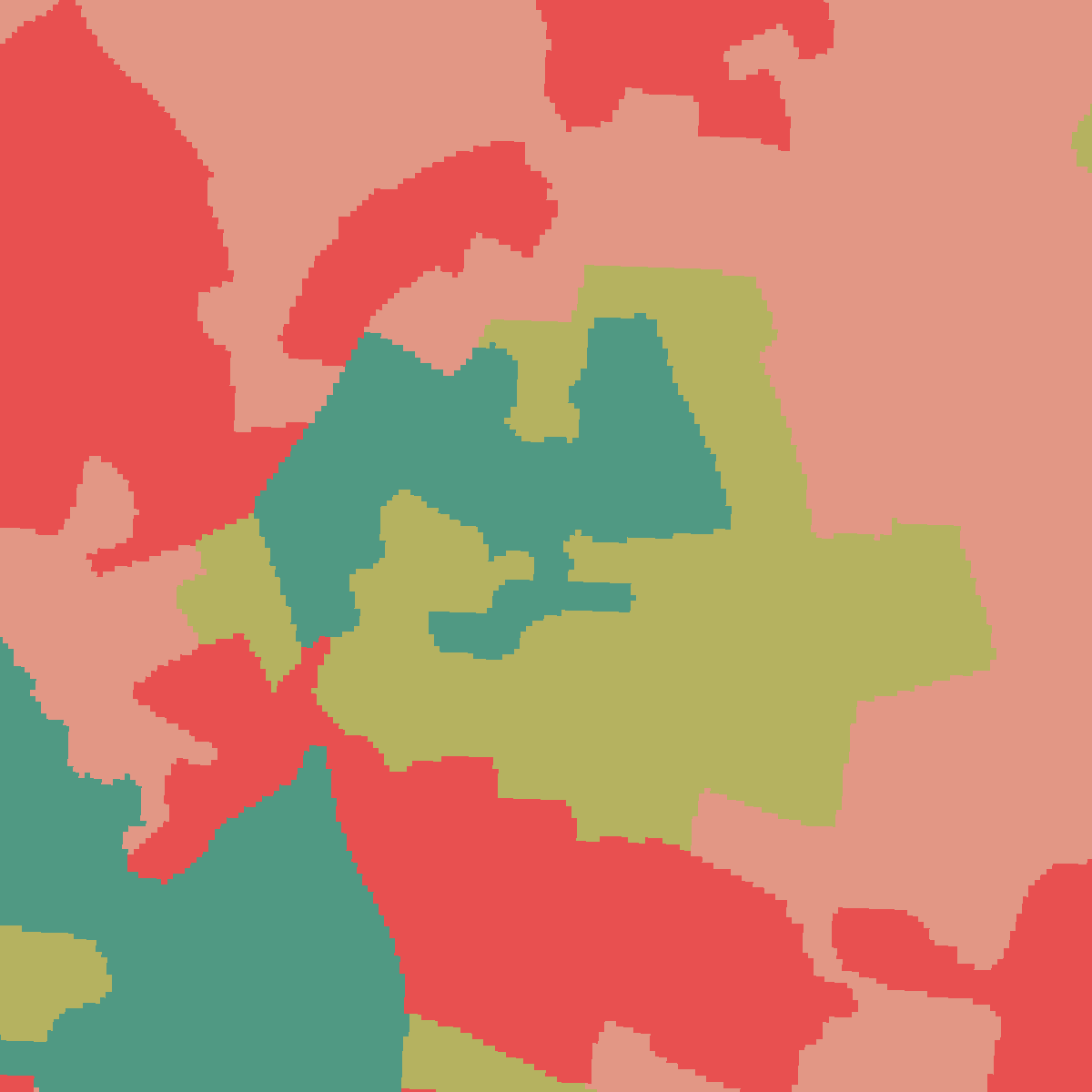}}}\hspace{1pt}
\subbottom[Land cover from OSM]{\label{lc-fig:LCstudy2OSM}
\resizebox*{4cm}{!}{\includegraphics{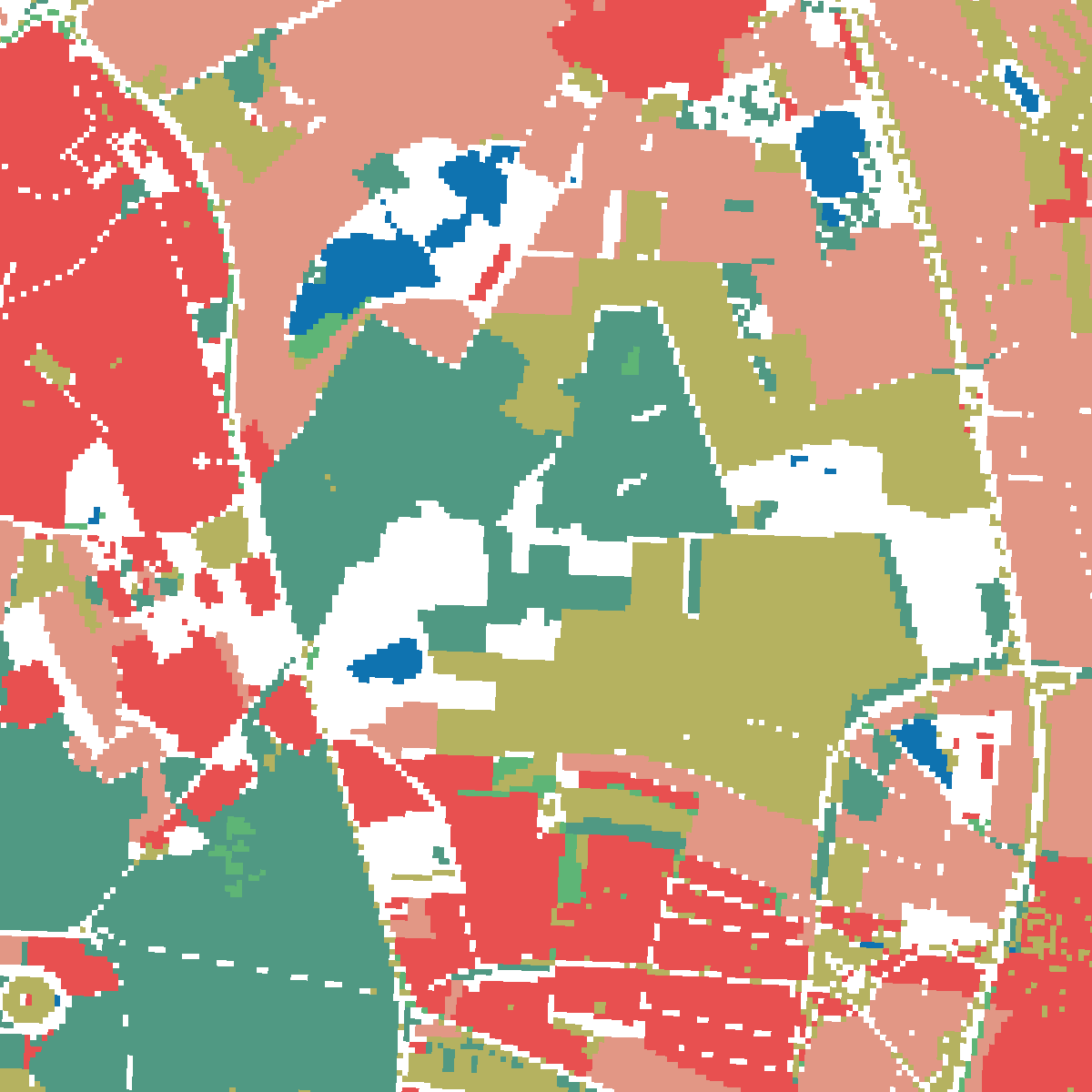}}}\hspace{1pt}
\subbottom[\correction{Legend}]{
\resizebox*{4.88cm}{!}{\includegraphics{chap4-figs/figs/scenarioLegend2.png}}}\hspace{1pt}
\caption{Validation study: Comparison of the different land cover datasets over a selected region in our validation scene. Due to inherent ambiguity in definition of classes such as grassland, some areas with conflicting patterns are observed in the original datasets (d) -- (f). Our approach (c) exhibits spatial smoothness while still capturing smaller details even in regions with conflicting information.} 
\label{lc-fig:LCexample2}
\end{figure}

The validation study represents six land cover classes with the most extensive coverage of artificial and agricultural areas. By applying the same approach, we produced a boosted land cover classification with the overall accuracy given in Table~\ref{lc-tab.accuracyStudy2}. Figure~\ref{lc-fig:LCexample2} introduces a comparison of different land cover datasets over the validation area, and Figure~\ref{lc-fig:compareAllstudy2} reports on the class-wise accuracy of our approach as well as the other datasets.

\begin{table}[h!]
    \centering
    \begin{tabular}{p{0.15\linewidth}p{0.11\linewidth}p{0.11\linewidth}p{0.17\linewidth}p{0.17\linewidth}p{0.11\linewidth}}
        Estimate &  CLC2006 & OSM & GlobeLand30 & Our Approach \\  \toprule
        Accuracy &  0.8302 & 0.8009 & 0.8401 & 0.8434 \\
        Kappa  &    0.6952 & 0.6886 & 0.7054 & 0.7111\\
    \end{tabular}
    \caption{Validation study: Accuracy assessment and comparison of the original datasets to the approach proposed.}
    \label{lc-tab.accuracyStudy2}
\end{table}

\begin{figure}[h!]
\centering
\begin{tikzpicture}[scale=0.7]
  \centering
  \begin{axis}[
        ybar, axis on top,
        height=10cm, width=20cm,
        bar width=0.4cm,
        ymajorgrids, tick align=inside,
        major grid style={draw=white},
        enlarge y limits={value=.1,upper},
        ymin=0, ymax=100,
        axis x line*=bottom,
        axis y line*=right,
        y axis line style={opacity=0},
        tickwidth=0pt,
        enlarge x limits=true,
        legend style={
            at={(0.5,-0.2)},
            anchor=north,
            legend columns=-1,
            /tikz/every even column/.append style={column sep=0.5cm}
        },
        ylabel={Class-wise balanced accuracy},
        symbolic x coords={
           Cultivated, Forest, Grassland, Shrubland,
           Water,
           Artificial},
       xtick=data,
       nodes near coords={
        \pgfmathprintnumber[precision=0]{\pgfplotspointmeta}
       }
    ]
    \addplot [draw=none, fill=IvoryTUM] coordinates {
        (Cultivated, 86.83)
        (Forest, 92.99)
        (Grassland, 85.032)
        (Shrubland, 54.63642)
        (Water, 82.5237)
        (Artificial, 89.66)};
   \addplot [draw=none,fill=Pantone542] coordinates {
        (Cultivated, 86.41)
        (Forest, 81.426)
        (Grassland, 81.1356)
        (Shrubland, 49.8)
        (Water, 54.50658)
        (Artificial, 84.49)};
   \addplot [draw=none, fill=GreenTUM] coordinates {
        (Cultivated, 85.01)
        (Forest, 79.414)
        (Grassland, 73.5503)
        (Shrubland, 49.9046)
        (Water, 62.3603)
        (Artificial, 86.53)};
   \addplot [draw=none, fill=OrangeTUM] coordinates {
        (Cultivated, 84.73)
        (Forest, 83.171)
        (Grassland, 79.0354)
        (Shrubland, 49.8189)
        (Water, 64.2089)
        (Artificial, 84.68)};
    \legend{OSM, CLC2006, GlobeLand30, Boosted (validation study)}
  \end{axis}
  \end{tikzpicture}
    \caption{Validation study: The balanced, class-wise accuracy of our approach and the original land cover maps when compared to the ATKIS reference dataset.}
    \label{lc-fig:compareAllstudy2}
\end{figure}

\section{Discussion}
Generally, the results presented in Section~\ref{lc-sec:results} show that our boosting method can merge existing land cover classifications in a robust manner. Furthermore, it shows that we can generate an accurate uncertainty map of the boosted area, which is useful in the analysis of the boosted land cover map.

In this section, we further investigate these results and describe the capabilities and advantages of our method for land cover boosting, as well as describe its shortfalls.

\subsection*{Comments on the effect of priors}
Since priors are set based on expert knowledge and are subjective by nature, it is difficult to fully analyse the effect of various decisions on accuracy. However, we can make a few key observations based on the results we presented. From Figure~\ref{lc-fig:compare}, we can see that all scenarios which include all the datasets have similar accuracy. In contrast, scenario 3, which excludes OSM data altogether, has a slightly lower accuracy across many classes. However, the accuracy measure does not provide a complete picture of how this factor affects the overall boosted classification appearance. From Figure~\ref{lc-fig:LCmaps}, we can see that the overall map confidence factor does not significantly change the final boosted classification. In contrast, class-wise confidence factors have a somewhat larger effect on the final booster classification -- see Figures~\ref{lc-fig:LCmaps}(d), \ref{lc-fig:LCmaps}(e), and \ref{lc-fig:LCmaps}(g). By adapting class-wise confidence factors of the VGI dataset, we can better capture fine details which are well represented in human labelled datasets (such as river and water features). These types of features are usually only present in land cover maps created with a relatively small minimum mapping unit and are, therefore, not often represented by existing large-scale land cover datasets such as GlobeLand30 and CLC2006.

Table~\ref{lc-tab.accuracy} confirms that the inclusion of more detailed prior information leads to better overall accuracy, with a greater Kappa coefficient. However, the overall differences are relatively small and, therefore, can also be interpreted that the proposed cluster graph approach is robust to the setting of priors and can produce an overall accurate boosted classification for a range of prior configurations. This is likely due to the strong inference abilities of probabilistic graphical models, which allow this approach to determine the complex inter-dependencies between factors. This, coupled with our geographically centred design approach, appears to capture the most important relationships and thus forgoes the need for strong expert priors to generate a reasonable land cover classification.

Apart from the effect of priors, Table~\ref{lc-tab.accuracy} reveals that the inclusion of VGI has a positive effect on the overall accuracy of the final boosted classification. Thus, the inclusion of additional datasets, even if noisy and incomplete, might play a larger role in the overall accuracy than the configuration of priors.

\subsection*{Qualitative view of boosted classification accuracy}
We compared the boosted classifications from our test scenarios over a diverse sub-section of our study area using a purely qualitative approach. In Figure~\ref{lc-fig:LCmaps}, we can see the classification differences between our four test scenarios, the original land cover classifications and the ATKIS \correction{reference datums}.

Based on Figure~\ref{lc-fig:LCmaps} it is clear how the inclusion of various datum as well as the selection of priors affect the overall boosting process. Furthermore, we can observe how our approach preserves spatial consistency between neighbouring classes even when the input data sources are noisy, conflicting or incomplete. This observation is particularly clear in scenario 4, Figure~\ref{lc-fig:LCmaps}(g), where the river feature from the OSM dataset is included in the boosted classification, but the area around the feature remains spatially consistent with CLC2006 and GlobeLand30.

Figure~\ref{lc-fig:LCexample} depicts another sub-region of our test scene and the ability of our approach to perform reasonable inference in the presence of conflicting and missing data becomes quite apparent. The lower right area of the region has very sparse coverage in the OSM dataset and conflicting labels between GlobeLand30 and CLC2006. In this case, our approach can infer a suitable and accurate (with comparison to ATKIS and the aerial image) land cover mapping while still maintaining smaller features, such as lakes and grassland areas, as well as spatial consistency (smooth land cover mapping and reasonable neighbouring classes). Additionally, by referring to the produced land cover uncertainty map for this region in Figure~\ref{lc-fig:LCconfidence}, it is clear that the areas where land cover was inferred based on missing and conflicting evidence show higher uncertainty than areas where all three input data sources were in agreement. Thus our cluster graph approach can be said to be reasoning in a rational manner.

Furthermore, the results of our validation study, Figure~\ref{lc-fig:LCexample2}, once again show how spatial consistency is preserved, and how even in the presence of noisy VGI and conflicting information, our cluster graph approach can boost the classifications to generate a detailed land cover classification. One particular area of interest is the area between the forest, artificial, and cultivated classes to the left of the centre of the image. The OSM data is incomplete in this region, and the CLC2006 and GlobeLand30 classifications are conflicting. However, even in the presence of this, our boosting approach can infer a classification for this region in a spatially consistent manner. While the classification for the region in CLC2006 is grassland, and in GlobeLand30 is artificial, our boosted classification labels the area as cultivated. This labelling is likely the result of the strong preference for preserving Tobler's first law of geography, as well as strong priors for inter-class relationships. Upon inspection of the aerial image of this region, it can be seen that while our classification overlooks a small artificial region, the remainder of the classification is feasible.

While our approach does lose some granularity in smaller spatial areas, this could be considered a reasonable trade-off, given that some of the finer features are not present in more than one of the input data sources. Additionally, this smoothing effect could possibly be a consequence of the patch-based processing approach we employed, as detailed in Figure~\ref{lc-fig:Stitching}.

\subsection*{Quantitative view of boosted classification accuracy}
Referring to the overall accuracy of our approach on the test and validation study areas in Table~\ref{lc-tab.accuracyAll} and Table~\ref{lc-tab.accuracyStudy2}, our boosted land cover classification exhibits the highest overall accuracy compared to our ATKIS reference dataset. Furthermore, the Kappa coefficient of our land cover classification is significantly higher in both cases. In general, our approach far exceeds the accuracy of OSM and CLC2006 land cover and has a small improvement over GlobeLand30. As overall accuracy does not provide a complete picture of land cover classification accuracy, we further investigate and analyse the balanced class-wise accuracy of the prior data and our boosted classification.

Based on the respective balanced class-wise accuracy of our test scene, as depicted in Figure~\ref{lc-fig:compareAll}, it can be seen that our approach shows reasonable performance in all classes. While the accuracy of OSM does exceed our approach for some classes, note that our approach is never affected by poor class accuracy in any of the datasets. For instance, concerning bareland, shrubland, and cultivated classes, the accuracy of our approach is always better or on par with the other sets of input data. This property could be argued to be of more significance than being able to always achieve the best accuracy in each class. The reason for this is that our approach can perform at a consistent level, even in the presence of noisy input data and, therefore, can provide a higher quality land cover classification overall.

By examining the class-wise accuracy for our validation scene in Figure~\ref{lc-fig:compareAllstudy2}, the same observations can be made. While in the validation scene, OSM performs the best overall, it should be noted that the confidence factor for the OSM dataset was not adjusted. Thus the evidence from the OSM dataset was down-weighted in the cluster graph. However, even with this low confidence in OSM, our approach was still able to extract value from the high accuracy OSM data to improve its accuracy over CLC2006 and GlobeLand30. This once again shows the robustness of a cluster graph approach in the presence of ill-defined priors. As the OSM dataset for this region is significantly less sparse than for our test region, the confidence factor should have been adjusted upwards. Furthermore, the region contains large areas of cultivated land, which is known to be a source of conflict among CLC2006 and GlobeLand30. Thus, prior to boosting the land cover classification, expert knowledge should have been used to adjust the confidence factors and priors for the region.

While class accuracy does not depict a large improvement over existing methods, the power of our proposed approach is in its ability to select and fuse the existing approaches in such a way as to have an overall better land cover classification than each of the individual land cover maps that were boosted. Furthermore, our approach can perform boosting in the presence of noisy, incomplete, and conflicting input data while preserving spatial consistency and producing an overall reasonable, detailed and still diverse land cover classification. 

\subsection*{Comments on classification uncertainty}
Perhaps one of the largest benefits of the proposed approach is that it provides  the probability for each class occurring at each pixel. These class-wise probabilities can easily be exploited to generate uncertainty maps of the boosted land cover classification as well as for the original datasets. Due to the nature of the aerial imagery, which is often used to generate land cover classifications, inherent inter-class ambiguities exist due to the lack of height information. The generated uncertainty maps can help develop better algorithms for land cover classification by either forming part of the optimisation function or providing experts with clues as to which areas are often misclassified and why.

By comparing the uncertainty map, Figure~\ref{lc-fig:LCconfidence}, to the corresponding classification maps, Figure~\ref{lc-fig:LCexample}, it is clear that the uncertainty for each region tends to agree with intuition about the nature of certainty across the datasets. For example, regions that present conflicting information in two inputs and are missing information in the third are deemed to be more uncertain in the boosted map, while areas with agreement present a very low uncertainty. Perhaps one interesting observation is the low uncertainty in some regions where information is conflicting; this is likely due to the expert priors that enforce self-similarity based on Tobler's law of Geography.

Uncertainty maps can provide useful inputs into ecological and climactic research where uncertainty about land cover classifications can help improve models of land use dynamics and ecosystem stability. Furthermore, the class-wise probabilities could open the door for manual intervention where the top $n$ classes which exhibit similar probabilities could be presented to practitioners for disambiguation and thus further improvement of the overall land cover map. This process could further be expanded to fine-tune the inter-class priors and thus improve the overall performance of the proposed approach.

\section{Conclusion}
In this chapter, we presented a probabilistic graphical model approach to boosting of land cover classification maps. The formulation of the proposed solution took the form of a cluster graph that used observation and relational factors, along with expert knowledge to perform inference across multiple existing land cover classification data products. The study is applied to land cover classifications derived from remote sensing data, as they are among the crucial inputs to environmental analysis that supports research on topics such as climate change, deforestation, urban change, and population growth. Additionally, we made use of incomplete but accurate volunteered geographic information (VGI), namely OpenStreetMap (OSM), as an additional set of evidence for land cover classification boosting. Furthermore, we analysed how confidence factors could benefit from accurately labelled regions of data while reducing the effect of inaccurate and incomplete areas on boosting.

To improve the accuracy of land cover classification in the study region of Garmisch-Partenkirchen, Germany, our approach exploits existing expert knowledge and constraints such as Tobler's first law of geography. Taking expert knowledge into account enables a classification boosting process with more flexibility and robustness. Furthermore, this approach allows practitioners to customise the tool to their needs while still being robust enough to compensate for poor assumptions and initialisation.

Using the cluster graph approach, we produced a feasible, diverse, and spatially-consistent boosted land cover classification based on GlobeLand30, CLC2006, and OSM data. Our boosted classification exhibited an overall accuracy improvement of around 1.4\% when compared against a reference land cover classification map of our test region. Furthermore, our approach was applied to a validation region without adjusting the priors and was shown to perform well even when initialised with sub-optimal priors.

In addition to producing accurate boosted land cover classifications, the proposed approach can provide additional information on the uncertainty of the boosted classification and highlight commonly misclassified classes within our study region. These additional products are not available when using naïve boosting methods or learned ensemble methods and can provide important insights into better understanding land cover and land use dynamics.

\chapter{Strengthening PGMs: the purge-and-merge algorithm}\label{csp-chap}

\section*{Preface}
\renewcommand{\thefootnote}{\fnsymbol{footnote}}

In the previous chapters, we established the groundwork for graph colouring, cluster graph formulation, and formulating a problem as a PGM. This chapter presents \textit{Strengthening PGMs: The Purge-and-merge Algorithm}~\cite{streicher2021strengthening}\footnote[2]{Sections of this work have been published in:
    \\
    S. Streicher and J. du Preez, “Strengthening Probabilistic Graphical Models: The Purge-and-merge
    Algorithm,” \textit{IEEE Access}, vol. 9, pp. 149 423–149 432, 2021.
}, which builds upon the previous work.

This publication expresses graph colouring from Chapter~\ref{gc-chapter} as constraint satisfaction and shows how to formulate general CSPs as PGMs. This process is illustrated on a few logic puzzles such as Sudoku, Fill-a-pix, Calcudoku, and Kakuro. 

The main contribution of the publication is a technique called purge-and-merge, which makes it possible to systematically simplify inference on complex CSPs. The algorithm is a combination of three established probabilistic techniques from Chapters~\ref{pgm-chapter} and~\ref{lc-chap}. More specifically, purge-and-merge extends on 
\begin{itemize}
    \item the LTRIP algorithm for building cluster graphs,
    \item applying belief propagation on graph colouring (and CSP) factors, and 
    \item merging factors into a joint space.
\end{itemize}
Furthermore, the technique successfully outperformed the PGM-based approaches mentioned in Chapter~\ref{gc-chapter} and some other approaches from the probabilistic reasoning literature.

\renewcommand{\thefootnote}{\arabic{footnote}}

\section*{Abstract}
    Probabilistic graphical models (PGMs) are powerful tools for solving systems of complex relationships over a variety of probability distributions. However, while tree-structured PGMs always result in efficient and exact solutions, inference on graph (or loopy) structured PGMs is not guaranteed to discover the optimal solutions~\cite[p391]{koller}. It is in principle possible to convert loopy PGMs to an equivalent tree structure, but this is usually impractical for interesting problems due to exponential blow-up~\cite[p336]{koller}. To address this, we developed the ``purge-and-merge'' algorithm. This algorithm iteratively nudges a malleable graph structure towards a tree structure by selectively \textit{merging} factors. The merging process is designed to avoid exponential blow-up by way of sparse structures from which redundancy is \textit{purged} as the algorithm progresses. We set up tasks to test the algorithm on constraint satisfaction puzzles such as Sudoku, Fill-a-pix, and Kakuro, and it outperformed other PGM-based approaches reported in the literature~\cite{BaukeH, GoldbergerJ, KhanS}. While the tasks we set focussed on the binary logic of CSP, we believe the purge-and-merge algorithm could be extended to general PGM inference.

\section{Introduction}
We have successfully created flexible probabilistic graphical model (PGM) structures to solve constraint satisfaction problems (CSPs) that cannot be solved with existing PGM inference techniques. This entailed the creation of an exact CSP solver that preserves all solutions. 

We did not set out to explore modern constraint satisfaction problem solving in general but rather to incorporate constraint satisfaction capabilities into PGMs. Central to this work is a PGM technique called purge-and-merge. It is the combination of three established probabilistic techniques: building cluster graphs, applying loopy belief propagation~\cite{dechter2010on}, and merging factors into a joint space. Together, these techniques enable purge-and-merge to allow the growth of factors via factor merging while also removing redundancies in the CSP problem space via loopy belief propagation. We can thus solve a range of CSPs that would be too intricate for either loopy belief propagation or factor merging. Our experimental study shows that purge-and-merge reliably solves problems too difficult for other belief-propagation approaches~\cite{streicher, BaukeH, GoldbergerJ, KhanS}.

Purge-and-merge provides higher-order reasoning for PGMs and constraint satisfaction. This technique would, therefore, be of benefit to any area that incorporates both these domains, such as
\begin{itemize}
    \item classification and re-classification problems -- e.g.\ image de-noising~\cite{koller}, scene classification~\cite{cvexample1}, and classification boosting in Chapter~\ref{lc-chap};
    
    \item image segmentation -- e.g.\ extracting superpixels using boundary constraints~\cite{zhao2020superpixels}; and
    
    \item hybrid reasoning -- e.g.\ solving the game of Clue by combining logical and probabilistic reasoning~\cite{clue2018} and solving a Sudoku visually by combining a handwriting input classifier with constraint satisfaction~\cite{mulamba2020hybrid}.
\end{itemize}

PGMs are tools that express intricate problems with multiple dependencies as graphs and PGM inference techniques such as message passing can be used to solve these graphs. As a result, PGMs are integral to a wide range of probabilistic problems~\cite{sucar2015probabilistic} such as medical diagnosis and decision making~\cite{wemmenhove2001inference}, object recognition in computer vision~\cite{cvexample1}, as well as speech recognition and natural-language processing~\cite{nlpexample}.

Constraint satisfaction, in turn, is classically viewed as a graph search problem that falls under the umbrella of NP-complete problems. It originated in the artificial intelligence (AI) literature of the 1970s, with early examples in Mackworth~\cite{Mackworth} and Laurière~\cite{Lauriere}. Broadly, a CSP consists of a set of variables $\mathcal{X} = \{X_1, X_2, \ldots, X_N\}$, where each variable must be assigned a value such that a given set of constraints (clauses) $\mathcal{C} = \{C_1, C_2, \ldots, C_M\}$ are satisfied. Typical applications of CSPs include resource management and time scheduling~\cite{cspplanning}, parity checking in error-correcting codes~\cite{paritycheck}, and puzzle games such as Sudoku, Killer-Sudoku, Calcudoku, Kakuro, and Fill-a-pix~\cite{csppuzzle}.

Many advances have been made in solving highly constrained PGMs, i.e.\ PGMs with a large number of prohibited outcomes specified in their factors. This includes PGMs constructed from CSP clauses. A popular approach is to transform such a PGM to another domain and then to solve it with tools specific to that domain. This includes converting PGMs to Boolean satisfiability problems (such as conjugate normal form (CNF)~\cite{cnfbool}), sentential decision diagrams (SDD)~\cite{sddchoi}, and arithmetic circuits (ACs)~\cite{acewebsite}. For example, the AC compilation and evaluation (ACE) system~\cite{acewebsite} compiles a factor graph or Bayes network into a separate AC, which can then be queried about the underlying variables. The drawback lies in the fact that an ACE query will yield a marginal for each queried variable, but it does not yield the joint distribution over all queried variables. ACE thus acts as a heuristic for selecting a single probable CSP outcome. 

In contrast, our CSP solver can return a joint solution to a CSP problem (i.e.\ find its joint distribution) using PGM techniques. 

There are trivial ways to formulate CSPs probabilistically and express them as PGMs \cite{streicher, MoonT, GoldbergerJ, KhanS, BaukeH, LakshmiA}, and although most of them are aimed at specific CSPs, they share the same basic approach. This amounts to (a) formulating the CSP clauses into PGM factors, (b) configuring the factors into a PGM graph structure, (c) applying belief propagation on this graph, and (d) using the most probable outcome as the solution to the CSP. \citet{dechter2010on} provide a bridge between CSPs and PGMs by proving that zero-belief conclusions made by loopy belief propagation reduce to an algorithm for generalised arc consistency in CSPs.

There are limitations, however. Goldberger~\cite{GoldbergerJ} highlights the difference between belief propagation (BP) with max-product and sum-product. They report that although max-product BP ensures the solution is preserved at all times, it is often hidden within a large spectrum of possibilities. This calls for additional search techniques. Meanwhile, sum-product BP acts as a heuristic to highlight a valid solution but can often highlight an incorrect one. Khan~\cite{KhanS} tries to improve on the success rate of sum-product BP by combining it with Sinkhorn balancing. Although they report an improvement, the system could still not reliably solve high-difficulty Sudoku puzzles. In Chapter~\ref{gc-chapter}, we suggested a sparse representation for factors and promoted the use of a cluster graph over the ubiquitous factor graph. However, although the cluster graphs improved the accuracy and execution time of the system, this approach is not reliable as a Sudoku solver or CSP solver in general.

The above approaches are all limited in one way or another. They are either ineffective in purging redundant search space -- due to their loopy PGM structure -- or they rely on an unreliable heuristic to select a probable solution. In this work, we propose techniques to sidestep these limitations and iteratively nudge the graph towards a tree-structured PGM while preserving the CSP solution.

Our proposed technique employs the purge-and-merge algorithm. Purge-and-merge starts by constructing a CSP into a cluster graph PGM with sparse factors. It then \textit{purges} redundancies from these factors by applying max-product belief propagation~\cite{dechter2010on} and thereby propagating zero-belief conclusions. Next, it \textit{merges} factors together to create cluster graphs that are closer to a tree structure. Finally, it constructs a new cluster graph from the factors. This process is repeated until a tree-structured cluster graph is produced. At this point, the exact solution to the CSP is found.

Purge-and-merge manages to reliably solve CSPs that are too difficult for the aforementioned approaches. We reason that a successful CSP approach, such as purge-and-merge, opens many new avenues for exploration in the field of PGMs. This may include hybrid models where rigid and soft constraints can be mixed. It may also be used in domains not previously suited for probabilistic approaches.

Our study is outlined as follows:
\begin{itemize}
    \item
    In Section~\ref{csp-sec-CSP}, we introduce CSP factors and show how they can be structured into a PGM. We also provide the design and techniques to build and solve a basic constraint satisfaction PGM.
    
    \item
    In Section~\ref{csp-sec:limitationsofpgms}, we investigate the limitations of PGMs as well as the trade-offs between the loopy-structured PGMs of small-factor scopes and the tree-structured PGMs of large-factor scopes.
    
    \item
    In Section~\ref{csp-sec:purgeandmerge-main}, we provide a factor clustering and merging routine along with the purging methods necessary for our purge-and-merge technique.
    
    \item
    In Section~\ref{csp-sec:experiments}, we evaluate purge-and-merge on a number of example CSPs such as Fill-a-pix and similar puzzles and compare them to the ACE system~\cite{acewebsite}.
\end{itemize}

We found that with the purge-and-merge technique, PGMs can solve highly complex CSPs. We, therefore, conclude that our approach is successful as a CSP solver and suggest further investigation into integrating constraint satisfaction PGMs as sub-components of more general PGMs.

\section{Constraint satisfaction using PGMs}\label{csp-sec-CSP}

In this section, we show how CSPs are related to PGMs. We express CSPs as factors, which can be linked in a PGM structure. We use graph colouring as an example and expand the idea to the broader class of CSPs.

Most constraint satisfaction problems are easily defined and verified, but they can be difficult to invert and solve. PGMs, by contrast, are probabilistic reasoning tools used to resolve large-scale problems in a computationally feasible manner. They are often useful for problems that are difficult to approach algorithmically -- CSPs being one such example.

\subsection{A general description of CSPs}
Constraint satisfaction problems are NP-complete. Nevertheless, they are of significant importance in operational research, and they are key to a variety of combinatorial, scheduling, and optimisation problems.

In general, constraint satisfaction deals with a set of variables $\mathcal{X} = \{X_1, X_2, \ldots, X_N\}$ and a set of constraints $\mathcal{C} = \{C_1, C_2, \ldots, C_M\}$. Each variable needs to be assigned a value from the variable's finite domain $\text{dom}(X_n)$, such that all constraints are satisfied. For example, if $X_n$ represents a die roll, then a suitable domain would be $\{1,2,3,4,5,6\}$. Furthermore, if we define a CSP where two die rolls, $X_1$ and $X_2$, are constrained to sum to a value of $10$, then the CSP solution $(X_1, X_2)$ consists of the possible value assignments $(4,6)$, $(5,5)$, and $(6,4)$.

\begin{figure}[h!]
	\centering
	\includegraphics[width=0.70\columnwidth]{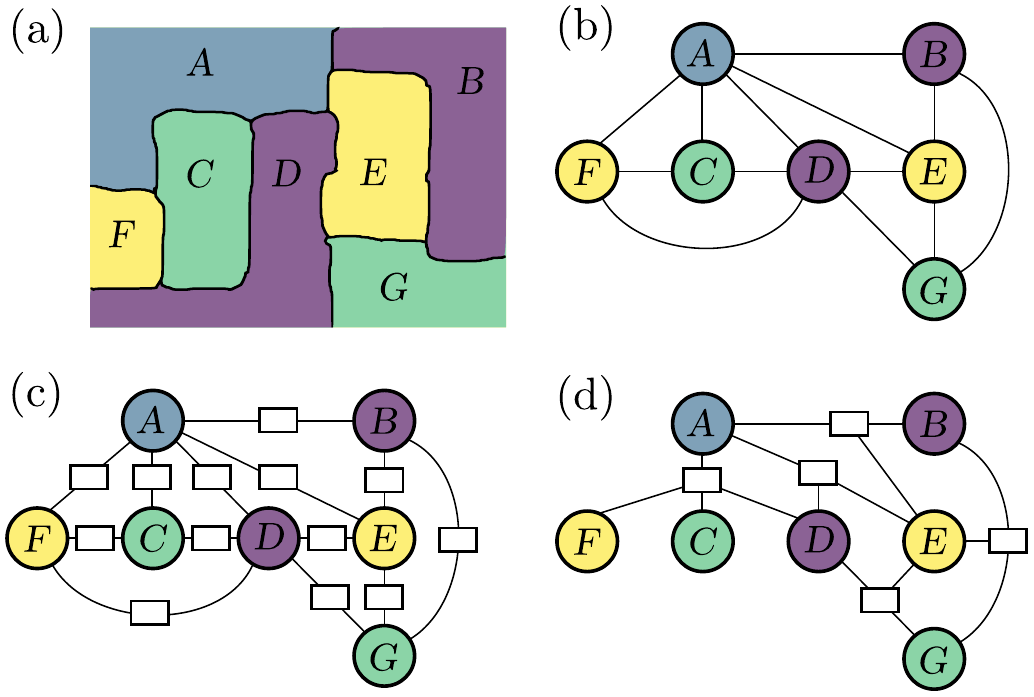}
	\caption{ (a) A map colouring example, with (b) its graph colouring representation, and (c) and (d) two different factor graph representations using rectangles to represent CSP clauses.
	}\label{csp-fig-mapcolouring}
\end{figure}

The CSP constraints can be visualised through a factor graph. This is a bipartite graph where the CSP variables are represented by variable nodes (circles) and the CSP clauses by factor nodes (rectangles). The edges of the graph are drawn between factor nodes and variable nodes, such that each factor connects to all the variables in its scope. The scope of a factor is the set of all random variables related to that factor. To illustrate, we present the map colouring example in Figure~\ref{csp-fig-mapcolouring}.
\begin{itemize}
    
    \item
    Figure~\ref{csp-fig-mapcolouring}(a) shows a map with bordering regions. These regions are to be coloured using only four colours such that no two bordering regions may have the same colour.
    
    \item
    In Figure~\ref{csp-fig-mapcolouring}(b), we represent this map as a graph colouring problem, where the regions are represented by nodes and the borders by edges.
    
    \item
    Figure~\ref{csp-fig-mapcolouring}(c) shows a factor graph where the factors represent the CSP clauses. Note that each of these factors has a scope of two variables.
    
    \item
    In Figure~\ref{csp-fig-mapcolouring}(d), we show that the problem can also be expressed equivalently by combining factors differently. Here we have multiple constraints captured by a single clause. As a result, we have fewer factors but larger factor scopes. (This example uses the maximal cliques in (b) as factors.)
    
\end{itemize}

\subsection{Factor representation}
\begin{figure}[h]
    \centering
    \hspace{-1em}\includegraphics[scale=0.9]{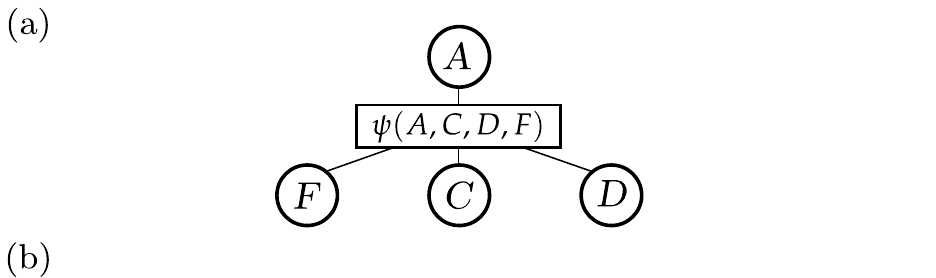}
    \\$\ $\\
    \vspace{-0.82em}
    \begin{tabular}{r c c c c |c }
        ~& $A$ & $C$ & $D$ & $F$ & $\psi(A,C,D,F)$  \\
        \cline{2-6}
        & 1\Tstrut & 2 & 3 & 4 & 1 \\
        & 1 & 2 & 4 & 3 & 1 \\
        & 1 & 3 & 2 & 4 & 1 \\
        & 1 & 3 & 4 & 2 & 1 \\
        & $\vdots$ & $\vdots$ & $\vdots$ & $\vdots$ & $\vdots$ \\
        & 4 & 3 & 2 & 1 & 1 \\
        \cline{2-6}
        & \multicolumn{4}{c|}{elsewhere\Tstrut} & 0 
    \end{tabular}
    \vspace{0.1em}
    \caption{(a) The map colouring clause $\{A,B,D,F\}$ from Figure~\ref{csp-fig-mapcolouring}, with factor $\psi(A, C, D, F)$, variables $A$,  $C$, $D$, $F$, and variable domains $\text{dom}(A)$ = $\text{dom}(B)$ = $\text{dom}(C)$ = $\text{dom}(D)$ = $\text{dom}(F)$ = $\{1,2,3,4\}$. (b) A sparse table explicitly listing the non-zero entries in $\psi(A, C, D, F)$, and assigning all other entries to be zero.}
    \label{csp-fig-probtable}
\end{figure}

A factor graph representation will only show the clauses, the variables, and the relationship between the clauses and variables. The details of these relationships, however, are suppressed. To fully represent the underlying CSP, each factor must also express the relationships implied by the associated constraint. We do so by assigning a potential function to each clause in order to encode all valid local assignments concerning that clause. These assignments are captured by sparse probability tables. The tables list each local possibility as a potential solution and assign a value to that possibility. For CSPs specifically, we work with binary probabilities, ascribing ``1'' to any (valid) possibility and ``0'' to any impossibility enforced by the constraint. As an example, see Figure~\ref{csp-fig-probtable} for the sparse table representing the factor scope $\{A,C,D,F\}$ from Figure~\ref{csp-fig-mapcolouring}(d). (The use of sparse tables as PGM factors is also referred to as flattening~\cite{dechter2010on}.)

It is worth noting that the factor graphs presented here are then not only a visually appealing representation for CSPs but are, in fact, PGMs. As such, PGM inference techniques such as loopy belief propagation and loopy belief update can be directly applied to these factor graphs. 

In order to perform belief propagation using sparse tables, it is important to implement some basic factor operations; most importantly, see Section~\ref{chap:basic-concepts} on factor multiplication, division, marginalisation, conditioning, and damping.

\subsection{PGM construction}\label{csp-sec:ltrip}
In essence, a PGM is a compact representation of a probabilistic space as the product of smaller, conditionally independent distributions. When we apply a PGM to a specific problem, we need to (a) obtain factors to represent these distributions, (b) construct a graph from them, and (c) use inference on this graph. In this section, we focus on graph construction.

A cluster graph is an undirected graph where:
\begin{itemize}
    \item each node $i$ is associated with cluster $\mathbf{C}_i \subseteq \mathcal{X}$, 
    \item each edge $(i, j)$ is associated with a separation set (sepset) $\mathbf{S}_{i,j} \subseteq \mathbf{C}_i \cap \mathbf{C}_j$, and 
    \item the graph configuration satisfies the running intersection property (RIP)~\cite[p347]{koller}.
\end{itemize}
The running intersection property requires that for all pairs of clusters containing a common variable, $X \in \mathbf{C}_i$ and $X \in \mathbf{C}_j$, there must be a unique path of edges, $(\hat{\imath}, \hat{\jmath})$, between $\mathbf{C}_i$ and $\mathbf{C}_j$ such that $X \in \mathbf{S}_{\hat{\imath}, \hat{\jmath}}\, \forall\, (\hat{\imath}, \hat{\jmath})$.

\begin{figure}[h]
    \centering
    \includegraphics[width=0.75\linewidth]{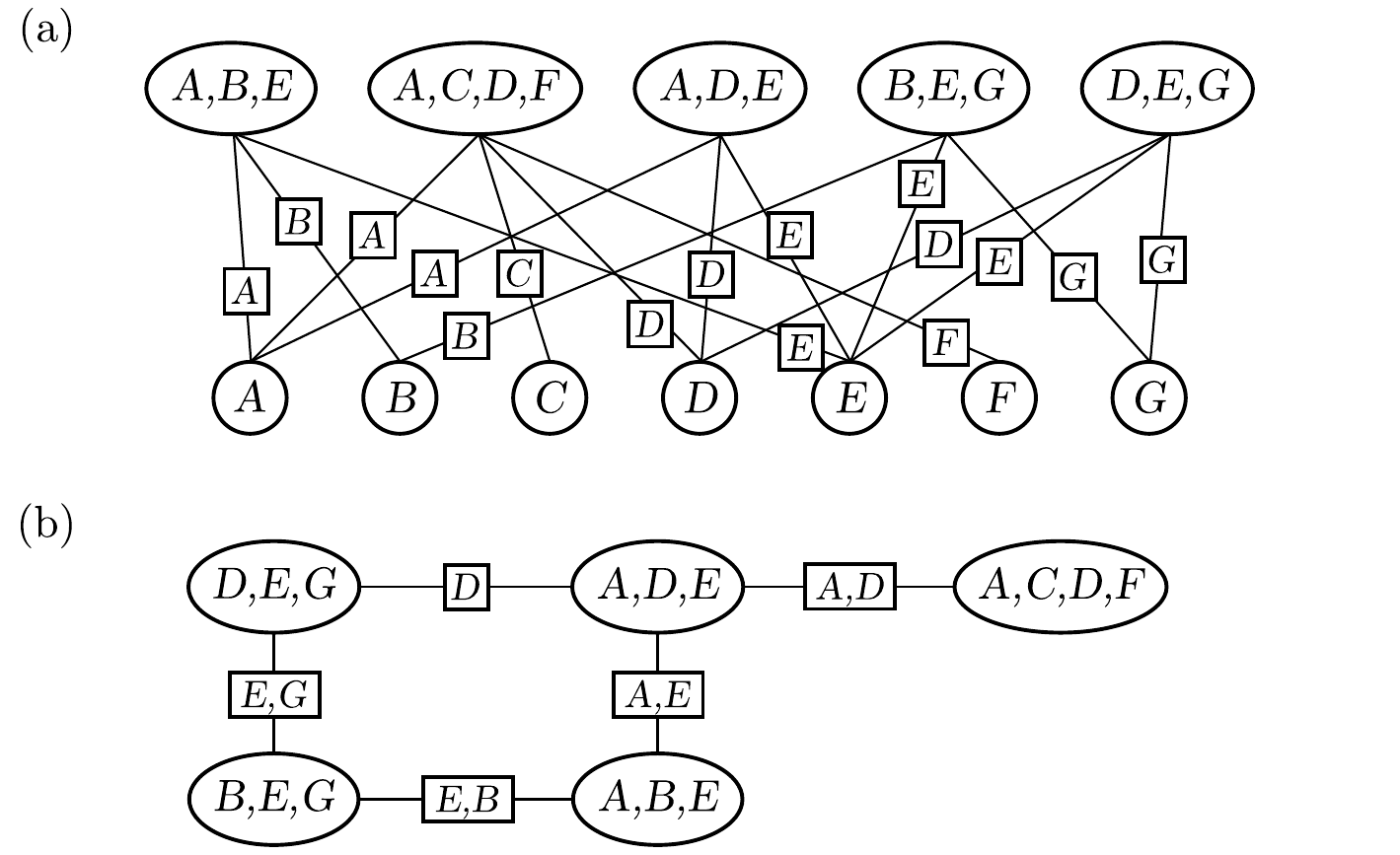}
    \caption{Two different cluster graph configurations for the map colouring example in Figure~\ref{csp-fig-mapcolouring}. (a) A Bethe graph configuration that satisfies RIP by connecting all CSP clauses via single-variable sepsets to singe-variable clusters. (b) A graph configuration generated by LTRIP with fewer edges and multivariate sepsets}
    \label{csp-fig:betheltrip}
\end{figure}

Figure~\ref{csp-fig:betheltrip} provides two examples of a cluster graph configuration for the CSP clauses in Figure~\ref{csp-fig-mapcolouring}. In (a), we have a trivial connection called a Bethe graph. This is a cluster graph with univariate sepsets, an equivalent of the factor graph in Figure~\ref{csp-fig-mapcolouring}(d). In (b) we have a cluster graph with multivariate sepsets. This graph is generated from the same factors as the Bethe graph but using the LTRIP algorithm. The result is also referred to as a join graph~\cite{dechter2010on}.

In Section~\ref{gc-sec:resultsandinterpretation}, we came to the conclusion that cluster graphs with multivariate sepsets have superior inference characteristics to factor graphs in terms of both speed and accuracy. The same is argued by Koller~\cite[p406]{koller}, where it is shown that cluster graphs are a more general case of factor graphs without the limitation of passing messages only through univariate marginal distributions. With factor graphs, correlations between variables are lost during belief propagation, which can have a negative impact on the accuracy of the posterior distributions and on the number of messages required for convergence.

The LTRIP algorithm (Algorithm~\ref{gc-alg:ltrip} from Section~\ref{gc-sec:ltrip}) is designed to configure factors into a valid cluster graph by following the RIP constraints. For each variable $X \in \mathcal{X}$, LTRIP builds a subgraph out of all clusters $\mathbf{C}_i$, where $X \in \mathbf{C}_i$. These subgraphs are then superimposed in order to construct the sepsets of the final graph. In summary, the algorithm states that for each variable $X \in \mathcal{X}$, do the following:
\begin{itemize}
    \item find all clusters $\mathbf{C}_i$ such that $X \in \mathbf{C}_i$,
    \item construct a complete graph over clusters $\mathbf{C}_i$,
    \item assign edge weights $w_{\hat{\imath}, \hat{\jmath}}$ to represent the similarity between neighbouring clusters,\footnote{The implementation from Section~\ref{gc-sec:ltrip} uses cluster intersections as edge weights: $w_{\hat{\imath}, \hat{\jmath}} = |\mathbf{C}_{\hat{\imath}} \cup \mathbf{C}_{\hat{\jmath}}|$. Other suggestions include mutual information or the entropy over the shared variables.}
    \item connect the graph into a maximum spanning tree by using an algorithm such as the Prim-Jarn\'{i}k algorithm~\cite{prim}, and
    \item populate the tree's edges with intermediate sepset results $\mathbf{S}^X_{\hat{\imath}, \hat{\jmath}} = \{X\}$.
\end{itemize}
After the sepset results are populated for each variable, the sepsets $\mathbf{S}_{\hat{\imath}, \hat{\jmath}}$ of the final graph are taken as the union of the intermediate sepset results  $\mathbf{S}_{\hat{\imath}, \hat{\jmath}} = \cup_{X \in \mathcal{X}} \mathbf{S}^X_{\hat{\imath}, \hat{\jmath}}$.

\begin{figure}[h]
    \centering
    \includegraphics[width=0.75\linewidth]{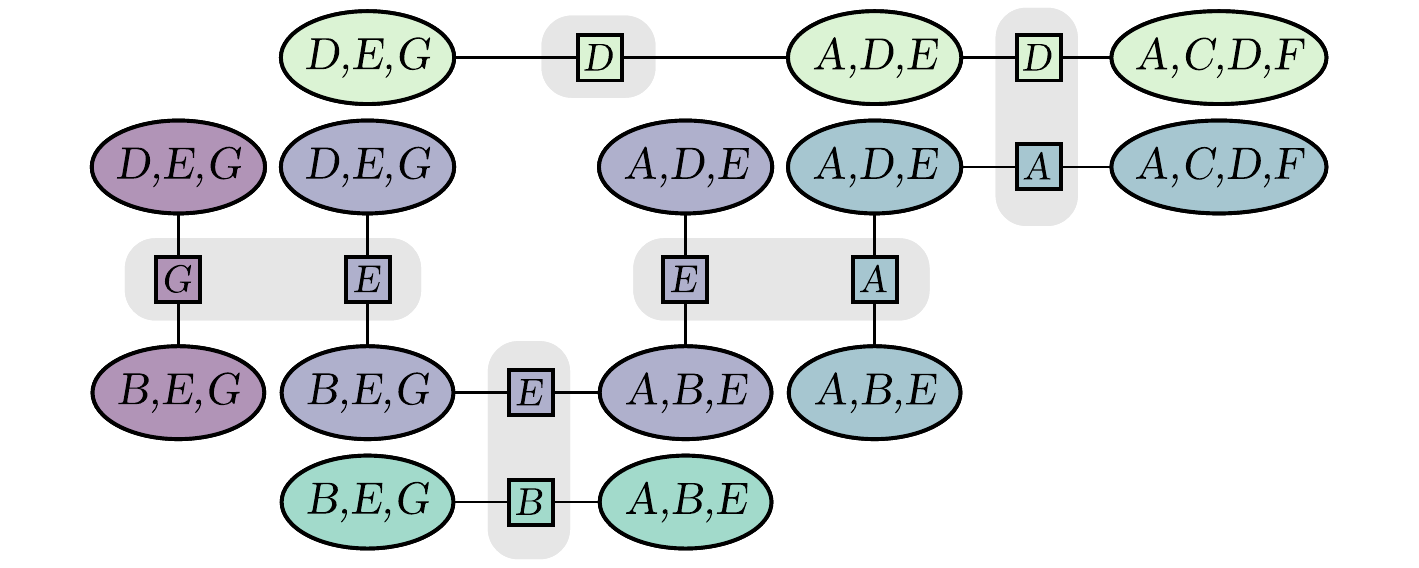}
    \caption{An example of applying the LTRIP algorithm in order to achieve the cluster graph construction from Figure~\ref{csp-fig:betheltrip}(b). For each variable $A, B, C, D, E, F,$ and $G$, a maximum spanning tree is constructed from its associated clusters and is populated with univariate sepsets. The resulting cluster graph is then created by taking the superposition of these intermediate trees.}
    \label{csp-fig:ltrip_construction}
\end{figure}

An example of the LTRIP algorithm for the graph in Figure~\ref{csp-fig:betheltrip}(b) can be seen in Figure~\ref{csp-fig:ltrip_construction}.

\subsection{PGM inference}\label{csp-sec:basig_pgm}
Our PGM approach extends the work in Chapters~\ref{pgm-chapter} and \ref{gc-chapter} and of flattened belief networks (i.e.\ sparse tables)~\cite{dechter2010on}. The specific design choices for our PGM implementation are as follows:
\begin{enumerate}
    \item
    The factors consist of sparse tables similar to those of Figure~\ref{csp-fig-probtable}.
    
    \item
    Graph construction is done using the LTRIP procedure.
    
    \item
    We use inference via belief \emph{update} (BU) message passing, also known as the Lau\-rit\-zen-Spie\-gel\-halter algorithm~\cite{lauritzen1988local}.
    
    \item
    We use the Kullback-Leibler divergence as a comparative
    metric (and deviation error metric) between distributions.
    
    \item
    Message passing schedules are set up according to residual belief propagation~\cite{elidan2012residual}. Messages are prioritised according to the deviation between a new message and the preceding message at the
    same location within the graph.
    
    \item
    Convergence is reached when the largest message deviation falls below a chosen threshold.
    
    \item
    Throughout the system, we use max-normalisation and max-marginalisation, as opposed to their summation equivalents.
\end{enumerate}

\citet{dechter2010on} proved that zero-belief conclusions made by loopy belief propagation are correct and equal to inducing arc consistency. This is true in the case of using both sum or max operations~\cite{GoldbergerJ, dechter2010on}. This means that the basic PGM approach does not guarantee a solution to the CSP, but it does guarantee that all possible solutions are preserved.

We found convergence to be faster with the max operations than with sum operations. Furthermore, the max operations maintain a unity potential for all non-zero table entries. This is in line with the constraint satisfaction perspective, where outcomes are either possible or impossible. Alternatively, if one is interested in a more dynamic distribution, the sum operations provide varying potentials that can be used as likelihood estimations~\cite{GoldbergerJ}.

Lastly, note that alternative message passing techniques such as warning propagation and survey propagation~\cite{braunstein2005survey} are available. These two approaches attempt to elevate the solution from the problem space but cannot guarantee that the solution is retained. Our interest is in pursuing an approach where the full solution space is preserved.

\section{The limitations of PGMs}\label{csp-sec:limitationsofpgms}

\begin{figure}[h]
    \centering
    \hspace{-1em}\includegraphics[scale=0.9]{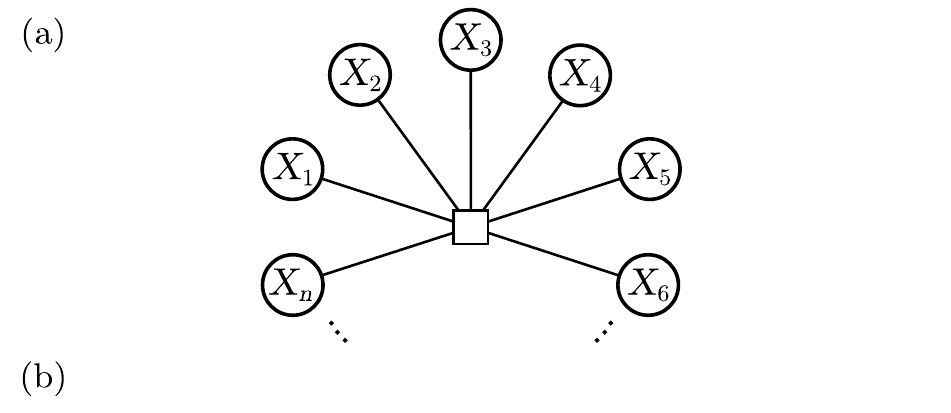}
    \\$\ $\\
    \vspace{-0.82em}
    \begin{tabular}{ c c c c c |c }
        $X_1$ & $X_2$ & $\cdots$ & $X_{n-1}$ & $X_n$ & $\psi(X_1,X_2,\cdots)$  \\
        \cline{1-6}
        1\Tstrut & 2 & $\cdots$ & n-1 &  n & 1 \\
        1 & 2 & $\cdots$ & n & n-1 & 1 \\
        $\vdots$ & $\vdots$ & $\ddots $ & $\vdots$  & $\vdots$ & $\vdots$ \Bstrut  \\
        n & n-1 & $\cdots$ & 2 & 1 & 1 \\
        \cline{1-6}
        \multicolumn{5}{c|}{elsewhere\Tstrut} & 0
    \end{tabular}
    \vspace{0.1em}
    \caption{ An example of an ill-suited problem with (a), its factor graph and (b), its sparse table containing $n!$ entries.}\label{csp-fig-factorlimit}
\end{figure}

One of the main limitations of constructing CSP potential functions is the resources required to encode them. If the potential functions are encoded as probability tables, then at least all the non-zero potentials need to be listed. Such a list can grow exponentially with the number of factor variables. Therefore, not all CSPs are suitable to be expressed as sparse tables. A trivial example of an ill-suited problem would be a graph colouring problem with $n$ fully connected nodes, as in Figure~\ref{csp-fig-factorlimit}. The full space of the problem is $n^n$ with $n!$ entries in the probability table. 

Inference on loopy graphs is non-exact; it cannot guarantee a complete reduction to the solution space of a CSP~\cite{dechter2010on}. In exchange, however, loopy graphs provide a great advantage: the ability to handle problems that would have required infeasibly large probability tables if constructed into tree-structured PGMs.

\begin{figure}[h!]
    \centering
    \includegraphics[width=0.80\columnwidth]{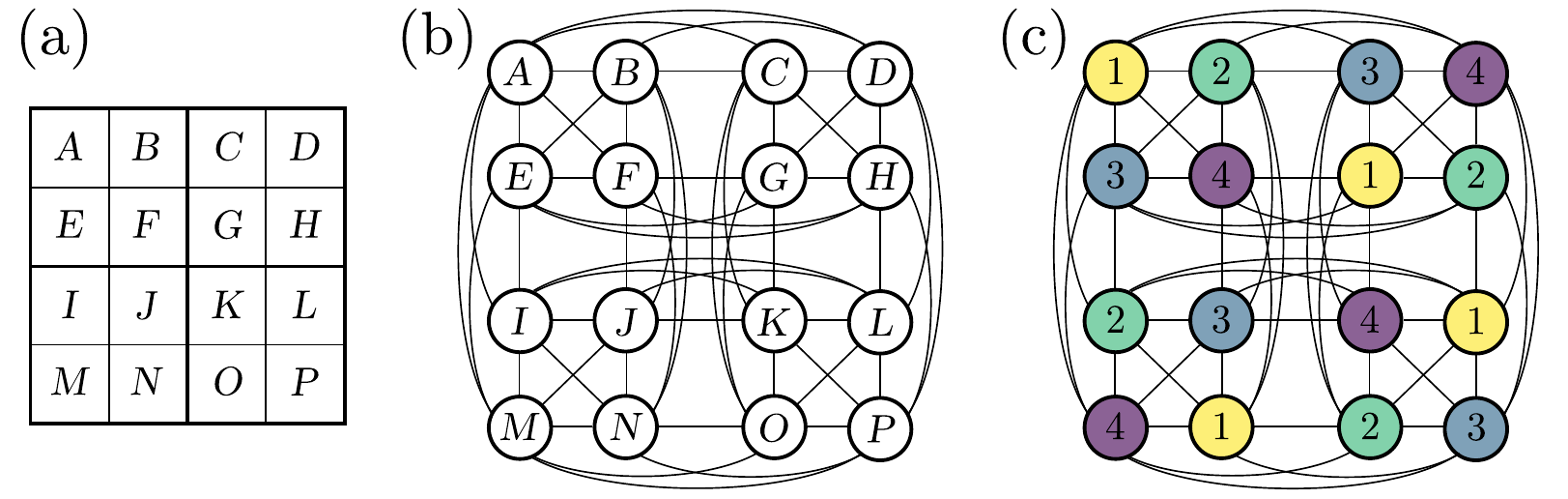}
    \caption{An example of a Sudoku puzzle as a graph colouring problem: (a) is a $4\times 4$ version of a Sudoku puzzle, (b) connects the Sudoku variables in an undirected graph, and (c) shows one solution (of many) to this particular problem.
    }
    \label{csp-fig:sudoku_colourgraph}
\end{figure}

Consider the Sudoku puzzle. A player is presented with a $9\times 9$ grid (with 81 variables) where each variable may be assigned a value of ``1'' to ``9'' and is constrained by the following 29 clauses: each row, each column, and each 3x3 non-overlapping subgrid may not contain any duplicates. Furthermore, a valid puzzle is partially filled with values such that only one solution exists. In Figure~\ref{csp-fig:sudoku_colourgraph}, we show the Sudoku-puzzle constraints as a graph colouring problem.

Since a valid Sudoku should only have a single solution, the posterior distribution, $P(X_1, X_2, \ldots, X_{81})$, can be expressed by a sparse table that covers the full variable scope and contains only a single entry. Each marginal distribution, $P(X_1,\ldots,X_9)$, $P(X_{10},\ldots,X_{18})$, $\dots$, would then also hold only a single table entry. Yet, after the pre-filled values are observed, and each factor is set up according to its local constraint (the no-duplicated rule), the prior distributions can result in tables with as many as $9!$ entries. Therefore, a Sudoku solver would have to reduce these large initial probability tables into single-entry posteriors. Before we attempt such a solver, let us first consider two cases -- (1) a loopy structure with small-factor scopes, and (2) a tree structure with large-factor scopes:
\begin{enumerate}
    
    \item
    For model (1), we build a loopy-structured PGM directly from the prior distributions. Each factor, therefore, has the same variable scope as one of the clauses. For an inference attempt to be successful, factors should pass information around until all sparse tables are reduced to single entries. In practice, however, the sparse tables are often reduced by very little. This is because inference on a loopy structure does not guarantee convergence to the final solution space~\cite{dechter2010on}.
    
    \item
    For model (2), we use the most trivial tree structure: multiply all factors together to form a structure with a single node. The resulting factor will now contain one single table entry as the solution. This approach will often (or rather usually) fail in practice since we cannot escape exponential blow-up. In the process of multiplying factors together, the intermediate probability tables first grow exponentially large before the system settles on this single-entry solution.
    
\end{enumerate}

Since we are confronted by the limitations of both small- and large-factor scopes, we propose a technique in the next section that mitigates these limitations.

\section{Purge-and-merge}\label{csp-sec:purgeandmerge-main}
In this section, we consider the various methods for purging factors and merging factors and combine these methods into a technique called purge-and-merge. It concludes with a detailed outline of the technique in Algorithm~\ref{csp-alg:purgeandmerge}.

\subsection{Factor merging}\label{csp-sec:factorclustering}
Our aim in merging factors is to build tree-structured PGMs and to be able to perform exact inference. This can result in exponentially larger probability tables, so it is necessary to approach this problem carefully.

One approach is to cluster the factors into subsets that will merge to reasonably sized tables. To pre-calculate the table size of a factor product is, unfortunately, as memory-inefficient as performing the actual product operation. While we, therefore, cannot use exact table size as a clustering metric, we have investigated three alternative metrics. We propose (1) variable overlap, as in the number of overlapping variables between factors, (2) an upper-bound shared entropy metric, and (3) a gravity analogy that is built on entropy metrics. These methods are experimentally tested in Section~\ref{csp-sec:experiments} (Figure{~\ref{csp-fig-histogram}), with the gravity analogy showing the most potential.

\subsubsection*{Variable overlap}
    For variable overlap, we define the attraction between two factors, $\psi_i$ and $\psi_j$, as
    \begin{equation*}
        a_{i, j} = \left| \mathcal{X}_i \cap \mathcal{X}_j \right|,
    \end{equation*}
    with $\mathcal{X}_i$ and $\mathcal{X}_j$ the scopes of $\psi_i$ and $\psi_j$ respectively. Note the 
    symmetrical relationship, in the sense that the attraction of  $\psi_j$ towards $\psi_i$ can be defined as $a_{i,j} = a_{i \leftarrow j} = a_{j \leftarrow i}$.

\subsubsection*{Upper-bound shared entropy}
    Upper-bound shared entropy is proposed as an alternative metric to variable overlap. The definition for the entropy of a set of variables $\mathcal{X}$ is
    \begin{equation*}
        \text{H}(\mathcal{X}) = \sum_{x \, \in \, \text{domain}(\mathcal{X})} -p(x) \log_2 p(x),
    \end{equation*}
    with a maximum upper bound achieved at the point where the distribution over $\mathcal{X}$ is uniform. This upper bound is calculated as
    \begin{equation*}
        \hat{\text{H}}\left(\mathcal{X}\right) = \log_2\left| \text{domain}(\mathcal{X}) \right|.
    \end{equation*}

    We use this definition to define the attraction between clusters $i$ and $j$ as the upper-bound entropy of the variables they share:
    \begin{equation*}
        a_{i, j} = \hat{\text{H}}(\mathcal{X}_i \cap \mathcal{X}_j),
    \end{equation*}
    with symmetrical behaviour $a_{i,j} = a_{i \leftarrow j} = a_{j \leftarrow i}$. Note that maximal entropy is used as a computationally convenient proxy for entropy since calculating shared entropy directly is as expensive as applying factor product.

\subsubsection*{Gravity method}
    For the gravity method, we use gravitational pull as an analogy for attraction:
    \begin{equation*}
        a_{i\leftarrow j} \propto m_i/ r_{i,j}^2.
    \end{equation*}
    The idea is to relate mass $m_i$ to how informed a factor is about its scope and distance $r_{i,j}$ to concepts regarding shared entropy.

    \vspace{0.3em}\noindent 
    \textbf{Pseudo-mass:}\, 
    Mass equation $\text{m}(\psi_i) = m_i$  is based on how informed factor $\psi_i$ is about its scope $\mathcal{X}_i$. To parse this into a calculable metric, we use the Kullback–Leibler divergence of the distribution of $\psi_i$ compared to a uniform distribution over $\mathcal{X}_i$.
    
    \vspace{0.3em}\noindent 
    \textbf{Pseudo-distance:}\, 
    As a distance metric, we want to register two factors as \textit{close together} if they have a large overlap and \textit{far apart} if they have little overlap. We also do not want this metric to be influenced by a factor's size or mass. 

    We arrived at a metric using the entropy of the joint distribution, normalised by the entropy of the variables shared between the factors. By using upper-bound entropy in our calculations, we arrive at distance
    \begin{equation*}
        r_{i,j} = \text{r}(\mathcal{X}_i, \mathcal{X}_j) = \log_2\left(\frac{\hat{\text{H}}(\mathcal{X}_i \cup \mathcal{X}_j)}{\hat{\text{H}}(\mathcal{X}_i \cap \mathcal{X}_j)}\right).
    \end{equation*}
    
    \vspace{0.3em}\noindent 
    \textbf{Attraction:}\, 
    Finally, we define the attraction of $\psi_j$ towards $\psi_i$ as analogous to acceleration
    \begin{equation*}
        a_{i\leftarrow j} = \frac{m_i}{r^2_{i, j}}.
    \end{equation*}

    Using the above metrics, we formulate a procedure for clustering our factors according to the mergeability between factors, as shown in Algorithm~\ref{csp-alg:clustering}. Although the algorithm is specialised for the gravity method, it can easily be adjusted for different attraction metrics.

\begin{algorithm}[h!]
	\caption{\ Factor Clustering}
	\label{csp-alg:clustering}
	\vspace{0.5em}
	\capfnt{Input:}\ \, Factors $\psi_1, \ldots, \psi_n$ and threshold $\hat{H}_\tau$.\vspace{0.3em} \\
	\capfnt{Output:}\ \, Clustered sets of factors $\mathcal{C}_1, \ldots, \mathcal{C}_m$, with property $\hat{\text{H}}(\text{vars in }\mathcal{C}_i) \leq  \hat{H}_\tau\, \forall\, \mathcal{C}_i$.
	\vspace{0.5em}
	\begin{algorithmic}[1]
		
		\State {\color{blue1}// Initialise clusters and attractions}
		\For {each factor index $i$}
		\State $\mathcal{C}_i$ := $\{\psi_i\}$
		\State $\mathcal{X}_i$ := variables of $\psi_i$
		\State $m_i$ := m$(\psi_i)$
		\EndFor
		\For {each $i,j$ pair where $\left| \mathcal{X}_i \cup \mathcal{X}_j \right| > 0$}
		\State $a_{i\leftarrow j}$ := ${m_i}/{\text{r}(\mathcal{X}_i, \mathcal{X}_j)}$
		\State $a_{j\leftarrow i}$ := ${m_j}/{\text{r}(\mathcal{X}_j, \mathcal{X}_i)}$
		
		\EndFor
		\State {\color{blue1}// Dynamically merge clusters together}
		\While {any $a_{i \leftarrow j}$ are still available}
		\newcommand*{\ii}{\hat{\imath}} %
		\newcommand*{\jj}{\hat{\jmath}} %
		\State $\ii, \jj$ := $\argmin_{i, j}(a_{i \leftarrow j})$
		\If {$\hat{\text{H}}(\mathcal{X}_{\ii} \cup \mathcal{X}_{\jj}) \leq \hat{H}_\tau$}
		\State $\mathcal{C}_{\ii}$ := $\mathcal{C}_{\ii} \cup \mathcal{C}_{\jj}$
		\State $\mathcal{X}_{\ii}$ := $\mathcal{X}_{\ii} \cup \mathcal{X}_{\jj}$
		\State $m_{\ii}$ :=  $m_{\ii} + m_{\jj}$
		\For{each $k\hspace{-0.17em}\neq\hspace{-0.17em}{\jj}$ where $\left| \mathcal{X}_{\ii} \cup \mathcal{X}_k \right|>0$}
		\State $a_{{\ii}\leftarrow k}$ := ${m_{\ii}}/{\text{r}(\mathcal{X}_{\ii}, \mathcal{X}_k)}$
		\State $a_{k\leftarrow {\ii}}$ := ${m_k}/{\text{r}(\mathcal{X}_k, \mathcal{X}_{\ii})}$
		\EndFor
		
		\State remove
		$\mathcal{C}_{\jj}$,
		$\mathcal{X}_{\jj}$ and
		$m_{\jj}$
		\State remove $a_{{\jj} \leftarrow l}$ and
		$a_{l \leftarrow {\jj}}$, for any index $l$
		\Else \textbf{\ then}
		\State remove $a_{\ii \leftarrow \jj}$
		\EndIf
		\EndWhile
		\State
		\Return all remaining $\mathcal{C}_i, \mathcal{C}_j, \ldots$
	\end{algorithmic}
	\vspace{0.3em}
\end{algorithm}

    Via this procedure, we can cluster factors $\psi_1, \psi_2, \ldots, \psi_n$  into clusters $\mathcal{C}_1, \ldots, \mathcal{C}_m$, where $m \leq n$. These clusters can then be incorporated into a PGM by calculating new PGM factors ${\psi}'_1, \ldots, {\psi}'_m$, by simply merging each cluster ${\psi}'_i = \prod_{\psi_j \in \mathcal{C}_i}\psi_j$.
    
    \subsection{Factor purging}\label{csp-sec:factorpurging}
    In this section, we show some methods for purging the probability tables of a constraint satisfaction PGM. We use the inference techniques from Section~\ref{csp-sec:basig_pgm} along with some additional purging techniques:  
        
    \textbf{Reducing variables:}\,  
    If for any factor $\psi_j$, a variable $X$ is uniquely determined to be $x_i$, i.e.\ there are no non-zero potentials with $X\neq x_i$ in that factor, then observe $X{=}x_i$ throughout all factors and remove $X$ from their scopes. This is a trivial case of node consistency~\cite{constraintnetworks}.
    
    \textbf{Reducing domains:}\, 
    Likewise, if any domain entry $x_i \in \text{dom}(X)$ has a zero probability to occur in a factor $\psi_j$, i.e.\ $P(X\equalss x_i|\psi_j)=0$ for any factor with $X$ in its scope, remove $x_i$ from $\text{dom}(X)$, and remove all probability table entries from the system that allowed for $X{=}x_i$.
    
    \textbf{Propagating local redundancies:}\, 
    For any two factors, $\psi_i$ and $\psi_j$, which have common variables, say $\{A, B, \ldots \}$, any zero outcome in $\psi_i$, i.e.\ $P(A\equalss a, B\equalss b, \ldots|\psi_i)=0$, should also be zero for $\psi_j$, i.e.\  $P(A\equalss a, B\equalss b, \ldots|\psi_j)=0$. The PGM inference from Section~\ref{csp-sec:basig_pgm} is, in fact, an algorithm to enforce this relationship, as \citet{dechter2010on} proved this to be an algorithm for generalised arc consistency.

    We can now combine these techniques along with our merging techniques to build a PGM-based CSP solver.
    
    \subsection{The purge-and-merge procedure}
    Having outlined all the building blocks needed for purge-and-merge, we can now describe the overall concept in more detail.
    
    We start our model with factors of small-variable scopes by using the CSP clauses directly. We then incrementally transition towards a model with larger-factor scopes by clustering and merging factors. More specifically, we start with a PGM of low-factor scopes, purge redundancies from this model, progress to a model of larger-factor scopes, and purge some more redundancies. We continue this process until our PGM is tree-structured and thus yields an exact solution to the CSP.

    This incremental-factor growth procedure dampens the exponential blow-up of the probability tables and allows the model to incrementally reduce the problem space. The full procedure is outlined in Algorithm~\ref{csp-alg:purgeandmerge}.
    
    \begin{algorithm}[h!]
        \caption{\ Purge-and-Merge}
        \label{csp-alg:purgeandmerge}
        \vspace{0.5em}
        \capfnt{Input:}\ \, Set of factors $\mathcal{F} =\{\psi_1, \ldots, \psi_n\}$.         \vspace{0.3em}\\
        \capfnt{Output:}\ \, Solved variables $\mathcal{X}_s = \{x_i, \ldots \}$ and
            solved factors $\mathcal{F}' = \{{\psi}'_i, \ldots \}$.
        \vspace{0.5em}
        \begin{algorithmic}[1]
            \State $\mathcal{X}_s = \{\}$
            \While{return conditions not met}
            \State $\hat{H}_\tau$ := an increasingly larger threshold
            \State {\color{blue1}// Factor clustering from Algorithm~\ref{csp-alg:clustering}: }
            \State $\mathds{C}$ := Factor-Clustering($\mathcal{F}$, $\hat{H}_\tau$)
            \State $\mathcal{F}'$ := $\{(\prod_{\psi_i \in \mathcal{C}} \psi_i)\,$ for each $\mathcal{C} \in \mathds{C} \}$
            \State {\color{blue1}// LTRIP and LBU from Section~\ref{csp-sec:basig_pgm}:}
            \State $\mathcal{G}$ := LTRIP($\mathcal{F}'$)
            \State $\mathcal{F}'$ := Loopy-Belief-Update($\mathcal{G}$)
            \State {\color{blue1}// Domain reduction from Section~\ref{csp-sec:factorpurging}:}
            \State Reduce-Domains($\mathcal{F}'$)
            \State {\color{blue1}// Variable reduction from Section~\ref{csp-sec:factorpurging}:}
            \State $\mathcal{X}$ := Reduce-Variables($\mathcal{F}'$)
            \State $\mathcal{X}_s$ := $\mathcal{X}_s \cup \mathcal{X}$\ \, {\color{blue1}// add solved variables}
            \If {$\mathcal{G}$ is a tree structure}
            \State return $\mathcal{X}_s$, $\mathcal{F}'$
            \EndIf
            \State $\mathcal{F}$ := $\mathcal{F}'$
            \EndWhile
        \end{algorithmic}
    \end{algorithm}

    \subsection{Algorithmic consistency}
    As a final reflection on purge-and-merge, we state that all steps taken by this algorithm are correct and result in a consistent CSP solver. Constraint satisfaction falls in the problem class of NP-complete~\cite{constraintnetworks}, and any general CSP solvers such as purge-and-merge must, therefore, be at least NP-complete.
    
    \citet{dechter2010on} provide a prove that loopy belief propagation performed on cluster graphs with flattened tables and hard constraints reduces to generalised arc consistency. They also prove that zero-belief conclusions converge within $\mathcal{O}(n \cdot t)$ iterations of loopy belief propagation and that the algorithm results in a complexity of $\mathcal{O}(r^2 t^2 \log t)$ -- where $n$ is the number of nodes in the cluster graph and $t$ bounds the size of each sparse table. These metrics are, however, not particularly useful for purge-and-merge since the values $n$ and $t$ are expected to change as the algorithm progresses.
    
    The merging of factors is exponential in time complexity, with some merge orders performing better than others~\cite[p287]{koller}. Finding an optimal merge order is, unfortunately, an NP-complete problem and is as difficult as the actual inference~\cite{cozman2000generalizing}. Some algorithms, such as variable elimination, can aid in this process~\cite[p287]{koller}. With variable elimination, the merging order is determined according to the marginalisation of each variable from the factors and the predicted effect it will have on the system as a whole. It should be noted that factor multiplication is commutative, and any merge order converges to the same solution.
    
    Since purge-and-merge is a combination of belief propagation and factor merging, the full algorithm is bounded by factor merging and is thus also NP-complete. Purge-and-merge does, however, mitigate between loopy belief propagation and factor merging with the aim of preventing exponential blow-up in the factor-merging process.

    In the next section, we investigate the performance of this procedure by applying it to a large number of CSP puzzles.  
    
    \section{Experimental study of purge-and-merge}\label{csp-sec:experiments}
    In this section, we investigate the reliability of the purge-and-merge technique by solving a large number of constraint satisfaction puzzles. To compare results, we include Sudoku datasets used in other constraint satisfaction PGM reports~\cite{MoonT, GoldbergerJ, KhanS, BaukeH, LakshmiA}, as well as the most difficult Sudoku puzzles currently available~\cite{champagne}.
    
    \subsection{Puzzle dataset}
    The Sudoku community has developed a large database of the hardest 9x9 puzzles~\cite{champagne} known to literature. The result is a unique collaborative research effort, spanning over a decade, using the widely accepted criterion of the Sudoku explainer rating (SER). This rating is applied by solving a puzzle using an ordered set of 96 rules ranked according to complexity. From the combination of rules required to solve the puzzle, the most difficult one of these rules is used as a hardness measure. The validity of SER as a difficulty rating is discussed by Berthier~\cite{csppuzzle}. They found that SER highly correlates to external pure-logic ratings and can thus be used as a proxy for puzzle complexity. They do note that SER is not invariant to puzzle isomorphisms, i.e.\ two puzzles from the same validity-preserving group~\cite{russell2006mathematics} can result in two different ratings.
    
    In addition to the above set, we also compiled a database of constraint puzzles from various other sources to be used as tests. We verified each puzzle to have valid constraints using either PicoSat~\cite{picosat} or Google's OR-Tools~\cite{ortools}. All puzzles are available on GitHub~\cite{Streichergithub}, and were sourced as follows:
    \begin{itemize}
        \item
        1000 \textbf{Sudoku} samples from the SER-rated set of the most difficult Sudoku puzzles, curated by Champagne~\cite{champagne},
        
        \item
        all 95 \textbf{Sudokus} from the Sterten set~\cite{sterten} used in Khan~\cite{KhanS} and in Section~\ref{gc-sec:experiments},
        
        \item
        all 49151 \textbf{Sudokus} with 17-entries from the Royle's 2010 set~\cite{RoyalG} (an older subset of roughly 350000 puzzles was available to Goldberg~\cite{GoldbergerJ} and Bauke~\cite{BaukeH}),
        
        \item
        10000 \textbf{Killer Sudokus} from \url{www.krazydad.com} (labelled according to five difficulty levels).
        
        \item
        4597 \textbf{Calcudokus} of size $9\mkern-3mu \times \mkern-3mu 9$ from \url{www.menneske.no},
        
        \item
        6360 \textbf{Kakuro} puzzles from \url{www.grandgames.net},
        
        \item
        2340 \textbf{Fill-a-Pix} puzzles from \url{www.grandgames.net}, and
        
        \item
        a mixed set of fairly high difficulty, with one of each of the above puzzle types.
    \end{itemize}

    \subsection{Clustering metrics}
    Section~\ref{csp-sec:factorclustering} listed three metrics for the purge-and-merge procedure, namely (1) variable overlap, (2) upper-bound shared entropy, and (3) the gravity method. In order to select a well-adapted clustering method, we compared these three metrics on the Champagne dataset. Our approach was to allow purge-and-merge to run for $10s$ under the different clustering conditions and to then report on the largest table size for that run.
    
    Under the naïve variable overlap metric, none of the puzzles came to convergence. When investigating further and re-running the first 10 puzzles without time restriction, they all ran out of physical memory. This is not surprising, as this metric does not account for the domain sizes of the variables, which can have a considerable impact on table size.
    
    Of the remaining metrics, the gravity method had a $100\%$ convergence rating within the $10s$ threshold, whereas upper-bound shared entropy had a $53\%$ rating. Compared to upper-bound shared entropy, the gravity method also resulted in a smaller maximum table size in $74.7\%$ of cases. A histogram representing the maximum table size for each run can be seen in Figure~\ref{csp-fig-histogram}.
    
    \begin{figure}[h!]
        \centering
        \includegraphics[width=0.80\columnwidth]{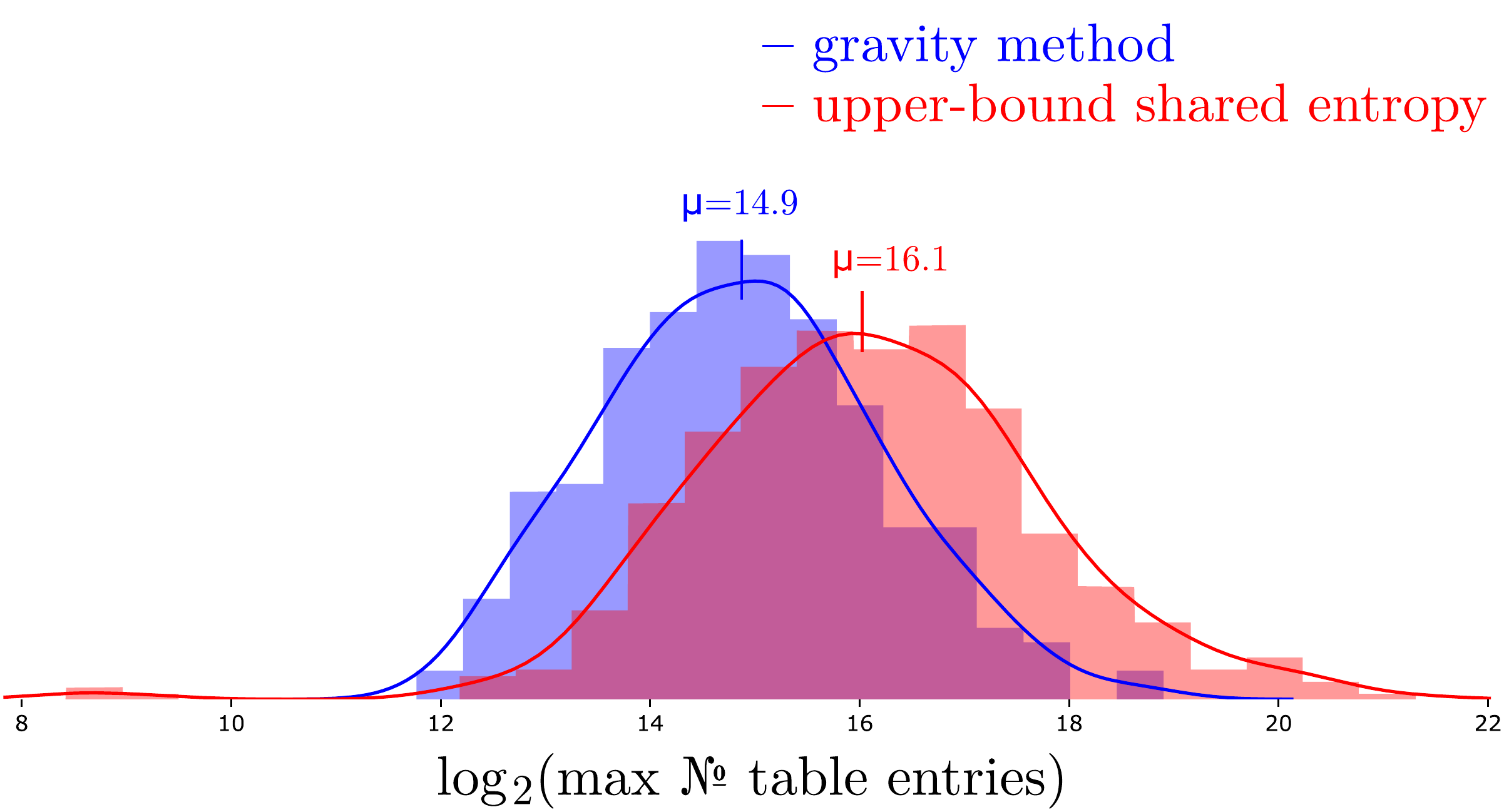}
        \caption{The maximum table produced in a purge-and-merge run using two different clustering methods: the upper-bound shared entropy method and the gravity method. All tests are run on the Champagne dataset. Only runs where both methods resulted in a convergence within $10s$ are displayed.}\label{csp-fig-histogram}
    \end{figure}
    
    Since the gravity method performed better than the other metrics, we opted to use it in all further purge-and-merge processes.
    
    \subsection{Purge-and-merge}
    All tests were executed single-threaded on an Intel$^\text{\textregistered}$  Core$^\text{TM}$ i7-3770K, with a rating of 3.50GHz and 4 cores / 8 threads in total. The purge-and-merge algorithm is available on GitHub~\cite{Streichergithub} as a Linux binary with a command-line interface for running any of the puzzles used in our tests.

    \begin{figure}[h!]
        \centering
        \includegraphics[width=1\columnwidth]{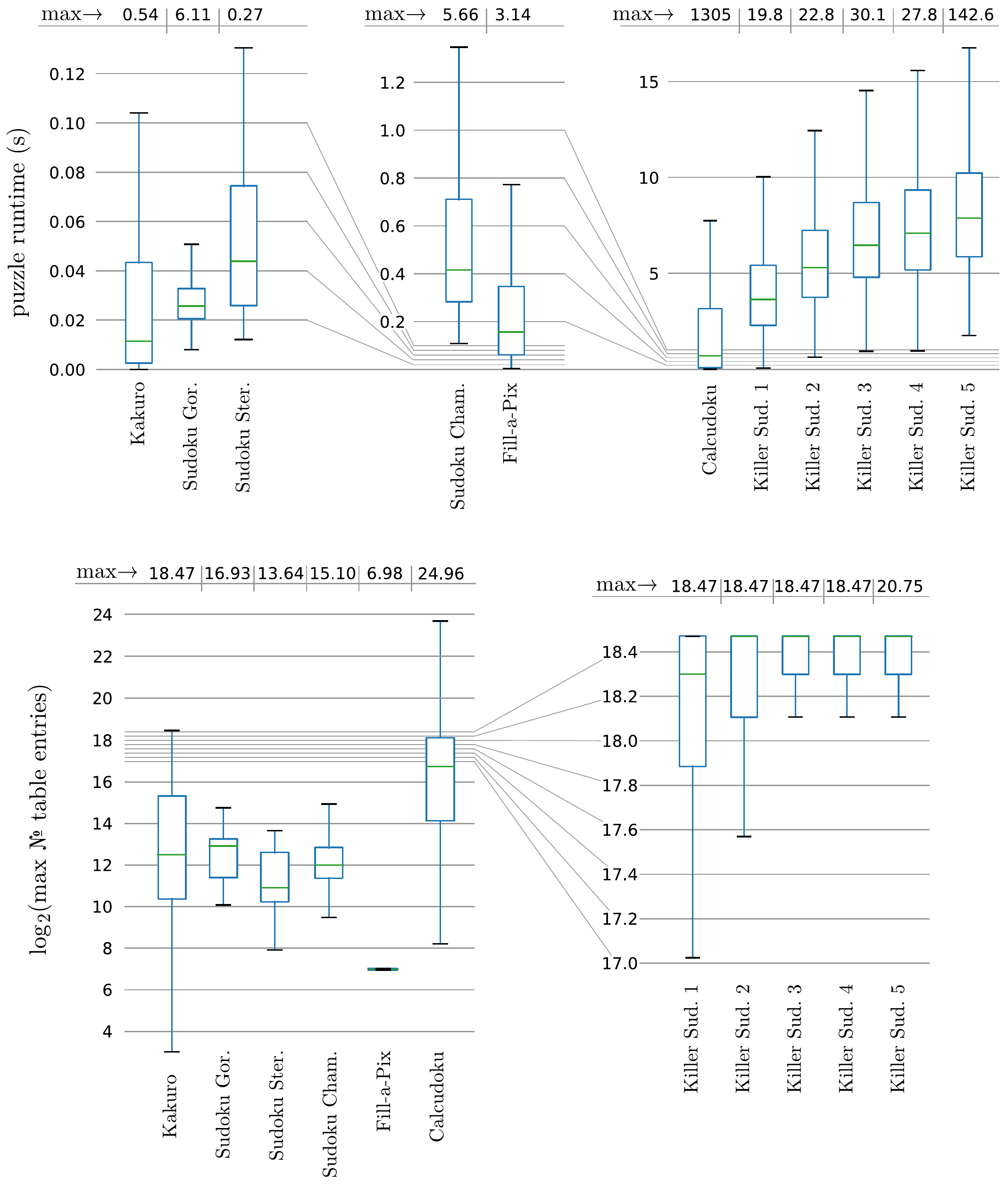}
        \caption{Runtime and size metrics for the purge-and-merge approach. The size metric indicates maximum entropy for any given factor during a purge-and-merge run, that is $\log_2(\text{maximum factor entries})$. The different Killer Sudoku sets are split according to reported difficulty, and the $1.4\%$ of unsuccessful Calcudoku runs are not included in these plots. \vspace{-0.7em}
        }\label{csp-fig-plotresults}
    \end{figure}
    
    Purge-and-merge can solve all the Sudoku, Killer Sudoku, Kakuro, and Fill-a-Pix puzzles we have provided. In the case of Calcudoku, $1.4\%$ of the puzzle instances reached the machine's physical RAM limitation of $32$Gb. This indicates that purge-and-merge deals better with large numbers of small factors, as with Kakuro, rather than a small or medium number of large factors, as with Calcudoku and Killer Sudoku. Runtime metrics for the various puzzles can be seen in Figure~\ref{csp-fig-plotresults}.

    Compared to the other available PGM approaches, purge-and-merge is the only method to achieve a $100\%$ success rate with all the Sudoku puzzles it encountered. Moreover, purge-and-merge was tested on more cases than what is reported in any of the comparable literature.
    
    If we compare purge-and-merge to the results in Section~\ref{gc-sec:experiments} for the Sterten~\cite{sterten} set, purge-and-merge is slower (see Figure~\ref{csp-fig-plotresults}). However, that approach is equivalent to a single ``purge'' step in purge-and-merge. The full purge-and-merge method obtained a success rate of $100\%$, whereas the success rate in Section~\ref{gc-sec:experiments} is only $36.8\%$.
    
    In comparison with the other Sudoku PGM literature, purge-and-merge and Chapter~\ref{gc-chapter} are the only PGM approaches that ensure the full CSP solution space is preserved (i.e.\ no valid solutions are lost). Additionally, purge-and-merge allows the scope of the solver to increase up to the point where only the solution space is left.
    
    The solution space is not preserved in the PGM approaches of Khan~\cite{KhanS}, Goldberger~\cite{GoldbergerJ}, and Bauke~\cite{BaukeH}; instead, they use sum-product BP to seek out a single likely solution from the problem space. Khan~\cite{KhanS} provides us with a comparison between these three approaches, as shown in Table~\ref{csp-tab-comparison}. From this table, it is clear that these reported PGM approaches are not well suited for Sudoku puzzles of medium and higher difficulty.
    
    \begin{table}[h!]
        \centering
            \begin{tabular}{|c|c|c|}
                \cline{1-3}
                Research &  Approach & \makecell{Reported accuracy \\ for $9\times 9$ Sudokus}  \\ \cline{1-3}
                
                \multicolumn{1}{ |c  }{\multirow{2}{*}{Bauke~\cite{BaukeH}} } &
                \multicolumn{1}{ |c| } {Sum-Product} & $53.2\%$      \\ %
                \multicolumn{1}{ |c  }{} &
                \multicolumn{1}{ |c| } {Max-Product} & $70.6\%$      \\ \cline{1-3}

                \multicolumn{1}{ |c  }{\multirow{3}{*}{Goldberger~\cite{GoldbergerJ}} } &
                \multicolumn{1}{ |c| } {Sum-Product} & $71.3\%$   \\ %
                \multicolumn{1}{ |c  }{}                        &
                \multicolumn{1}{ |c| } {Max-Product} & $70.7\% - 85.6\%$   \\ %
                \multicolumn{1}{ |c  }{}                        &
                \multicolumn{1}{ |c| } {Combined Approach} &  $76.8\% - 89.5\%$  \\ \cline{1-3}

                \multicolumn{1}{ |c  }{\multirow{2}{*}{Khan~\cite{KhanS}} } &
                \multicolumn{1}{ |c| } {Khan with 40 iterations} & $70\%$     \\ %
                \multicolumn{1}{ |c  }{} &
                \multicolumn{1}{ |c| } {Khan with 200 iterations} & $95\%$      \\ \cline{1-3}
                
                \multicolumn{1}{ |c  }{\multirow{1}{*}{This paper}} &
                \multicolumn{1}{ |c| }{ Purge-and-merge} & $100\%$   \\ \cline{1-3}
            \end{tabular}
        \caption{The success rate of various PGM approaches on Sudoku puzzles, originally compiled by Bauke~\cite{BaukeH}. Note that this applies to puzzles far easier than our expansive set, which also includes the current most difficult Sudoku set~\cite{champagne}.} \label{csp-tab-comparison}
    \end{table}

    \section{Comparison to the ACE system}
    
    The ACE system~\cite{acewebsite} is a related system for solving constraint satisfaction PGMs. ACE works by compiling Bayes networks and other factor graph representations into an arithmetic circuit, which can then be used to answer queries about the input variables. 
    
    ACE focuses on the marginals of the variables of the system and not on finding the joint distribution of the system. This distinction is important -- purge-and-merge produces all the solutions to a CSP, whereas ACE will only report on the marginal of each variable. 
    
    To illustrate, if we take the first 10 puzzles from the Champagne dataset and arbitrarily remove a known entry, purge-and-merge finds $426$, $380$, $917$, $799$, $77$, $476$, $454$, $1754$, $777$, and $796$ answers for each puzzle, respectively. ACE, on the other hand, only reports on the domain of each unknown variable and is, therefore, unable to find any valid solution.
    
    ACE approaches the problem in two stages. It first compiles a network along with its unknown variables into an arithmetic circuit. Then it uses the compiled network to answer multiple queries with respect to the unknown variables. Note that a single ACE circuit to represent all Sudoku puzzles is too large to fit in 32GB of memory due to the large number of possible solutions $\approx 6.67 \times 10^{21}$~\cite{numberofsudokus}.
    
    To compare ACE with purge-and-merge, we parsed all the Champagne puzzles into ACE-compliant structures and then compiled each structure into an ACE circuit. In plotting the result in Figure~\ref{csp-fig-histogram_ace}, we discarded the loading and query times since all evidence was already incorporated into the structure.
    
    \begin{figure}[h!]
        \centering
        \includegraphics[width=0.80\columnwidth]{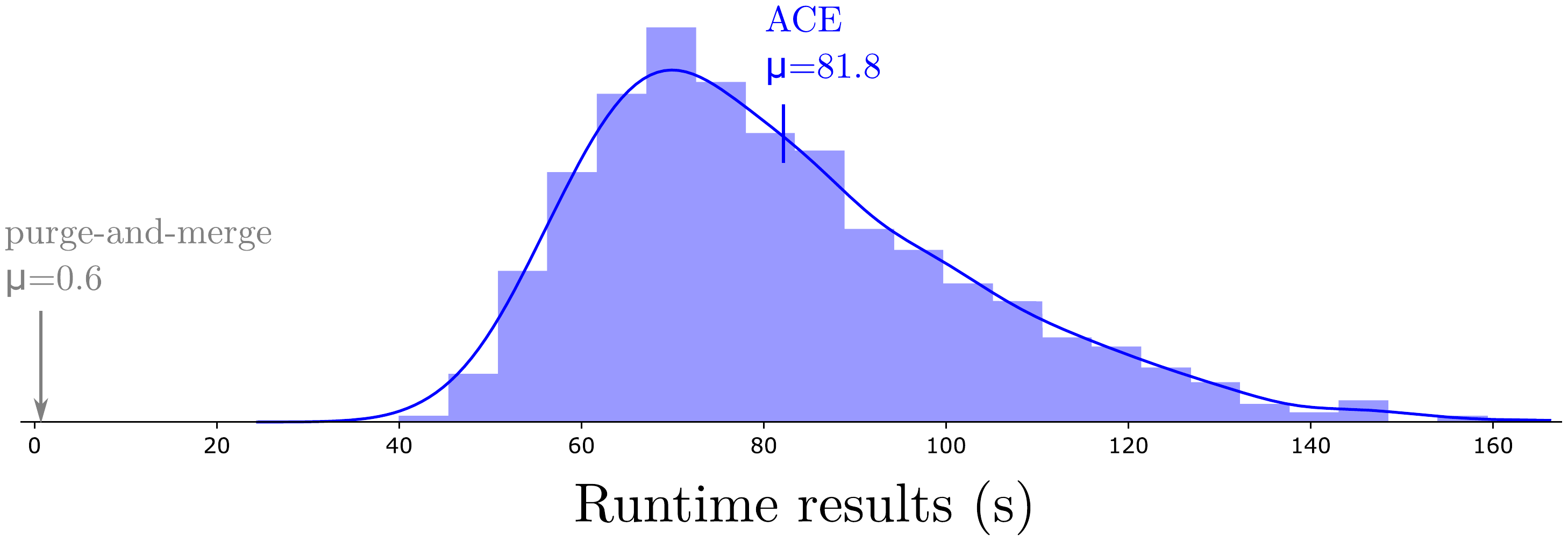}
        \caption{ACE runtime on the Champagne dataset. Only the network's compile times were recorded since the query times were negligible. For comparison, the average purge-and-merge runtime is indicated.}\label{csp-fig-histogram_ace}
    \end{figure}

    \section{Conclusion and future work}
    
    In general, the factors in a PGM can be linked up in different ways, resulting in different graph topologies. If such a graph is tree-structured, inference will be exact. However, more often than not, the graph structure will be loopy, which results in inexact inference~\cite[p381]{koller}. Transforming a loopy graph into a tree structure, unfortunately, is not always feasible -- in all but the simplest cases, the resultant hyper-nodes will exponentially blow up to impractical sizes. Hence we are usually forced to work with message passing on a loopy graph structure. 
    
    The ubiquitous factor graph is the structure most frequently encountered in the literature -- its popularity presumably stems from its simple construction. Previous work has shown that inference on factor graphs is often inferior to what can be obtained with more advanced graph structures such as cluster graphs. Nodes in cluster graphs typically exchange information about multiple random variables, whereas a factor graph is limited to sending only messages concerned with single random variables~\cite[p406]{koller}. The LTRIP algorithm enables the automatic construction of valid cluster graphs. Despite their greater potency, however, they might still be too limited to cope with complicated relationships. 
    
    In our current work, we extend the power of cluster graphs by dynamically reshaping the graph structure as the inference procedure progresses. Semantic constraints discovered by the inference procedure reduce the entropy of some factors. Factors with high mutual attraction can then be merged without necessarily suffering an exponential growth in factor size. The LTRIP algorithm reconfigures a new structure that becomes progressively more sparse over time. When the graph structure morphs into a tree structure, the process stops with an exact solution. We refer to this whole process as purge-and-merge. 
    
    Purge-and-merge is especially useful in tasks that, despite an initially huge state space, ultimately have a small number of solutions. By hiding zero-belief conclusions from memory, purge-and-merge can perform calculations on subspaces within an exponentially large state space.
    
    The purge-and-merge approach is not suited to tasks where the number of valid solutions would not fit into memory, as this would preclude a sufficient reduction in factor entropy. However, as the above results show, purge-and-merge enabled us to solve a wide range of problems that were previously beyond the scope of PGM-based approaches. 
    
    In comparison with ACE, we find purge-and-merge more suited to constraint sat\-is\-fac\-tion problems with multiple solutions, as well as puzzles with a problem space too large to be compiled into a single ACE network. 
    
    Our current approach relies on the increased sparsity of the resultant graphs to gradually nudge the system towards a tree structure. In future work, we intend to control that process more actively. This should result in further gains in efficiency, and it is our hope that it will conquer the couple of Calcudokus that still elude us.

\chapter{Conclusion and future implications}
The aim of this dissertation was to improve the PGM literature by providing accessible algorithms and tools for improved PGM inference. 
To this end, our main contributions are
\begin{itemize}
    \item a comparative study between cluster graphs and factor graphs,
    \item boosted land cover classification as a practical application for using PGMs in the field of cartography, and
    \item purge-and-merge as a PGM formulation and technique for solving constraint satisfaction problems too complex for the traditional PGM approach.
\end{itemize}

This dissertation presented an overview of probabilistic graphical models (Chapter~\ref{pgm-chapter}) and managed to build on that basis (Chapter~\ref{gc-chapter}) by configuring graph colouring factors into potential functions, providing a means for constructing cluster graphs from these factors, and comparing the performance of these graphs to the ones prevalent in the current literature. Through experimental results, we established that the cluster graphs produced by LTRIP have superior inference characteristics to the ones prevalent in the current literature.

Furthermore, these tools (established in Chapters~\ref{pgm-chapter} and~\ref{gc-chapter}) were shown to be effective in other domains. For example, we applied our PGM formulation to a problem in cartography, land cover classification boosting, in Chapter~\ref{lc-chap}. In this, we illustrated how to approach relational problems as probabilistic reasoning problems and formulated them using PGMs. Our approach managed to reclassify a test region in Garmisch-Partenkirchen, Germany, and boosted the overall classification accuracy compared to an independent reference land cover dataset. This formulation can effectively be applied to a wide range of other probabilistic reasoning problems. We suggest using it for image de-noising~\cite{koller}, extracting superpixels~\cite{zhao2020superpixels}, and natural-language processing~\cite{nlpexample}, for instance.
 
Lastly, the research reported in this dissertation contributed to the development of a more optimised approach for higher-order probabilistic graphical models applied to constraint satisfaction problems. This was presented in the form of an algorithm named purge-and-merge. Purge-and-merge extends the power of cluster graphs by dynamically reshaping a PGM graph structure as the inference procedure progresses. Inference is done via belief propagation and reduces the entropy of the factors. This allows factors to be merged without necessarily suffering an exponential growth in factor size. This routine is iteratively applied until a tree-structured graph is reached and exact inference is guaranteed. Purge-and-merge hides zero-belief conclusions from memory and can thereby perform calculations on subspaces within an exponentially large state space. It is especially useful for tasks that, despite an initial huge state space, ultimately have a small number of solutions. 

Purge-and-merge was tested on a number of constraint satisfaction problems, such as Sudoku, Fill-a-pix, Kakuro, and Calcudoku puzzles, and managed to outperform other PGM-based approaches reported in the literature~\cite{BaukeH, GoldbergerJ, KhanS}. Finally, purge-and-merge was compared to another probabilistic reasoning approach, the ACE system~\cite{acewebsite}, and was shown to be more robust when the problem domain had large factor sizes or multiple solutions.
 
\section*{Future work}
Currently, purge-and-merge is aimed at constraint satisfaction problems. The purge routine is done by loopy belief propagation, after which the merge routine is done by clustering factors and merging them. This process is then repeated by reconfiguring everything into a new cluster graph structure. We restrict our CSP potential functions to only allow potentials of a binary nature, i.e.\ values of 0 or 1. If we were to allow the potentials to be continuous, i.e.\ $0\leq \phi \leq 1$, then the posterior beliefs from belief propagation would not represent conditional distributions but rather that of marginal distributions. Since PGM inference is not designed for the use of marginal distributions as priors but rather for conditional distributions (corresponding to particular renditions of the chain rule in Equations~\ref{eq:list-of-prob-rules}), the following belief propagation step of purge-and-merge would effectively lead to overfitting of the beliefs established from the first iteration. We suggest a study into finding a scheme for calibrating the posterior beliefs of non-binary potential functions to be reused in subsequent rounds of belief propagation. This will allow purge-and-merge to be used in hybrid systems where probabilistic reasoning is mixed with constraint satisfaction.

Furthermore, it might also be possible to significantly reduce dimensionality for the purge-and-merge algorithm. 
Currently, the merge routine in purge-and-merge is performed via a factor product. This results in exponential blow-up in cases where the purge routine cannot reduce the valid system states effectively. We suggest exploring a more effective merging scheme, such as configuring factors into a junction tree (using variable elimination~\cite[p287]{koller}, for example) rather than performing a factor product. 
These junction trees can then be used as hyper-nodes within a larger cluster graph. 
This scheme would also require a system to calculate messages from junction trees and pass messages from one junction tree to another. It would also require running tree-based belief propagation on the junction tree nodes and loopy belief propagation on the hyper structured cluster graph.

{
	\backmatter
	\pagestyle{plain}%
    \if@openright\cleardoublepage\else\clearpage\fi
    \bibliographystyle{IEEETranN} %
    {\footnotesize\bibliography{
	    	biblio
    	}}
	\printindex
}
\pagestyle{plain}
\end{document}